\documentclass[journal]{IEEEtran}

\usepackage{times}
\usepackage{epsfig}
\usepackage{graphicx}
\usepackage{amsmath}
\usepackage{amssymb}
\usepackage{subfigure}
\usepackage{multirow}
\usepackage{array}
\usepackage{color}
\usepackage{booktabs}
\usepackage{caption}
\ifCLASSINFOpdf
\else
\fi
\begin{document}
%
\title{Two-Step Image Dehazing with Intra-domain and Inter-domain Adaptation}
%
%
%

\author{Xin~Yi, 
        Bo~Ma,~\IEEEmembership{Member,~IEEE,}
        Yulin~Zhang,
        Longyao~Liu,
        JiaHao~Wu
\thanks{The authors are with the Beijing Laboratory of Intelligent Information Technology, School of Computer Science and Technology, Beijing Institute of Technology, Beijing 100081, China.
(Email: yixin@bit.edu.cn; bma000@bit.edu.cn; zhangyulin@bit.edu.cn; roel\_liu@bit.edu.cn; wujiahao@bit.edu.cn)}
}

\maketitle

\begin{abstract}

Caused by the difference of data distributions, intra-domain gap and inter-domain gap are widely present in image processing tasks. In the field of image dehazing, certain previous works have paid attention to the inter-domain gap between the synthetic domain and the real domain. However, those methods only establish the connection from the source domain to the target domain without taking into account the large distribution shift within the target domain (intra-domain gap). In this work, we propose a Two-Step Dehazing Network (TSDN) with an intra-domain adaptation and a constrained inter-domain adaptation. First, we subdivide the distributions within the synthetic domain into subsets and mine the optimal subset (easy samples) by loss-based supervision. To alleviate the intra-domain gap of the synthetic domain, we propose an intra-domain adaptation to align distributions of other subsets to the optimal subset by adversarial learning. Finally, we conduct the constrained inter-domain adaptation from the real domain to the optimal subset of the synthetic domain, alleviating the domain shift between domains as well as the distribution shift within the real domain. Extensive experimental results demonstrate that our framework performs favorably against the state-of-the-art algorithms both on the synthetic datasets and the real datasets.
\end{abstract}


%
\IEEEpeerreviewmaketitle

\section{Introduction}\label{introduction}
%
%
%
%
%
%

\IEEEPARstart{H}{aze}, fog or smoke usually affects visibility and obscures key information of the images. To deal with this issue, image dehazing has been widely studied in recent years which aims to recover the clear images from their corresponding hazy images. The whole procedure can be formulated as
\begin{equation} \label{eq1}
	I(x) = J(x)t(x) + A(1-t(x))
\end{equation}
where $I(x)$ denotes the hazy image and $J(x)$ denotes the clear image, $x$ denotes a pixel position in the image, $A$ denotes the global atmospheric light and $t(x)$ denotes the transmission map. In homogeneous situation, the transmission map can be represented as $t(x) = e^{-\beta d(x)}$, where $\beta$ and $d(x)$ is the atmosphere scattering parameter and the scene depth, respectively.

\begin{figure}[t]
	\centering
	\subfigure[Input]{
		\centering
		\begin{minipage}[b]{0.31\linewidth}
			\includegraphics[width=1\linewidth]{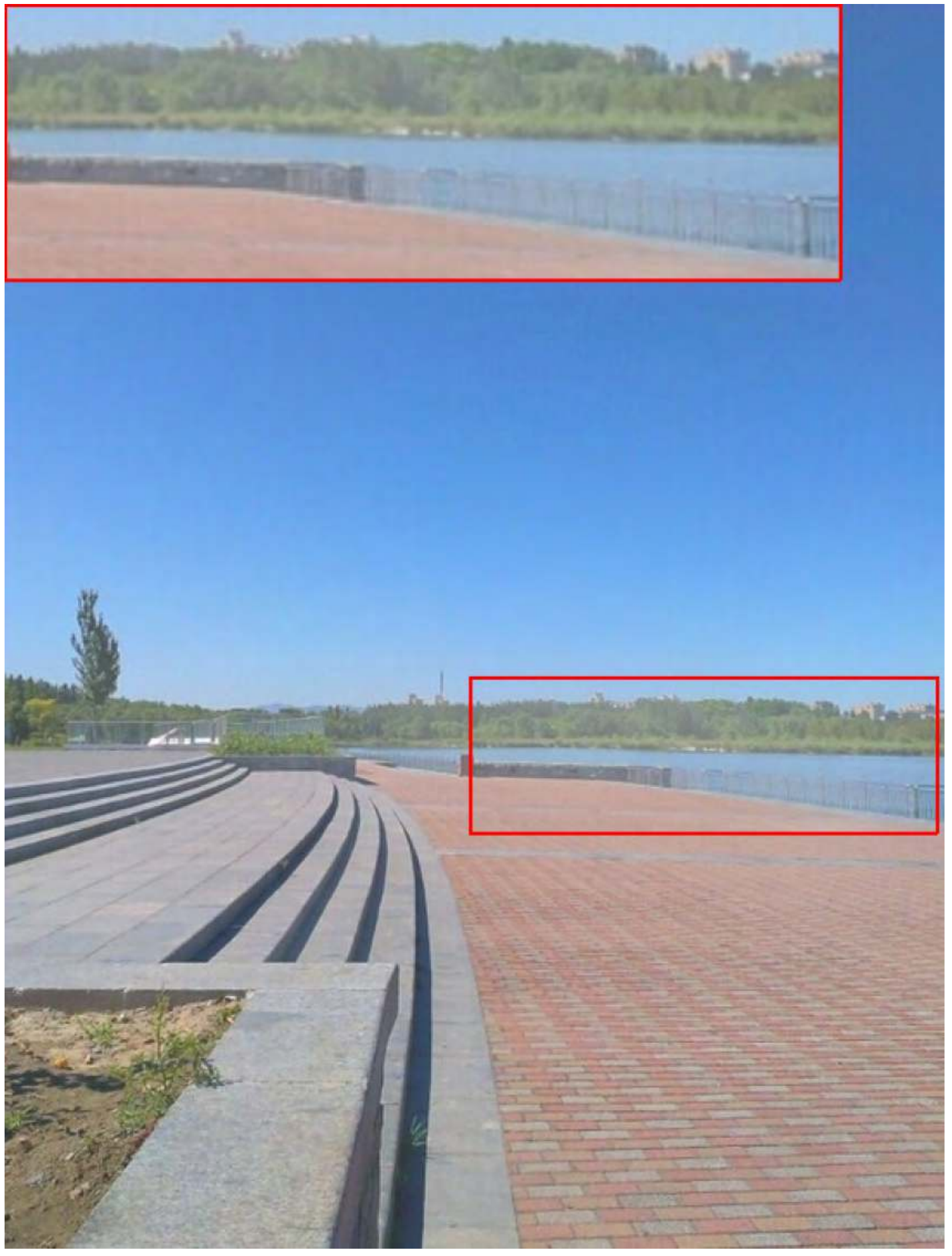}\vspace{2pt}
			\includegraphics[width=1\linewidth]{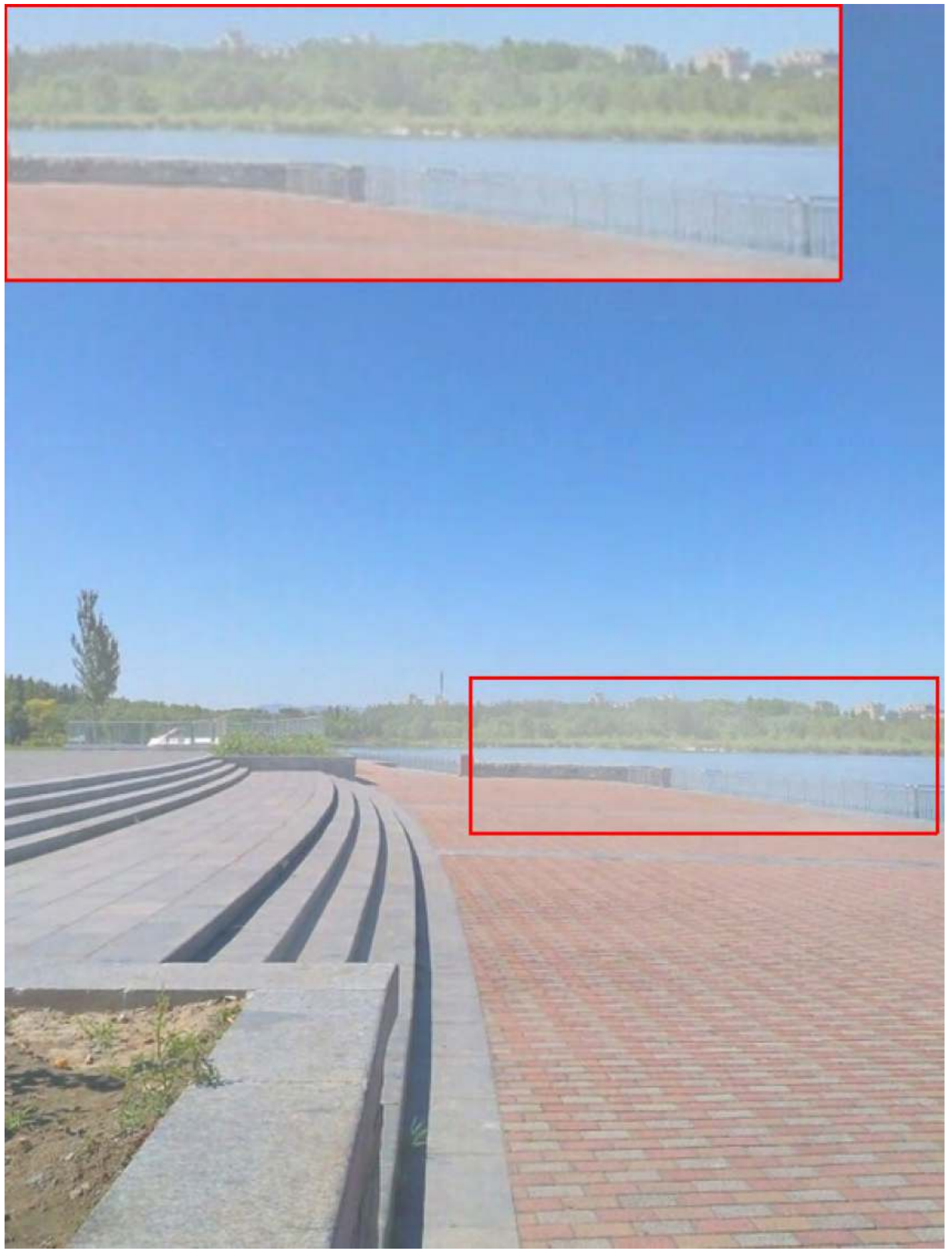}\vspace{2pt}
		\end{minipage}\hspace{-0.45em}
		\label{fig:head_a}}
	\subfigure[MSBDN-DFF~\cite{dong2020multi}]{
		\centering
		\begin{minipage}[b]{0.31\linewidth}
			\includegraphics[width=1\linewidth]{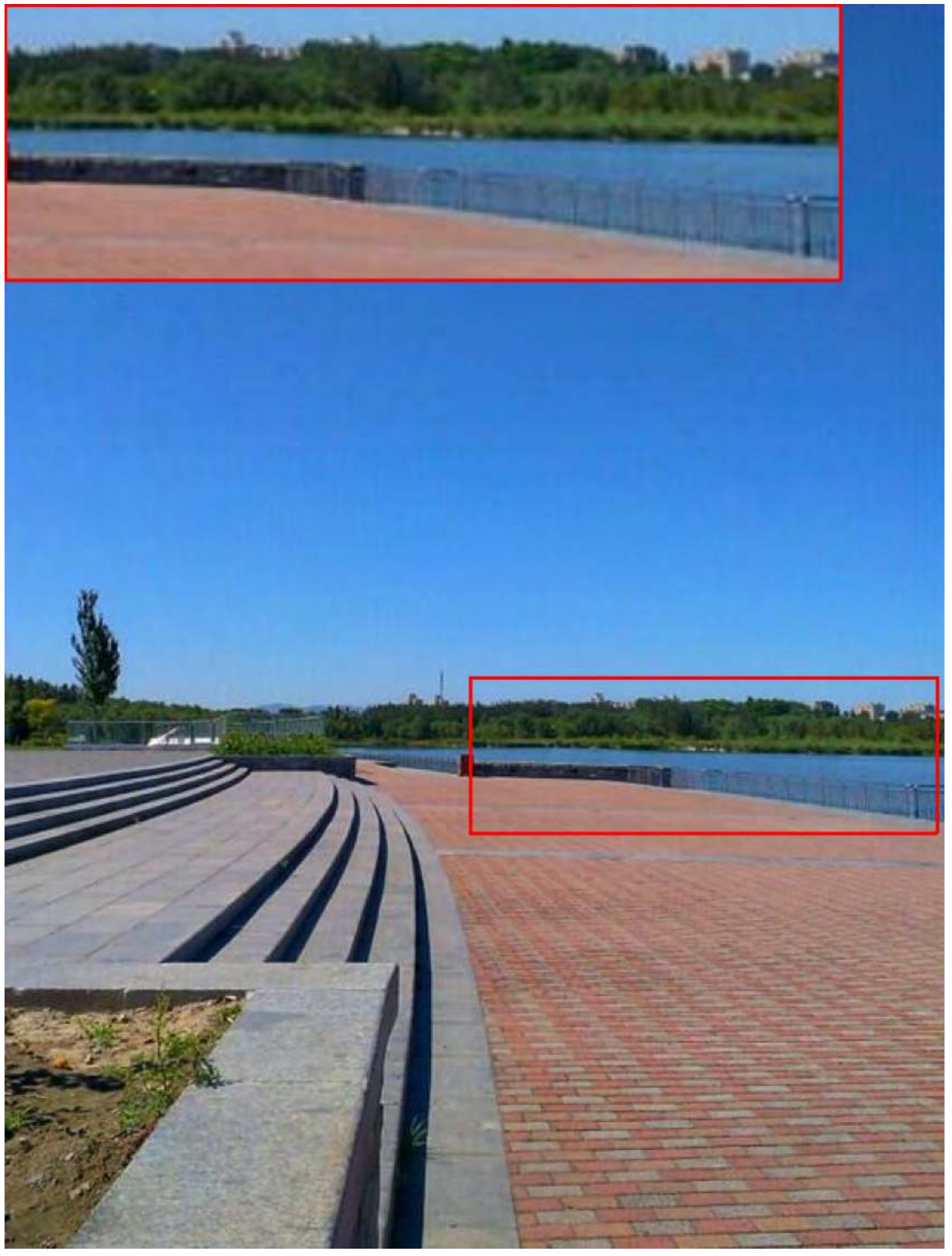}\vspace{2pt}
			\includegraphics[width=1\linewidth]{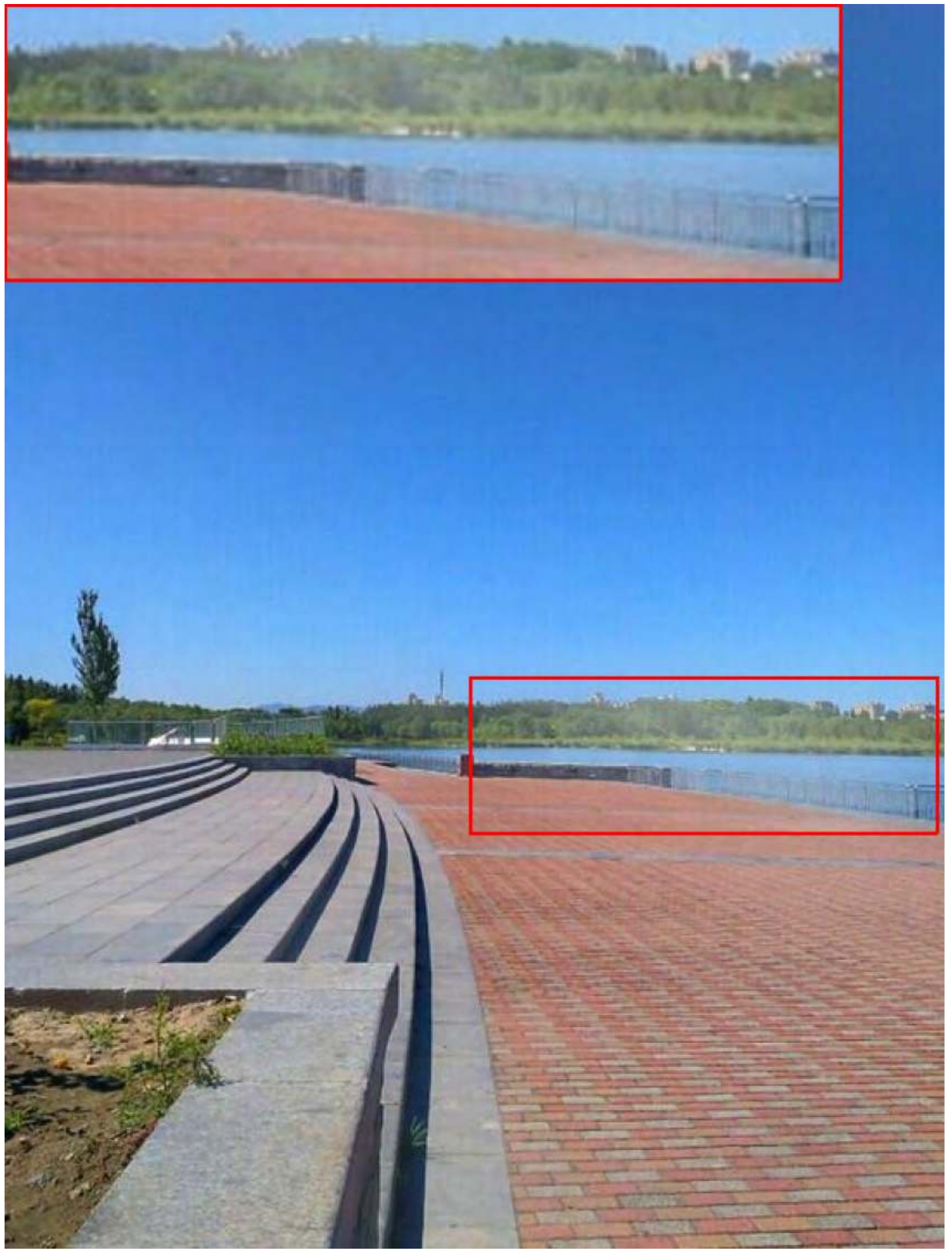}\vspace{2pt}
		\end{minipage}\hspace{-0.45em}
		\label{fig:head_b}}
	\subfigure[Ours]{
		\centering
		\begin{minipage}[b]{0.31\linewidth}
			\includegraphics[width=1\linewidth]{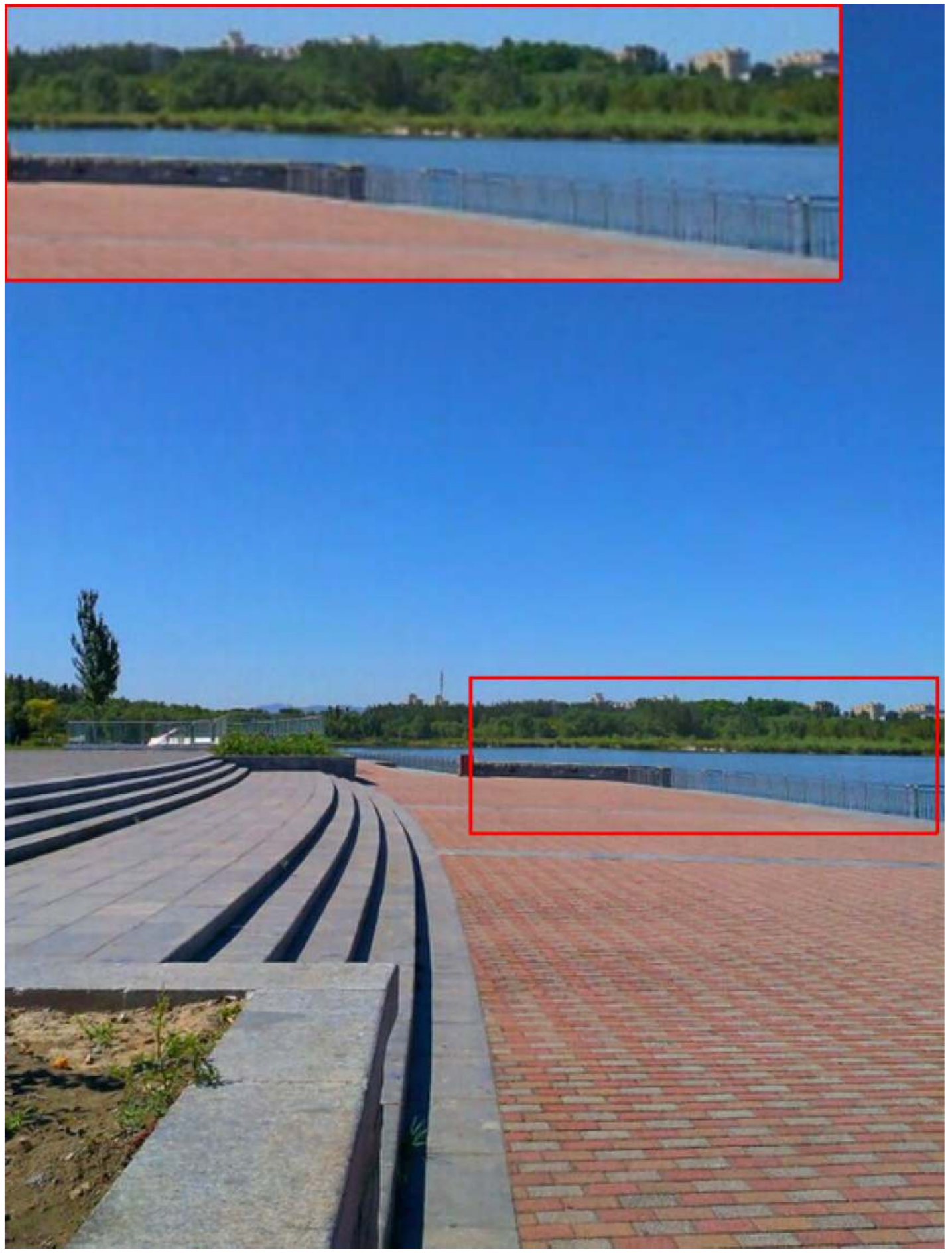}\vspace{2pt}
			\includegraphics[width=1\linewidth]{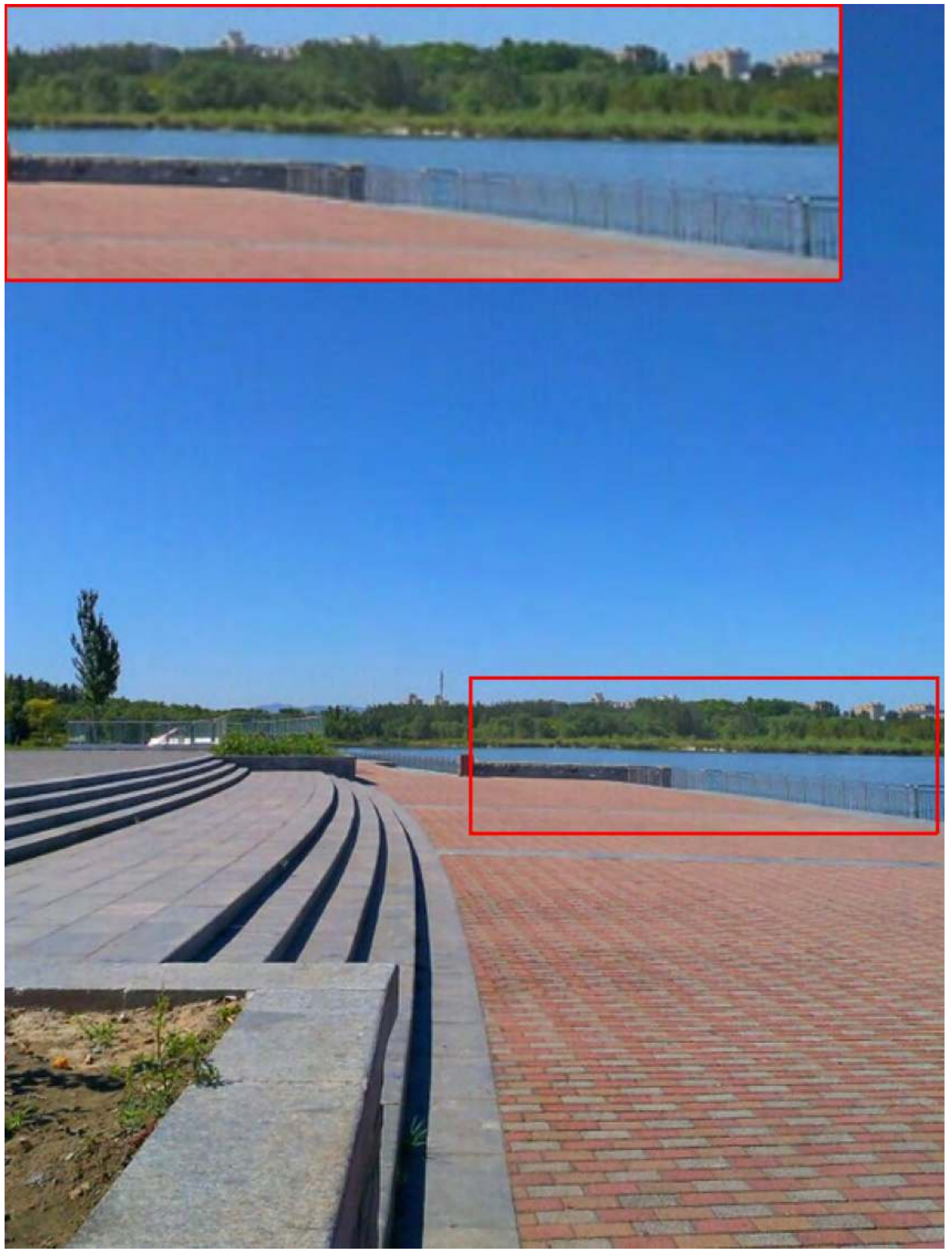}\vspace{2pt}
		\end{minipage}\hspace{-0.45em}
		\label{fig:head_c}}
	\caption{The dehazed results of images with haze distribution shift. Our method is more robust than MSBDN-DFF~\cite{dong2020multi} and can recover sharper images under different haze distributions. (a) Hazy images. (b) Results of MSBDN-DFF. (c) Our results.}
	\label{fig:head}
\end{figure}

Obviously, image dehazing is an ill-posed problem. Thus, many researchers try to transform this problem into a well-posed problem by estimating atmospheric light intensity and transmission map via certain priors~\cite{tan2008visibility, he2010single, zhu2015fast}. However, those methods are not robust and tend to fail in some scenes, especially when the color of the objects is similar to atmospheric light. More recently, to avoid hand-craft priors, other researchers apply convolutional neural network to directly predict atmospheric light intensity and transmission map from training data~\cite{cai2016dehazenet, zhang2018densely, zhang2019joint}. However, due to lack of intermediate supervision, the estimation of atmospheric light intensity and transmission map is inaccurate which leads to undesirable results. Therefore, certain follow-up works~\cite{li2018single, qu2019enhanced, qin2020ffa, dong2020multi} propose the end-to-end dehazing frameworks to circumvent the evaluation process of intermediate variables. Those methods utilize neural networks to learn a mapping from hazy images to clear images directly. However, fused information, including atmosphere information ($A$, $\beta$) and depth information ($d(x)$), is tightly coupled throughout the whole framework, causing instability in convergence and optimization. In addition, those methods do not consider the large intra-domain gap, which further reduces the robustness under different haze distributions, as illustrated in Figure~\ref{fig:head_b}. 


Although changes in haze distribution would cause changes in model performance, there are always certain easy samples, on which the model can achieve the best dehazing performance. Motivated by this phenomenon and existing deep learning methods including disentangled representation learning~\cite{chen2016infogan, higgins2016beta, tran2017disentangled, zhang2019gait, wang2020cross} and unsupervised domain adaptation (UDA)~\cite{ganin2015unsupervised, long2016unsupervised, saito2018maximum, kang2019contrastive, chang2019domain, tang2020unsupervised}, we propose an intra-domain adaptation and a constrained inter-domain adaptation in this work to address above issues. 

In intra-domain adaptation step, we design a multi-to-one dehazing framework to decouple fused information, and mine anchor distribution (easy sample) by loss-based deeply supervision. Then, we apply GAN-based adaptation to align other distributions to anchor distribution. By implementing such an information decoupling and adaptation, the difficulty of training each sub-network is reduced, e.g., the subsequent reconstruction network only needs to recover clear images from the anchor distribution, which promotes better and faster convergence of the model. More importantly, haze distribution shift within the synthetic domain is alleviated and performance under different haze distributions is improved.

Aforementioned domain adaptation within the synthetic domain achieves haze distribution invariant framework, but the gap between the real domain and the synthetic domain still exists. Thus, we propose the inter-domain adaptation based on the intra-domain adaptation to improve the generalization of the model under different domains. Although previous work~\cite{shao2020domain} has discussed the domain shift and developed a network to address it, the bridge they built from arbitrary real haze distributions to arbitrary synthetic haze distributions increases the difficulty of image dehazing when real haze distributions are aligned to hard samples of the synthetic domain. In our work, we only establish the connection from the real distributions to the anchor distributions in the synthetic domain (easy samples) instead. With this constraint, distributions of the real domain are aligned to the optimal subset of the synthetic domain, which alleviates the domain shift between domains along with the distribution shift in the real domain. In addition, this mechanism that imposing constraint on features is similar to normalization~\cite{ioffe2015batch}, making underlying optimization problem more stable and smooth.



For image dehazing on synthetic datasets and real datasets, our proposed two-step image dehazing network (TSDN) achieves state-of-the-art performance against previous algorithms. The contributions of this work are summarized as follows:

$\bullet$ We divide synthetic domain into subsets and mine the optimal subset (easy samples) by losses. By applying our proposed intra-domain adaption and information decoupling, we alleviate the distribution shift and make the optimization more stable.


$\bullet$ Based on intra-domain adaptation, we propose a constrained inter-domain adaptation between real domain and synthetic domain. By aligning real haze distributions to the optimal subset of synthetic haze distributions, we solve the domain shift between domains and the distribution shift within real domain.

$\bullet$ We conduct extensive experiments and comprehensive ablation studies on the synthetic datasets and the real datasets which validates the effectiveness of our proposed method.

$\bullet$ Our domain adaptation module can be integrated into existing dehazing frameworks for performance improvement.

\section{Related Works}


\subsection{Image Dehazing}
Previous image dehazing methods can be divided into prior based methods and learning based methods.

\subsubsection{Prior-based methods} Those methods recover clear images through statistics prior, e.g., the albedo of the scene in ~\cite{fattal2008single}. Recently, researchers have explored different priors for image dehazing~\cite{tan2008visibility, he2010single, fattal2014dehazing, zhu2015fast}. Specifically, based on the observation that clear images have higher contrast than hazy images, Tan et al.~\cite{tan2008visibility} enhance the visibility of hazy images by maximizing local contrast. He~\cite{he2010single} proposes dark channel prior (DCP) that the intensity of pixels in haze-free patches is very low in at least one color channel to achieve image dehazing. Besides, based on a generic regularity that small image patches typically exhibit a one-dimensional distribution in the RGB color space,  Fattal~\cite{fattal2014dehazing} proposes a color-lines approach to recover the scene transmission. Zhu et al.~\cite{zhu2015fast} propose color attenuation prior to recover the scene depth of the hazy image with a supervised learning method.

All above methods heavily rely on hypothetical priors. However, those priors tend to lose effectiveness in complex scene, leading to performance drop.

\subsubsection{Learning-based methods} Different from the above methods, learning-based methods use convolutional neural networks to recover clear images from hazy images directly~\cite{cai2016dehazenet,ren2016single,li2018single,qu2019enhanced,qin2020ffa, shao2020domain,dong2020multi}. Specifically, an end-to-end system for transmission estimation is proposed in ~\cite{cai2016dehazenet}. Ren et al.~\cite{ren2016single} design a multi-scale neural network for learning transmission maps from hazy images in a coarse-to-fine manner. Qiu et al.~\cite{qu2019enhanced} propose a pix2pix model with an enhancer block which reinforces the dehazing effect in both color and details. A multi-scale boosted decoder with dense feature fusion is proposed to restore clear images in~\cite{dong2020multi}. However, those methods do not take into account the intra-domain gap, resulting in less robustness in the case of haze distribution shift.

\subsection{Domain Adaptation}

The purpose of domain adaptation is to eliminate the distribution difference between labeled source domain and target domain. Recently, numerous domain adaptation approaches have been proposed, including aligning the source domain and target domain distributions, generating a mapping between two domains, or creating ensemble models~\cite{wilson2020survey}. The alignment based methods can be divided into pixel-level alignment~\cite{bousmalis2017unsupervised, shrivastava2017learning, zhu2017unpaired,kim2017learning,yi2017dualgan} and feature-level alignment~\cite{chen2019progressive,sun2016deep,tzeng2017adversarial}. The feature-level alignment methods mostly try to produce feature maps with the same distribution from images with different distributions. And the pixel-level alignment methods usually learn a transformation in the pixel space from one domain to the other~\cite{bousmalis2017unsupervised}.

With the introduction of GAN~\cite{goodfellow2014generative}, adversarial learning begins to be used in other computer vision tasks, e.g., image generation~\cite{Bao_2017_ICCV, yang2017lr, lin2019coco}, image-to-image translation~\cite{zhu2017unpaired, choi2018stargan, isola2017image, wang2018high}, etc. Among them, adversarial-based unsupervised domain adaptation (UDA) utilizes adversarial learning to learn domain invariant features. This framework usually consists of a generator and a discriminator, where they play min-max games to obtain the distribution migration from the source domain to the target domain.


In image dehazing field, Shao et al.~\cite{shao2020domain} propose a bidirectional translation network to bridge the domain gap. However, they only consider the inter-domain gap. In this work, we further minimize the intra-domain gap to achieve extra performance gains.

%

\subsection{Deeply Supervised Learning}

The deeply supervised learning is proposed in~\cite{lee2015deeply}. They apply the classifier on the deep feature layers of the neural networks to promote better convergence. Also, they draw a conclusion that more discriminative features will improve the final performance of the classifier. Recently, the deeply supervised learning is widely used in image classification~\cite{sun2019deeply}, semantic segmentation~\cite{zhang2018exfuse}, human pose estimation~\cite{newell2016stacked}.

In this work, we append auxiliary supervision branch on the feature layer. Unlike classification tasks which want to make features more discriminative, we want to make features of the same scene images less discriminative, i.e., eliminating haze distribution shift in feature space. Furthermore, our supervised learning is based on the dehazing loss so that we can ensure all features are aligned to the best one.

\begin{figure*}[t]
	\begin{center}
		\includegraphics[width=0.9\textwidth]{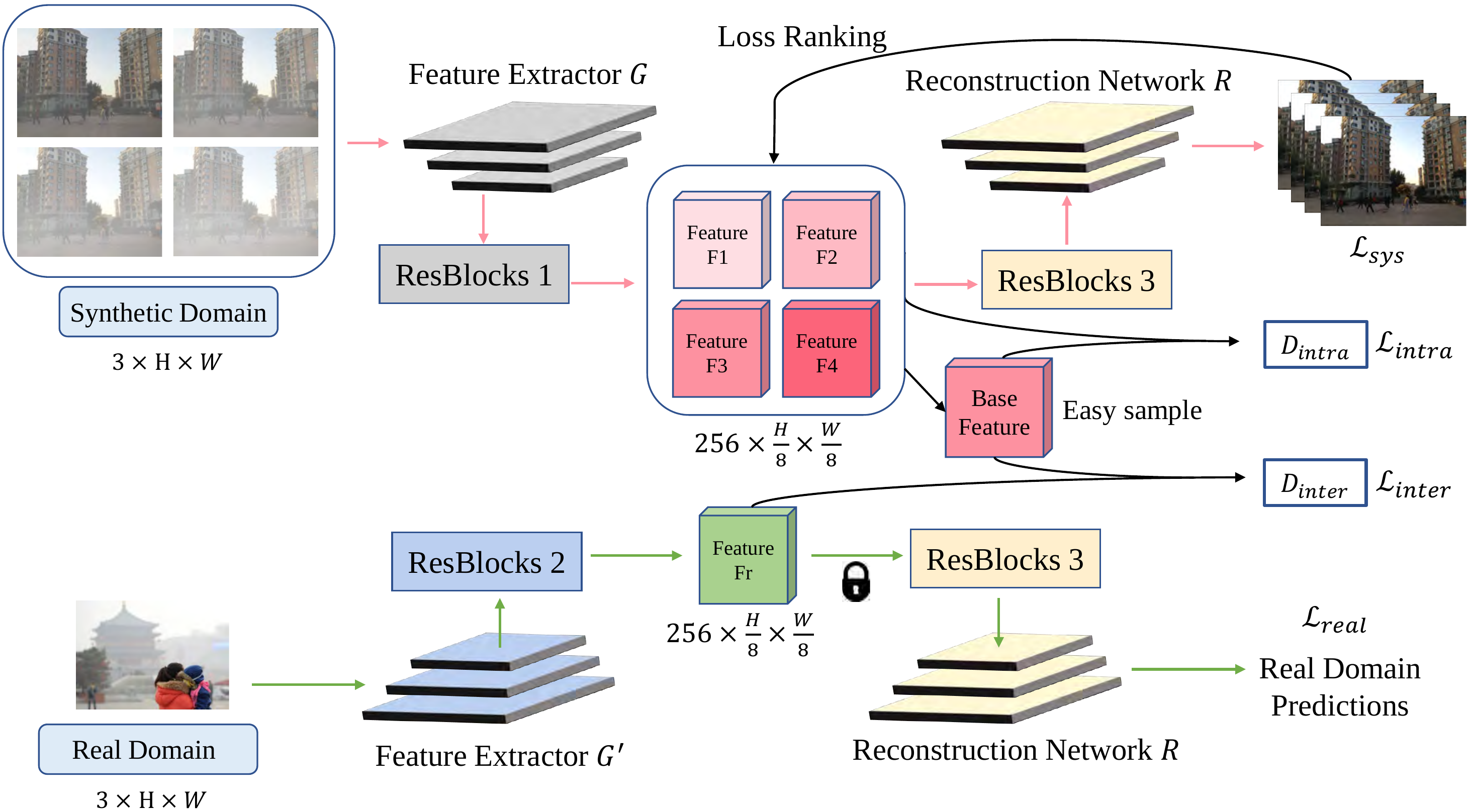}  
	\end{center}
	\caption{Illustration of our method. The overall framework consists of two main modules, the dehazing module and the domain adaptation module. The dehazing module that comprises feature extractors and reconstruction networks aims to recover clear images from haze, as depicted by red arrows and green arrows. The domain adaptation module comprises two steps, an intra-domain step and a constrained inter-domain step, aiming to close intra-domain gap and inter-domain gap. In intra-domain phase, we sort dehazing losses to mine the base feature (easy sample) and align other features to the base feature, alleviating haze distribution shift. In order to promote better convergence, this operation is performed on the same scene images to decouple fused information. In inter-domain phase, we align features of real domain only to the base features of synthetic domain, alleviating domain shift between domains as well as distribution shift in real domain.}
	\label{fig:framework}
\end{figure*}

\begin{figure}
	\begin{center}
		\includegraphics[width=0.9\linewidth]{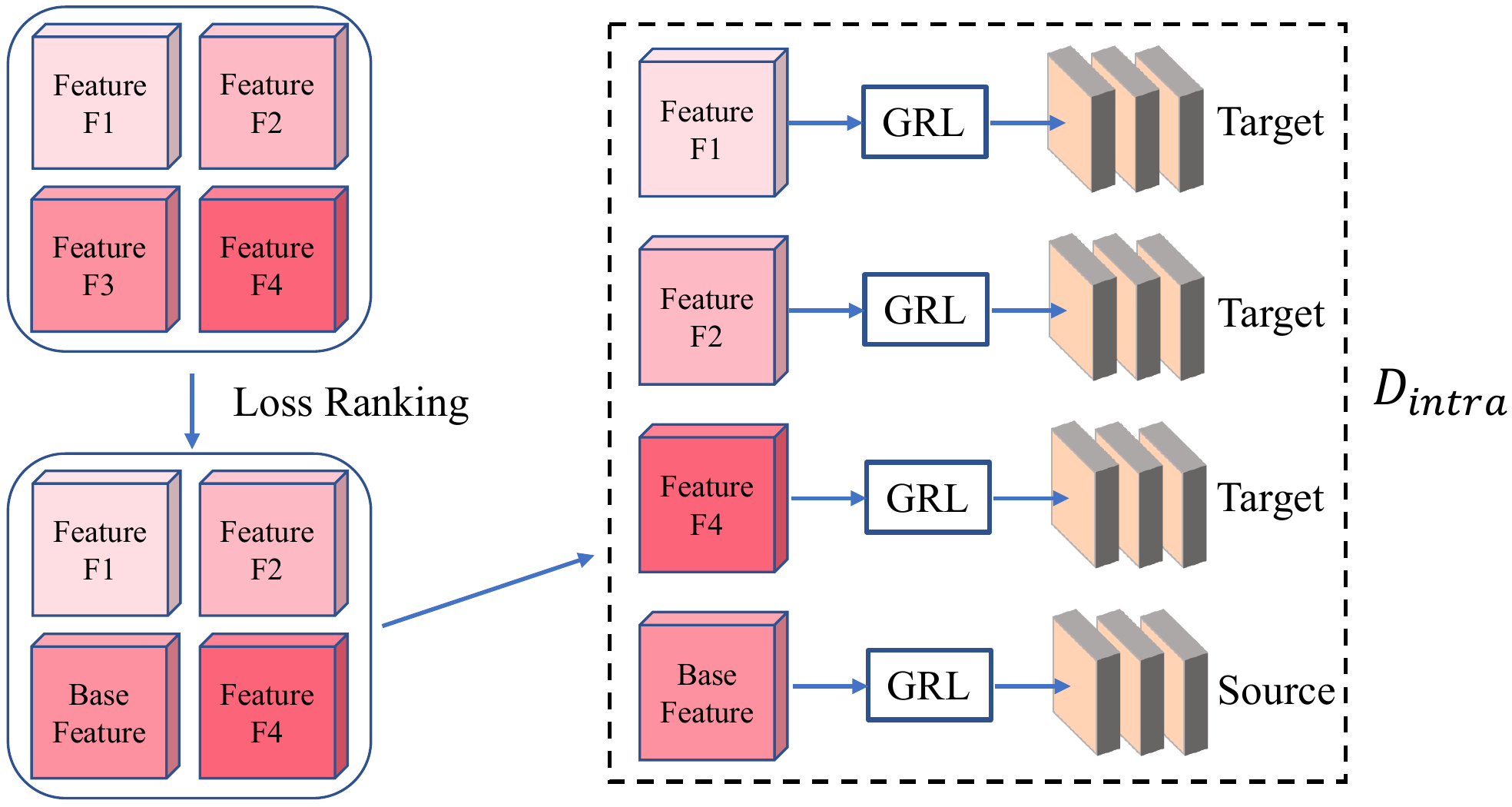}
	\end{center}
	\caption{Illustration of the intra-domain adaptation. First, we apply loss-based deep supervision to select the base feature $F_b$ (easy sample) from all features of images with the same depth information. Then, we align all other features to the base feature in order to alleviate haze distribution shift. This alignment is achieved by the intra-domain discriminator $D_{intra}$ and the GRL~\cite{ganin2015unsupervised} module, where $D_{intra}$ tries to distinguish $F_b$ from all features, and GRL module reverses the gradient so that feature extractor $G$ will generate similarly distributed features to confuse $D_{intra}$.}
	\label{fig:framework_sub}
\end{figure}

\section{Method}
In this section, we introduce our overall method in section \ref{overview}, intra-domain adaptation in section \ref{section:intra}, constrained inter-domain adaptation in section \ref{section:inter} and loss functions in section \ref{lossfunction}.  

\subsection{Method Overview}
\label{overview}

The overall framework of our work is illustrated in Figure \ref{fig:framework}. We design a multi-to-one dehazing framework to decouple fused information and mine easy samples. We append two auxiliary discriminators after feature extractor, guiding the network to learn distribution-invariant/domain-invariant fetaures based on those easy samples. Thus, all the features in the synthetic domain and the real domain would be aligned to the features of easy samples, which makes optimization more stable and performance more robust.  


We first minimize the intra-domain gap then the inter-domain gap. In the intra-domain step, to reduce the effect of excessive differences in depth information, we take a set of hazy images in the same scene but with different haze distributions as input and get their corresponding features by the feature extractor $G$. Then, we select base feature $F_b$ (easy sample) by comparing their dehazing losses and align all other features to the base feature $F_b$ by the intra-domain discriminator $D_{intra}$. Since those features possess the same depth information, this alignment is actually to decouple atmosphere information $A$ and $\beta$ from fused information and approximate them to the optimal $A^*$ and $\beta^*$, where $A^*$ and $\beta^*$ serve as the anchor of different haze distributions. Thus, distribution shift can be alleviated in the feature space and subsequent module only need to learn how to reconstruct clear images from the easy sample, which accelerates convergence and improves performance. Finally, for the subsequent inter-domain adaptation, we mark easy samples in all scenes as the optimal subset of the synthetic domain.




In the inter-domain step, we begin by obtaining features of synthetic domain and real domain using feature extractor $G$ and $G'$, respectively. The hazy images of the synthetic domain are all selected from the optimal subset (marked in the intra-domain step), so the features of them are all base features (easy samples). Then, we perform the inter-domain adaptation from features of real domain to those base features using the inter-domain discriminator $D_{inter}$. By adding above constraints to the targets, we resolve the inter-domain gap along with intra-domain gap in the real domain. 



\subsection{Intra-domain Adaptation}
\label{section:intra}

Generally, a clear image corresponds to multiple hazy images with different haze distributions. The goal of intra-domain adaptation is to align those hazy images and improve the performance on each image. To this end, we apply adversarial alignment in feature space via intra-domain discriminator $D_{intra}$.

Suppose we have $n$ hazy images $\{x_i\in \mathbb{R}^{H\times W\times 3}\}_{i=1}^{n}$ that belong to the same scene, we can extract $n$ features $\{F_i\in \mathbb{R}^{h\times w\times k}\}_{i=1}^{n}$ with a designed feature extractor network $G$. To find the base feature $F_b$ to which all other features are aligned, we apply deeply supervised learning based on the dehazing losses. Specifically, we input all features into the reconstruction module to get their corresponding haze-free predictions and dehazing losses $L_{sys}$. According to those losses, we pick the feature with the lowest dehazing loss as base feature $F_b$. Then, base feature $F_b$ along with other features  $F_j$ $(j=1,2,...,n \text{ and } j\neq b)$ are fed into a fully-convolutional network $D_{intra}$ to generate intra-domain classification score maps. The score map has the same spatial resolution as the feature where each pixel position represents the intra-domain prediction of the same position in the feature. The loss function between the predicted classification score map and the label is binary cross-entropy loss which can be written as:
\begin{equation}
	\label{formula:intra}
	\begin{aligned}
		\mathcal{L}_{intra} &= -\frac{1}{n-1}\sum_{j}\sum_{h,w}y\log(D_{intra}(F_b^{(h,w)})) \\ &+(1-y)\log(1-D_{intra}(F_j^{(h,w)}))
	\end{aligned}
\end{equation}
where $(h,w)$ denotes a pixel position in the feature, $y$ denotes intra-domain label and $n$ denotes the total number of features we extracted. In our work, we set label $y$ of source and target as 1 and 0, respectively. Correspondingly, base feature $F_b$ is the intra-domain source and other features $F_j$ are the intra-domain target. For the discriminator $D_{intra}$, we optimize it using  loss function Eq.(\ref{formula:intra}). For feature extractor $G$, we apply gradient reversal layer (GRL)~\cite{ganin2015unsupervised} to perform adversarial learning. The pipeline of the deeply supervised learning and the intra-domain adaptation is illustrated in Figure \ref{fig:framework_sub}. The discriminator $D_{intra}$ tries to distinguish $F_b$ from $F_j$ ($j=1,2,4$) while feature extractor $G$ tries to generate similarly distributed $F_b$ and $F_j$ to confuse $D_{intra}$.

\subsection{Constrained Inter-domain Adaptation}
\label{section:inter}
The goal of the inter-domain adaptation is to close the inter-domain gap between the synthetic and real domains. We propose a constrained inter-domain adaptation which builds a bridge between real haze distributions and the optimal subset of the synthetic domain instead of the whole synthetic domain.

We perform the inter-domain adaptation in feature space by adversarial learning. Particularly, given a synthetic hazy image marked as an easy sample $x_s\in \mathbb{R}^{H\times W\times 3}$ and a real hazy image $x_r\in \mathbb{R}^{H\times W\times 3}$, we extract their features $F_s\in \mathbb{R}^{h\times w\times k}$ and $F_r\in \mathbb{R}^{h\times w\times k}$ by extractor network $G$ and $G'$, respectively. Then, we obtain domain classification prediction maps of $F_s$ and $F_r$ by a fully-convolutional discriminator network $D_{inter}$. The prediction map has the same spatial shape as input and each position on it denotes domain label of the same position on input. We apply binary cross-entropy loss between the classification score map and the label, which can be written as:
\begin{equation}
	\begin{aligned}
		\mathcal{L}_{inter} &= -\sum_{h,w}z\log(D_{inter}(F_s^{(h,w)})) \\
		&+(1-z)\log(1-D_{inter}(F_r^{(h,w)}))
	\end{aligned}
	\label{formula:inter}	
\end{equation}
where $(h,w)$ denotes a pixel position in the feature and $z$ denotes inter-domain label. We set the synthetic domain as source and real domain as target, where source label is 1 and target label is 0. For feature extractor $G'$, we also apply gradient reversal layer (GRL)~\cite{ganin2015unsupervised} to perform adversarial learning. In addition, we freeze reconstruction network $R$ at the beginning of the inter-domain adaptation during training to ensure that the real domain features fall into the optimal subset of the synthetic domain.

\subsection{Loss Functions}
\label{lossfunction}
Given a synthetic dataset $D_{sys}$ and a real dataset $D_{real}$, where $D_{sys}$ consists of a hazy subset $I_{haze}=\{x_h\}_{h=1}^{N_h}$ and a clear subset $I_{clear}=\{x_c\}_{c=1}^{N_c}$ while $D_{real}$ only contains a hazy set $J_{haze}=\{x_r\}_{r=1}^{N_{r}}$, We adopt following loss functions in our framework.

\subsubsection{Domain Adversarial Losses} As described in section \ref{section:intra} and \ref{section:inter},  domain adversarial losses are generated by $D_{intra}$ and $D_{inter}$. On the scale of the entire dataset, the intra-domain loss can be written as: 
\begin{equation}
	\mathcal{L}_{1} = \sum_{c=1}^{N_c} \mathcal{L}_{intra}
	\label{formula:whole_intra}
\end{equation}
and the inter-domain loss can be written as:
\begin{equation}
	\mathcal{L}_{2} = \sum_{i=1}^{N_i} \mathcal{L}_{inter}
	\label{formula:whole_inter}
\end{equation}
where $N_i$ denotes minimum of $N_c$ and $N_{r}$.


\subsubsection{Image Dehazing Losses}Those losses measure the difference between the predicted images and the ground truth. In the synthetic domain, we apply $L1$ loss to make sure the dehazed results are close to the clear images. Since a clear image corresponds to multiple hazy images, we further define the hazy subset as $I_{haze}=\{x_h^{(c)}, c=1,2,...,N_c\}_{h=1}^{N_h}$, where $x_h^{(c)}$ represents the $h$-th hazy image corresponding to the $c$-th clear image, $N_c$ denotes the total number of clear images and $N_h$ denotes the total number of hazy images. Thus, the predicted clear images can be defined as $I_{pre}=\{y_h^{(c)},c=1,2,...,N_c\}_{h=1}^{N_h}$. The dehazing loss between $I_{pre}$ and $I_{clear}$ are defined as:
\begin{equation}
	\mathcal{L}_{sys} = \frac{1}{N_h} \sum_{h=1}^{N_h}\left\|y_h^{(c)}- x_c\right\|_1
\end{equation}
Besides, in order to improve the performance of our model in the real domain, we add the dark channel prior loss~\cite{he2010single} and the total variation loss on the predicted real images~\cite{shao2020domain}. We divide an image into $n$ patches and define the overall dark channel loss as:
\begin{equation}
	\mathcal{L}_{dc} = \frac{1}{n}\sum_{x}^{n}\left\|I^{dark}(x)\right\|_{1}
\end{equation}
where $x$ represents a patch and $I^{dark}(x)$ denotes the dark channel prior. The total variation loss is defined as:
\begin{equation}
	\begin{aligned}
		\mathcal{L}_{tv} &= \frac{1}{w}\sum_{i}^{w}\left\|I_{i+1,j}-I_{i,j}\right\|_1 \\
		&+\frac{1}{h}\sum_{j}^{h}\left\|I_{i,j+1}-I_{i,j}\right\|_1
	\end{aligned}
\end{equation}
where $i$ and $j$ denote the horizontal position and the vertical position of an image, respectively. $w$ is the width of the image and $h$ is the height. So, the image dehazing loss in the real domain can be written as:
\begin{equation}
	\mathcal{L}_{real} = \lambda_{dc} \mathcal{L}_{dc} + \lambda_{tv} \mathcal{L}_{tv}
\end{equation}
\subsubsection{Overall Loss} The overall loss is defined as weighted sum of all losses. In intra-domain training phase, the overall loss can be written as:
\begin{equation}
	\mathcal{L}_{altra} = \lambda_1 \mathcal{L}_{1} + \lambda_3 \mathcal{L}_{sys}
\end{equation}
while in inter-domain training phase, the overall loss can be written as:
\begin{equation}
	\mathcal{L}_{alter} = \lambda_2 \mathcal{L}_{2} + \lambda_3 \mathcal{L}_{sys} + \lambda_4 \mathcal{L}_{real}
\end{equation}

%

\section{Experiments}

We introduce related experiments and ablation studies in this section to verify our proposed method.

\begin{table*}[htb]
	\caption{Quantitative evaluation of the dehazing results on SOTS~\cite{li2019benchmarking} and HazeRD\cite{zhang2017hazerd} datasets.}
	\label{quantitative results}
	\resizebox{\textwidth}{!}{%
		\begin{tabular}{cccccccccccc}
			\hline
			\rule{0pt}{12pt} 
			&      & DCP~\cite{he2010single} & DehazeNet~\cite{cai2016dehazenet} & DCPDN~\cite{zhang2018densely} & EPDN~\cite{qu2019enhanced} & GFN~\cite{ren2018gated}  & GDN~\cite{liu2019griddehazenet} & DAdehazing~\cite{shao2020domain} & MSBDN-DFF~\cite{dong2020multi} & Ours \\ \hline
			\rule{0pt}{10pt} 
			\multirow{2}{*}{SOTS~\cite{li2019benchmarking}} & PSNR  & 15.49  & 21.14 & 19.39 & 23.82 & 22.30     & 31.51 & 27.76      & 33.79     & \textcolor{red}{35.26} \\
			& SSIM & 0.646  & 0.853   & 0.659  & 0.893  & 0.886      & 0.982 & 0.928       & 0.983     & \textcolor{red}{0.985} \\ \hline
			\rule{0pt}{10pt} 
			\multirow{2}{*}{HazeRD~\cite{zhang2017hazerd}} & PSNR  & 14.01  & 15.54 & 16.12 & 17.53 & 14.83     &  15.12 & 18.07      & 18.40     & \textcolor{red}{19.84} \\
			& SSIM & 0.390  & 0.432   & 0.407  & 0.593  & 0.802      & 0.833 & 0.632       & 0.881     & \textcolor{red}{0.892} \\ \hline
	\end{tabular}}
\end{table*}

\begin{figure*}[tb]
	\centering
	\subfigure[\scriptsize{Input}]{
		\begin{minipage}[b]{0.12\linewidth}
			\includegraphics[width=1\linewidth]{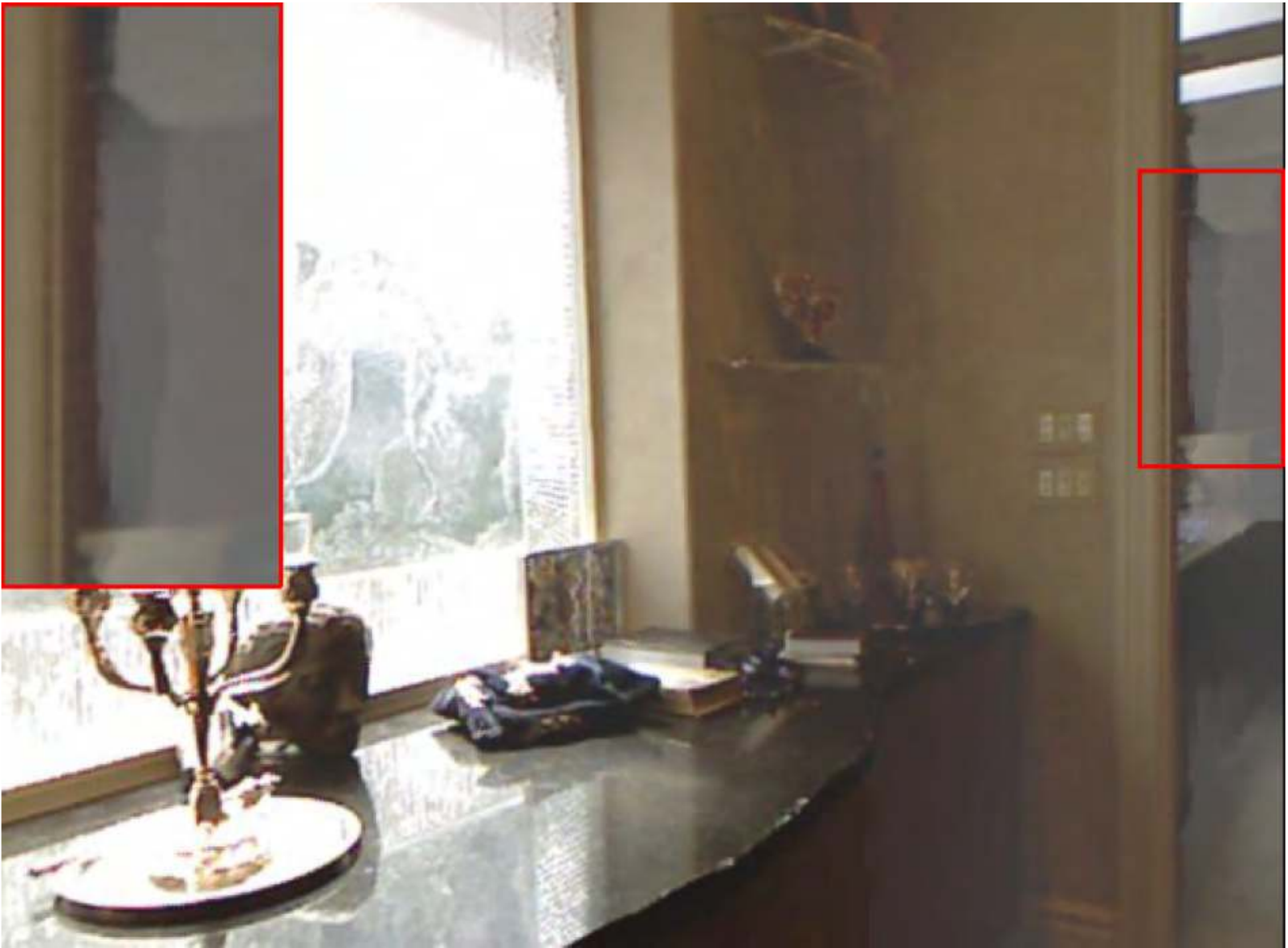}\vspace{2pt}
			\includegraphics[width=1\linewidth]{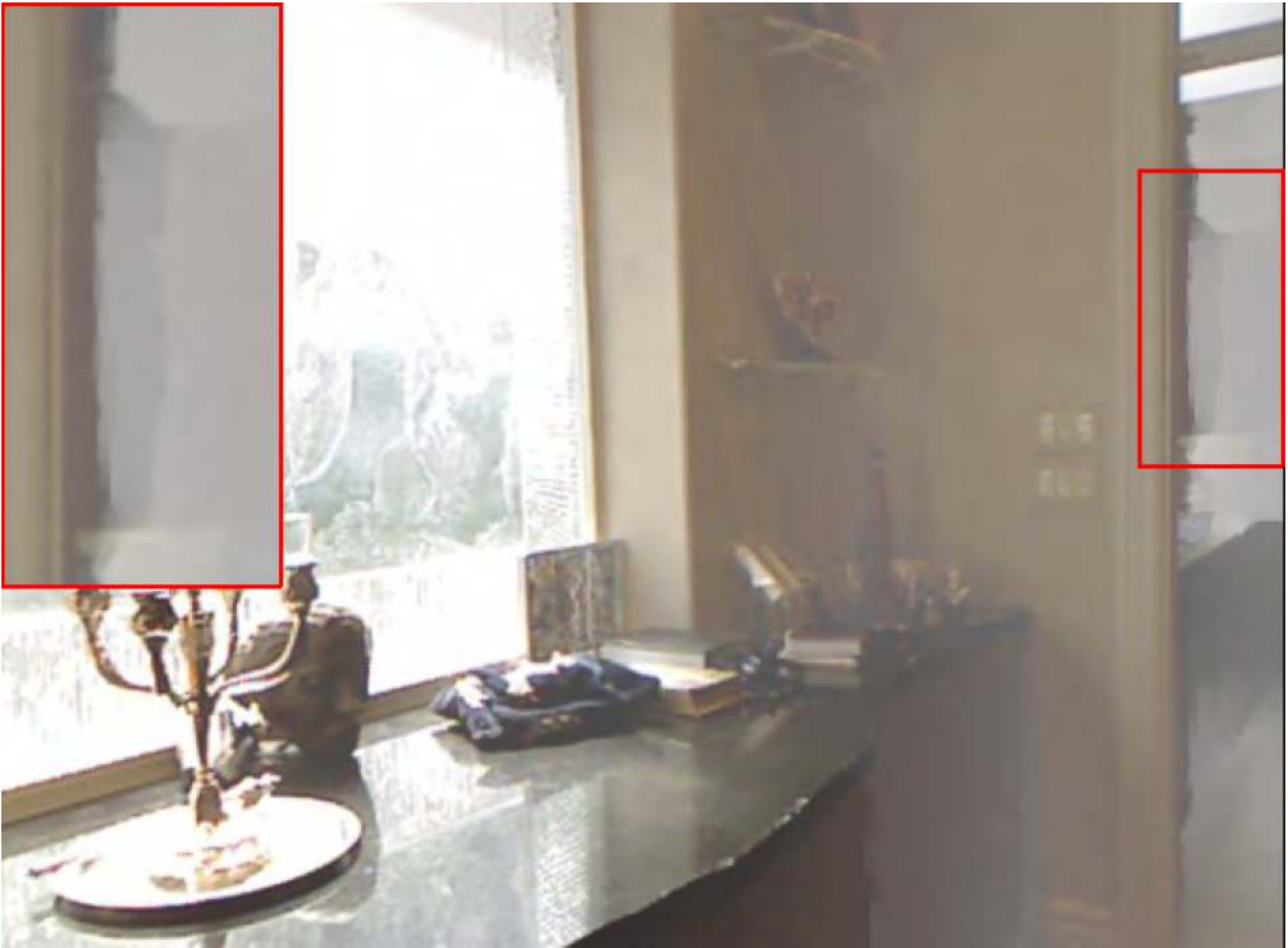}\vspace{2pt}
			\includegraphics[width=1\linewidth]{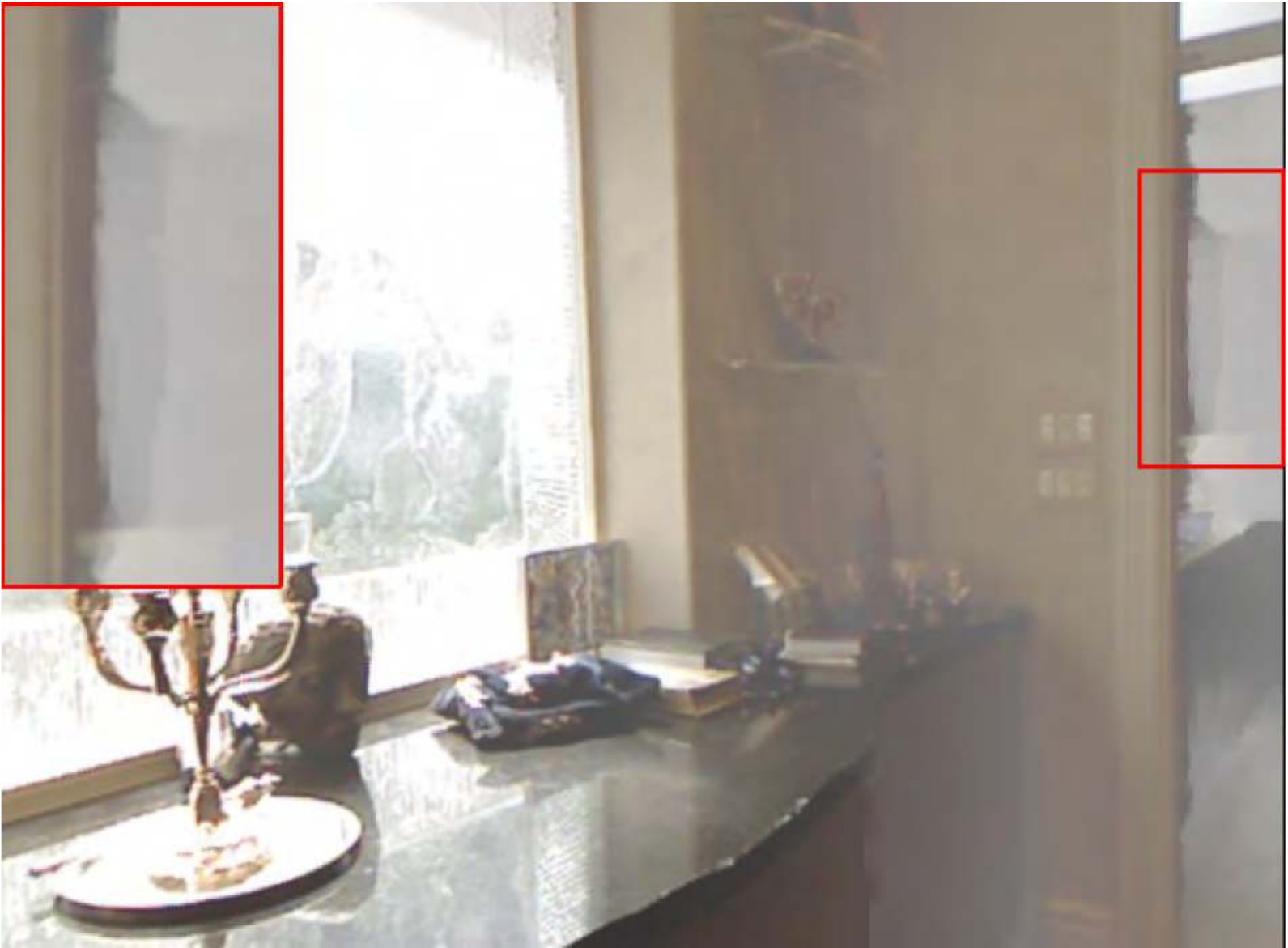}\vspace{2pt}
			\includegraphics[width=1\linewidth]{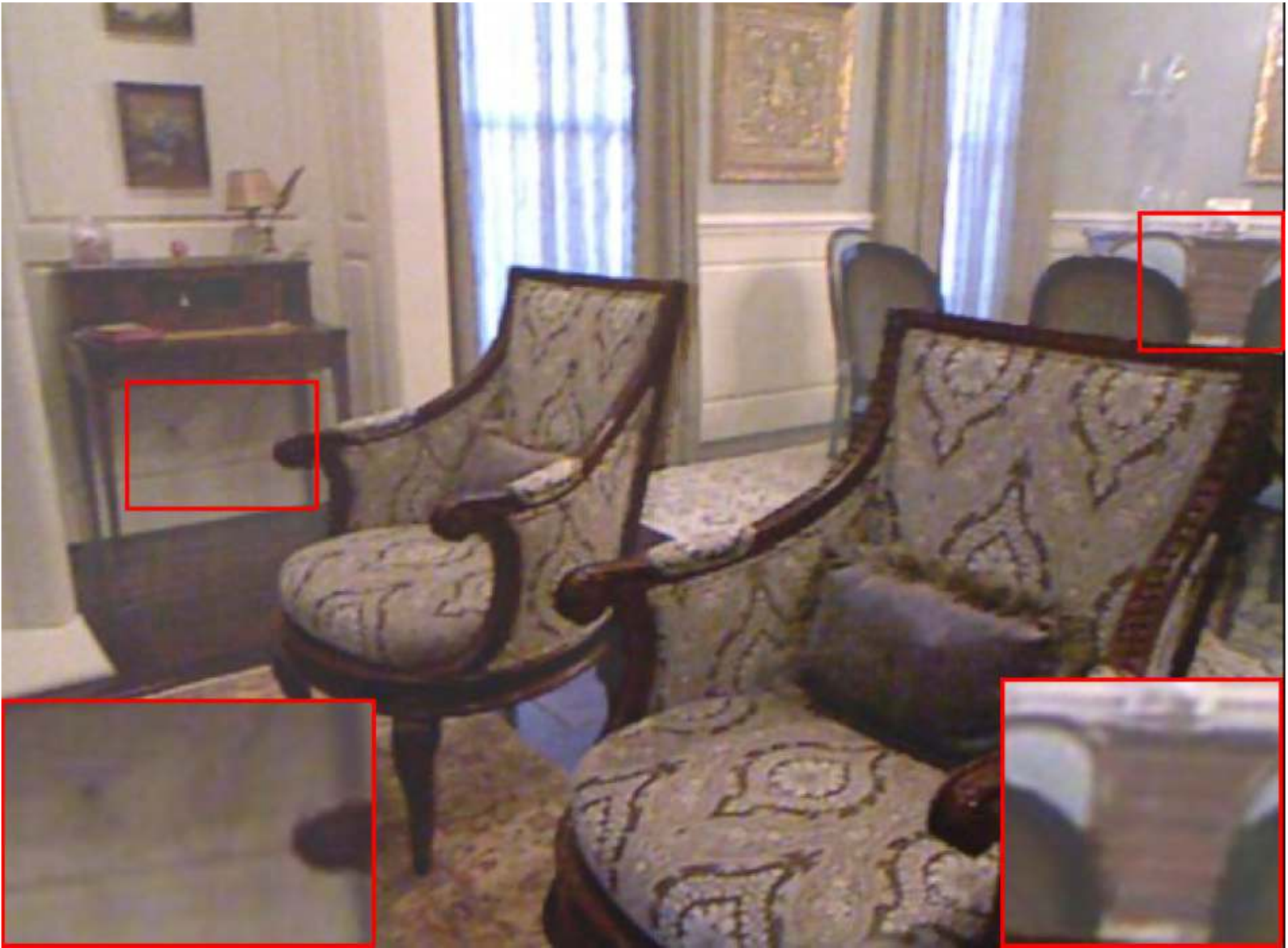}\vspace{2pt}
			\includegraphics[width=1\linewidth]{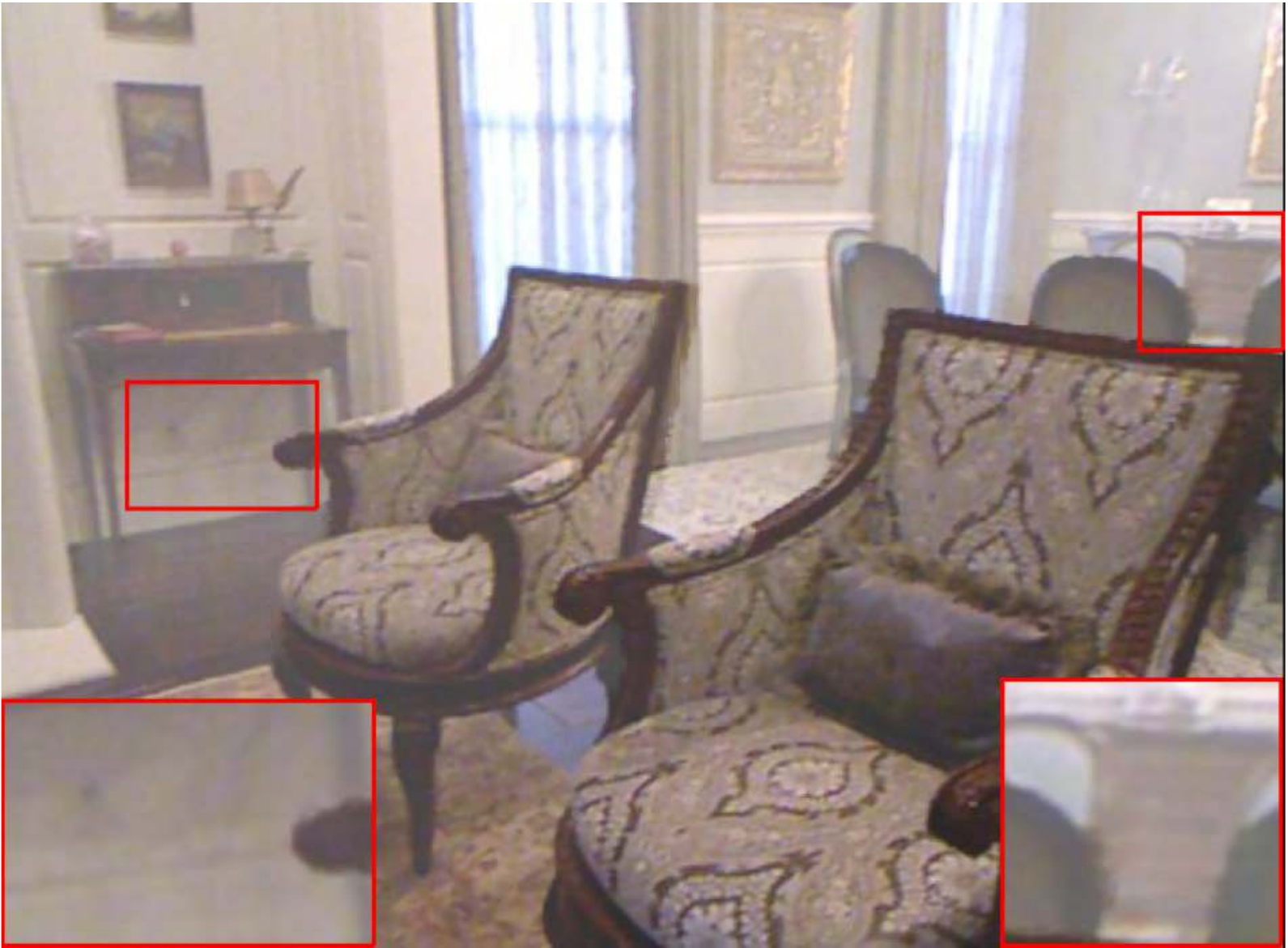}\vspace{2pt}
			\includegraphics[width=1\linewidth]{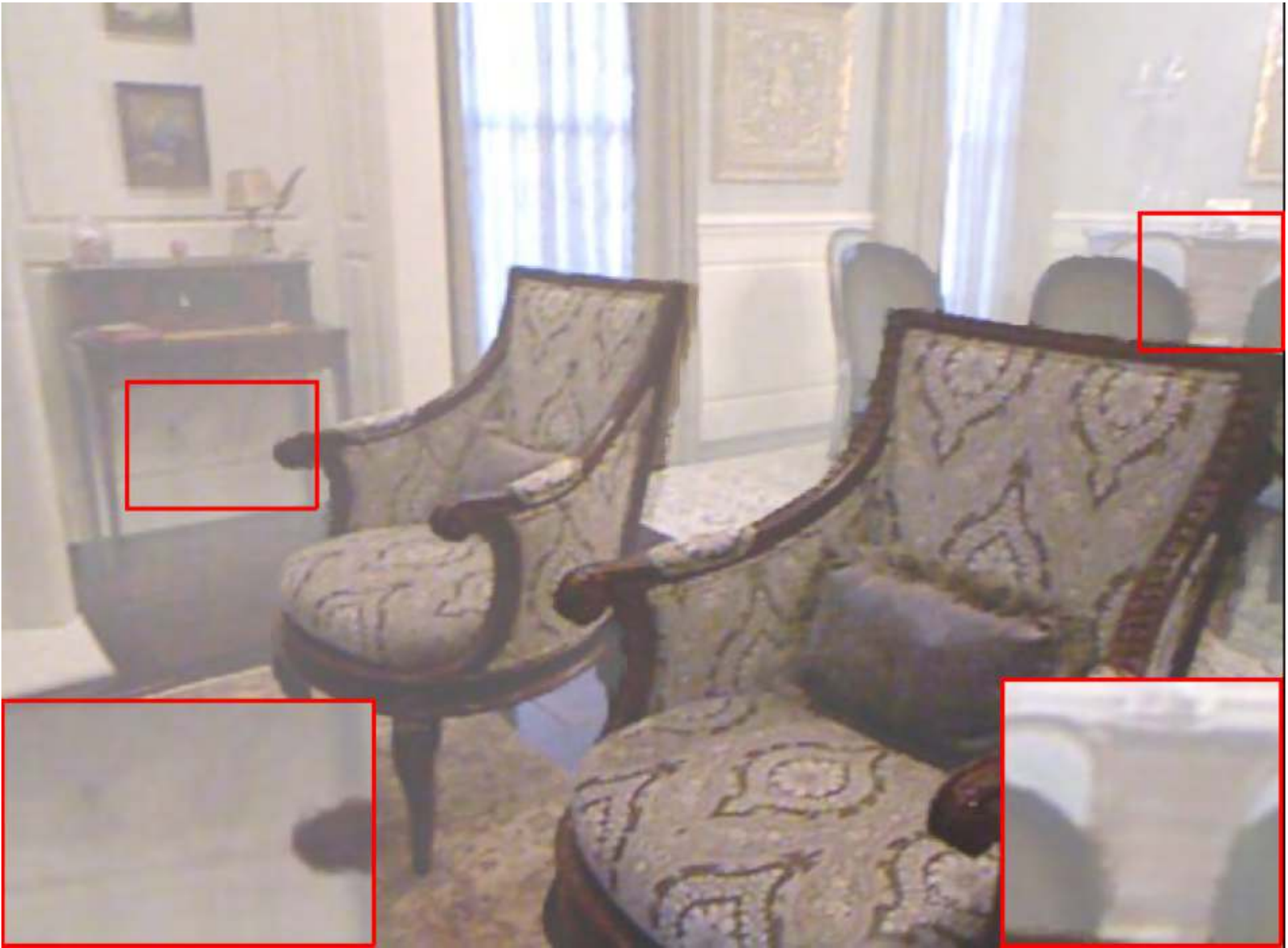}\vspace{2pt}
	\end{minipage}}\hspace{-0.45em}
	\subfigure[\scriptsize{DCP~\cite{he2010single}}]{
		\begin{minipage}[b]{0.12\linewidth}
			\includegraphics[width=1\linewidth]{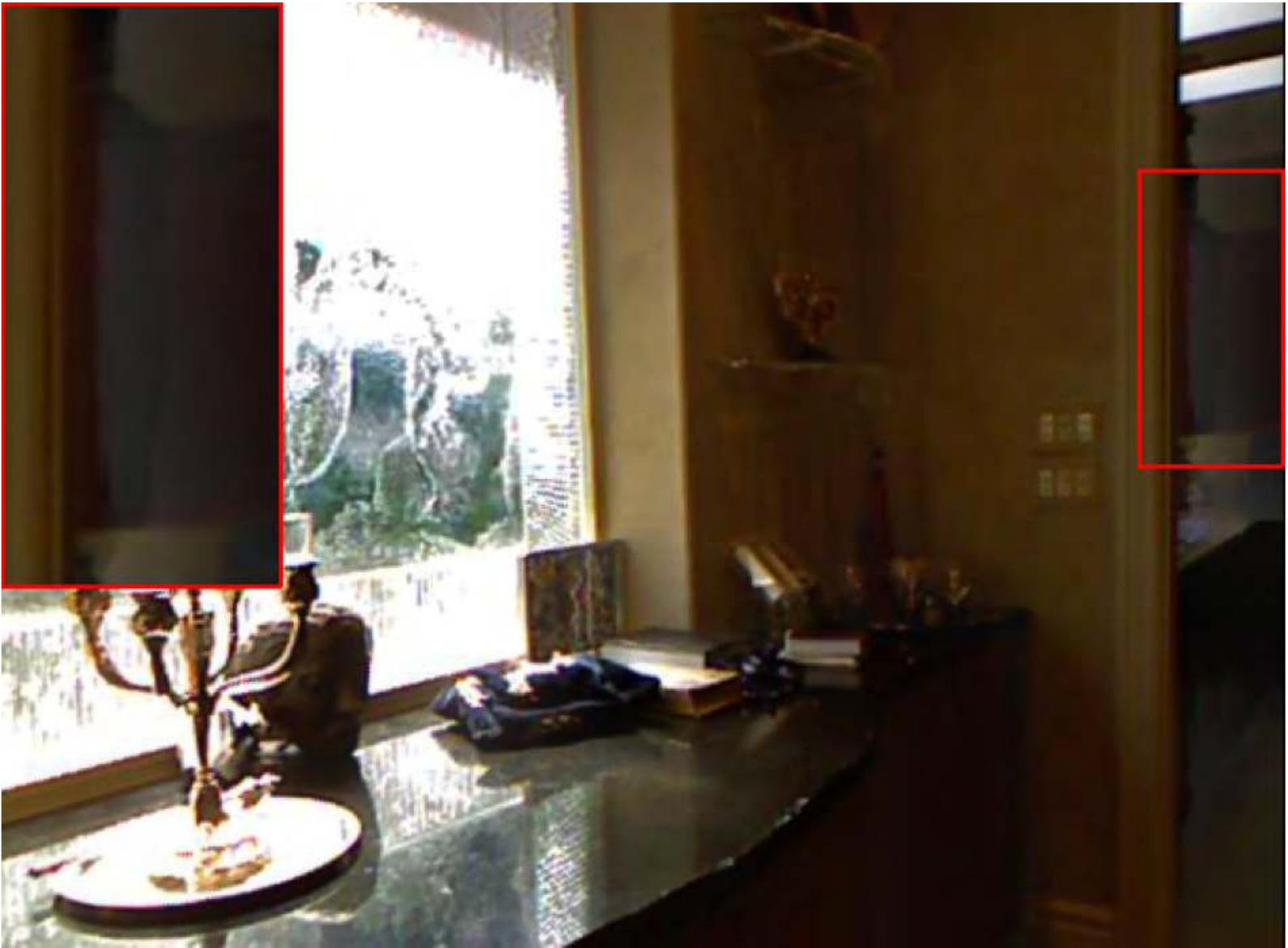}\vspace{2pt}
			\includegraphics[width=1\linewidth]{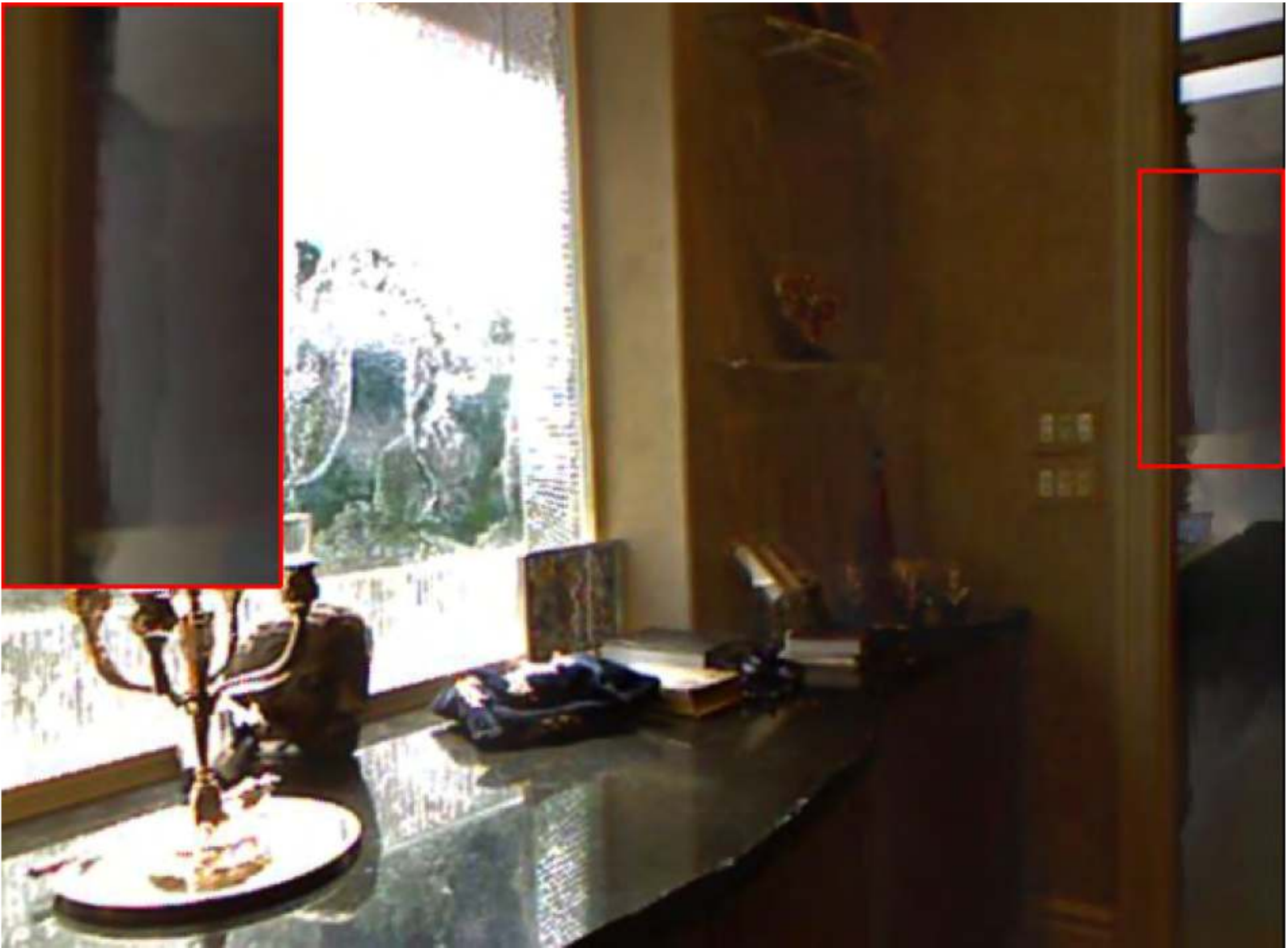}\vspace{2pt}
			\includegraphics[width=1\linewidth]{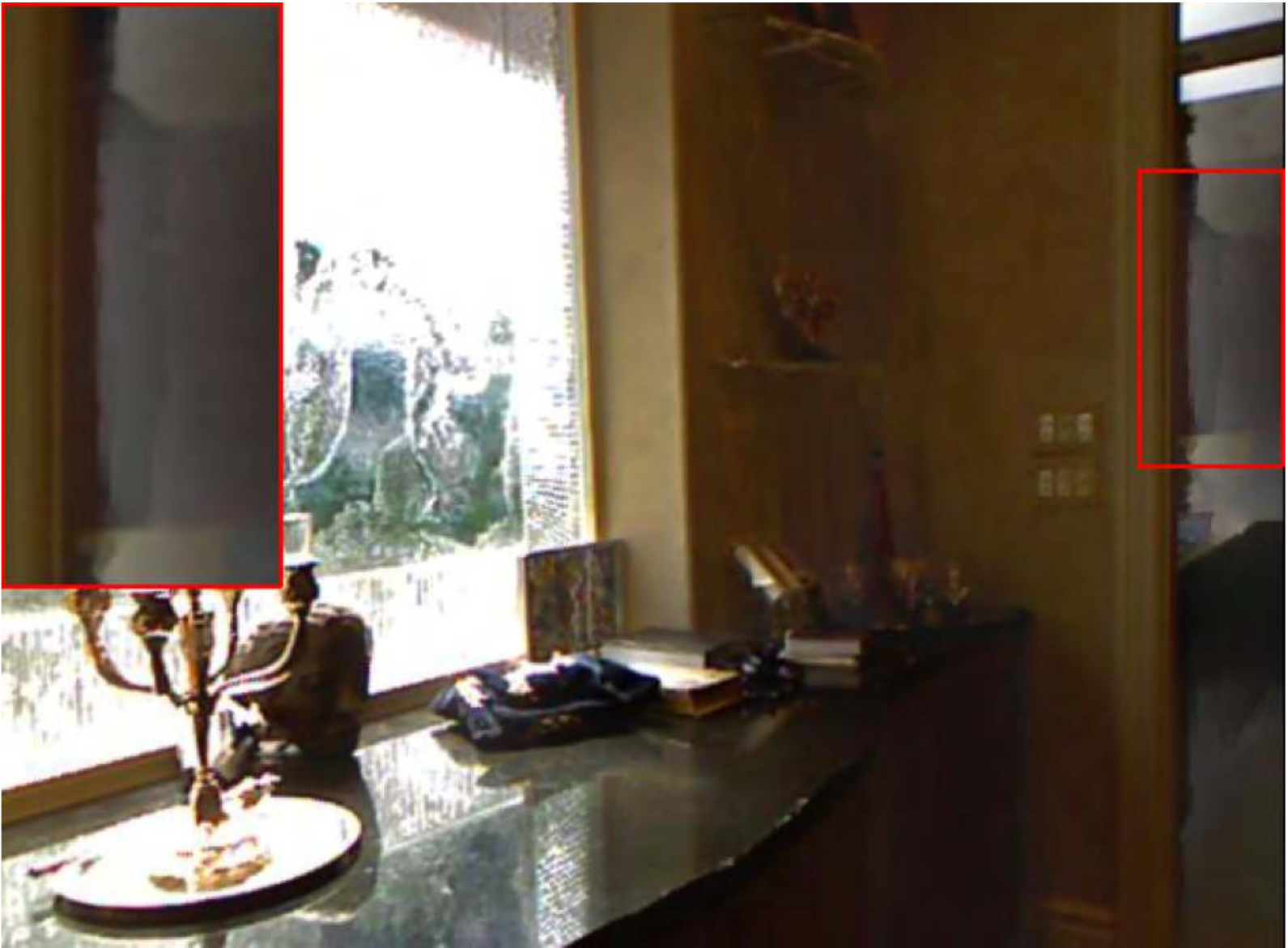}\vspace{2pt}
			\includegraphics[width=1\linewidth]{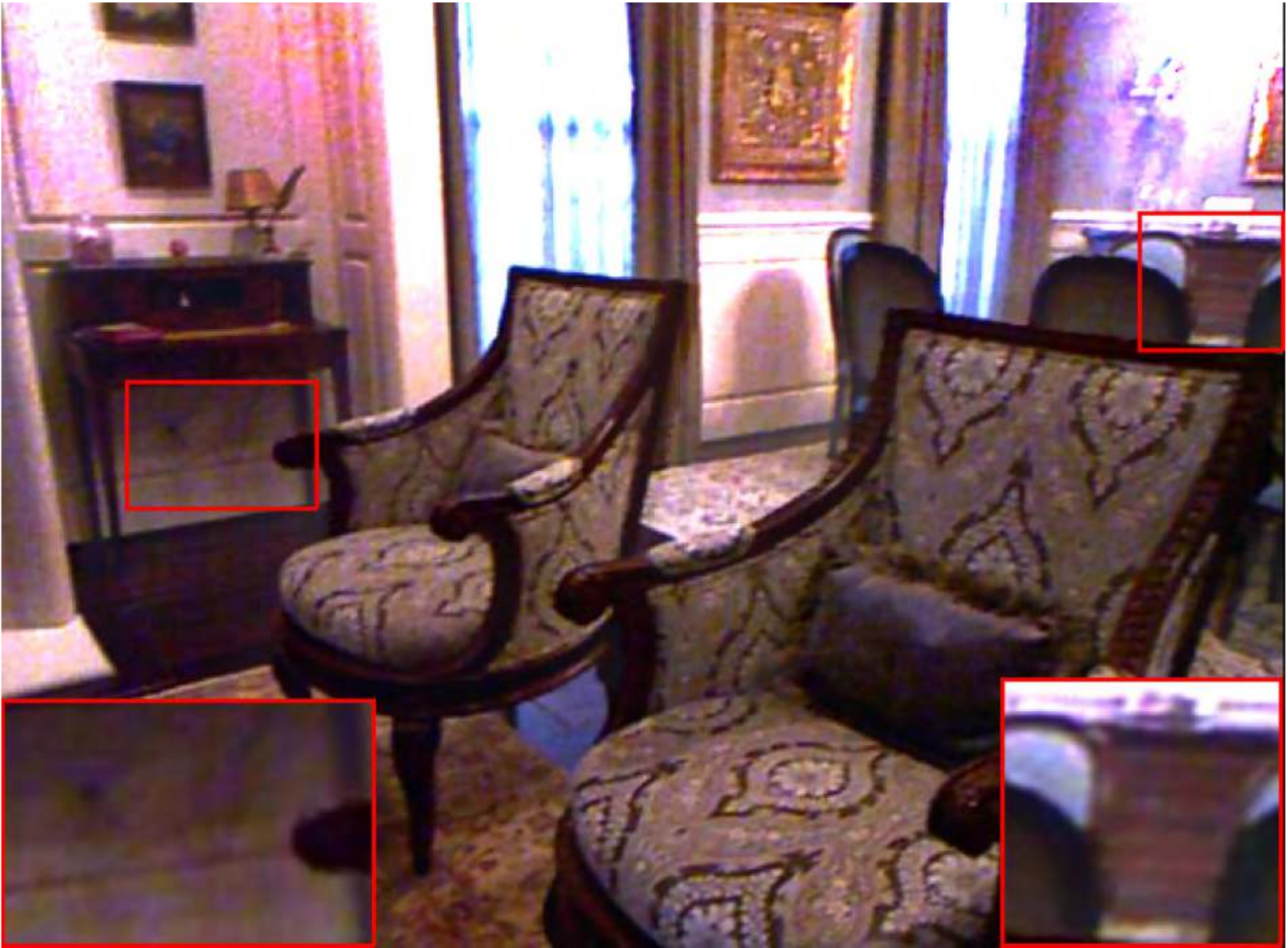}\vspace{2pt}
			\includegraphics[width=1\linewidth]{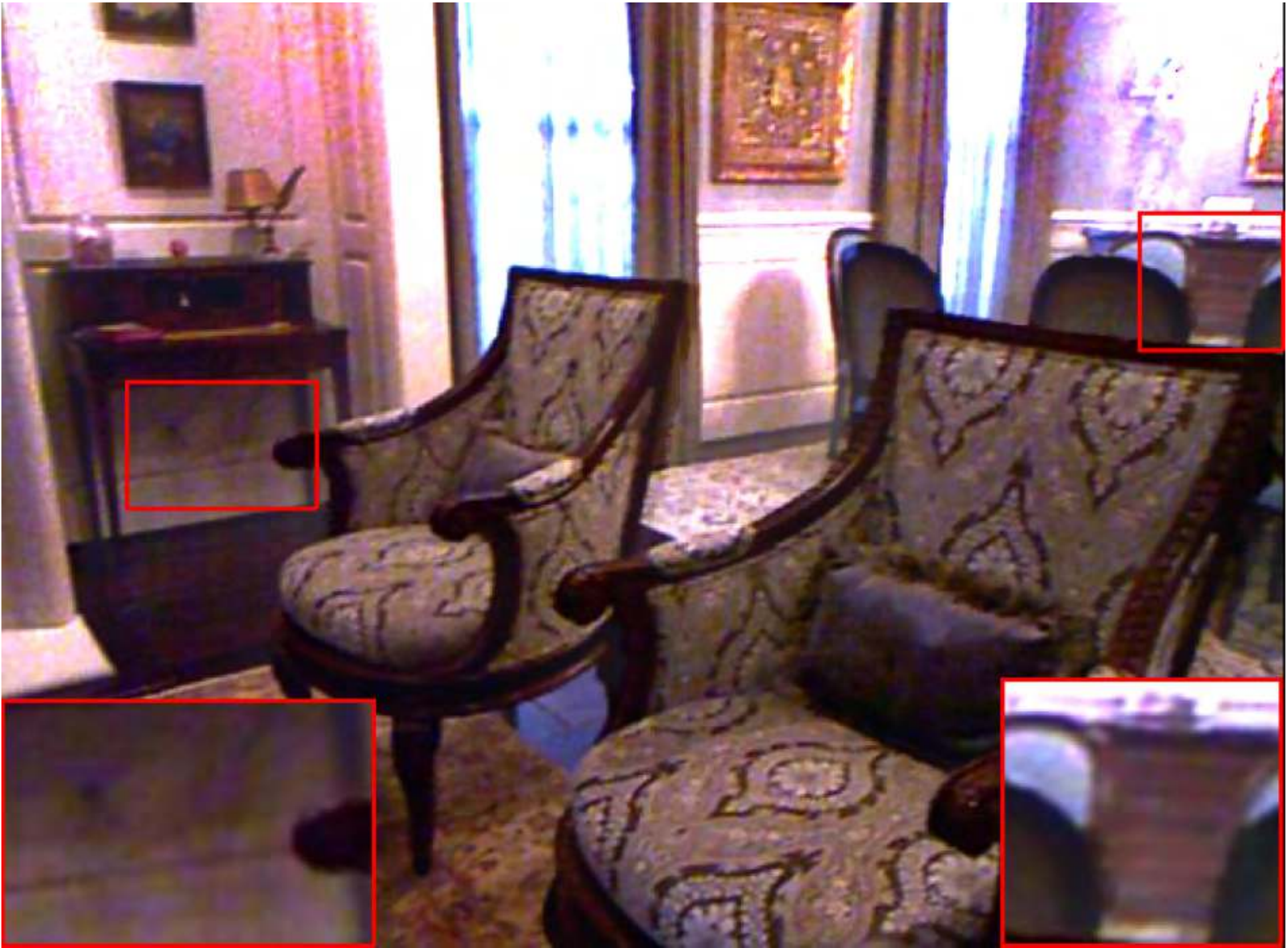}\vspace{2pt}
			\includegraphics[width=1\linewidth]{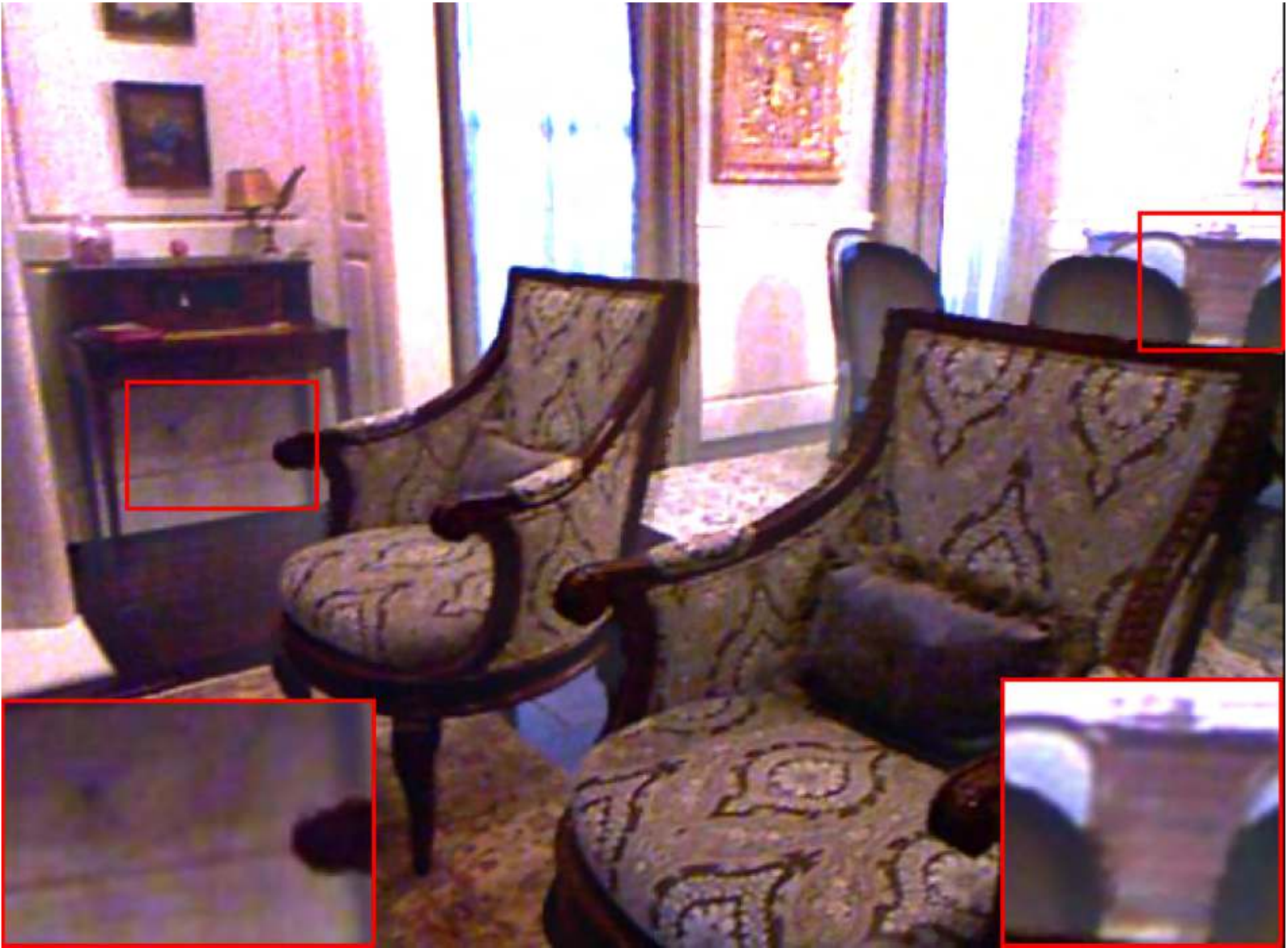}\vspace{2pt}
	\end{minipage}}\hspace{-0.45em}
	\subfigure[\scriptsize{DehazeNet~\cite{cai2016dehazenet}}]{
		\begin{minipage}[b]{0.12\linewidth}
			\includegraphics[width=1\linewidth]{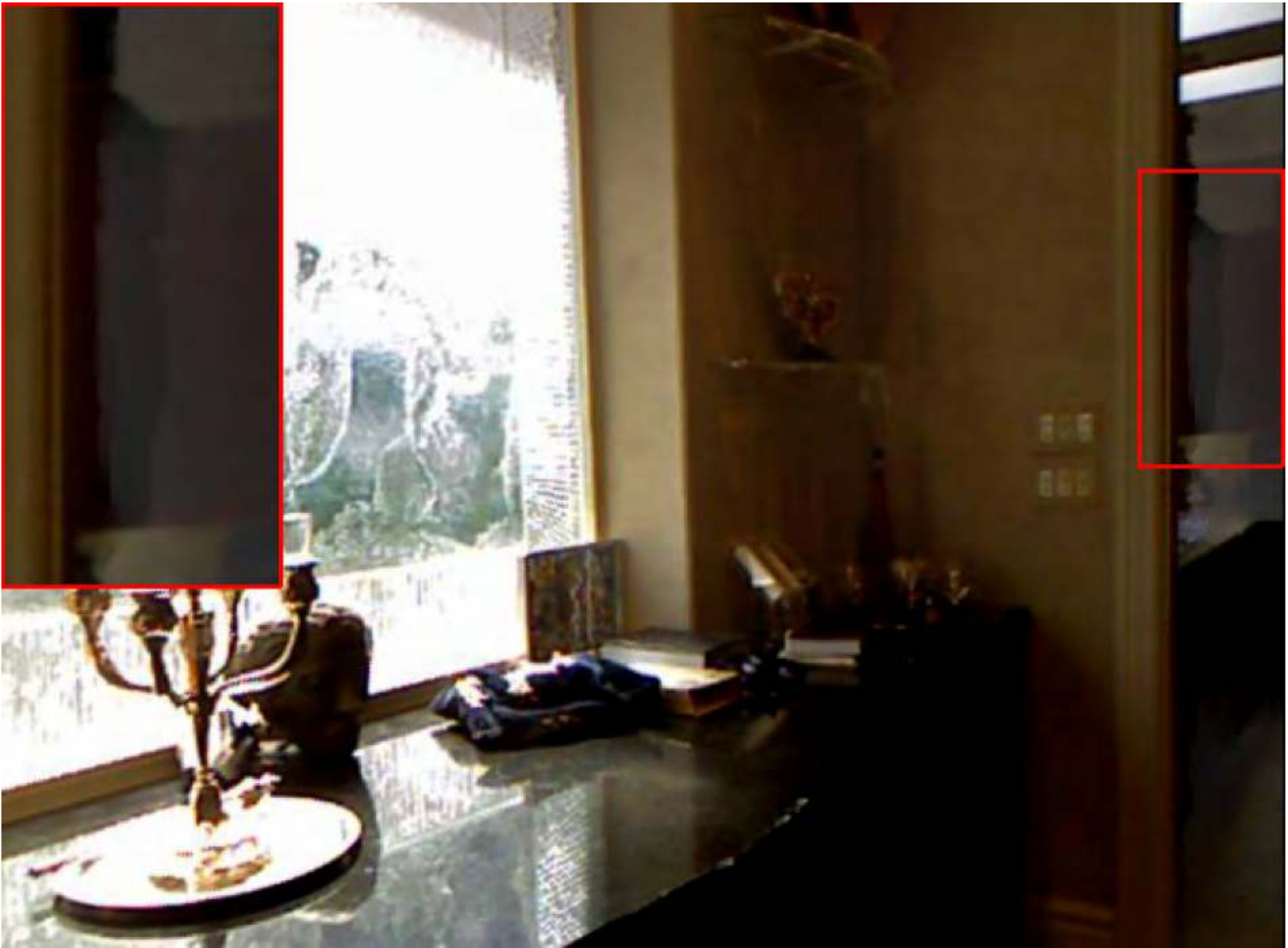}\vspace{2pt}
			\includegraphics[width=1\linewidth]{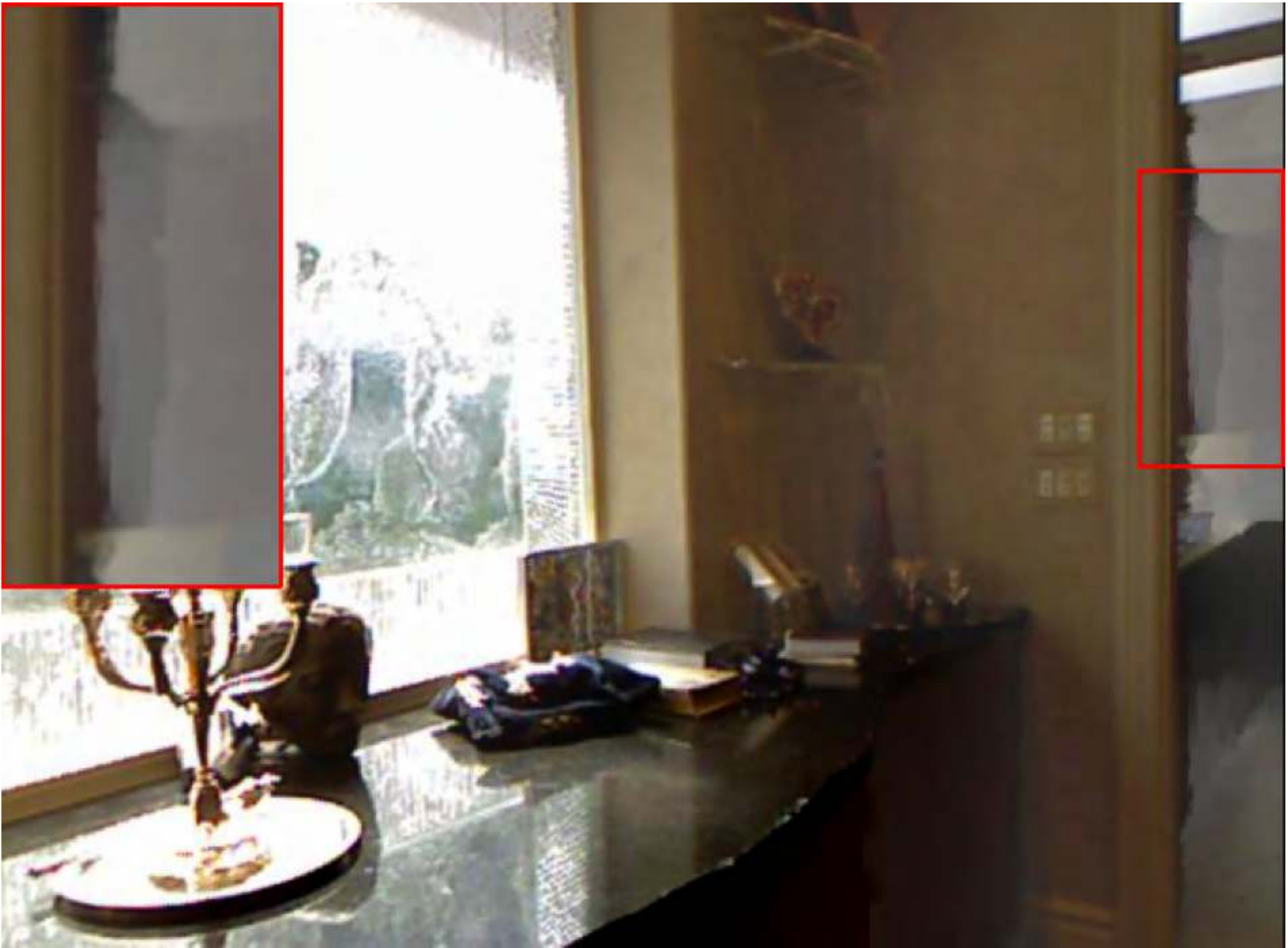}\vspace{2pt}
			\includegraphics[width=1\linewidth]{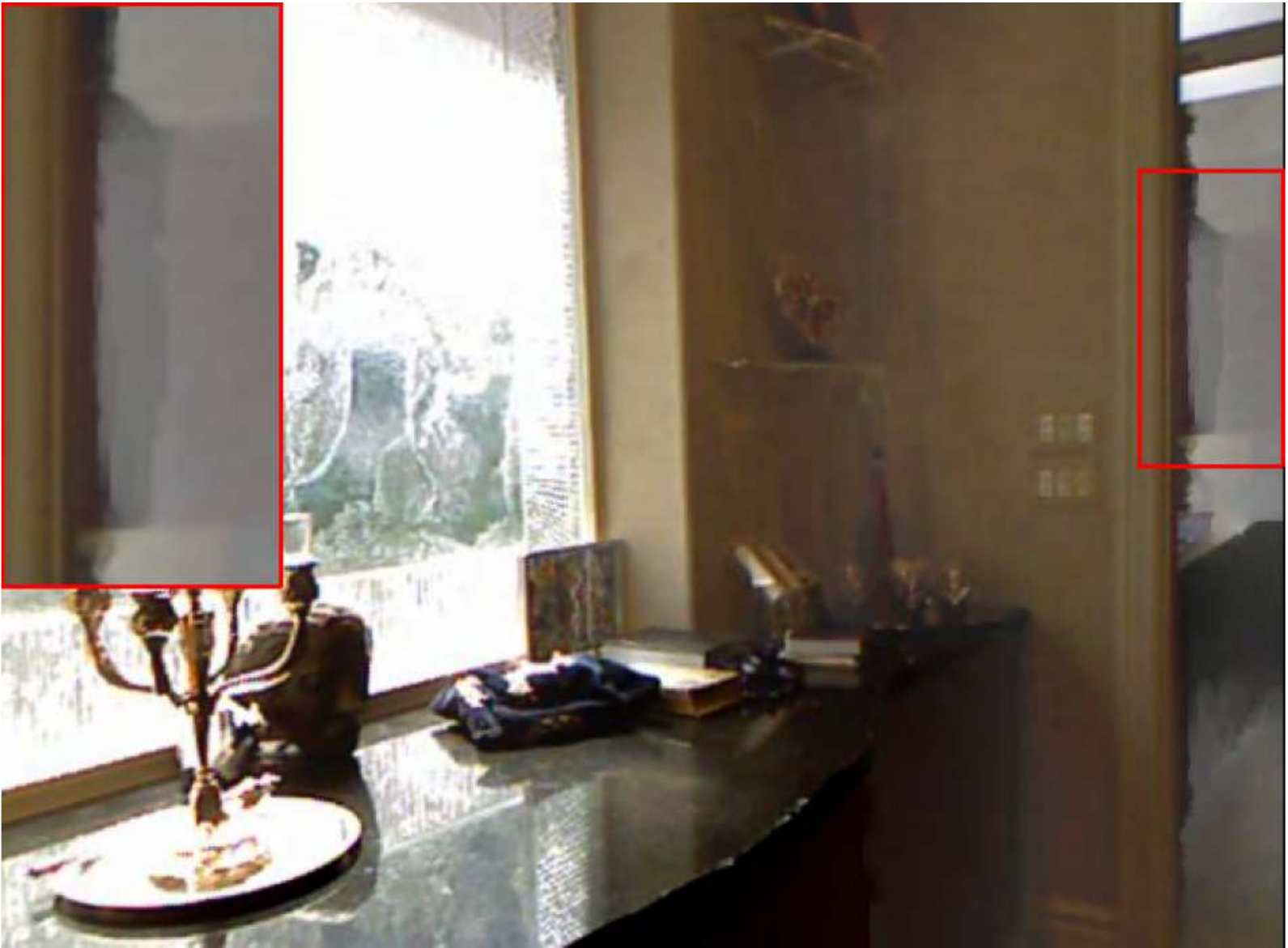}\vspace{2pt}
			\includegraphics[width=1\linewidth]{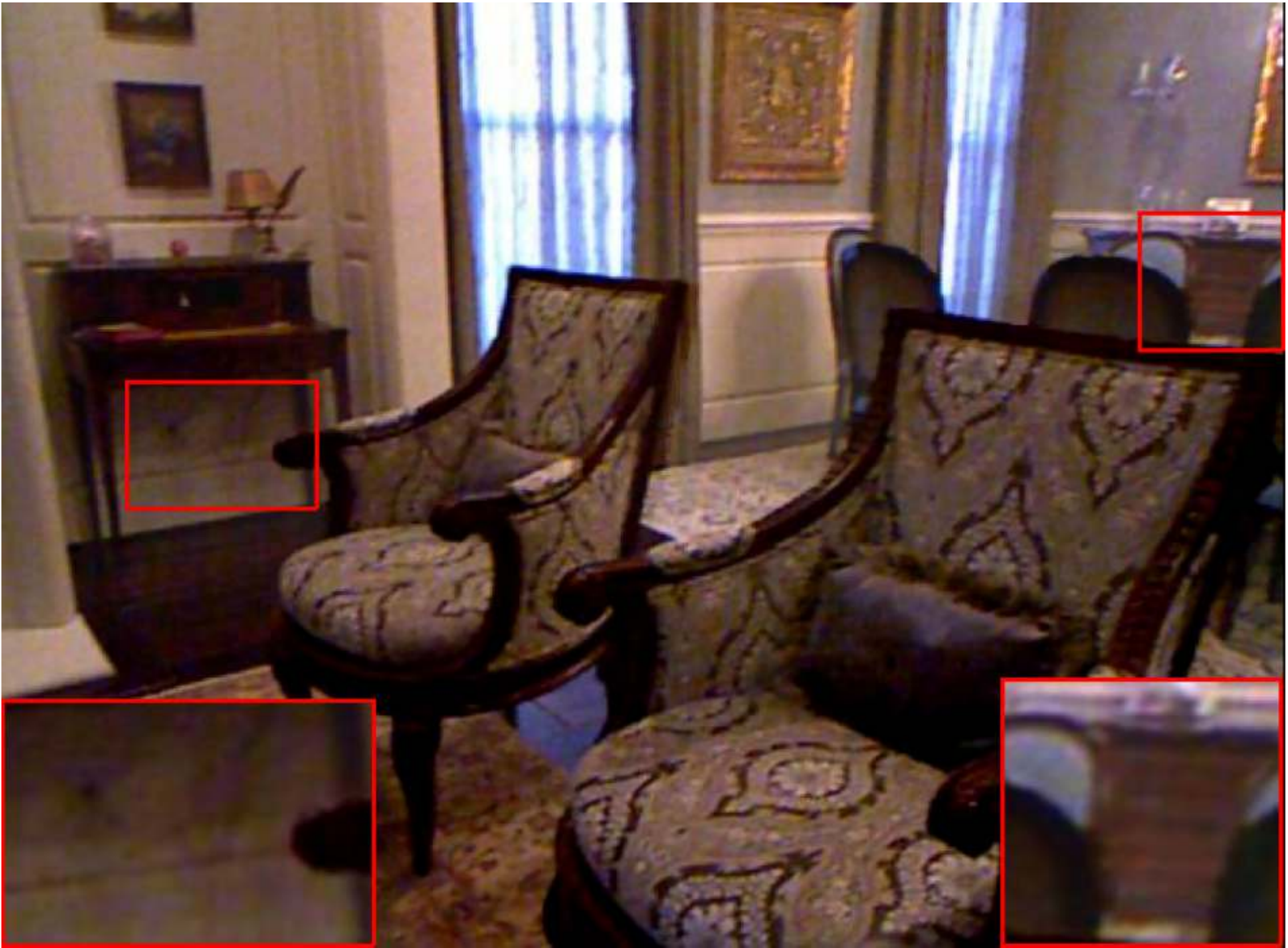}\vspace{2pt}
			\includegraphics[width=1\linewidth]{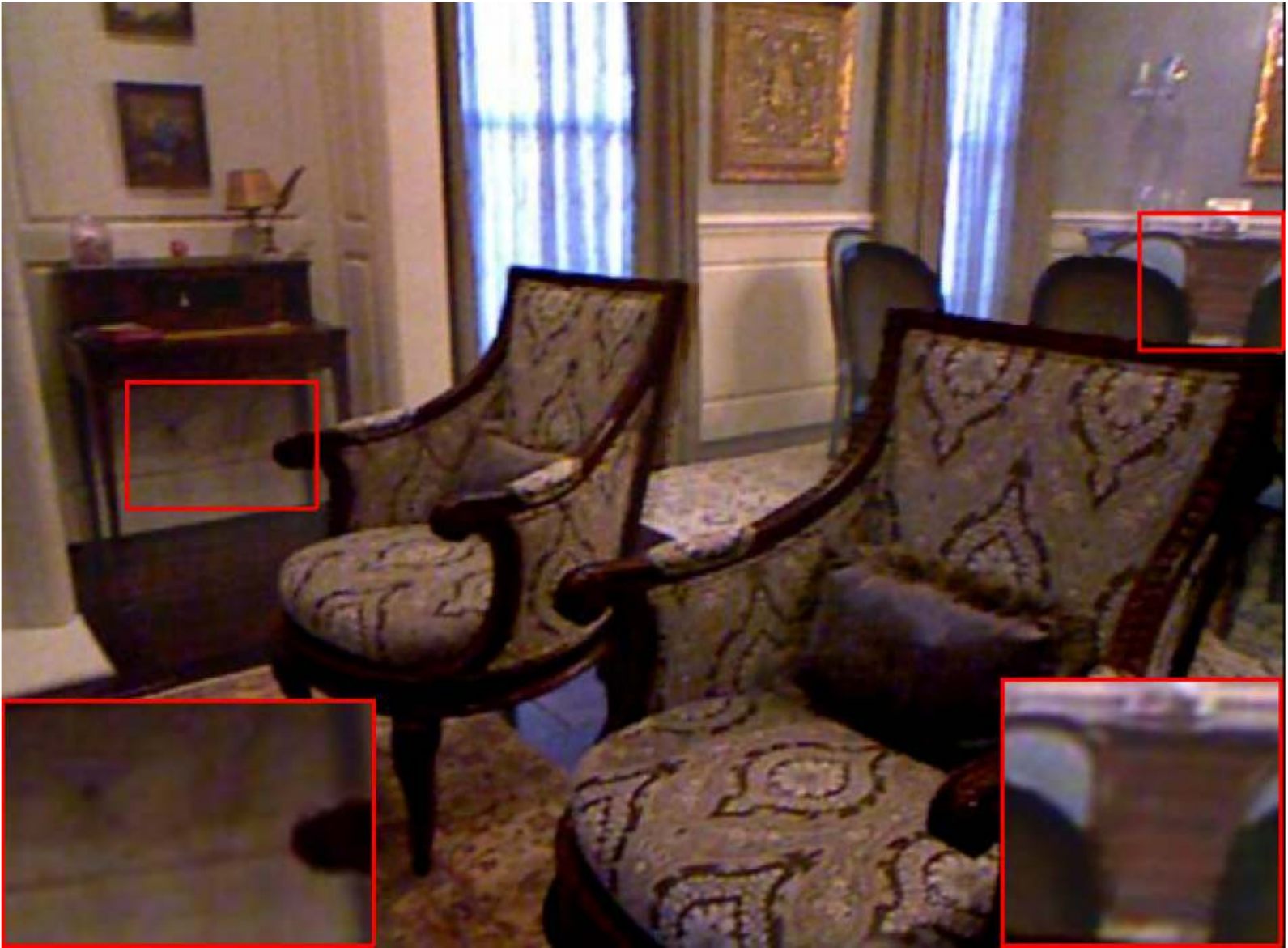}\vspace{2pt}
			\includegraphics[width=1\linewidth]{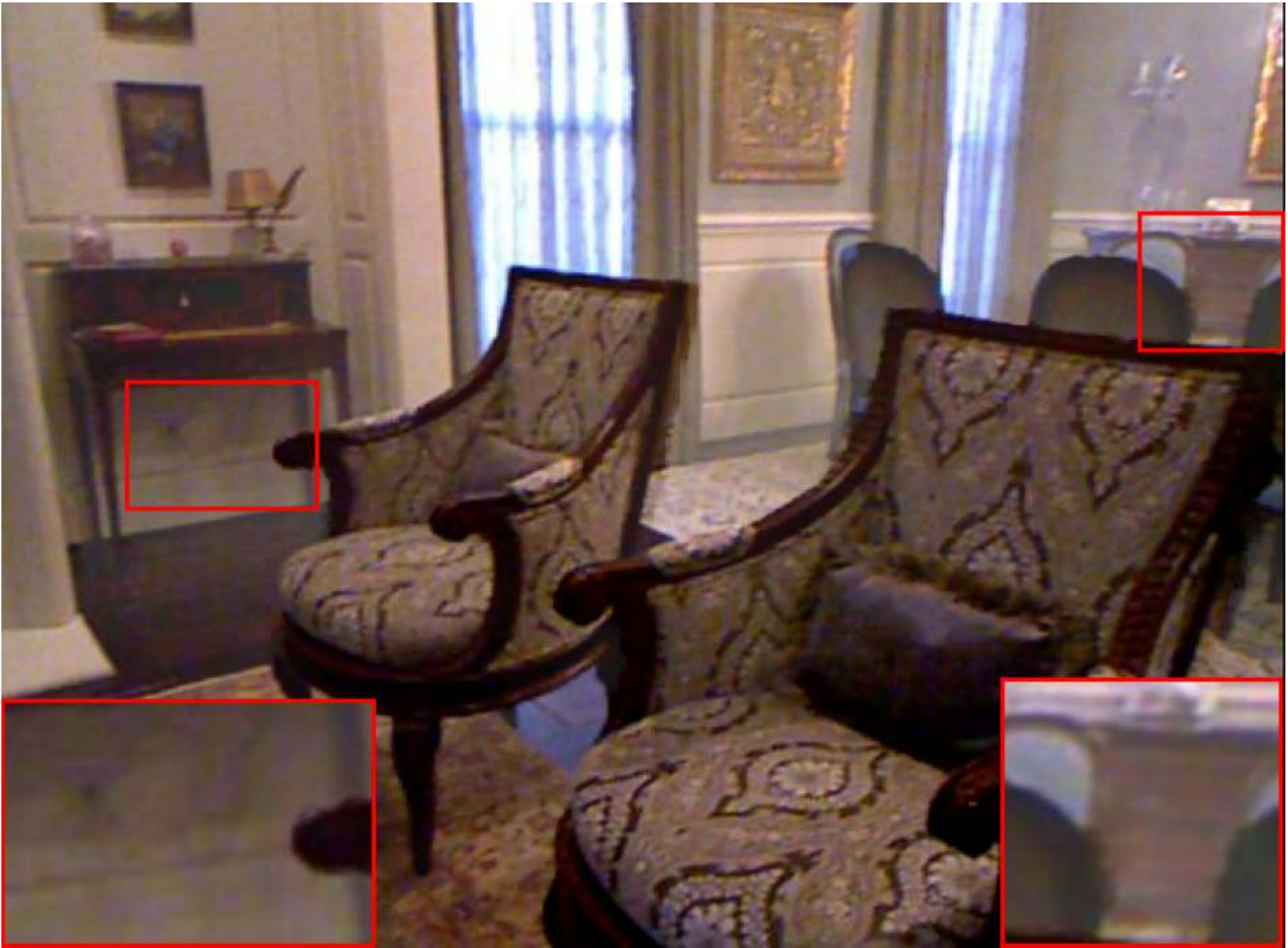}\vspace{2pt}
	\end{minipage}}\hspace{-0.45em}
	\subfigure[\scriptsize{EPDN~\cite{qu2019enhanced}}]{
		\begin{minipage}[b]{0.12\linewidth}
			\includegraphics[width=1\linewidth]{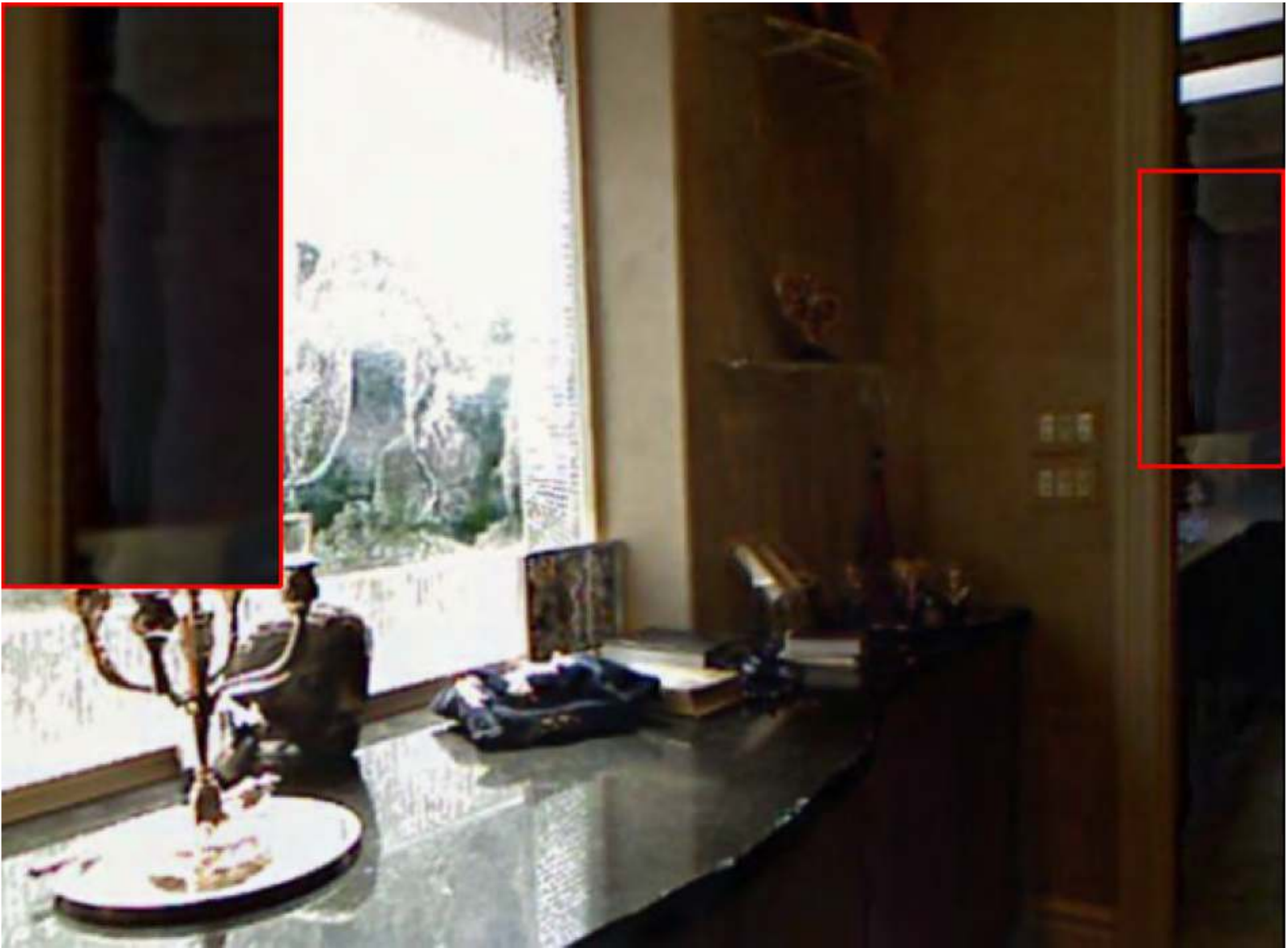}\vspace{2pt}
			\includegraphics[width=1\linewidth]{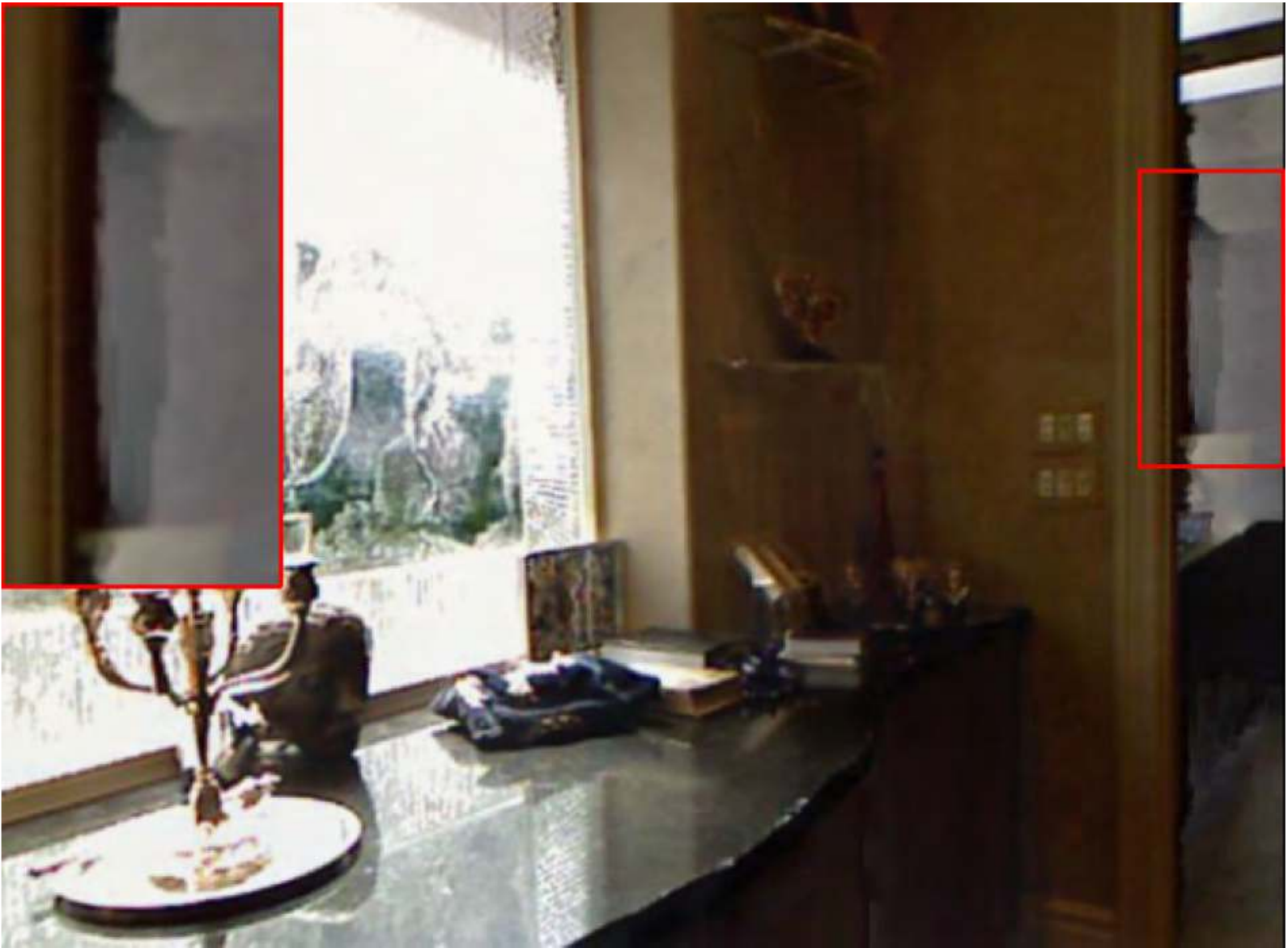}\vspace{2pt}
			\includegraphics[width=1\linewidth]{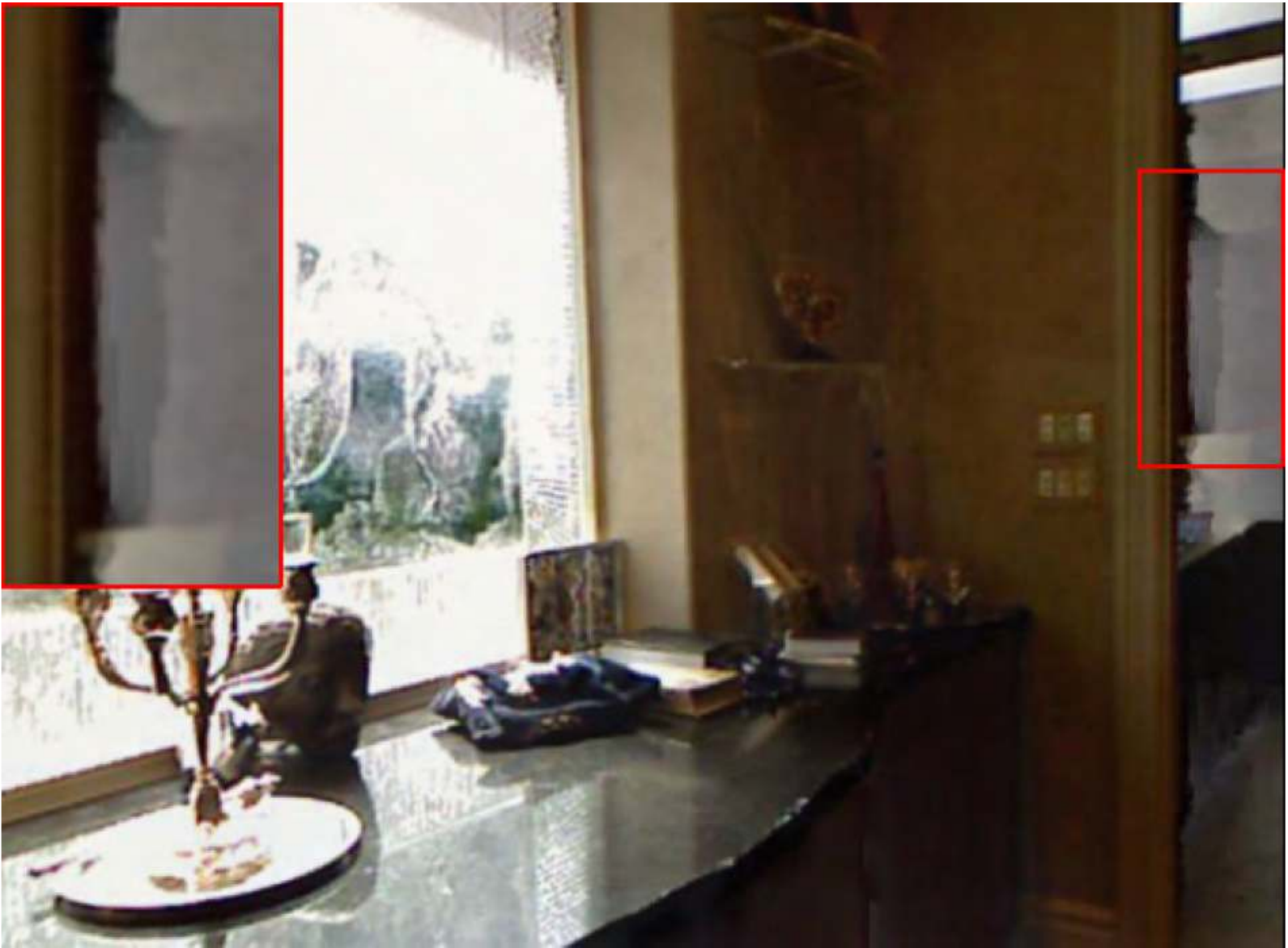}\vspace{2pt}
			\includegraphics[width=1\linewidth]{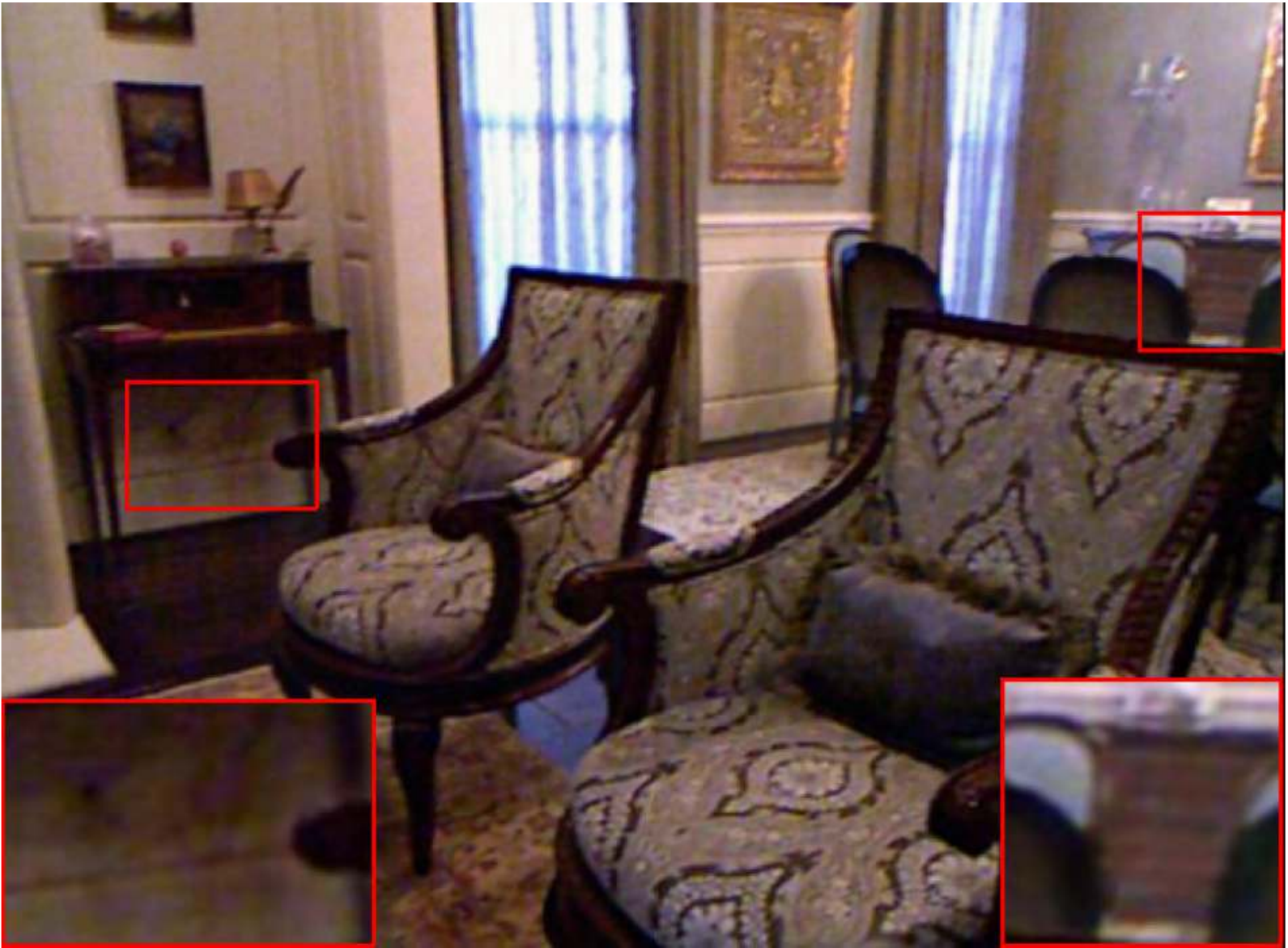}\vspace{2pt}
			\includegraphics[width=1\linewidth]{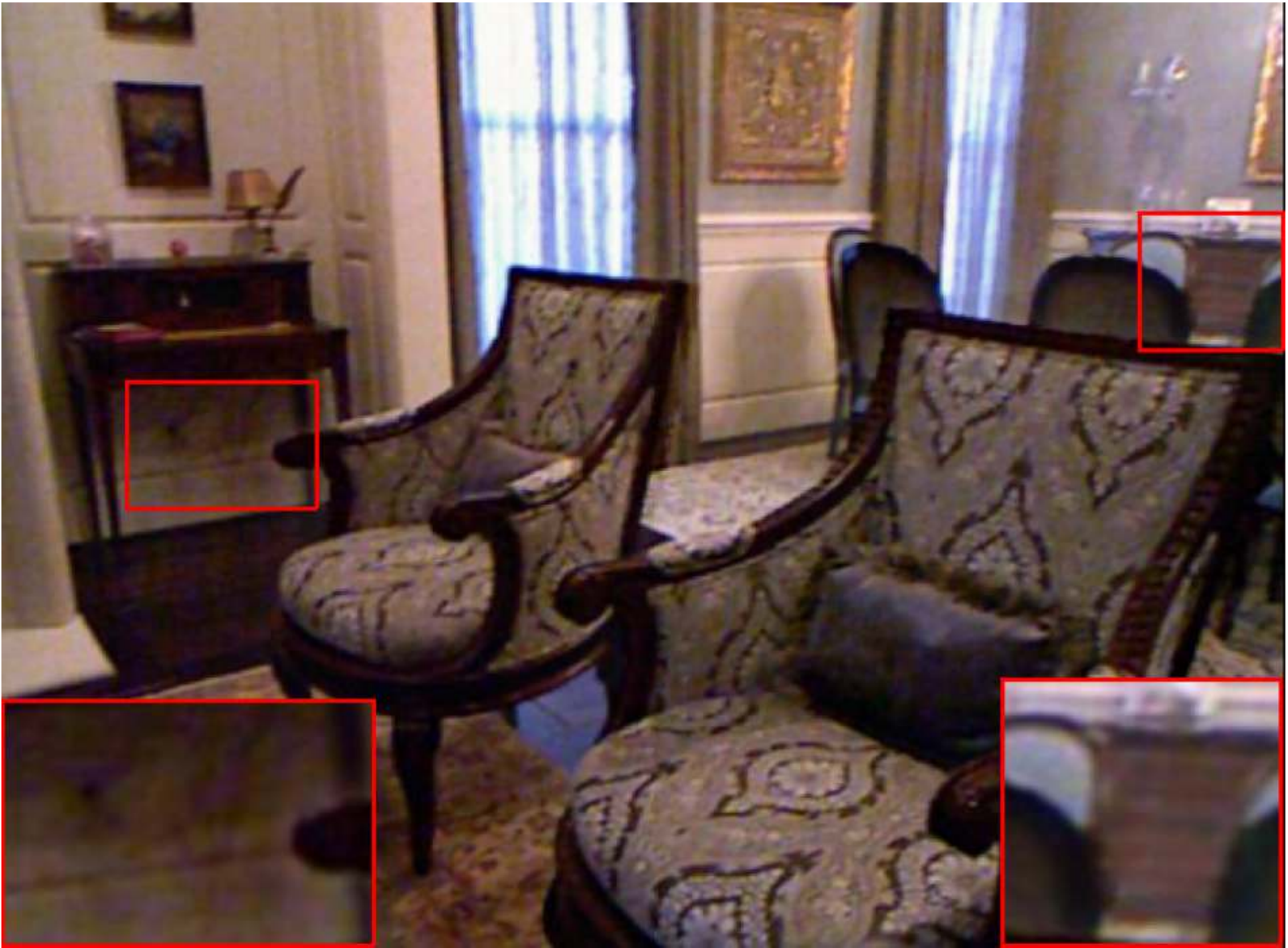}\vspace{2pt}
			\includegraphics[width=1\linewidth]{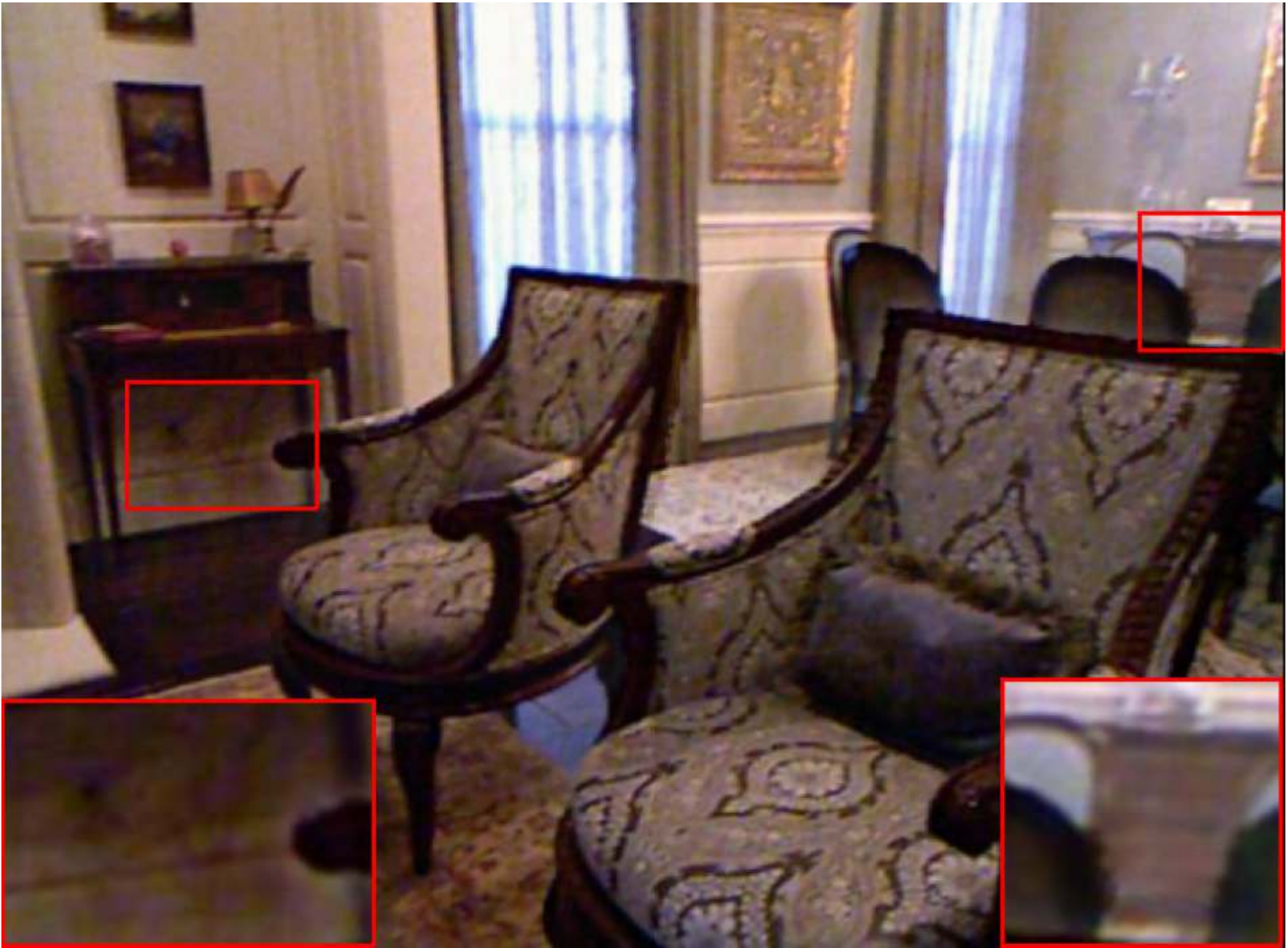}\vspace{2pt}
	\end{minipage}}\hspace{-0.45em}
	\subfigure[\scriptsize{DAdehazing~\cite{shao2020domain}}]{
		\begin{minipage}[b]{0.12\linewidth}
			\includegraphics[width=1\linewidth]{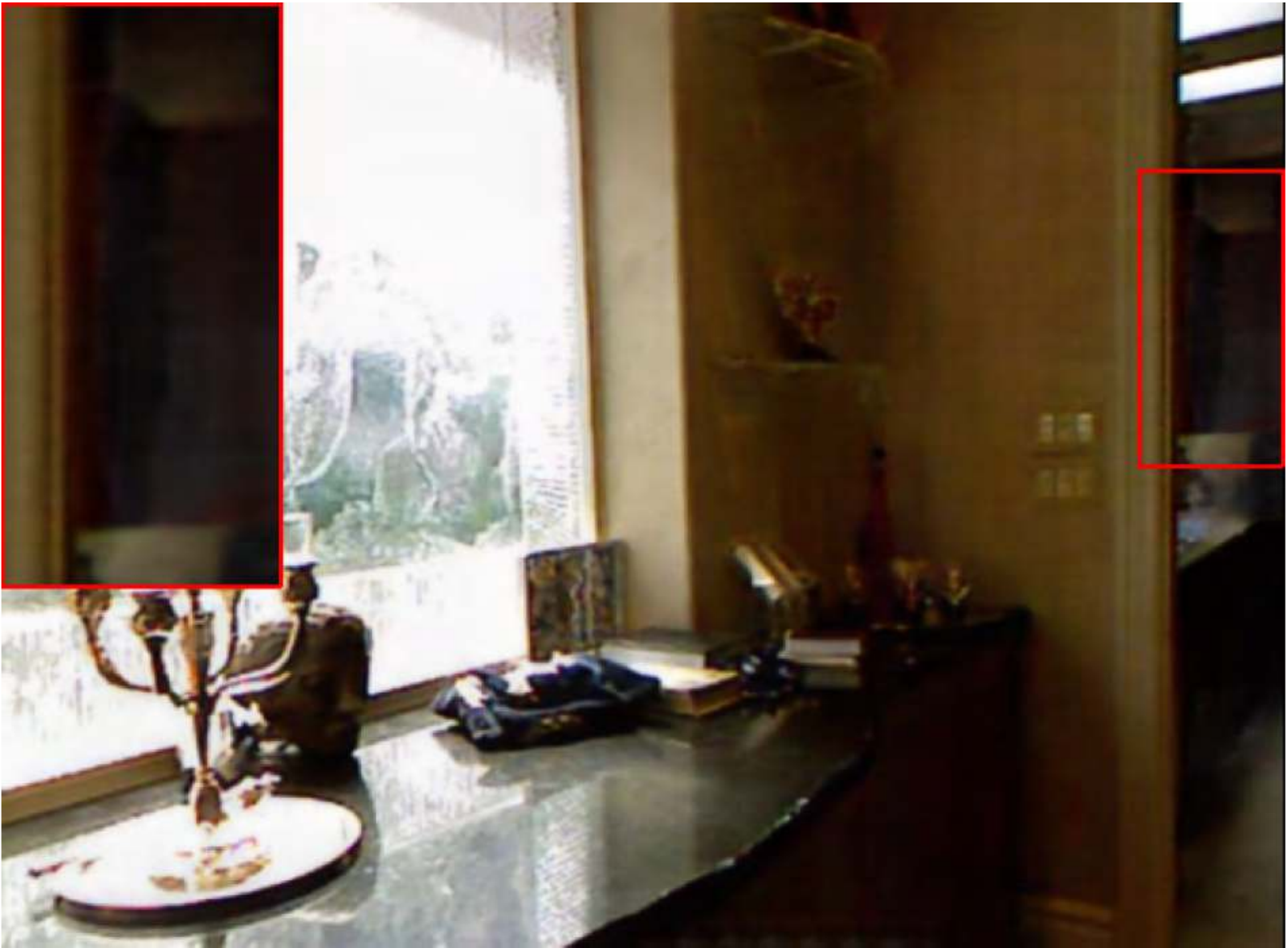}\vspace{2pt}
			\includegraphics[width=1\linewidth]{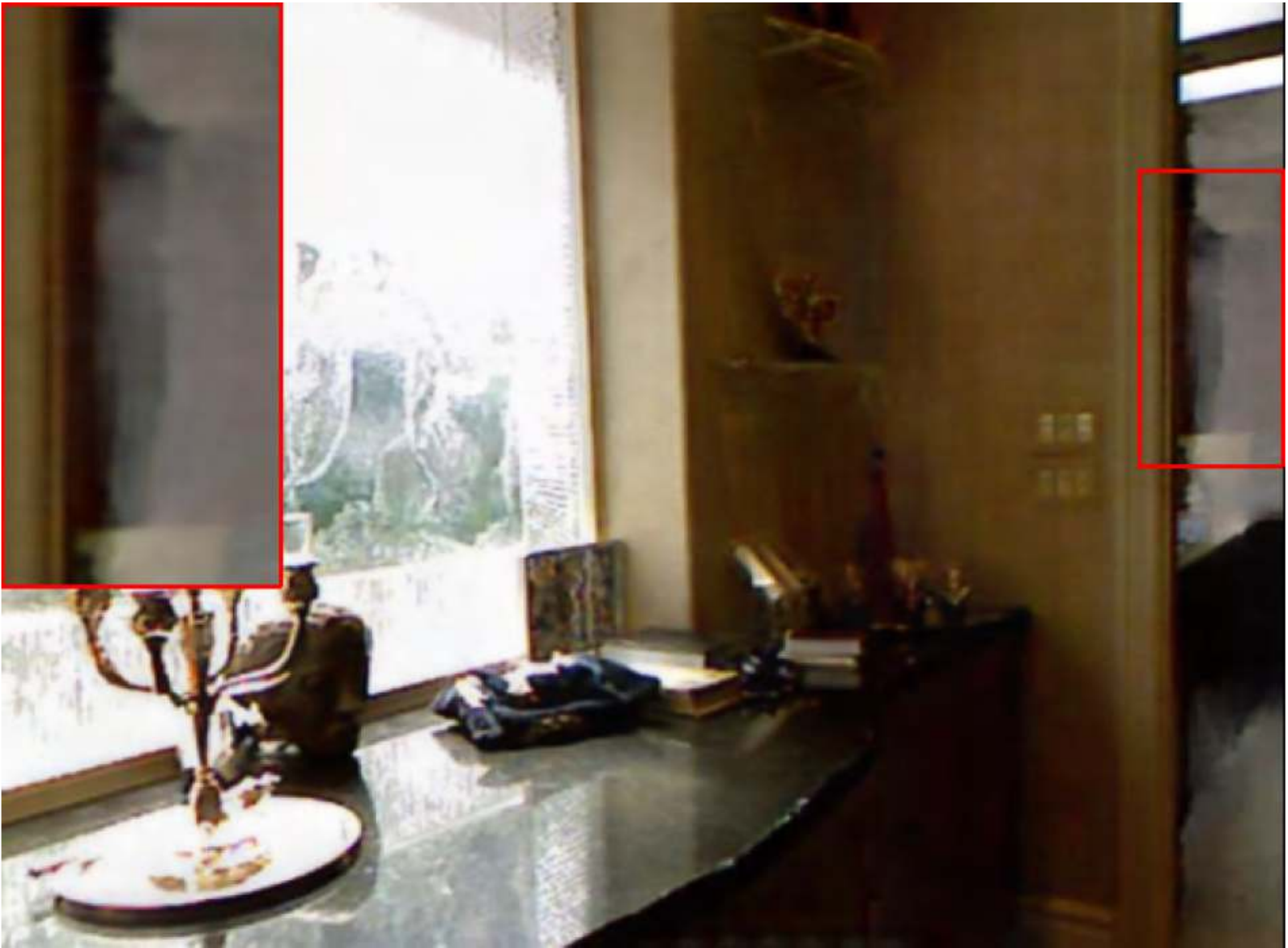}\vspace{2pt}
			\includegraphics[width=1\linewidth]{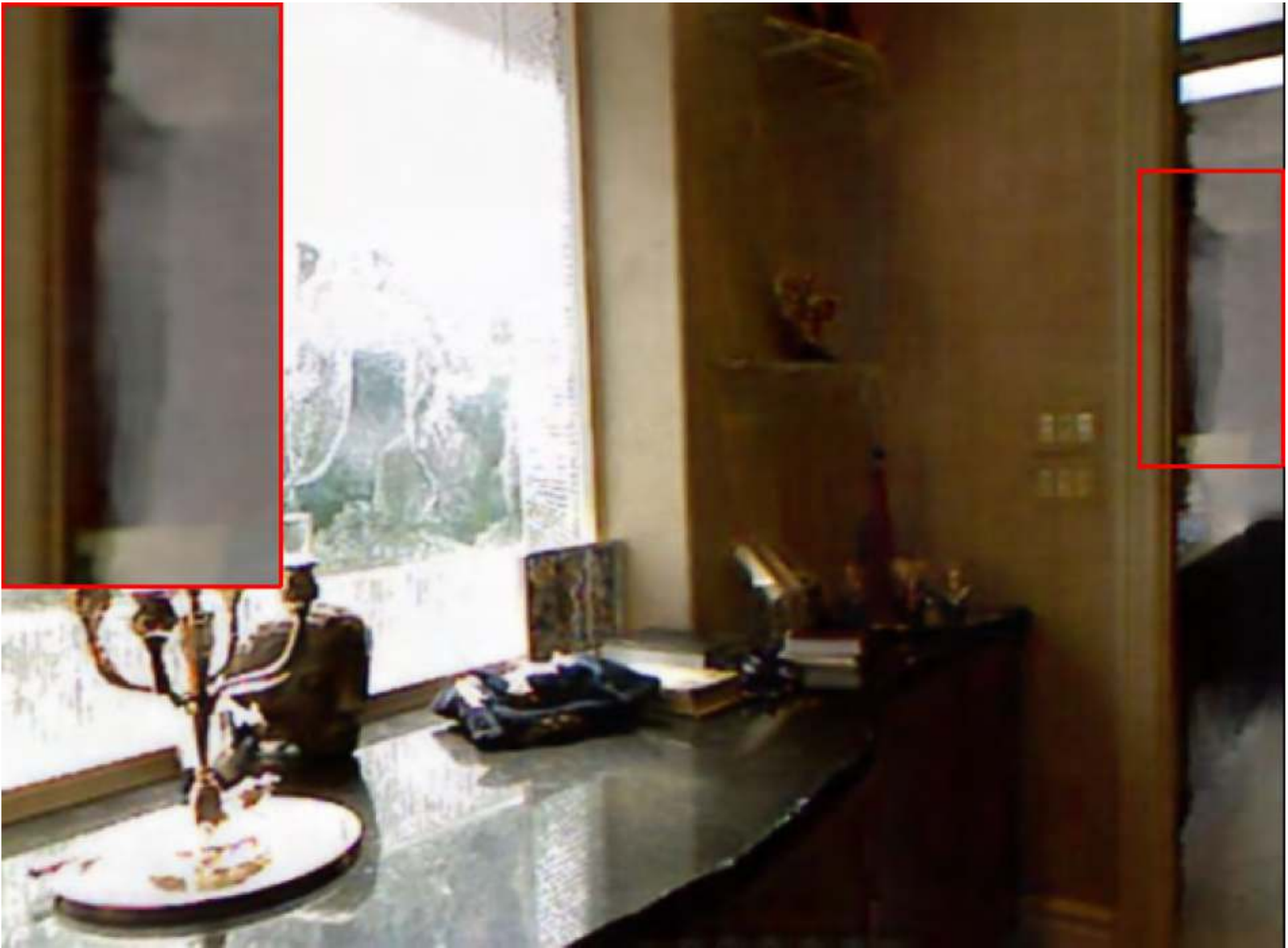}\vspace{2pt}
			\includegraphics[width=1\linewidth]{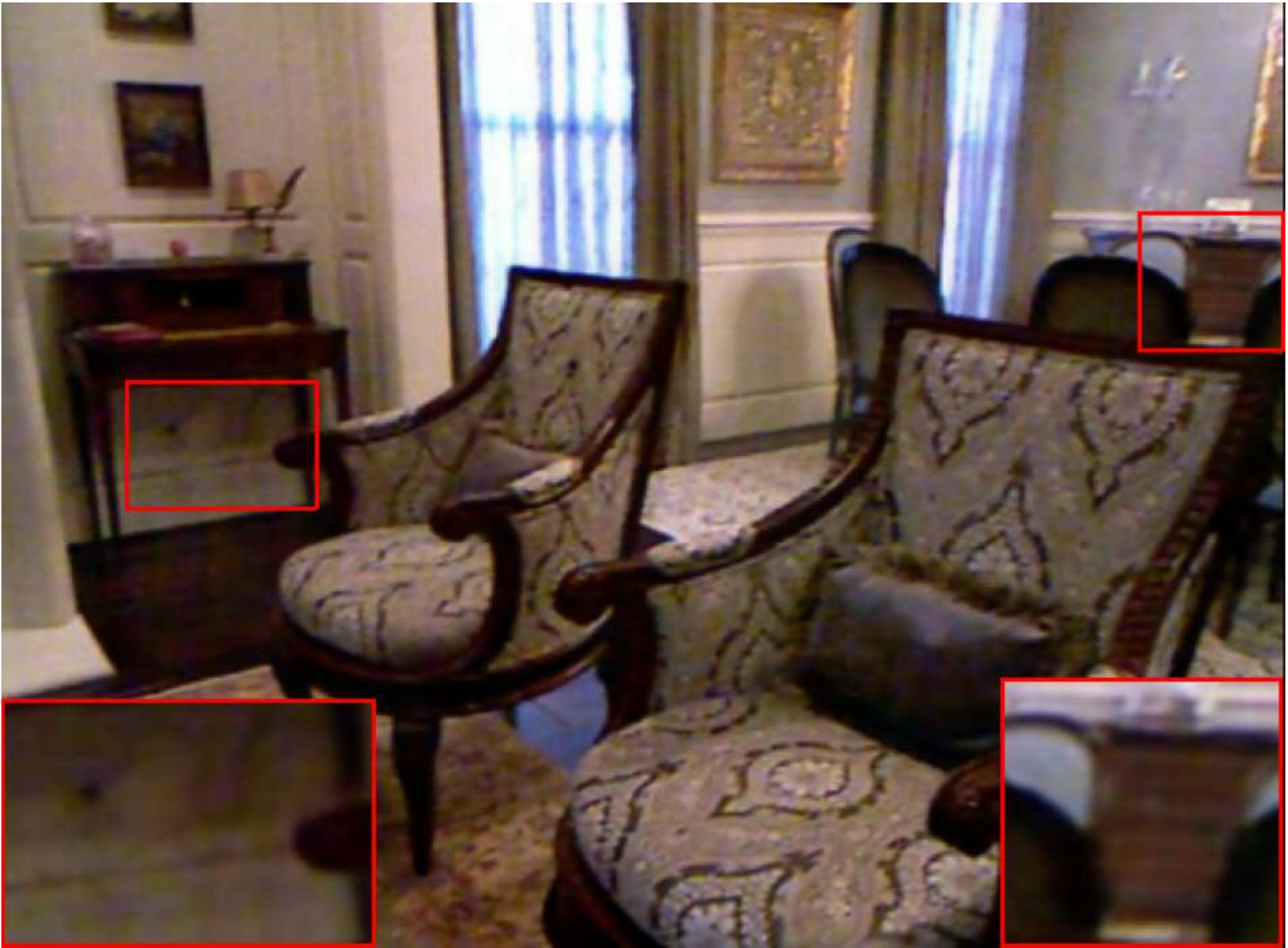}\vspace{2pt}
			\includegraphics[width=1\linewidth]{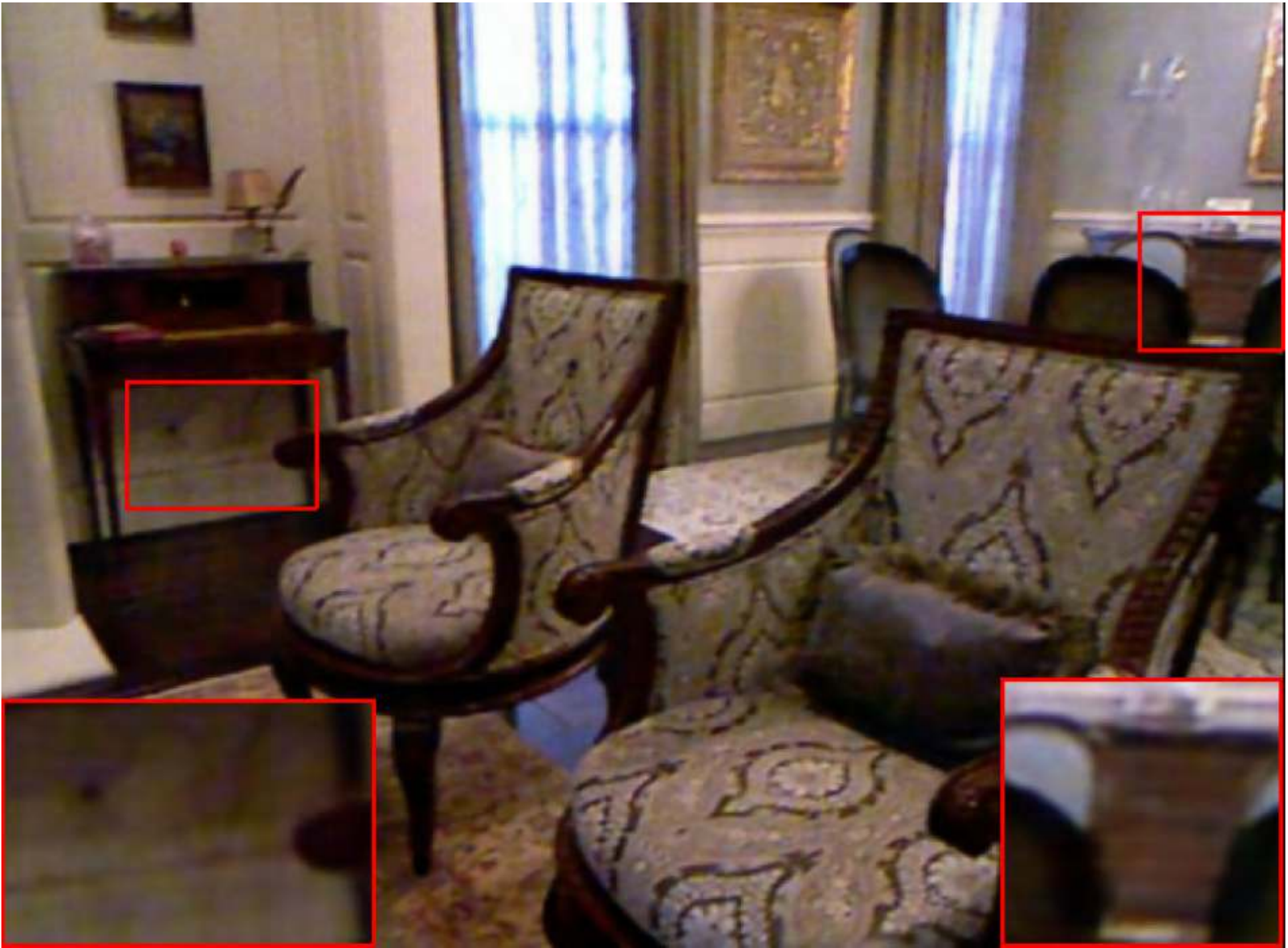}\vspace{2pt}
			\includegraphics[width=1\linewidth]{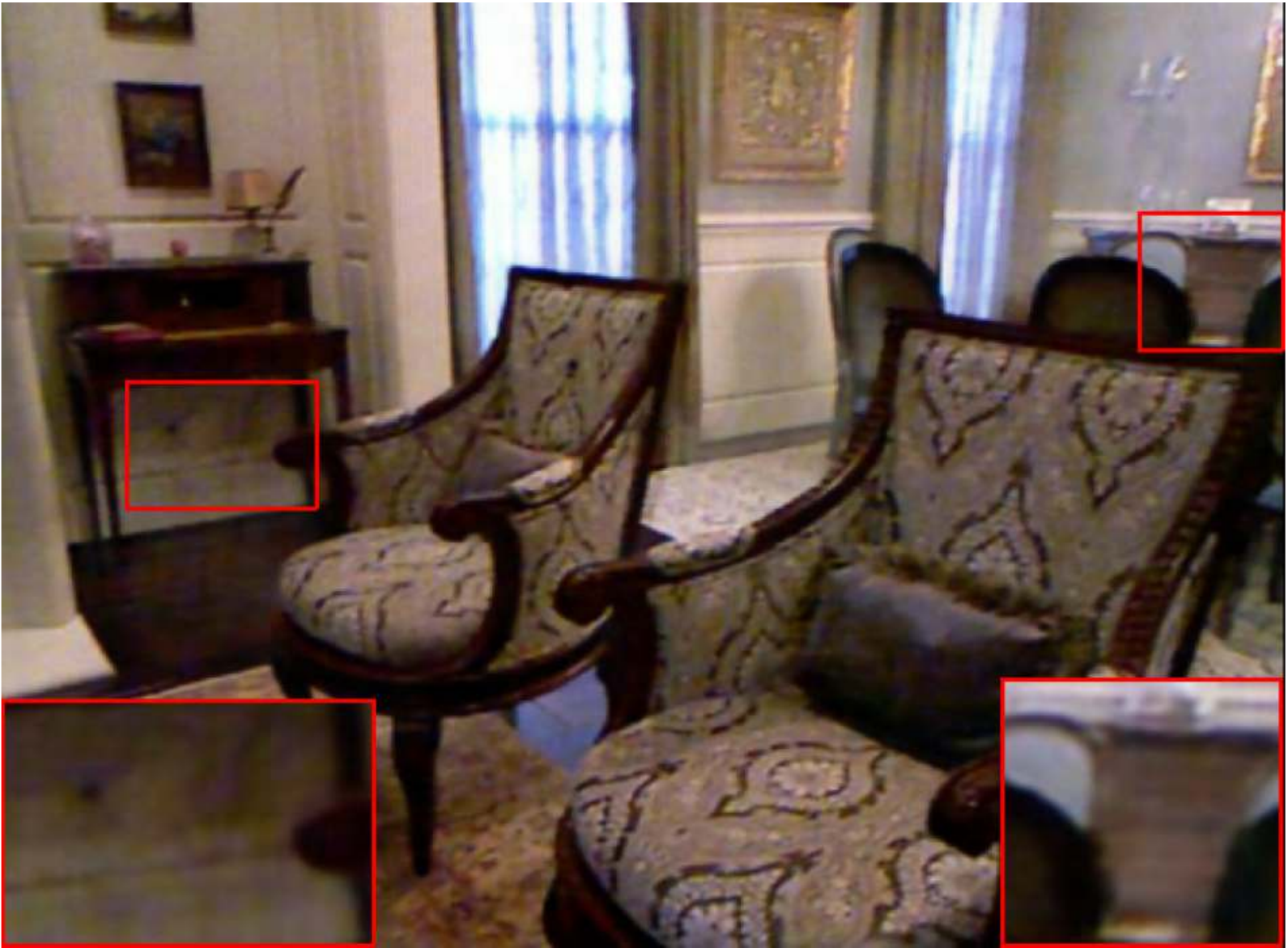}\vspace{2pt}
	\end{minipage}}\hspace{-0.45em}
	\subfigure[\scriptsize{MSBDN-DFF~\cite{dong2020multi}}]{
		\begin{minipage}[b]{0.12\linewidth}
			\includegraphics[width=1\linewidth]{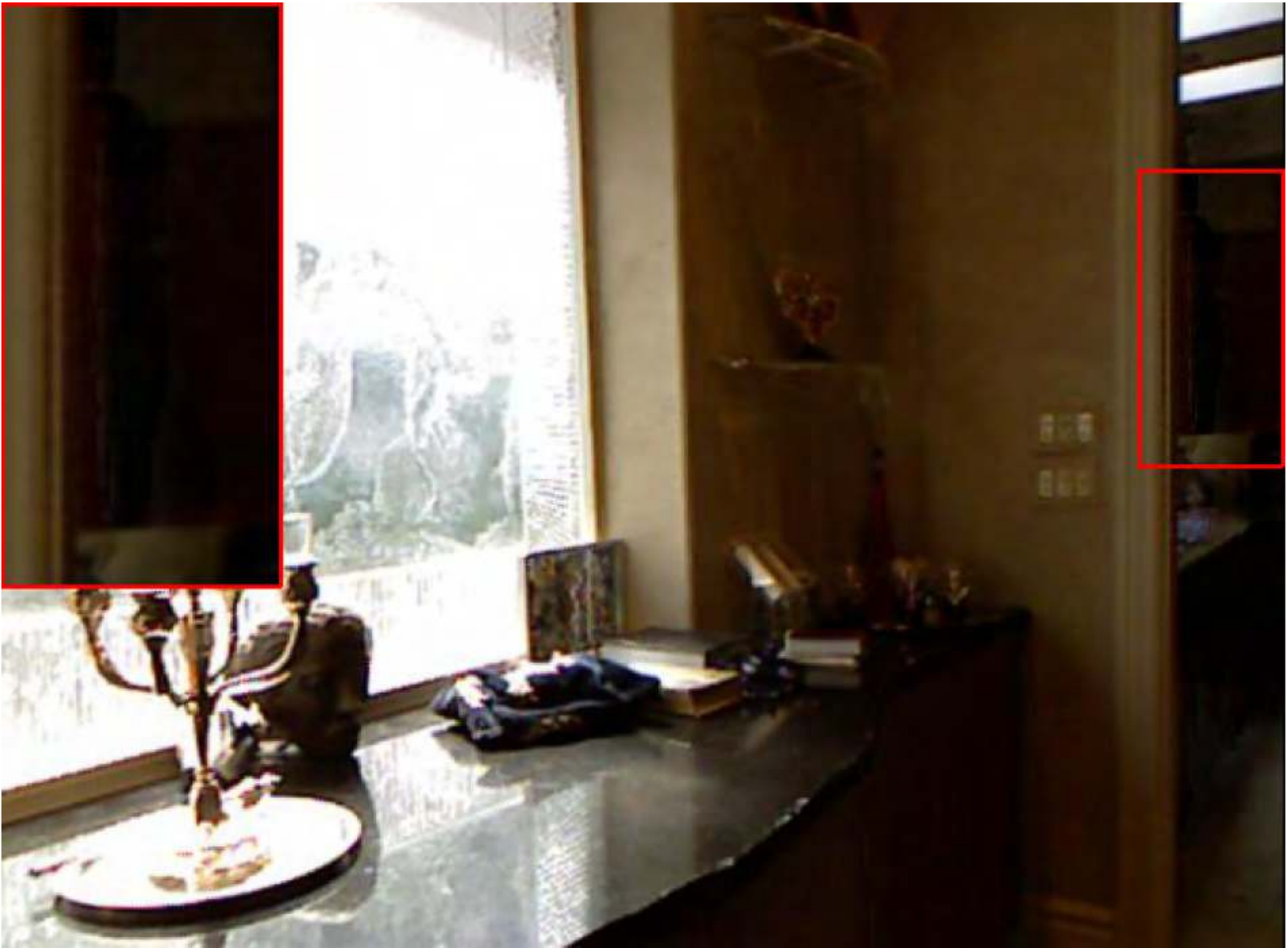}\vspace{2pt}
			\includegraphics[width=1\linewidth]{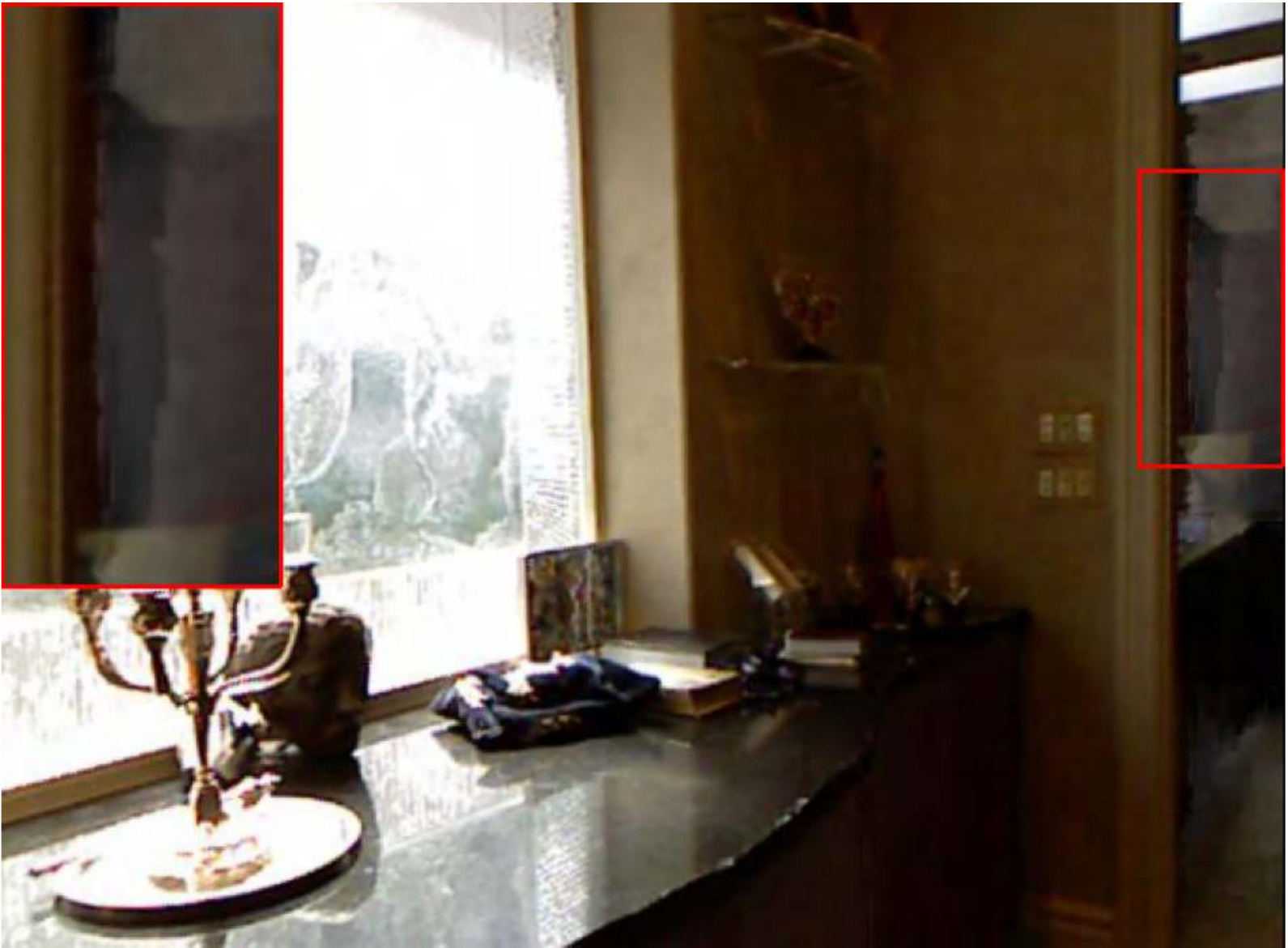}\vspace{2pt}
			\includegraphics[width=1\linewidth]{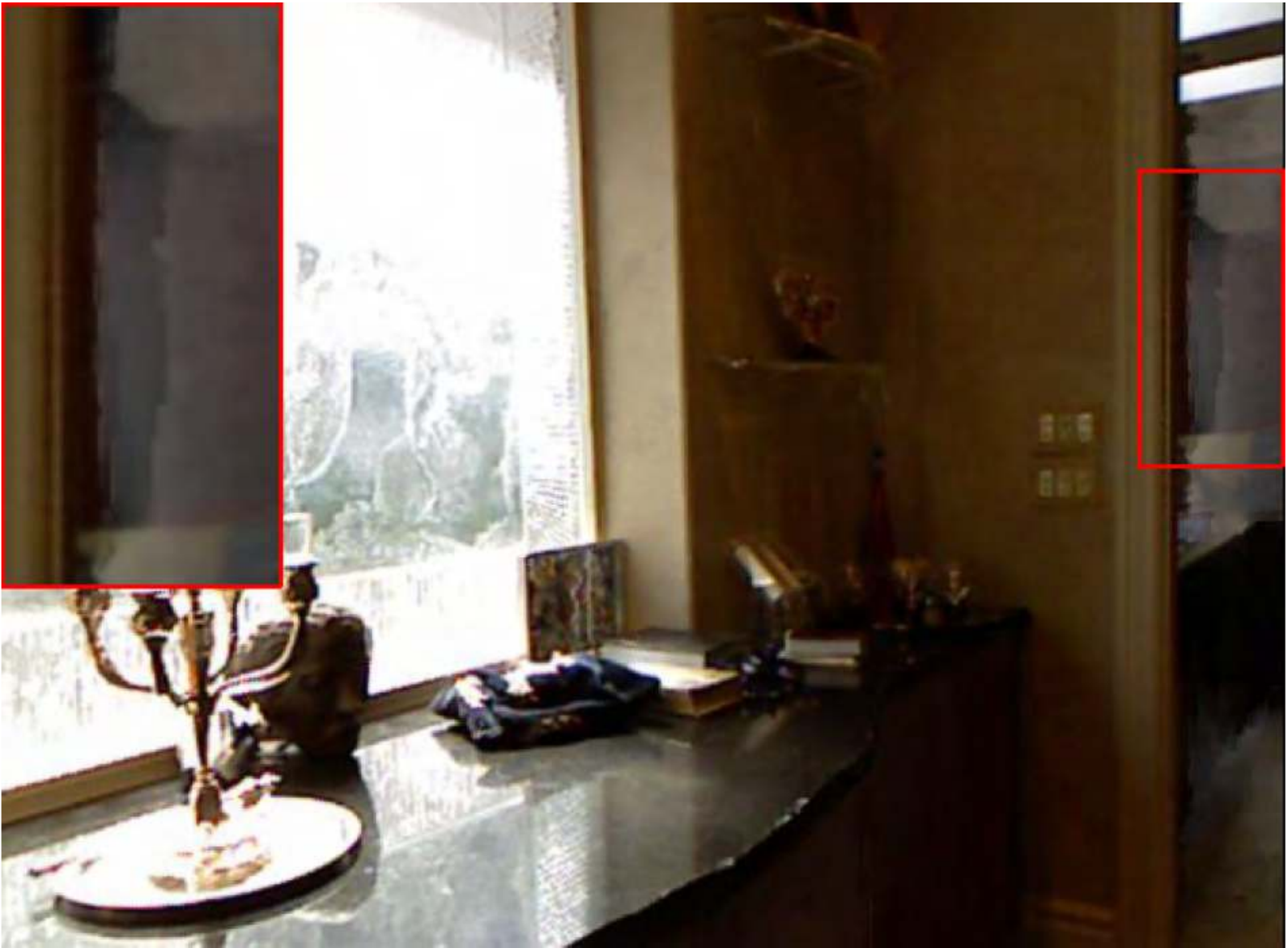}\vspace{2pt}
			\includegraphics[width=1\linewidth]{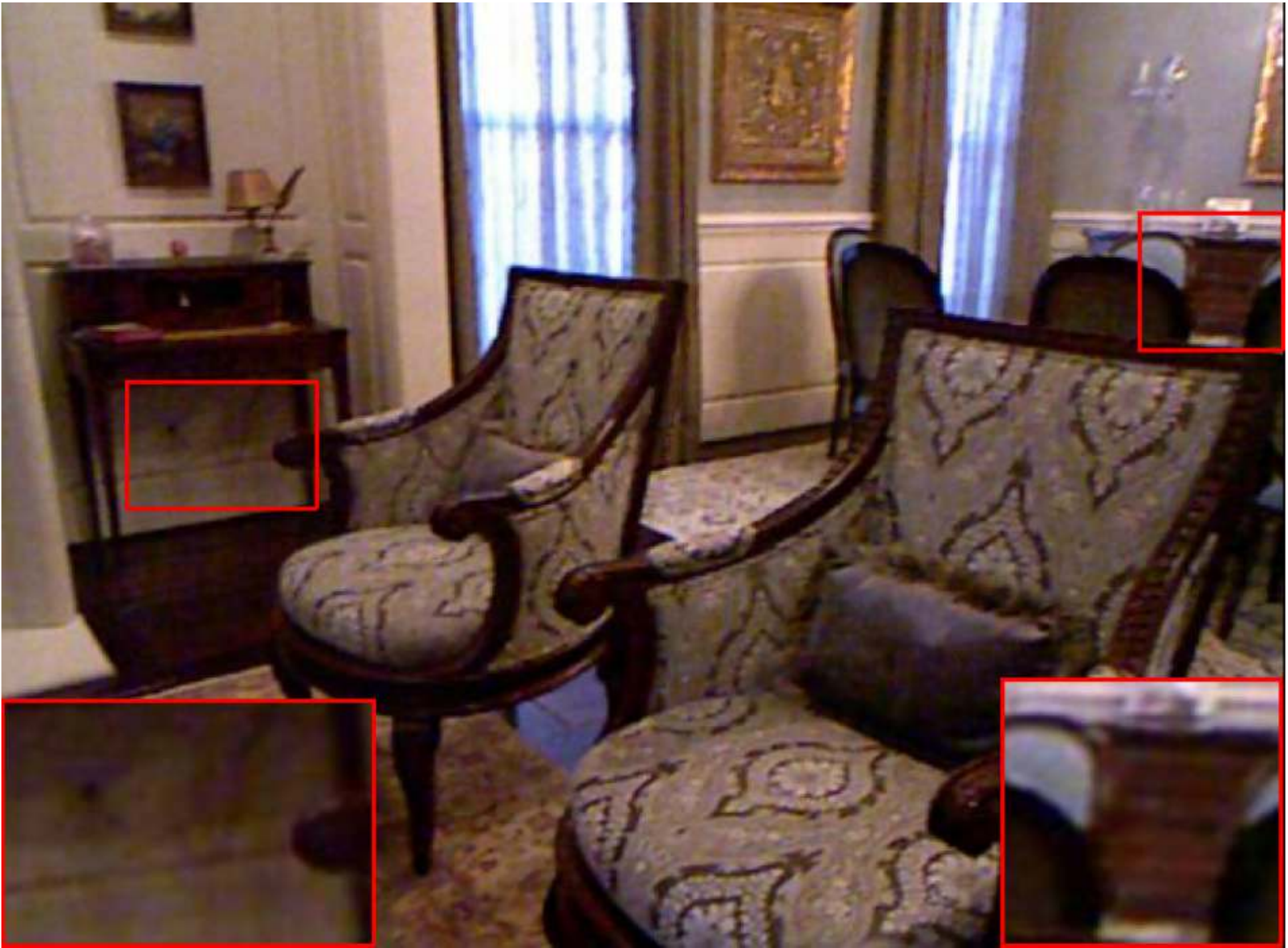}\vspace{2pt}
			\includegraphics[width=1\linewidth]{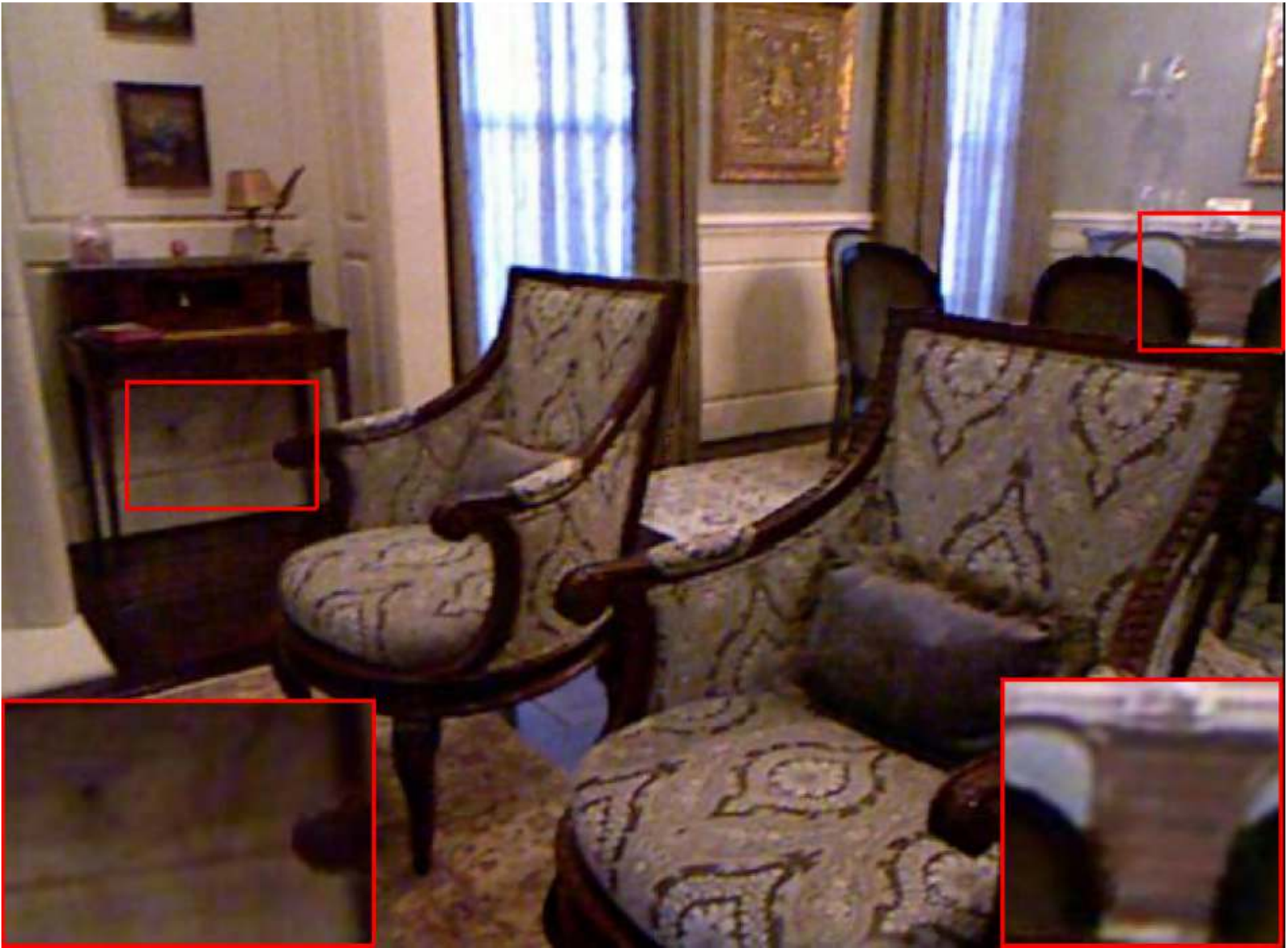}\vspace{2pt}
			\includegraphics[width=1\linewidth]{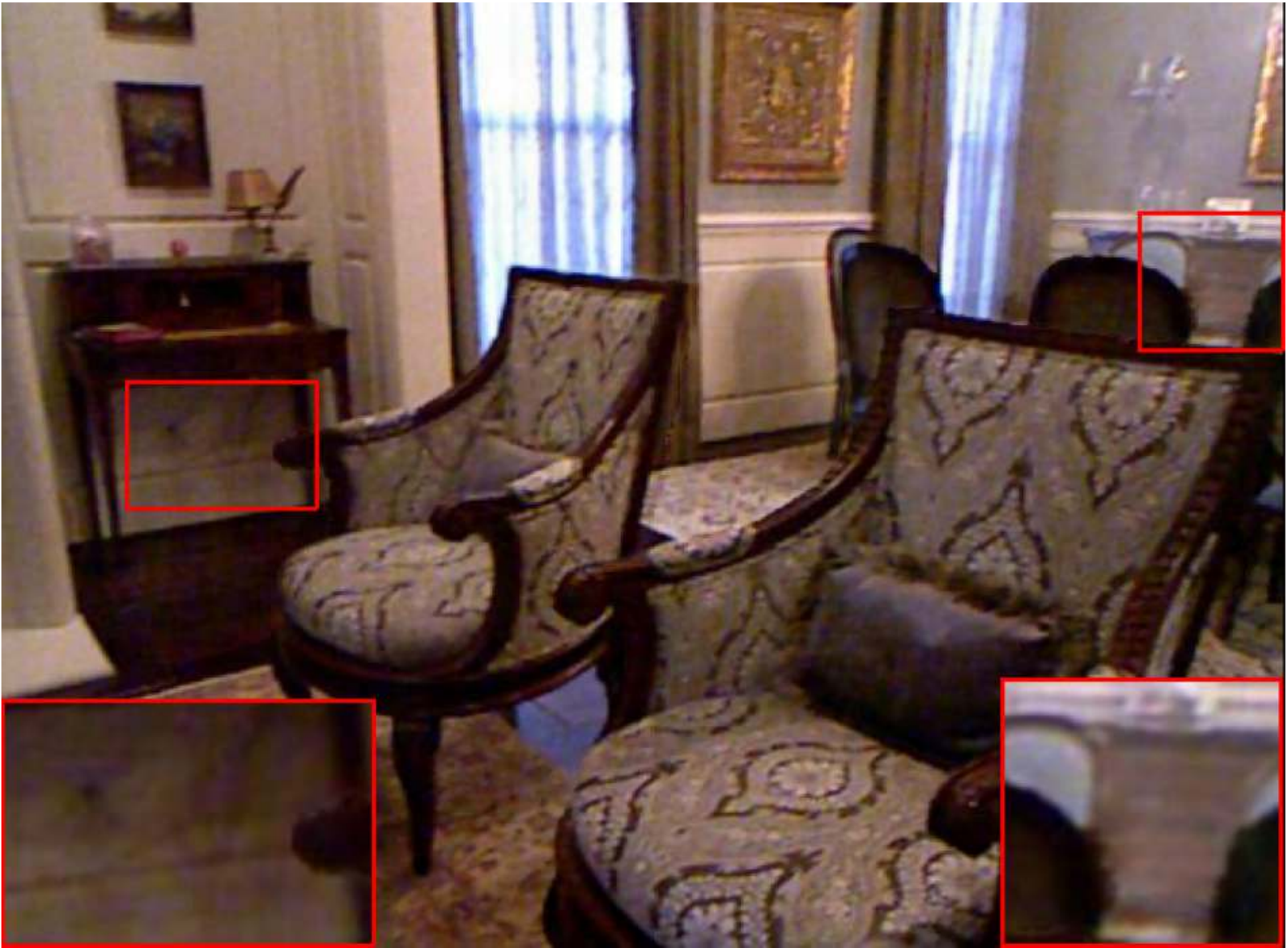}\vspace{2pt}
	\end{minipage}}\hspace{-0.45em}
	\subfigure[\scriptsize{Ours}]{
		\begin{minipage}[b]{0.12\linewidth}
			\includegraphics[width=1\linewidth]{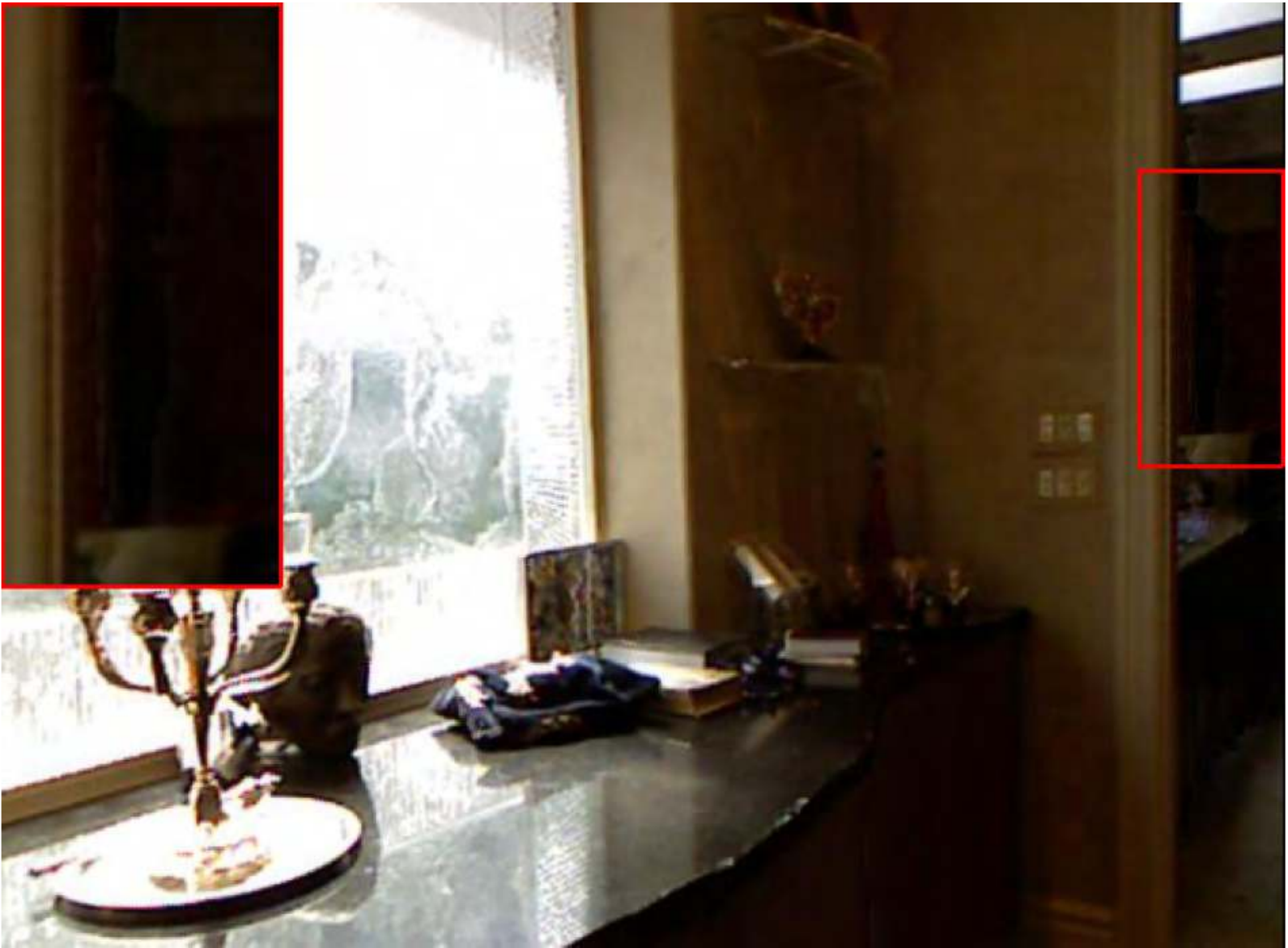}\vspace{2pt}
			\includegraphics[width=1\linewidth]{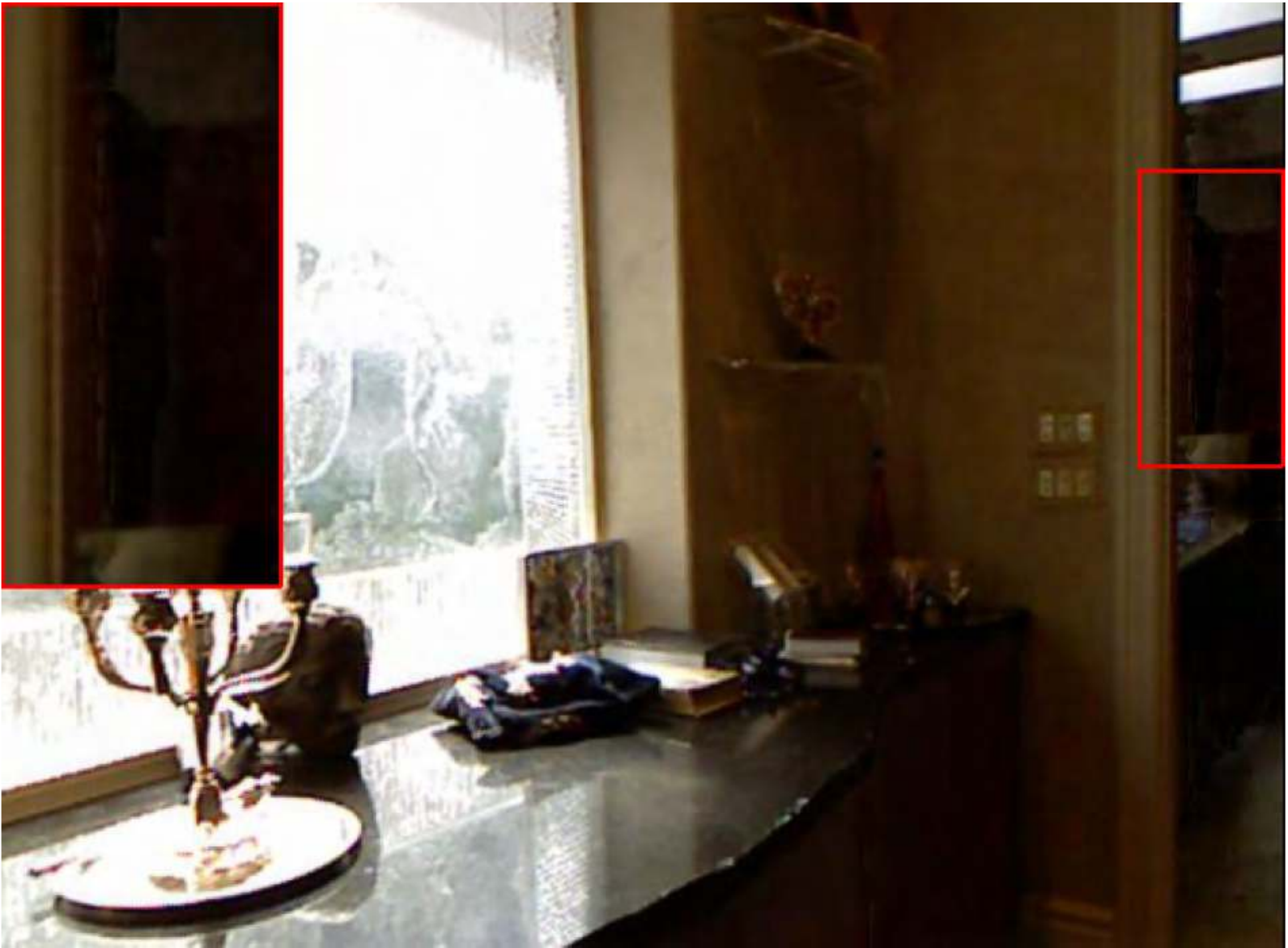}\vspace{2pt}
			\includegraphics[width=1\linewidth]{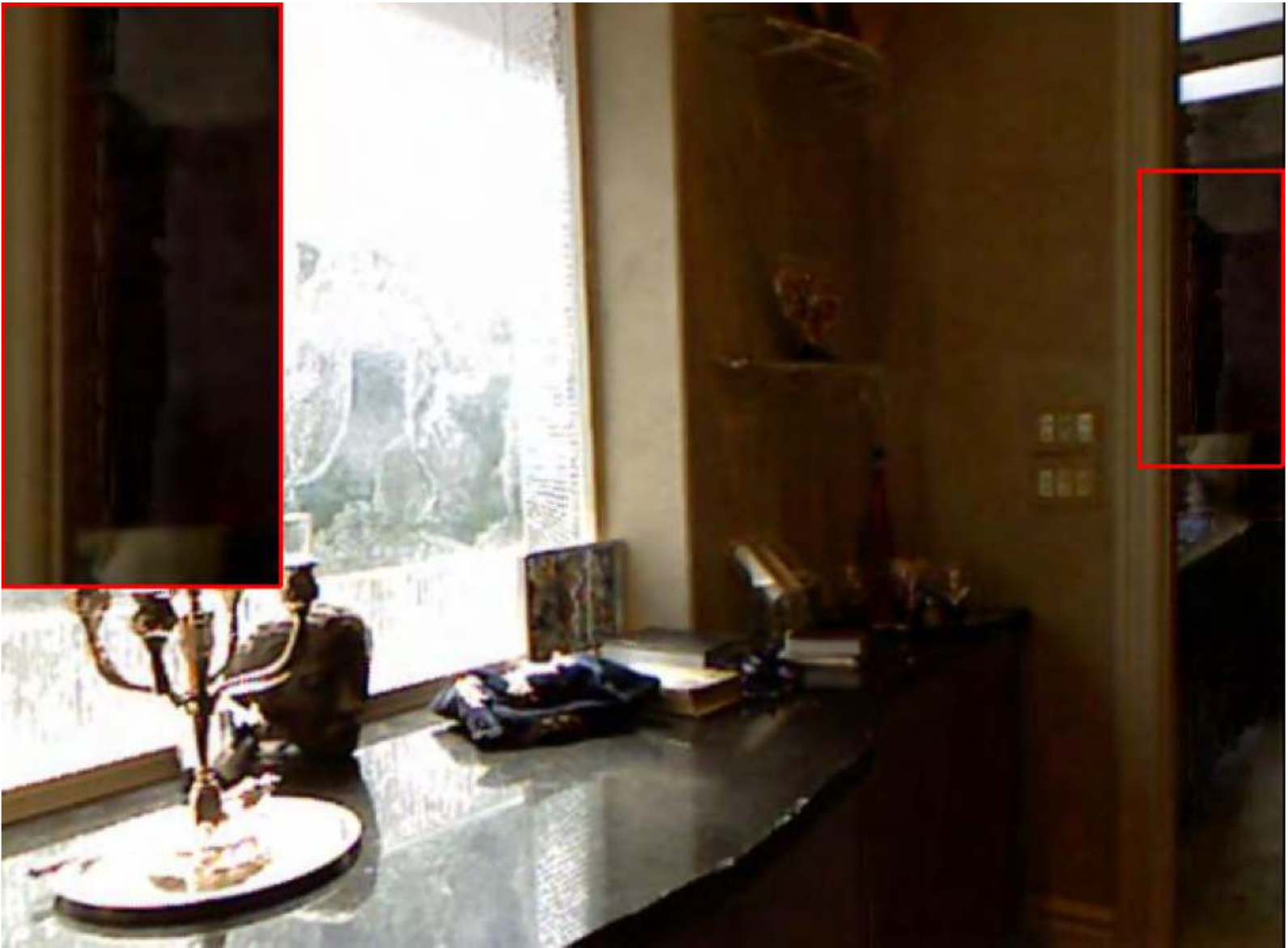}\vspace{2pt}
			\includegraphics[width=1\linewidth]{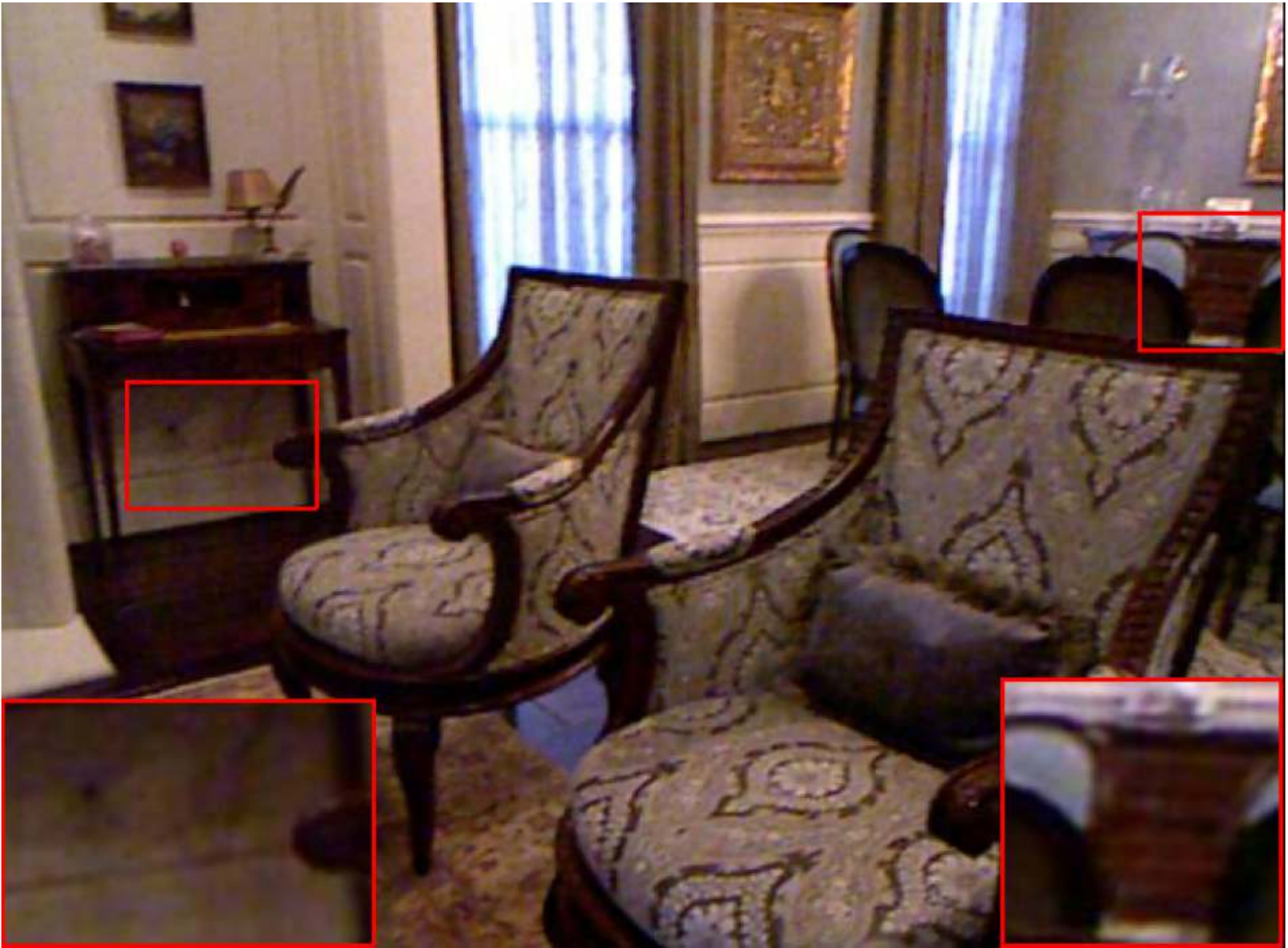}\vspace{2pt}
			\includegraphics[width=1\linewidth]{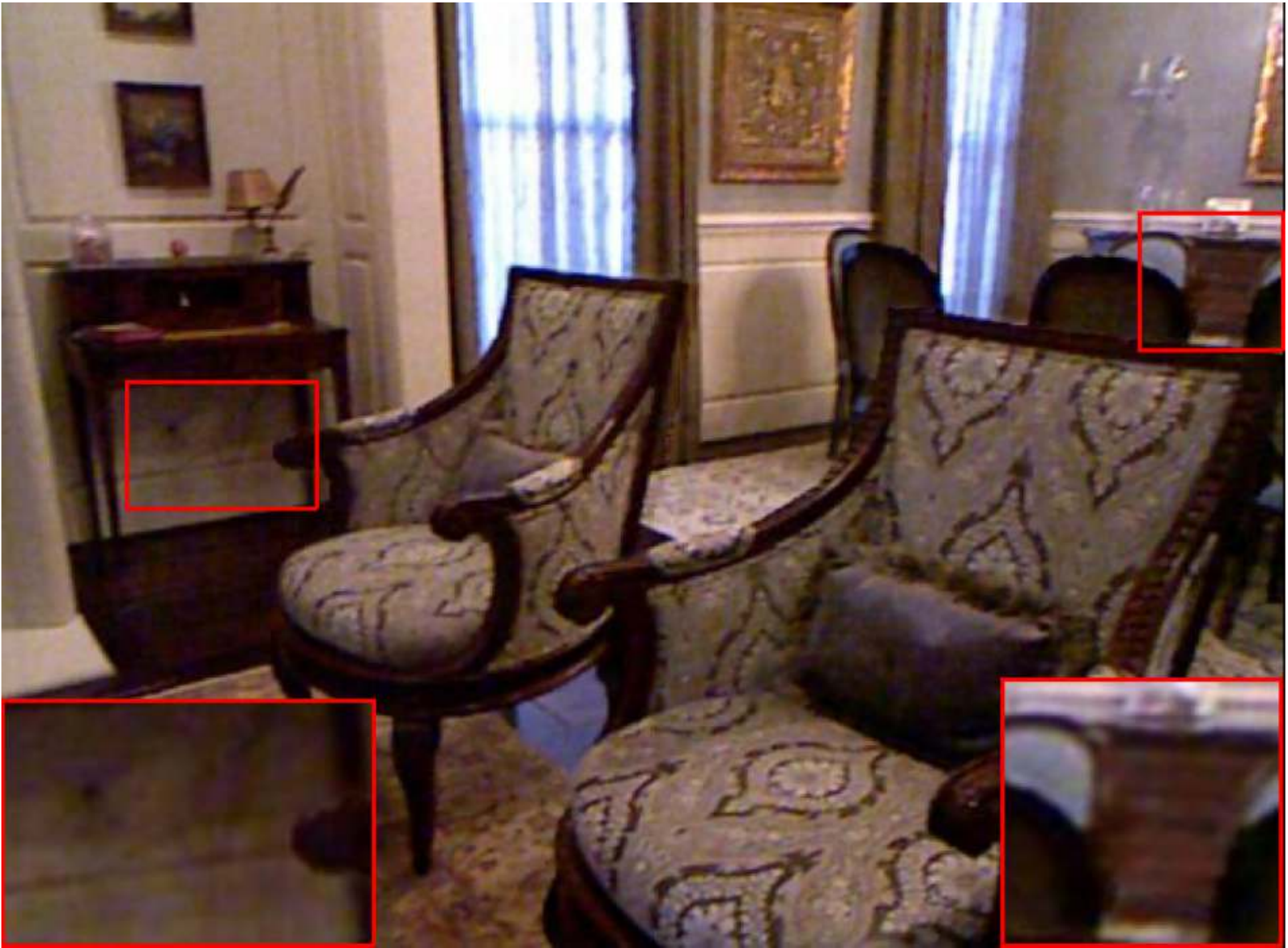}\vspace{2pt}
			\includegraphics[width=1\linewidth]{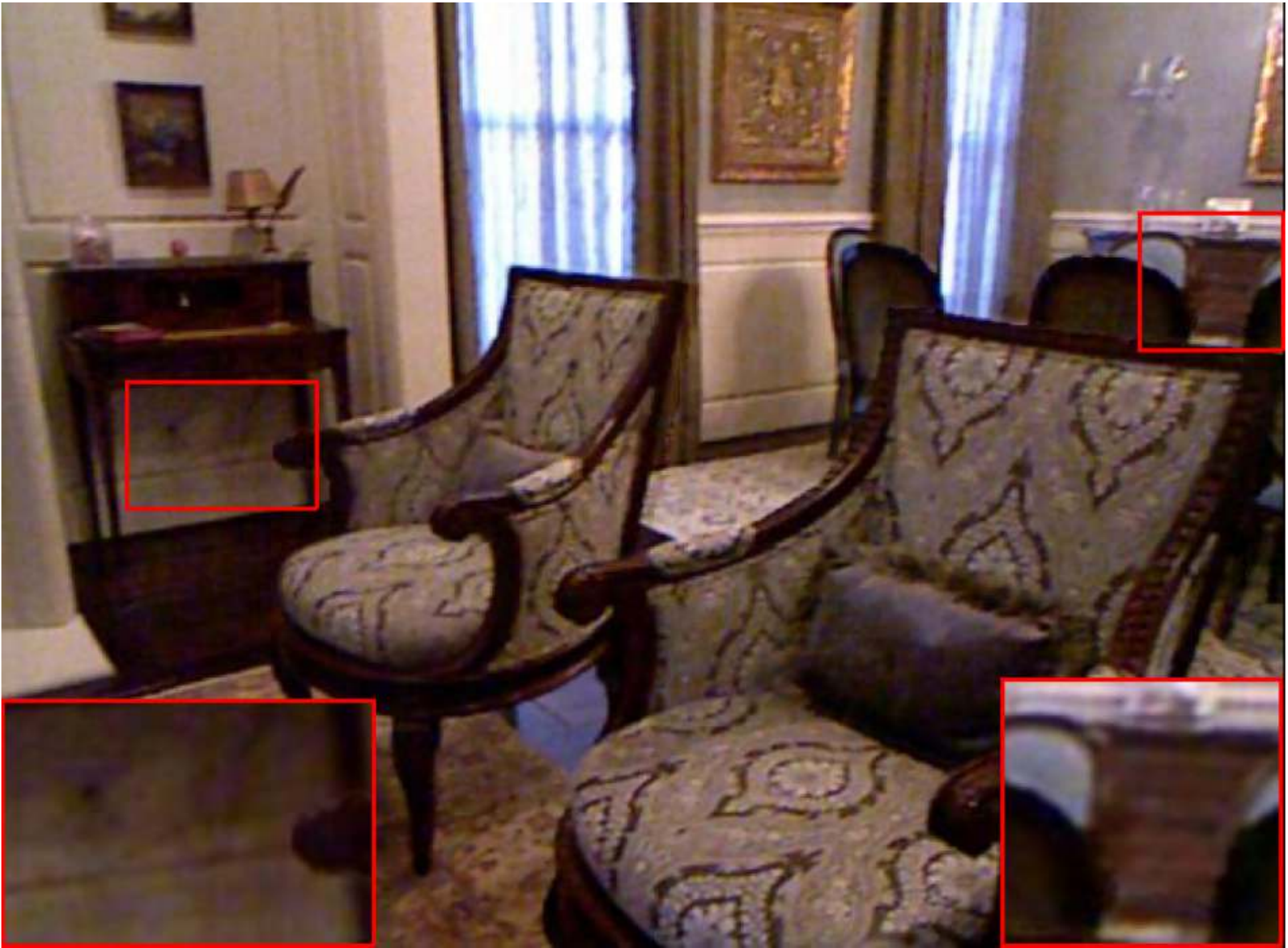}\vspace{2pt}
	\end{minipage}}\hspace{-0.45em}
	\subfigure[\scriptsize{GT}]{
		\begin{minipage}[b]{0.12\linewidth}
			\includegraphics[width=1\linewidth]{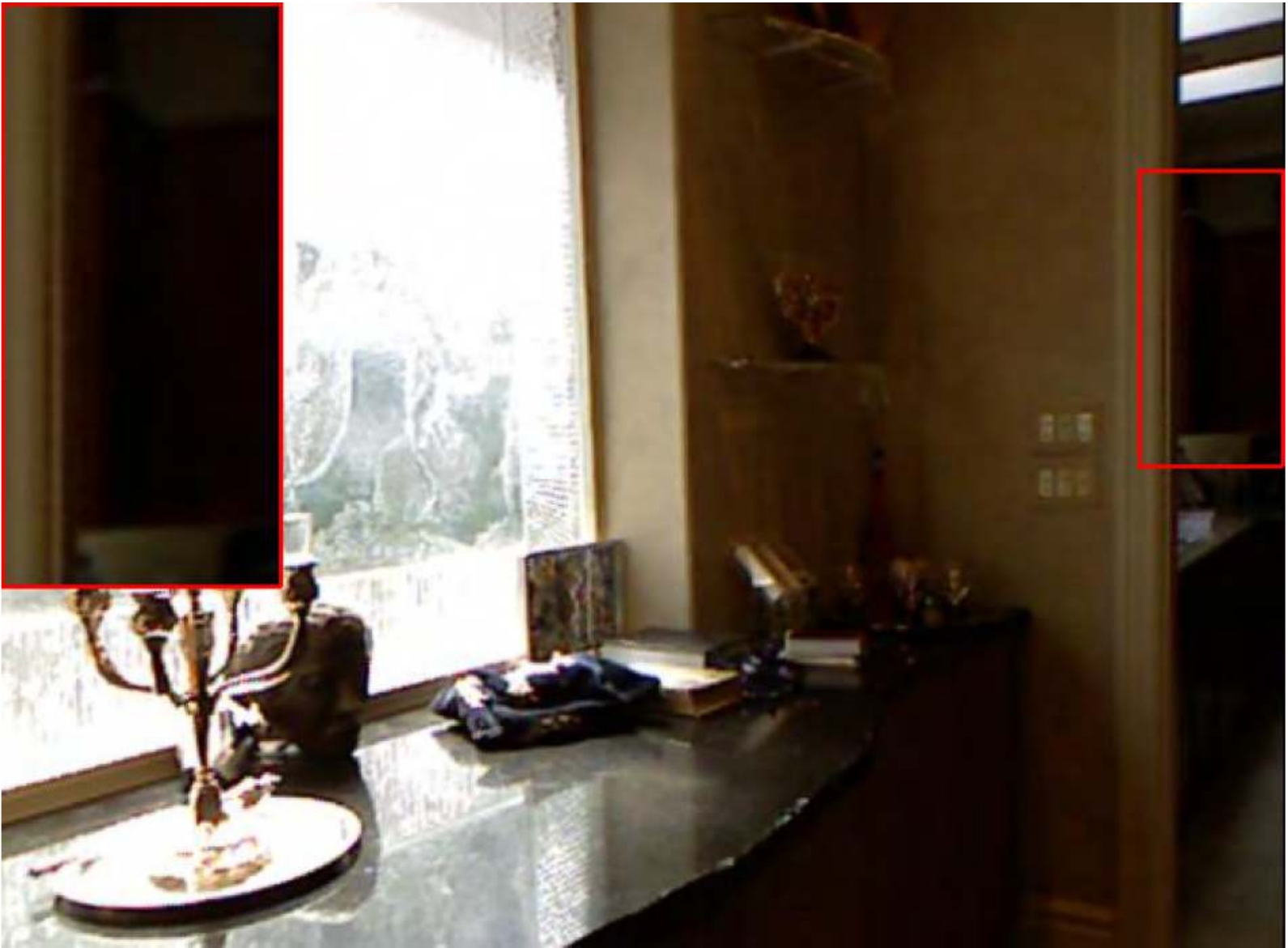}\vspace{2pt}
			\includegraphics[width=1\linewidth]{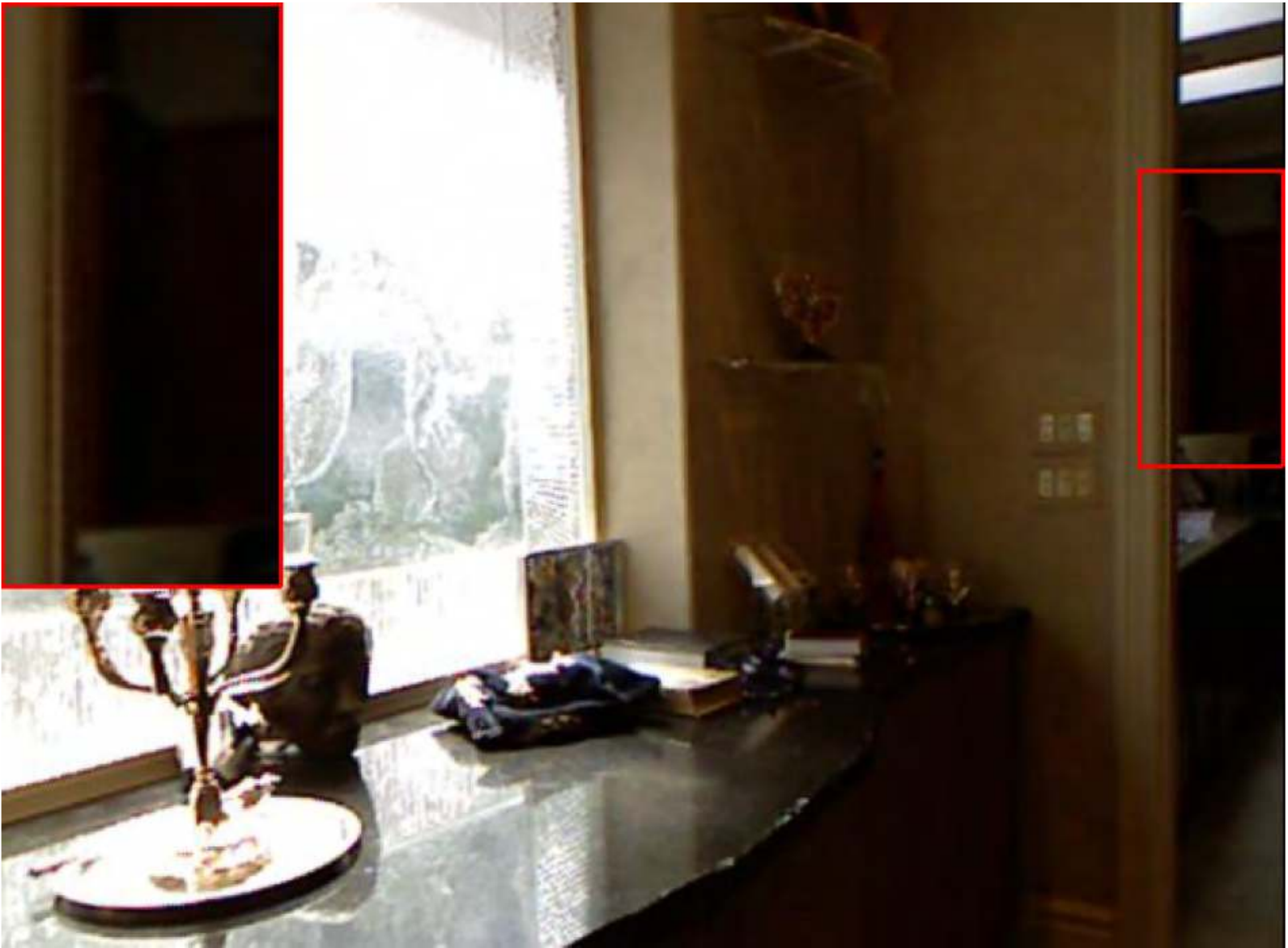}\vspace{2pt}
			\includegraphics[width=1\linewidth]{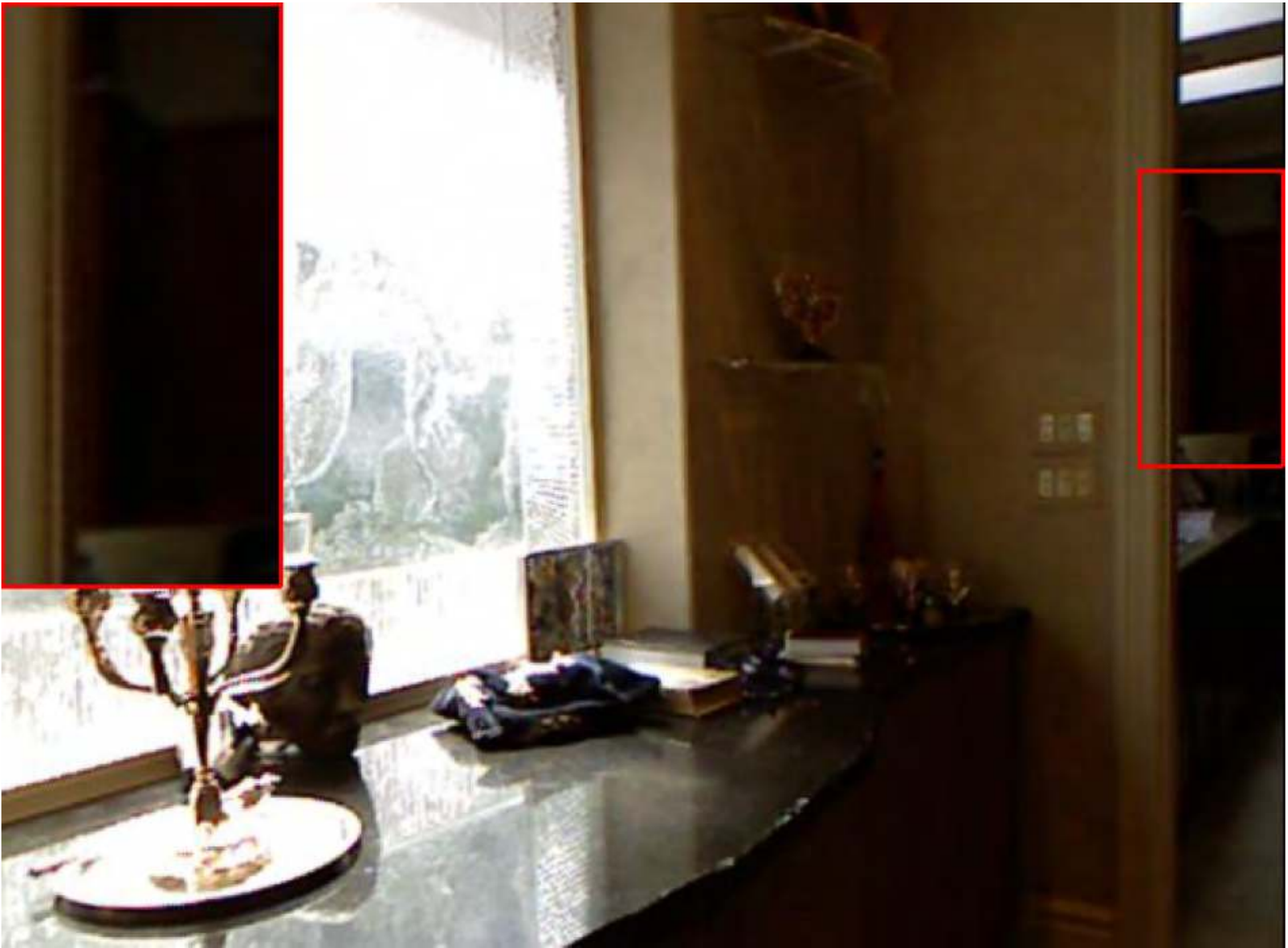}\vspace{2pt}
			\includegraphics[width=1\linewidth]{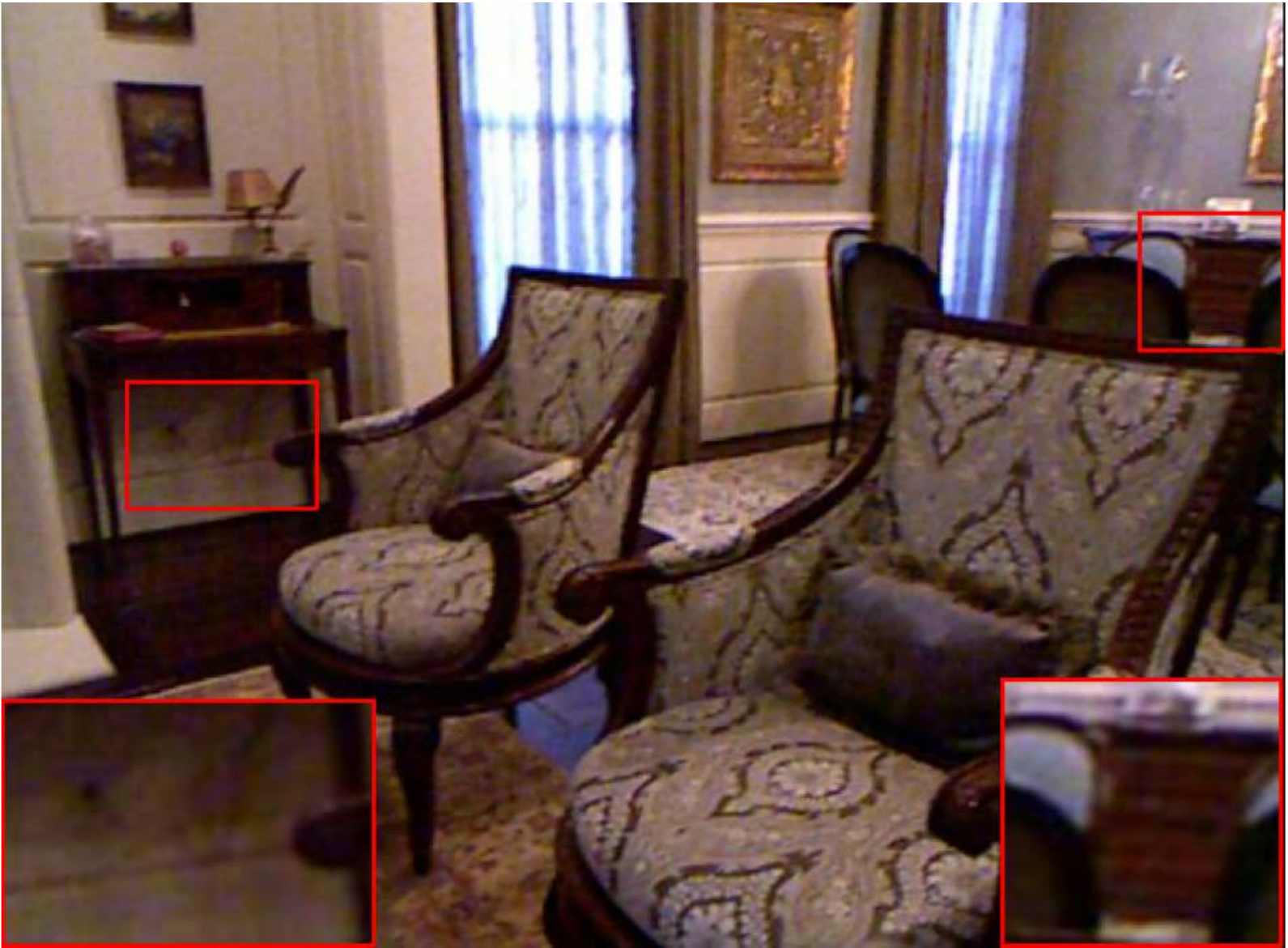}\vspace{2pt}
			\includegraphics[width=1\linewidth]{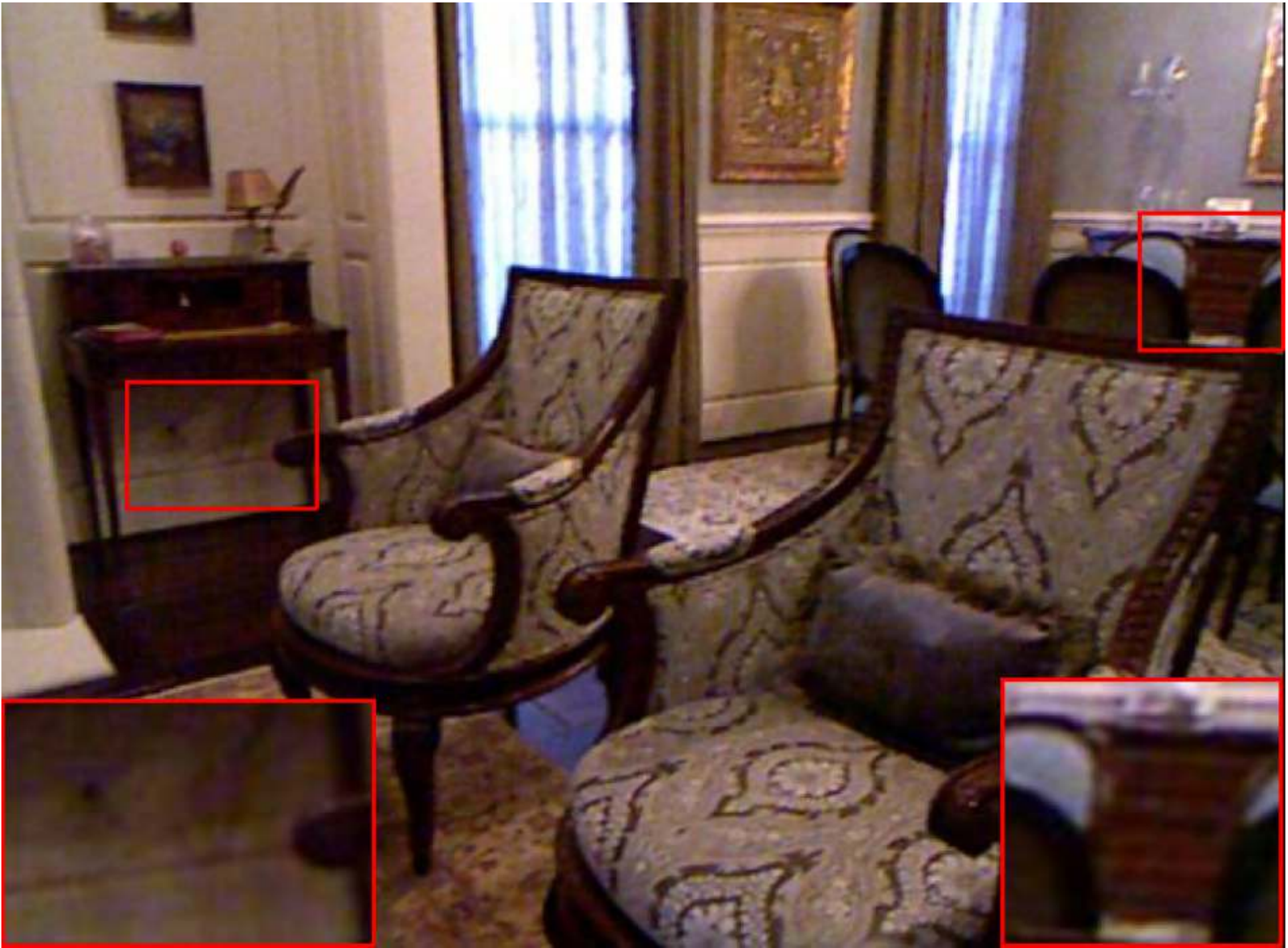}\vspace{2pt}
			\includegraphics[width=1\linewidth]{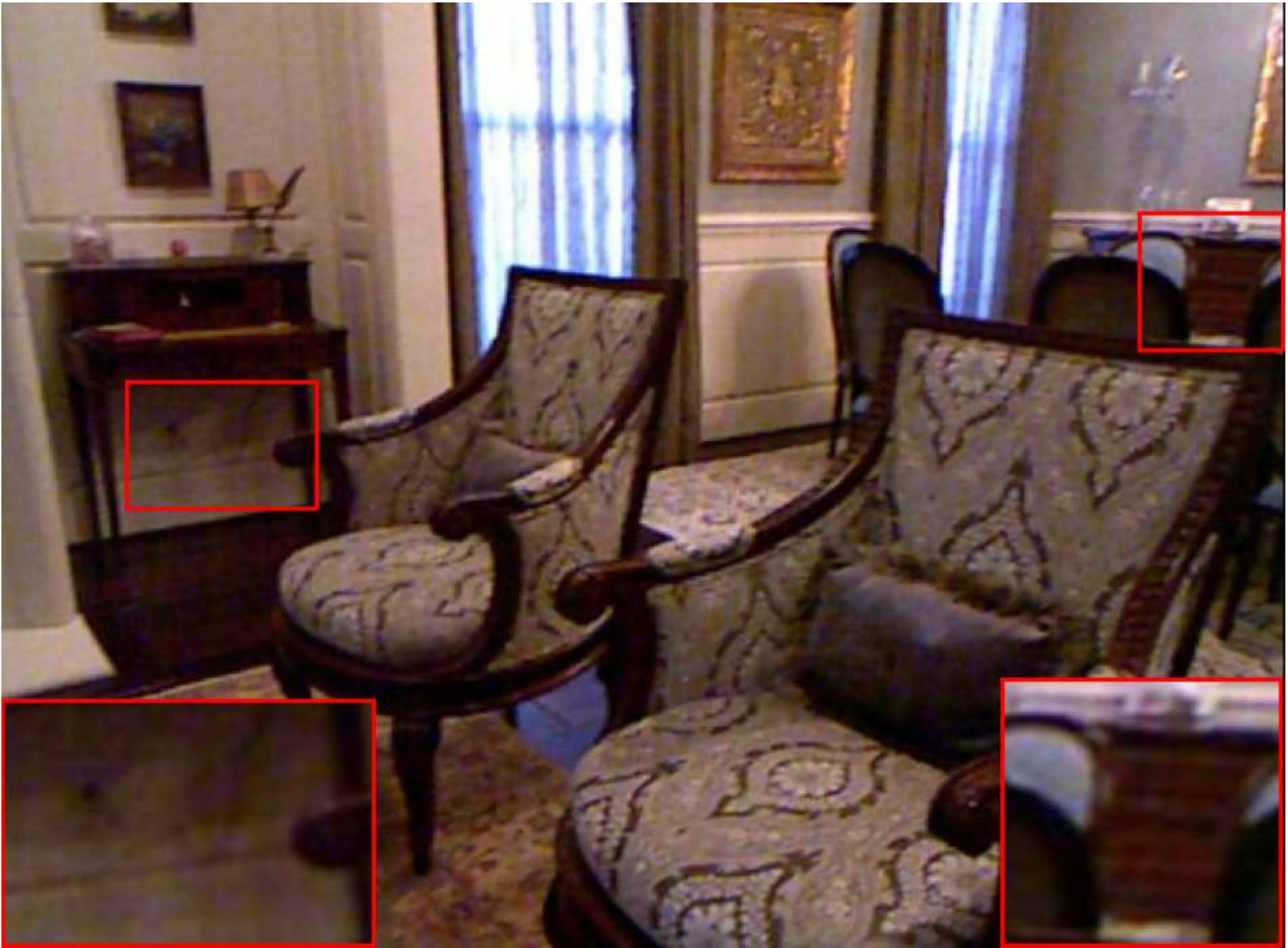}\vspace{2pt}
	\end{minipage}}
	\caption{Visual results of images with different haze distributions on SOTS~\cite{li2019benchmarking} dataset}
	\label{figure:syn_its_contrast}
\end{figure*}

\subsection{Experimental details}

\subsection{Datasets} We choose RESIDE~\cite{li2019benchmarking} dataset as our training dataset. For the intra-domain adaptation step, we randomly sample 8000 hazy images from ITS (Indoor Training Set) and 8000 hazy images from OTS (Outdoor Training Set). We utilize 4 same scene images with haze distribution shift to represent the varied haze distributions of the same scene and we explore 2000 indoor scenes and 2000 outdoor scenes to train our network. In other words, there are total 20,000 images (including clear labels) in the synthetic training set. For the inter-domain adaptation step, we randomly sample 3000 real hazy images from URHI (Unannotated Realistic Hazy Images). To achieve data augmentation, we randomly crop images to $256\times 256$ and randomly flip the cropped images horizontally during the training phase. Furthermore, we ensure that the crop areas and horizontal directions of the same scene images (four hazy images and one clear image) are consistent in each iteration.

\subsection{Implementation details} We implement our method using PyTorch~\cite{paszke2017automatic} framework, and we conduct experiments on both our designed base network (encoder-decoder architecture with residual blocks in Figure~\ref{fig:framework}) and MSBDN-DFF~\cite{dong2020multi} architecture. First, we train the dehazing module ($G$ and $R$) and the intra-domain discriminator $D_{intra}$ within the synthetic domain for 200 epochs. For the dehazing module, we apply the SGD~\cite{bottou2010large} optimizer with learning rate $1.25\times 10^{-4}$, momentum $0.9$ and weight decay $5\times 10^{-4}$. For the intra-domain discriminator, we apply the Adam~\cite{kingma2014adam} optimizer with learning rate $1\times 10^{-4}$, $\beta_1=0.9$ and $\beta_2=0.99$. We set parameter of reversed gradients as 0.1 in GRL module. Then, we adapt model to real hazy images by training network $G'$, $R$ and inter-domain discriminator $D_{inter}$ for 20 epochs. Since all the features need to fall into the optimal subset of the synthetic domain, we freeze reconstruction network $R$ for 15 epoch and fine-tuned it for 5 epoch. We apply SGD optimizer with learning rate $1\times 10^{-4}$ for the dehazing module and Adam optimizer with learning rate $1\times 10^{-4}$ for the inter-domain discriminator. 

%

\begin{figure*}[t]
	\centering
	\subfigure[\scriptsize{Input}]{
		\begin{minipage}[b]{0.12\linewidth}
			\includegraphics[width=1\linewidth]{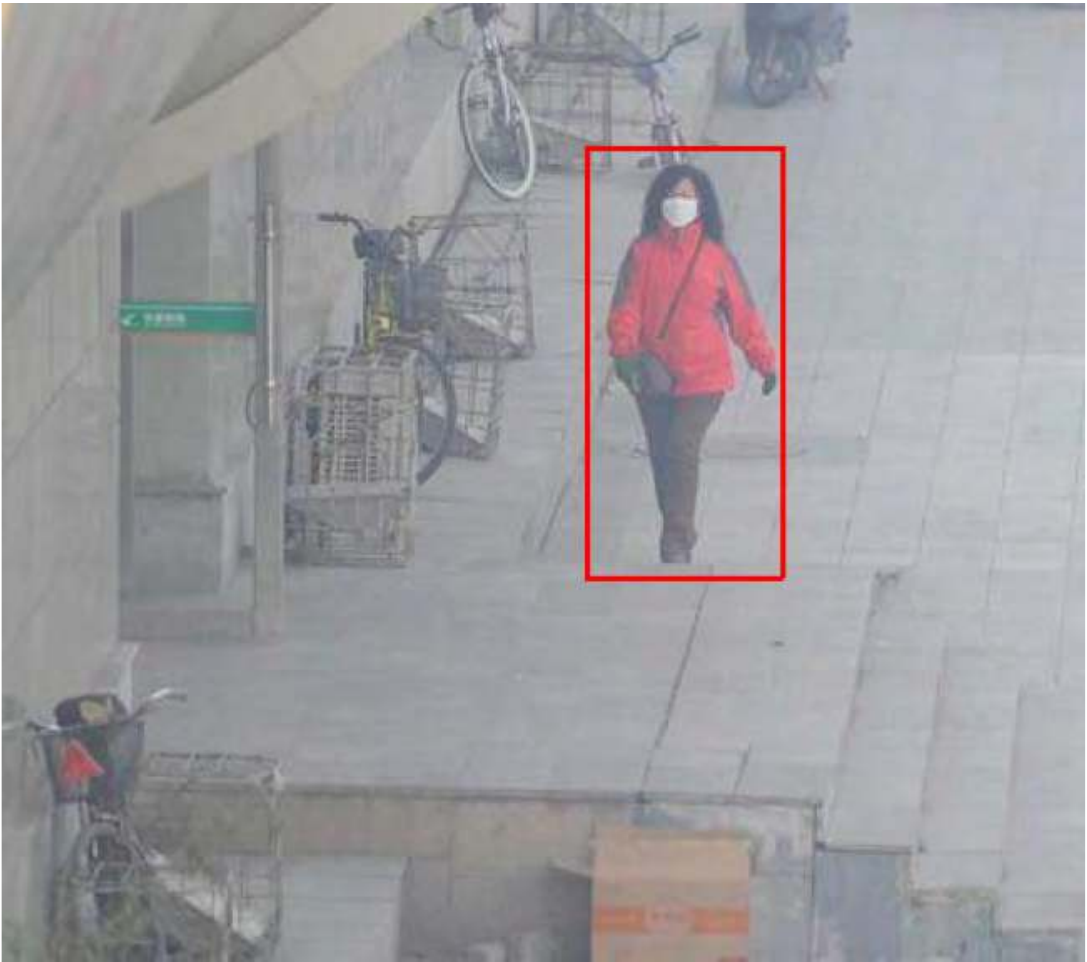}\vspace{2pt}
			\includegraphics[width=1\linewidth]{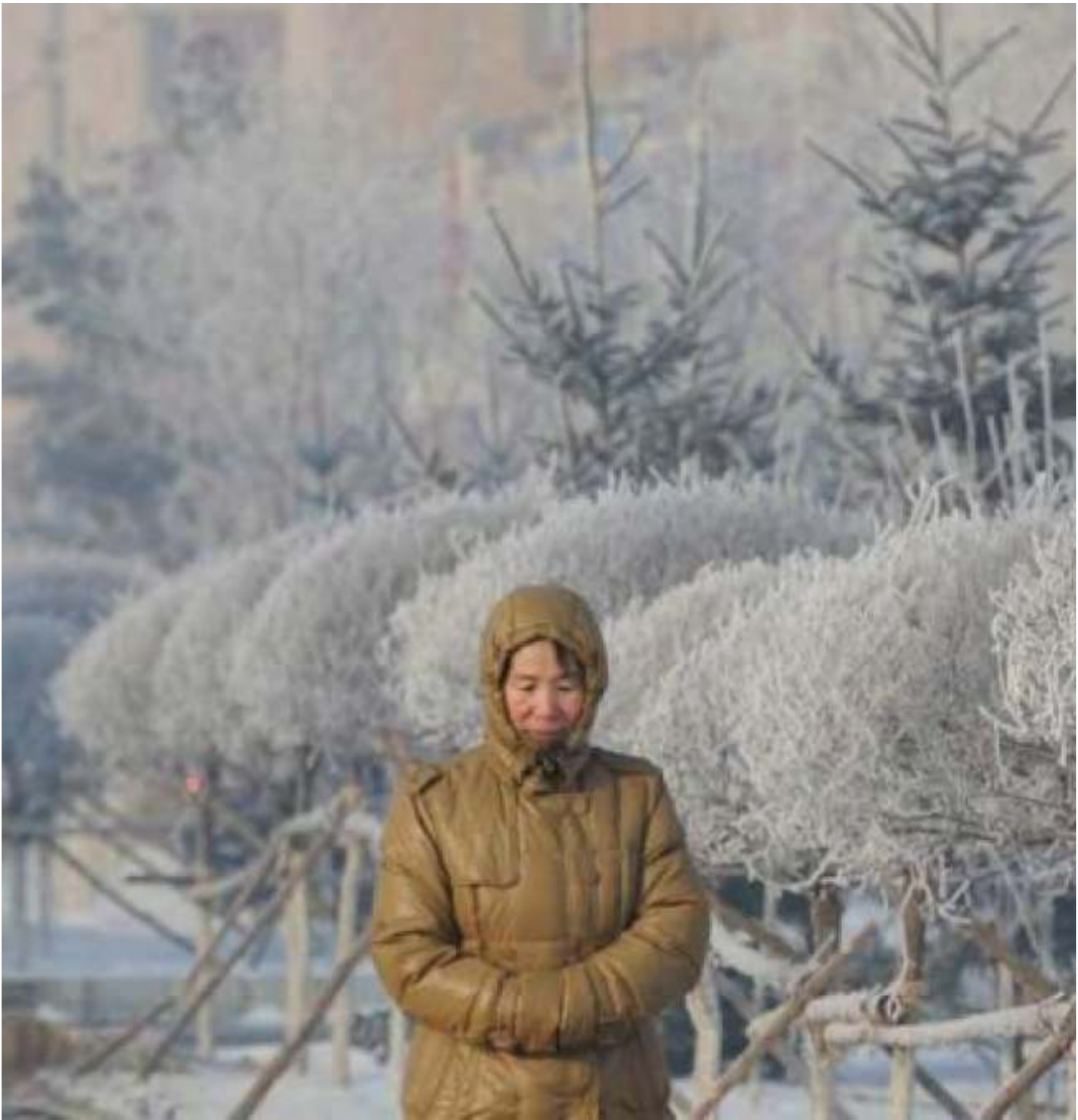}\vspace{2pt}
			\includegraphics[width=1\linewidth]{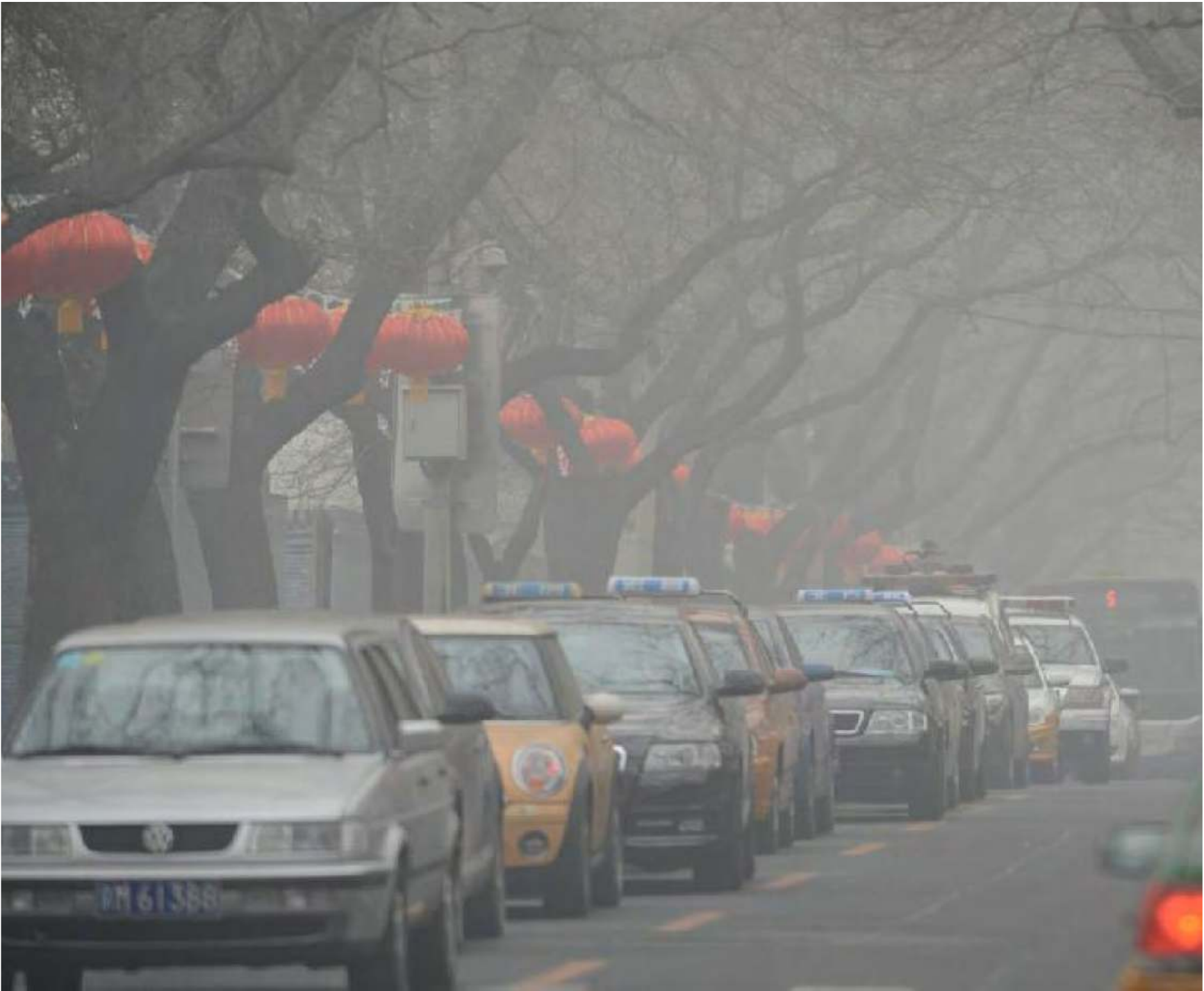}\vspace{2pt}
			\includegraphics[width=1\linewidth]{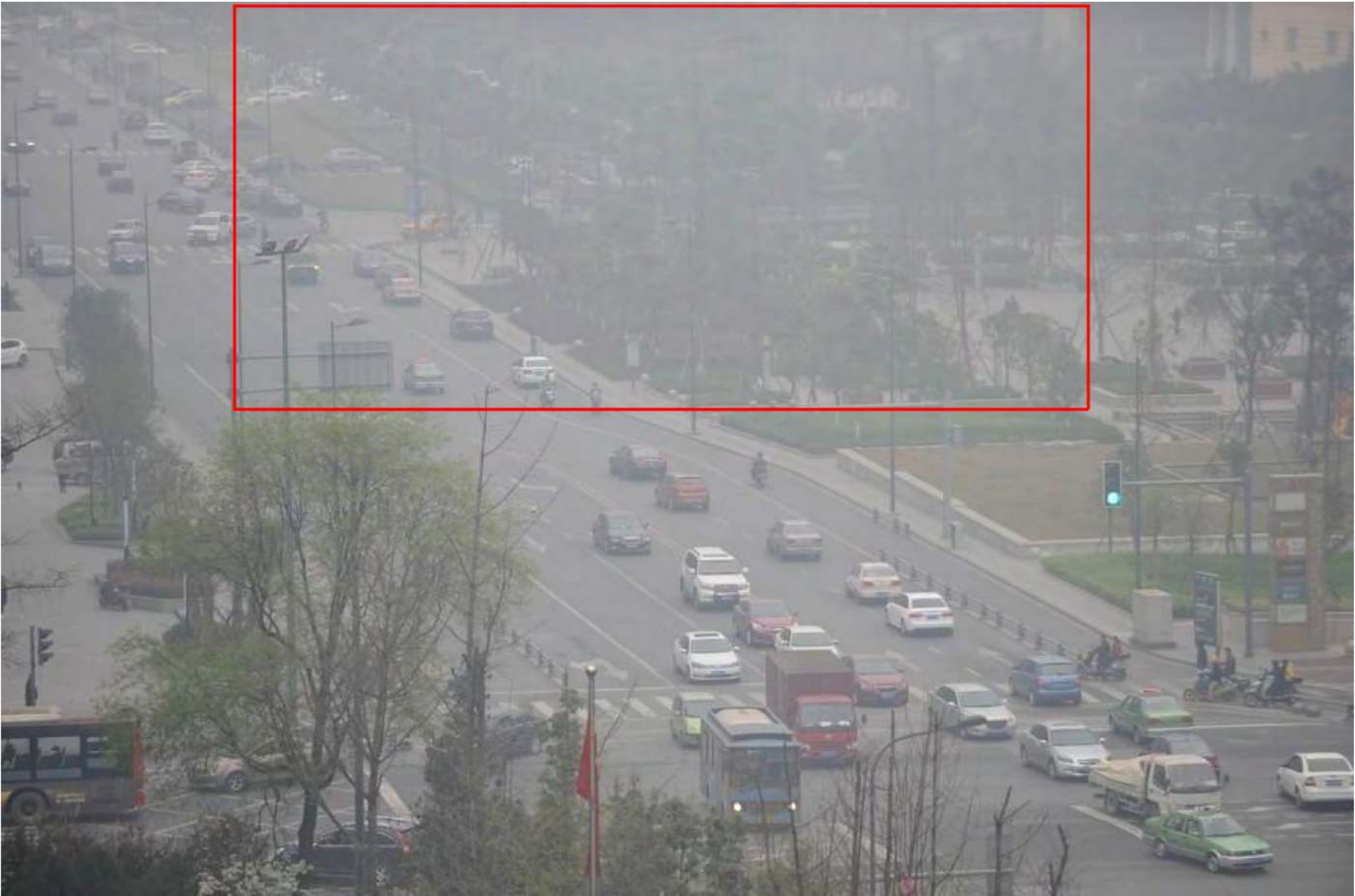}\vspace{2pt}
			\includegraphics[width=1\linewidth]{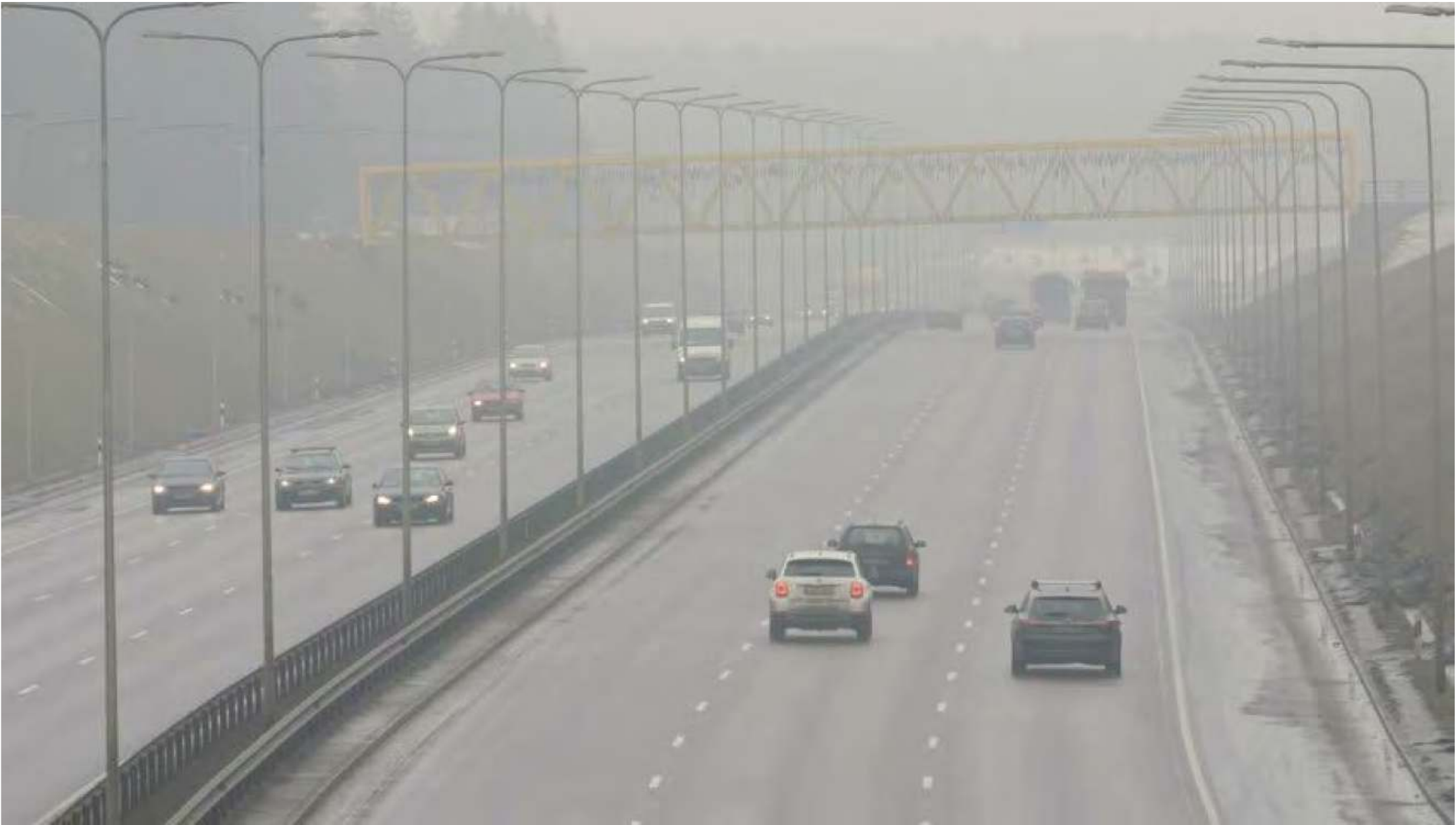}\vspace{2pt}
			\includegraphics[width=1\linewidth]{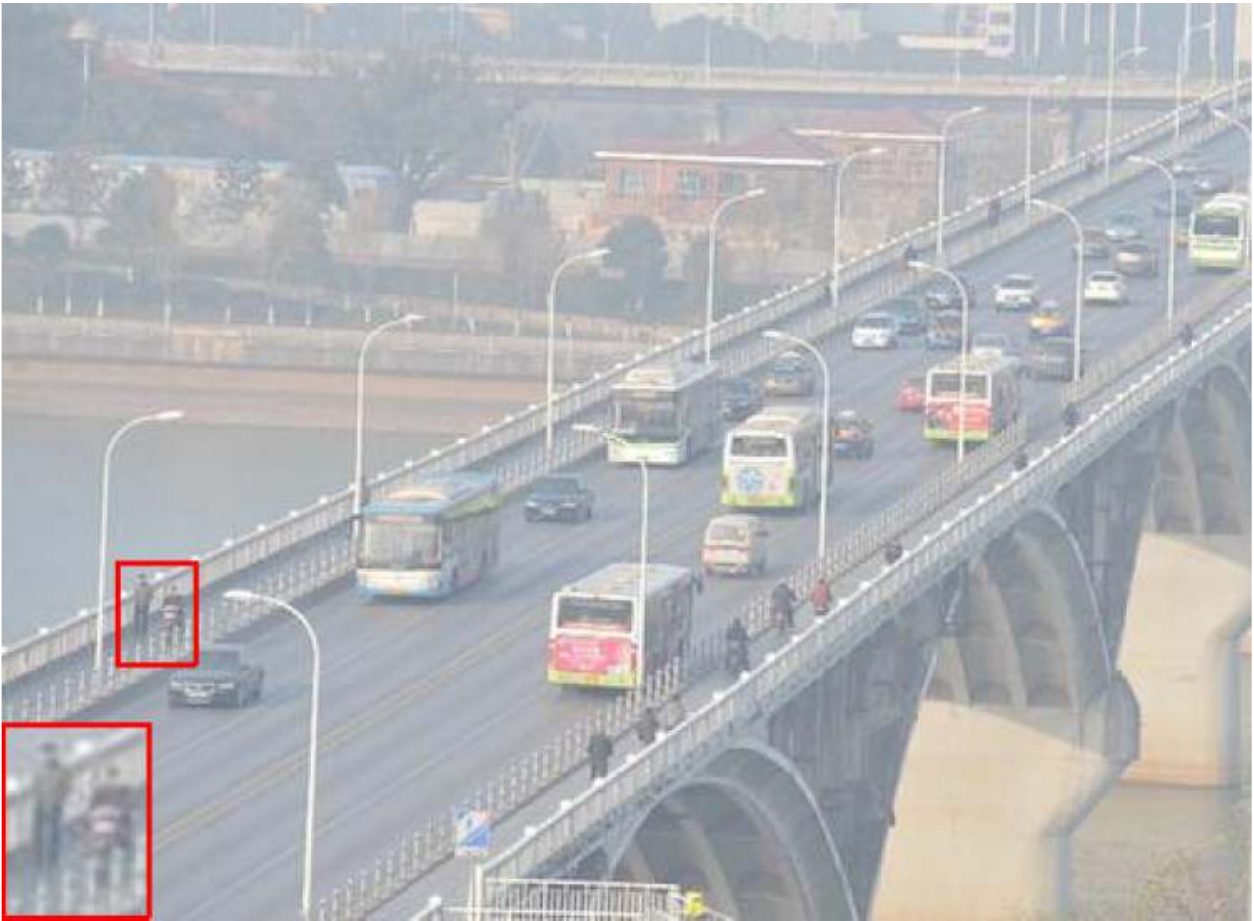}\vspace{2pt}
			\includegraphics[width=1\linewidth]{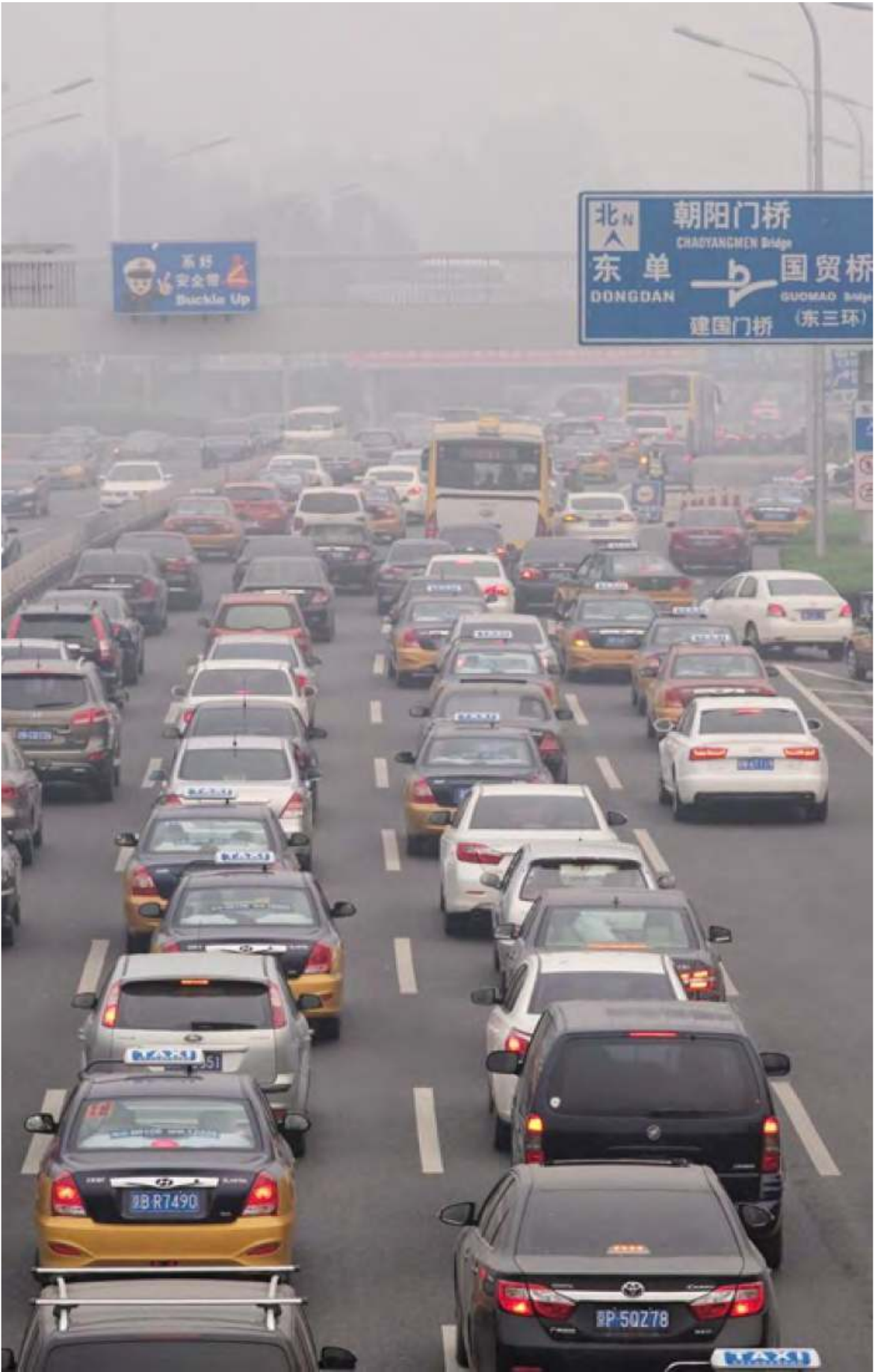}\vspace{2pt}
			\includegraphics[width=1\linewidth]{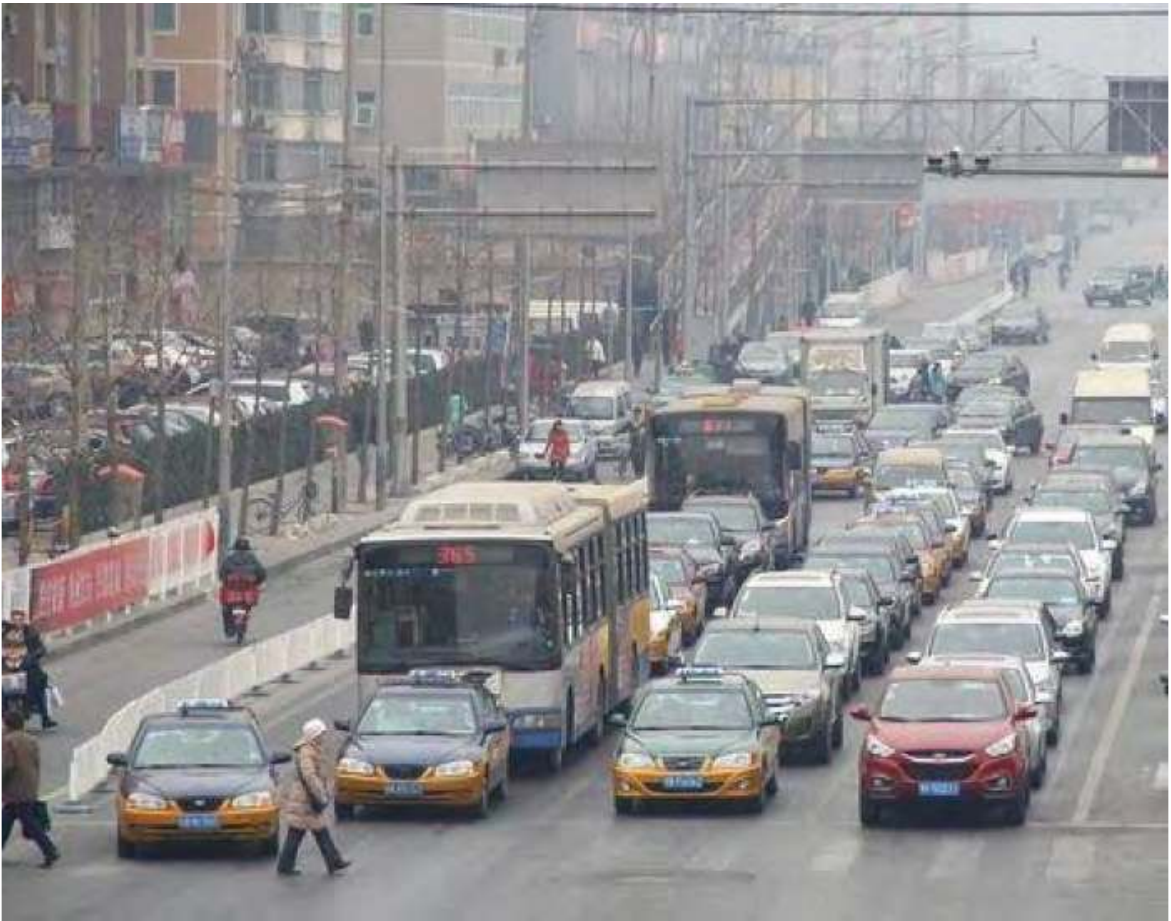}\vspace{2pt}
	\end{minipage}}\hspace{-0.45em}
	\subfigure[\scriptsize{DCP~\cite{he2010single}}]{
		\begin{minipage}[b]{0.12\linewidth}
			\includegraphics[width=1\linewidth]{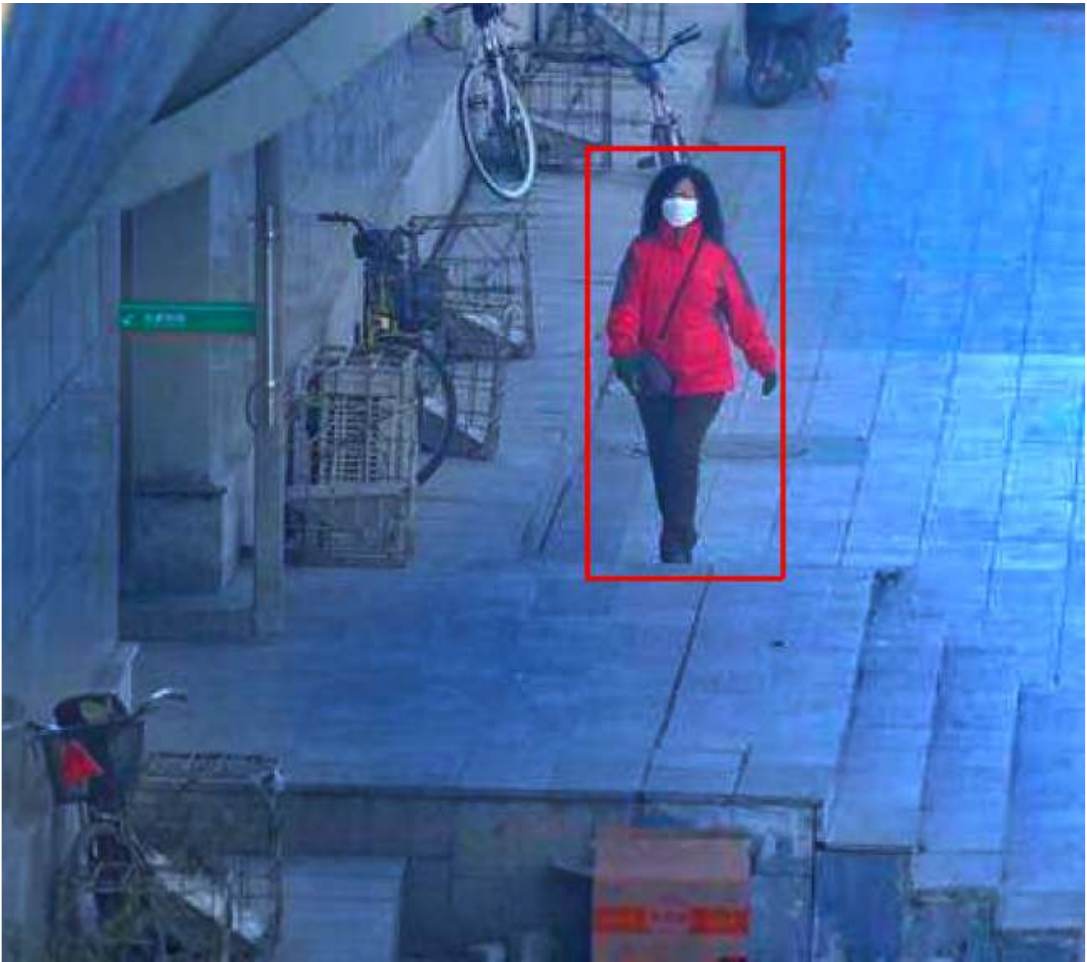}\vspace{2pt}
			\includegraphics[width=1\linewidth]{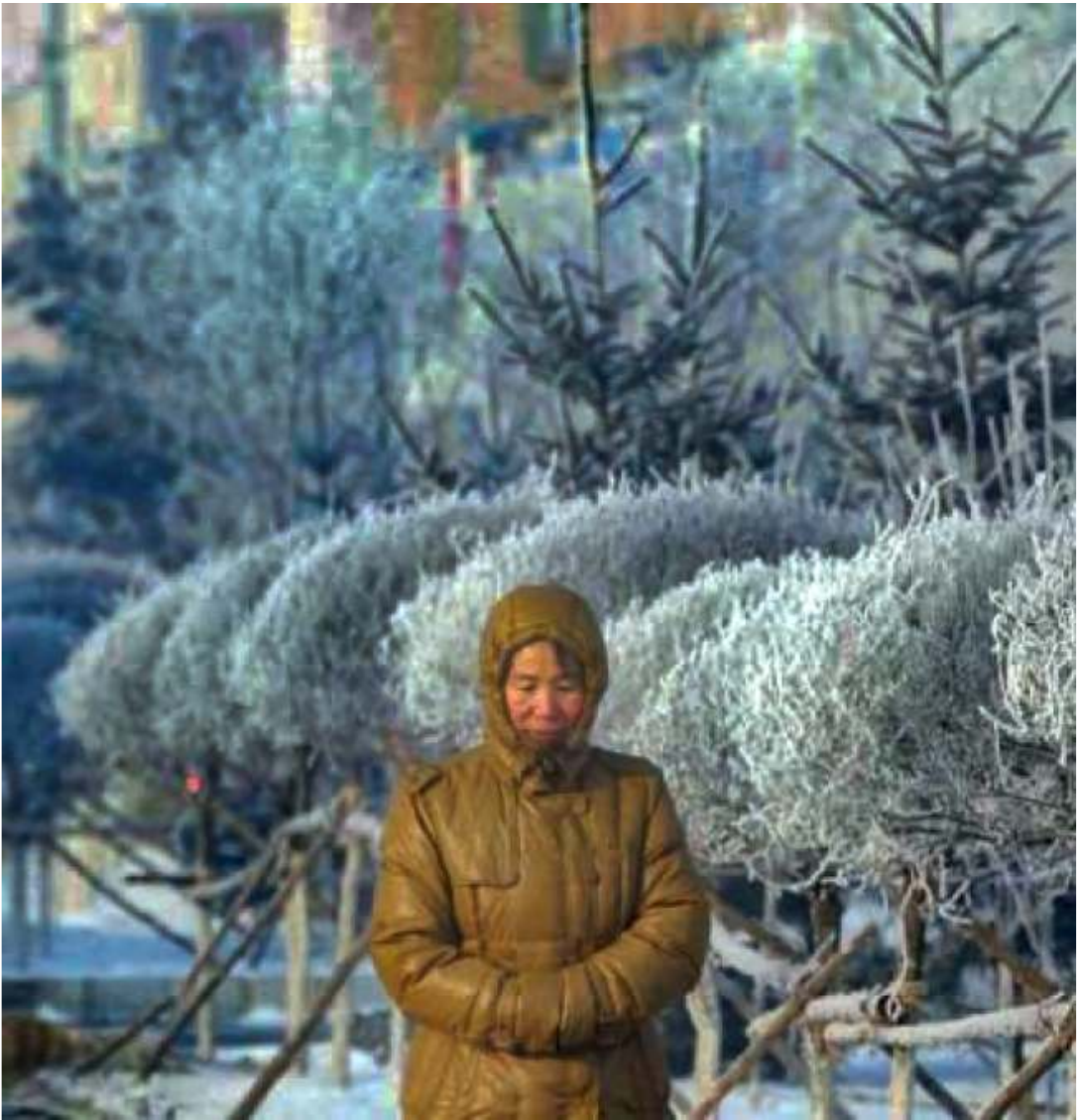}\vspace{2pt}
			\includegraphics[width=1\linewidth]{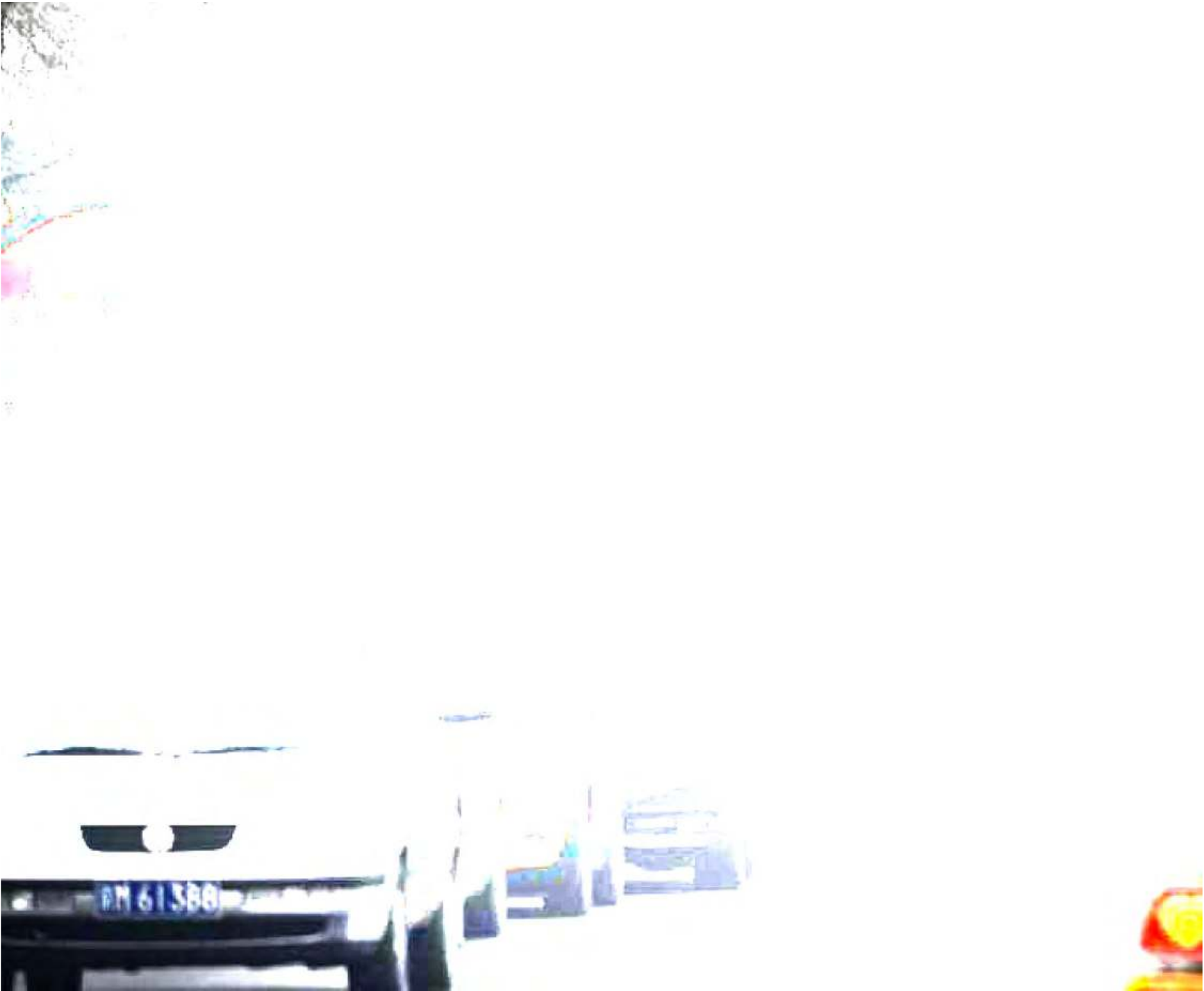}\vspace{2pt}
			\includegraphics[width=1\linewidth]{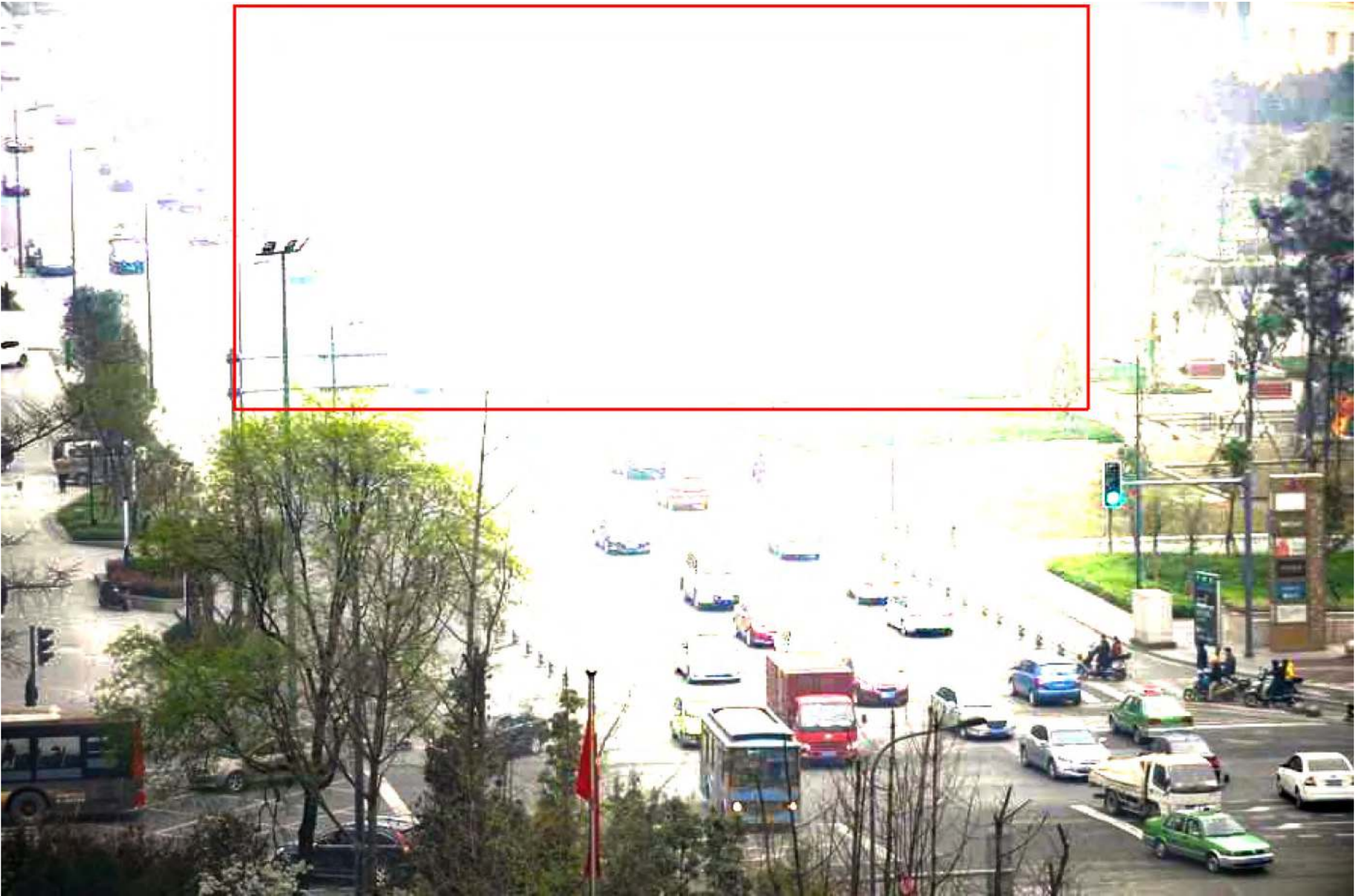}\vspace{2pt}
			\includegraphics[width=1\linewidth]{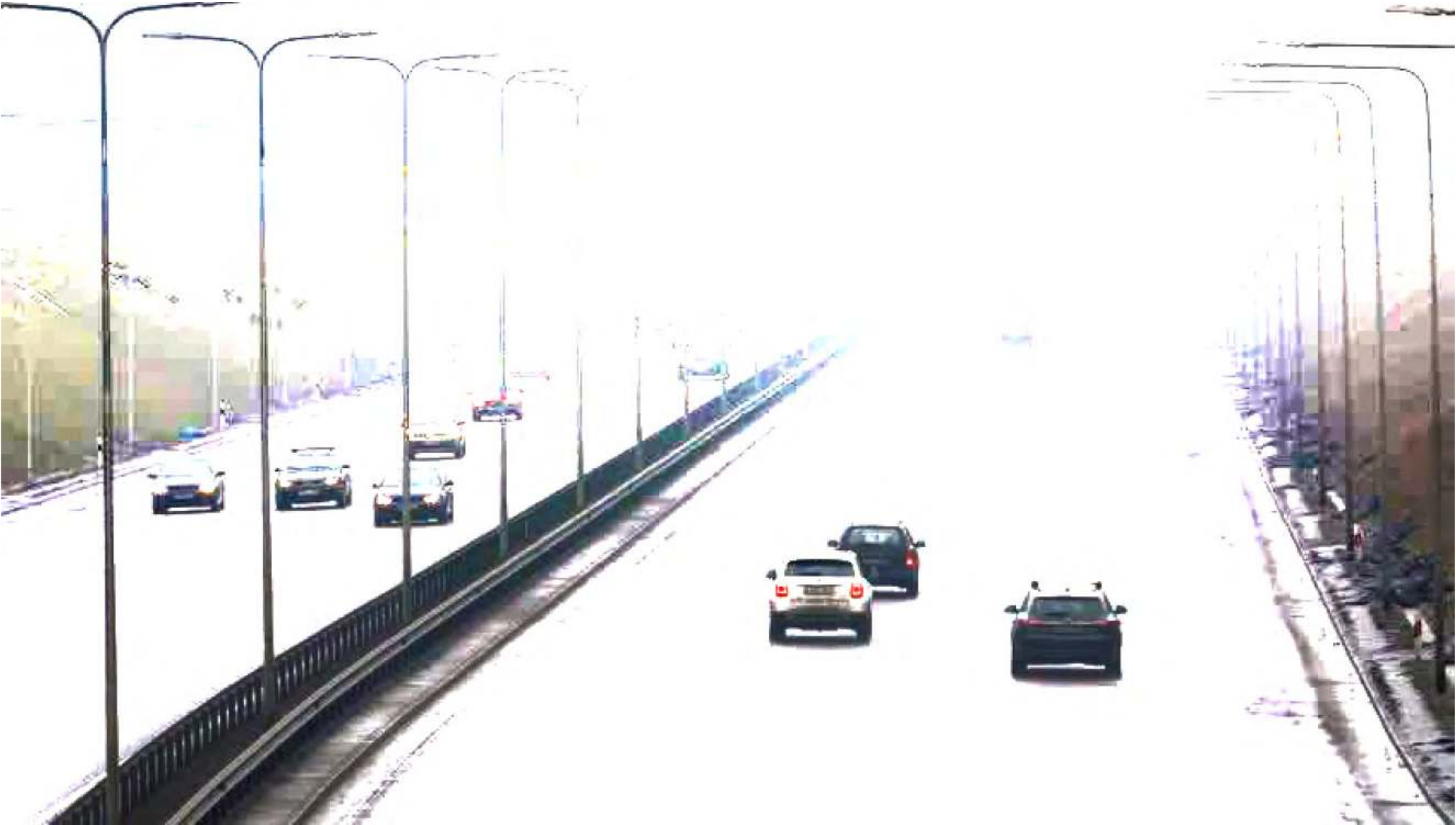}\vspace{2pt}
			\includegraphics[width=1\linewidth]{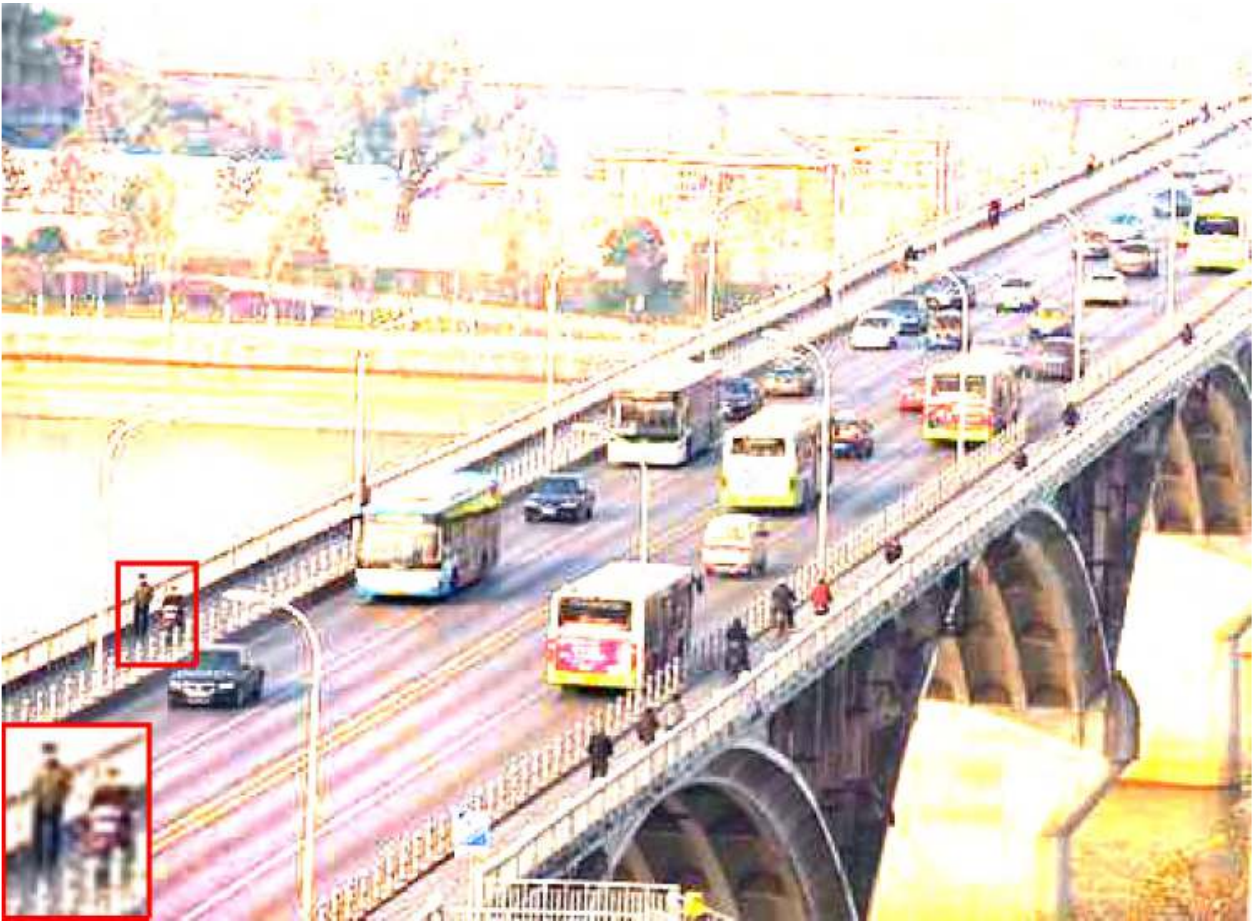}\vspace{2pt}
			\includegraphics[width=1\linewidth]{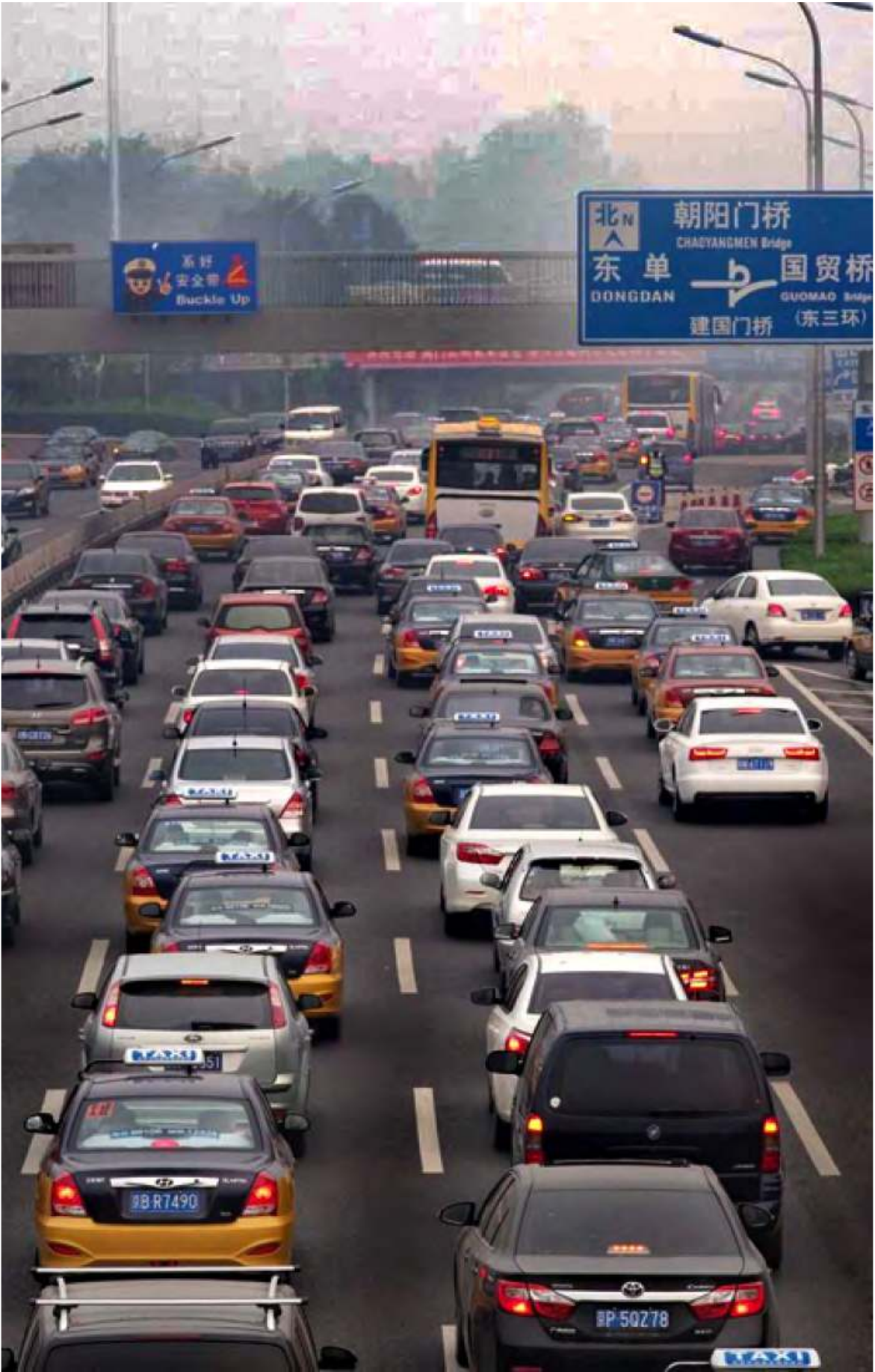}\vspace{2pt}
			\includegraphics[width=1\linewidth]{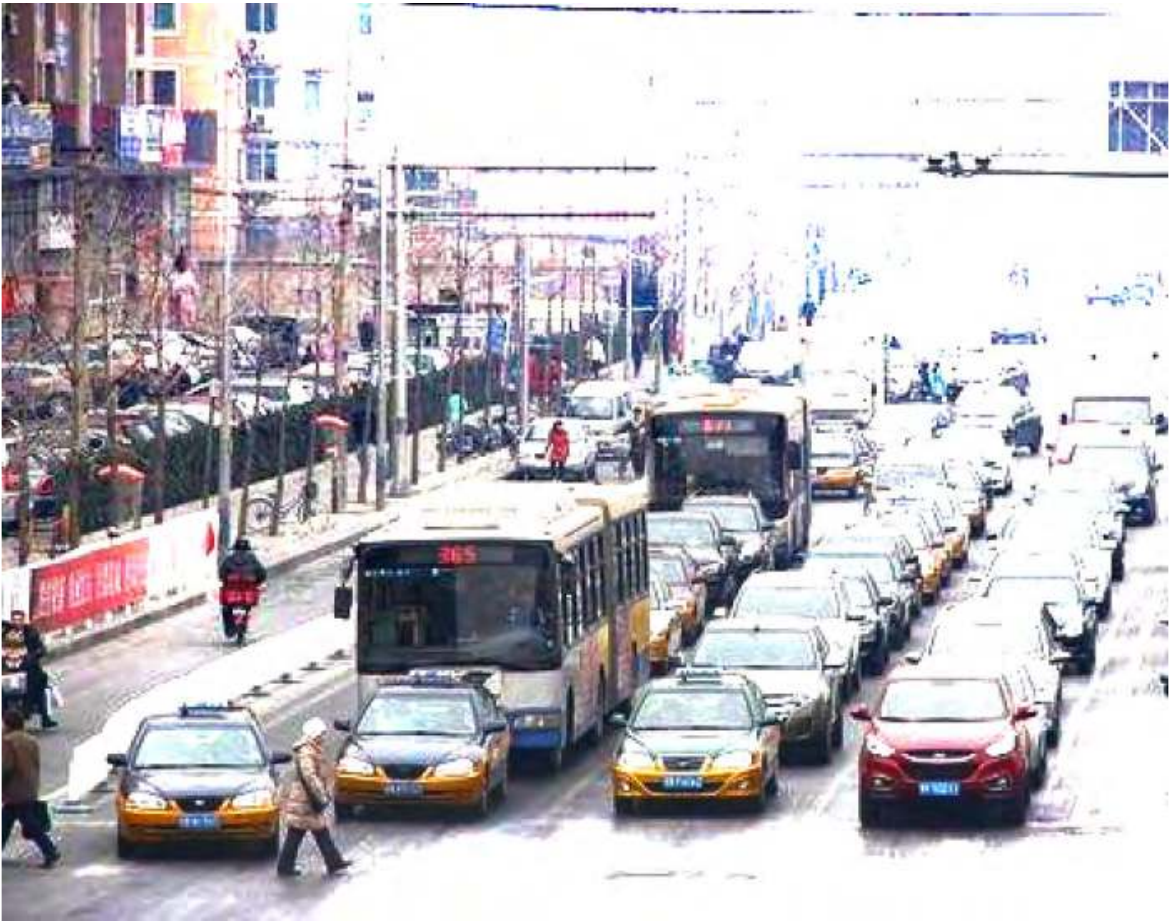}\vspace{2pt}
	\end{minipage}}\hspace{-0.45em}
	\subfigure[\scriptsize{DehazeNet~\cite{cai2016dehazenet}}]{
		\begin{minipage}[b]{0.12\linewidth}
			\includegraphics[width=1\linewidth]{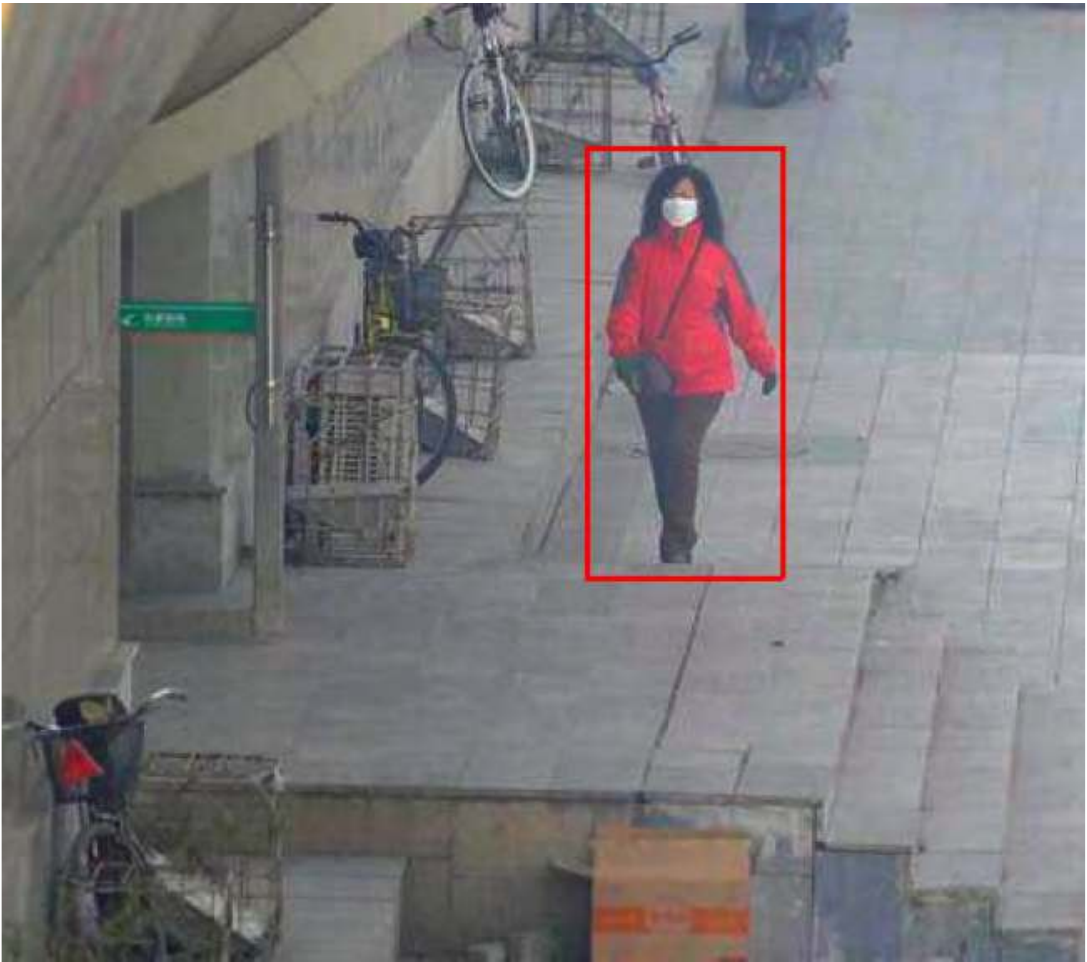}\vspace{2pt}
			\includegraphics[width=1\linewidth]{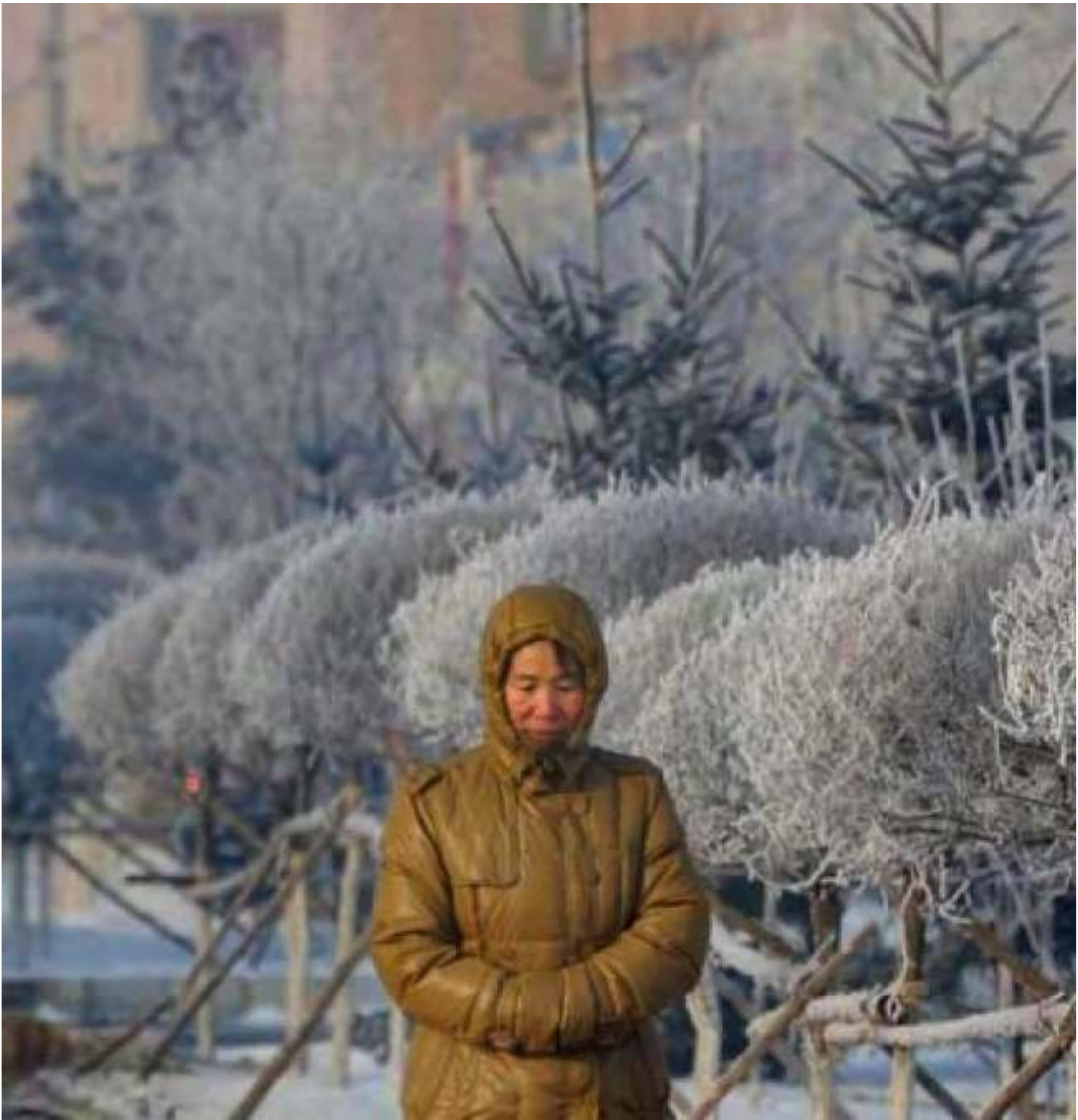}\vspace{2pt}
			\includegraphics[width=1\linewidth]{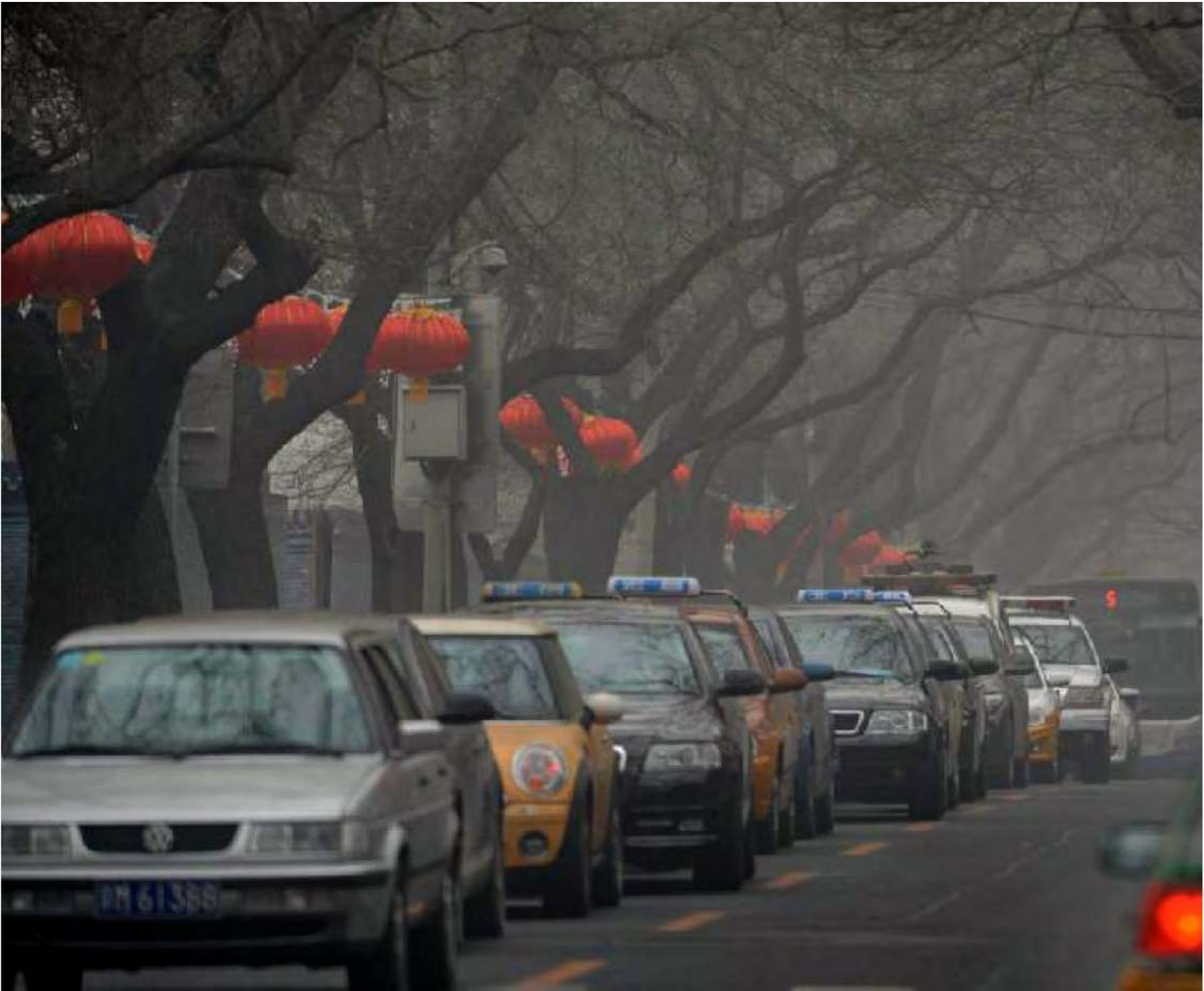}\vspace{2pt}
			\includegraphics[width=1\linewidth]{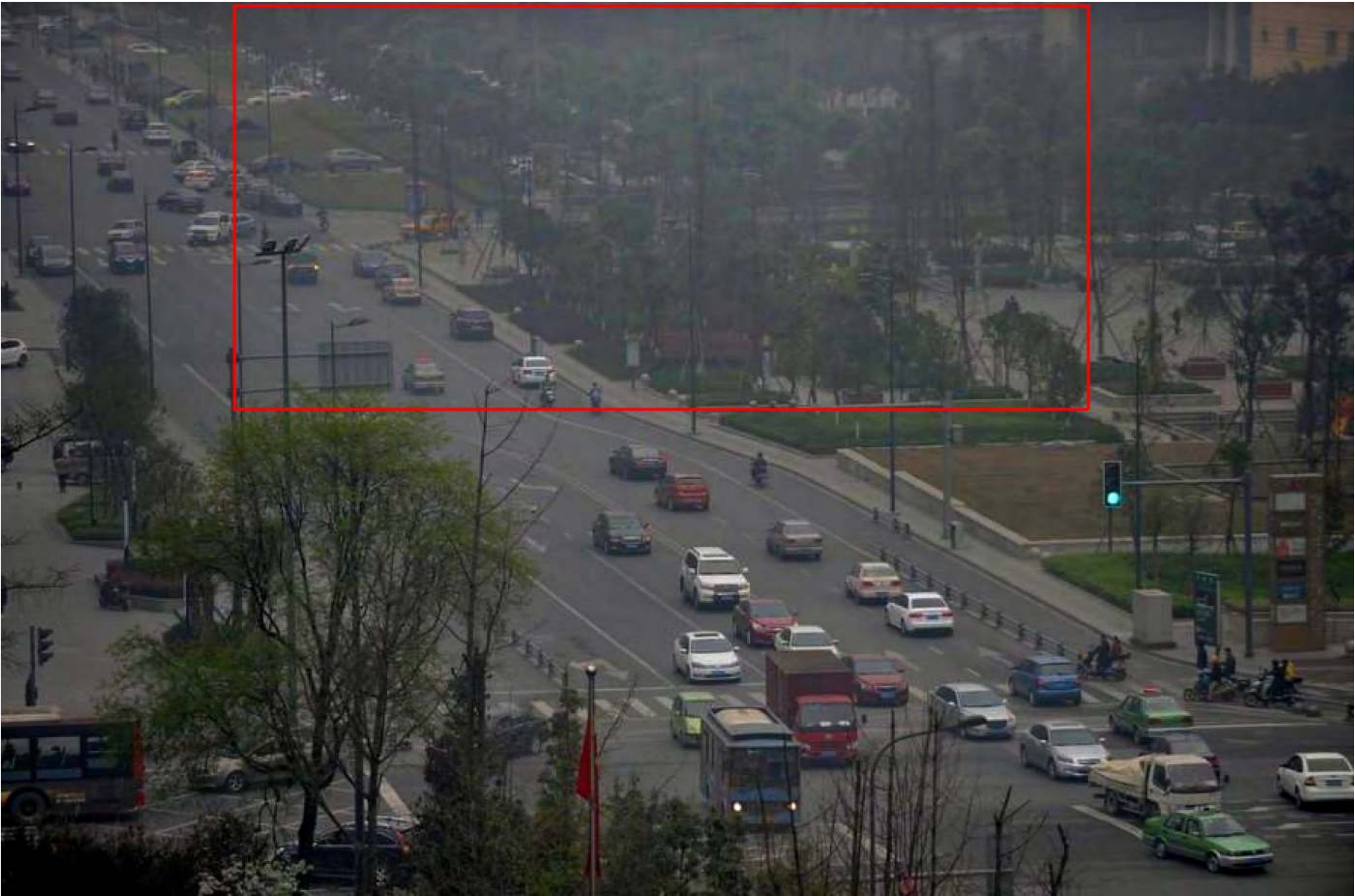}\vspace{2pt}
			\includegraphics[width=1\linewidth]{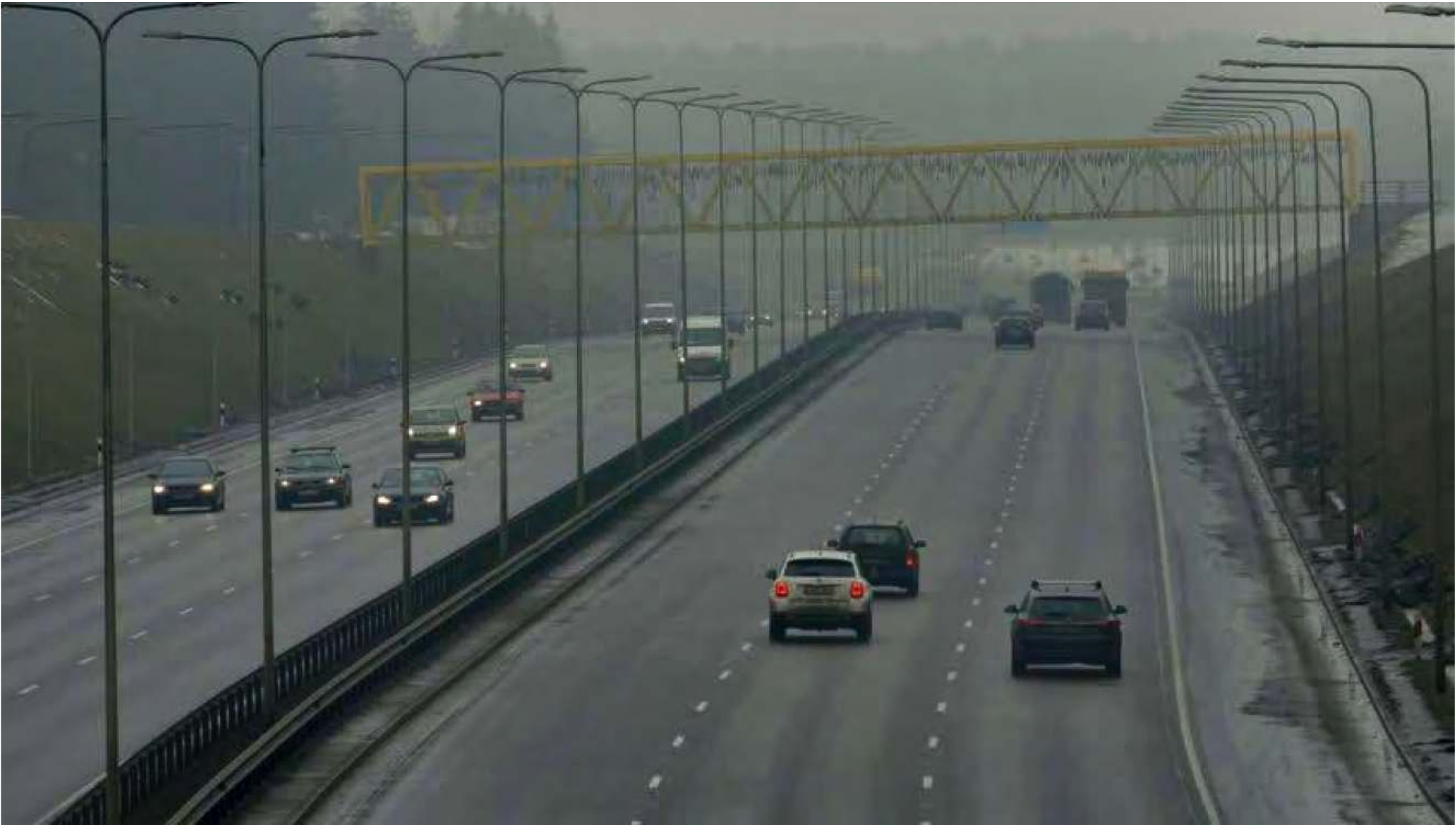}\vspace{2pt}
			\includegraphics[width=1\linewidth]{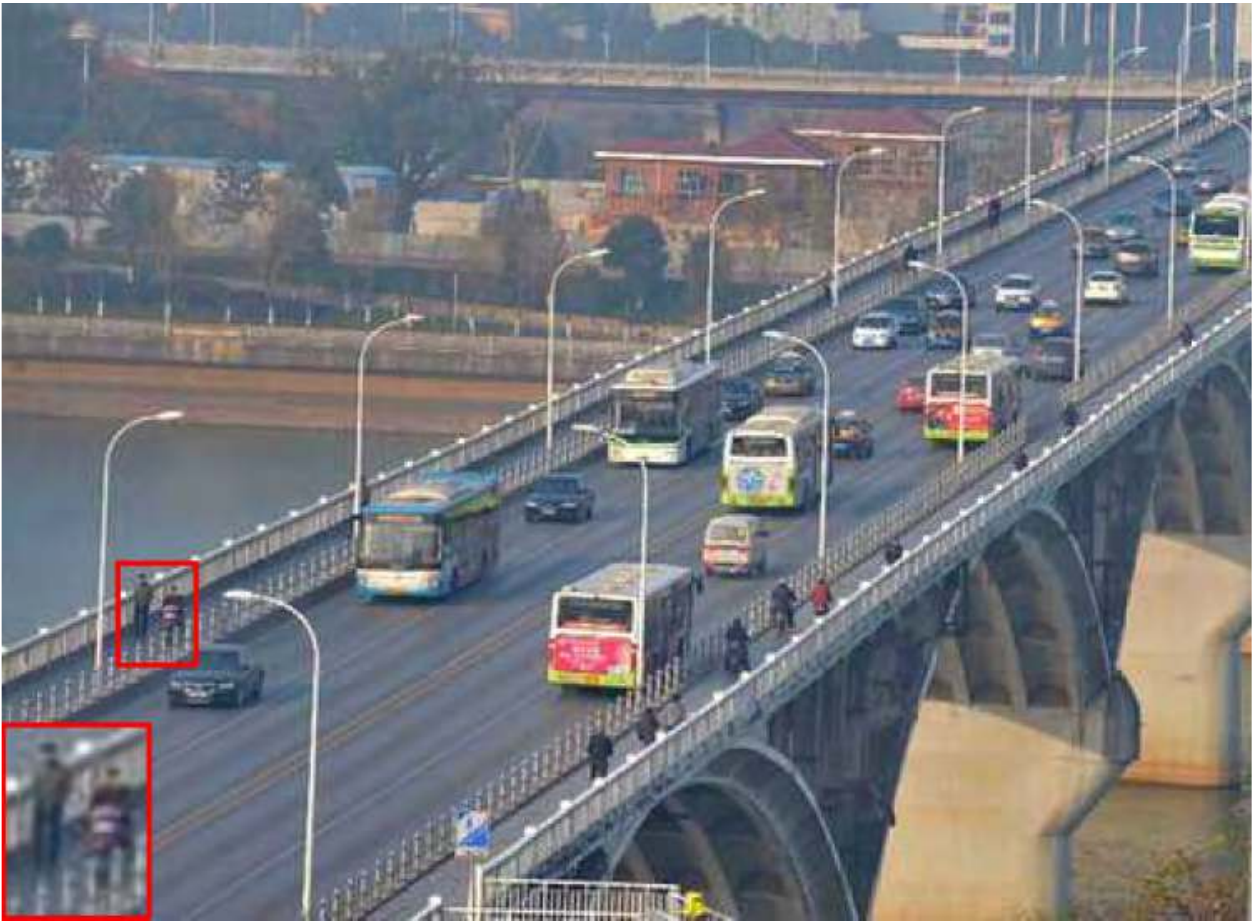}\vspace{2pt}
			\includegraphics[width=1\linewidth]{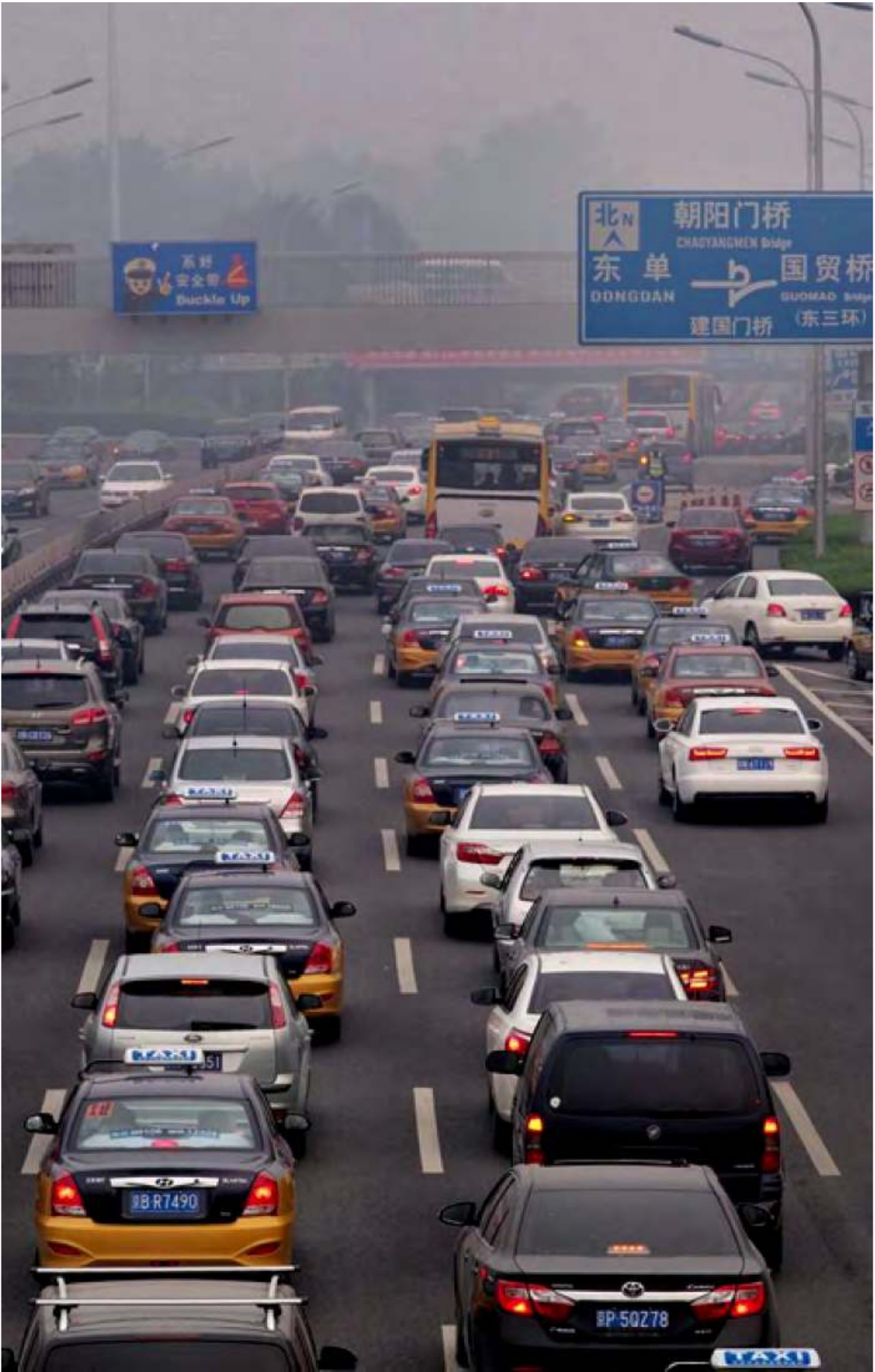}\vspace{2pt}
			\includegraphics[width=1\linewidth]{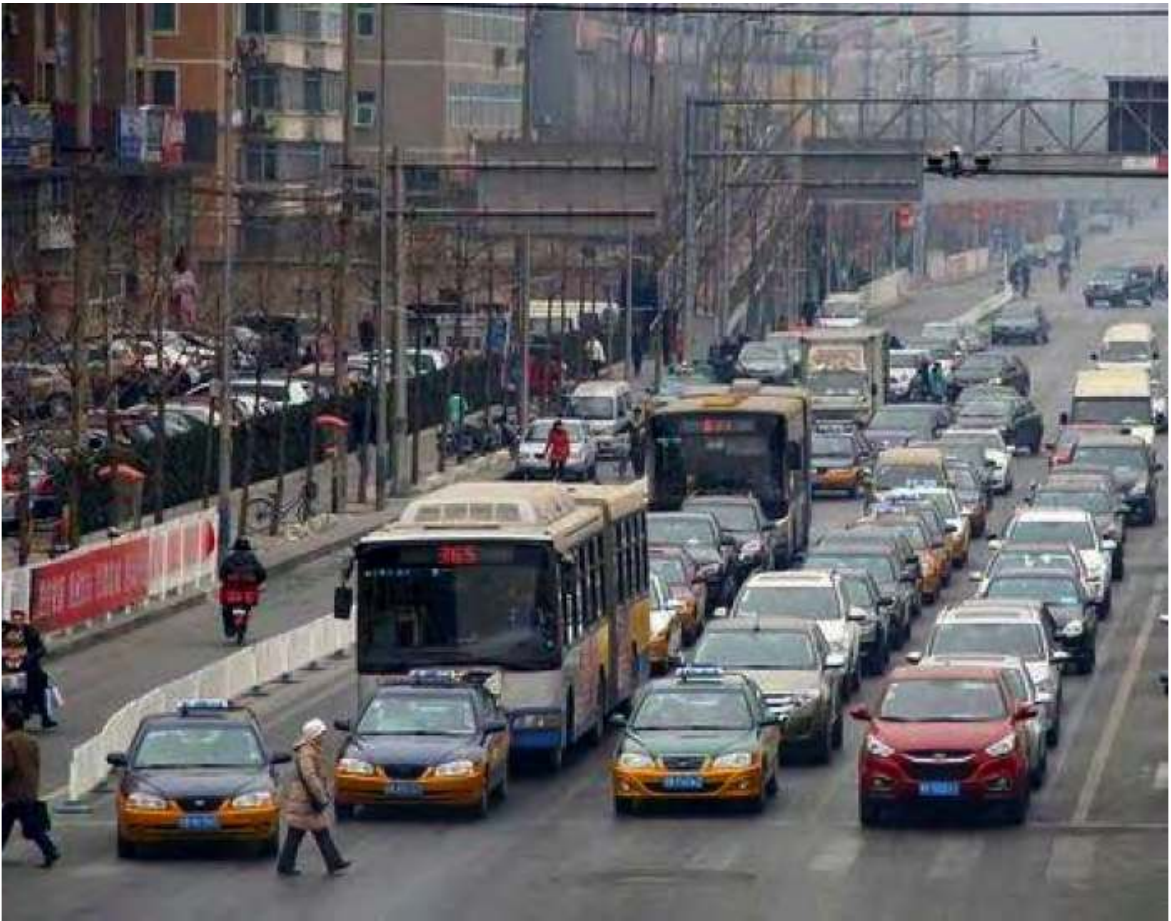}\vspace{2pt}
	\end{minipage}}\hspace{-0.45em}
	\subfigure[\scriptsize{EPDN~\cite{qu2019enhanced}}]{
		\begin{minipage}[b]{0.12\linewidth}
			\includegraphics[width=1\linewidth]{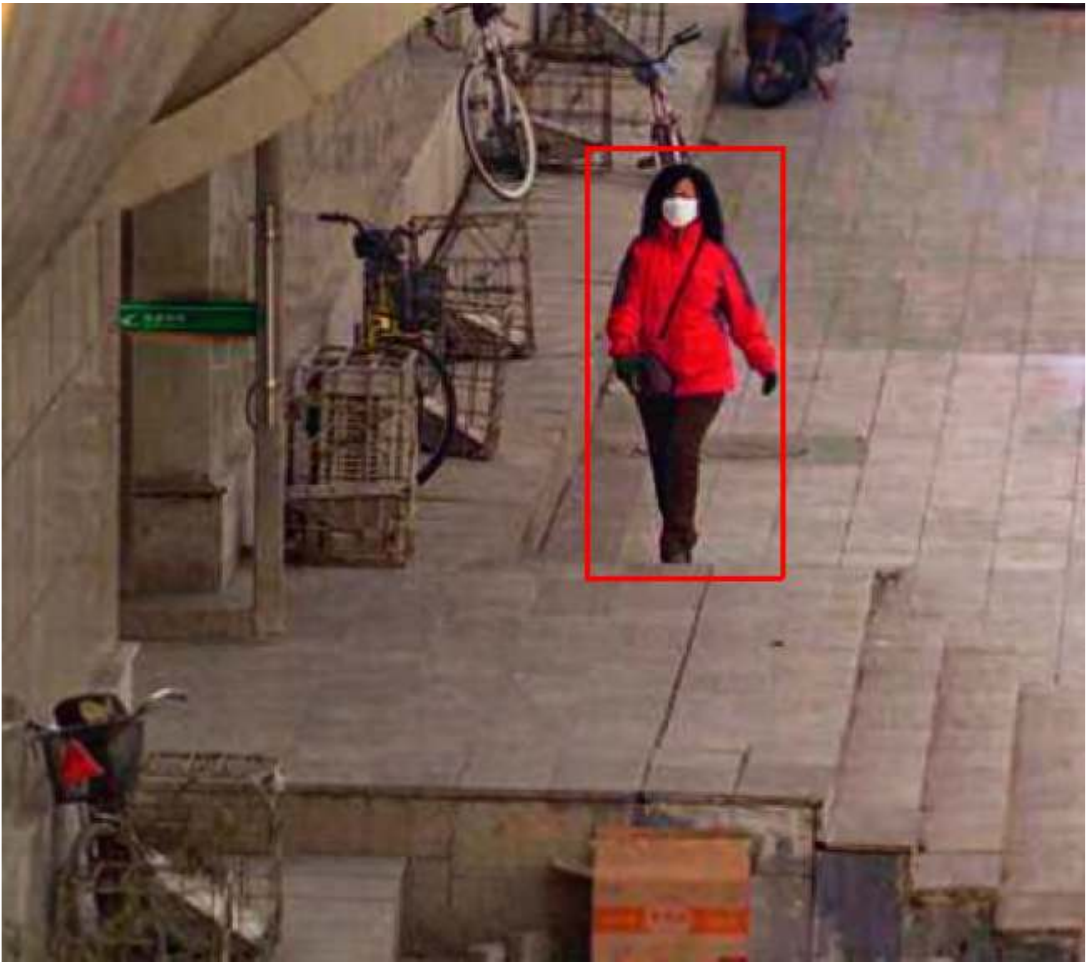}\vspace{2pt}
			\includegraphics[width=1\linewidth]{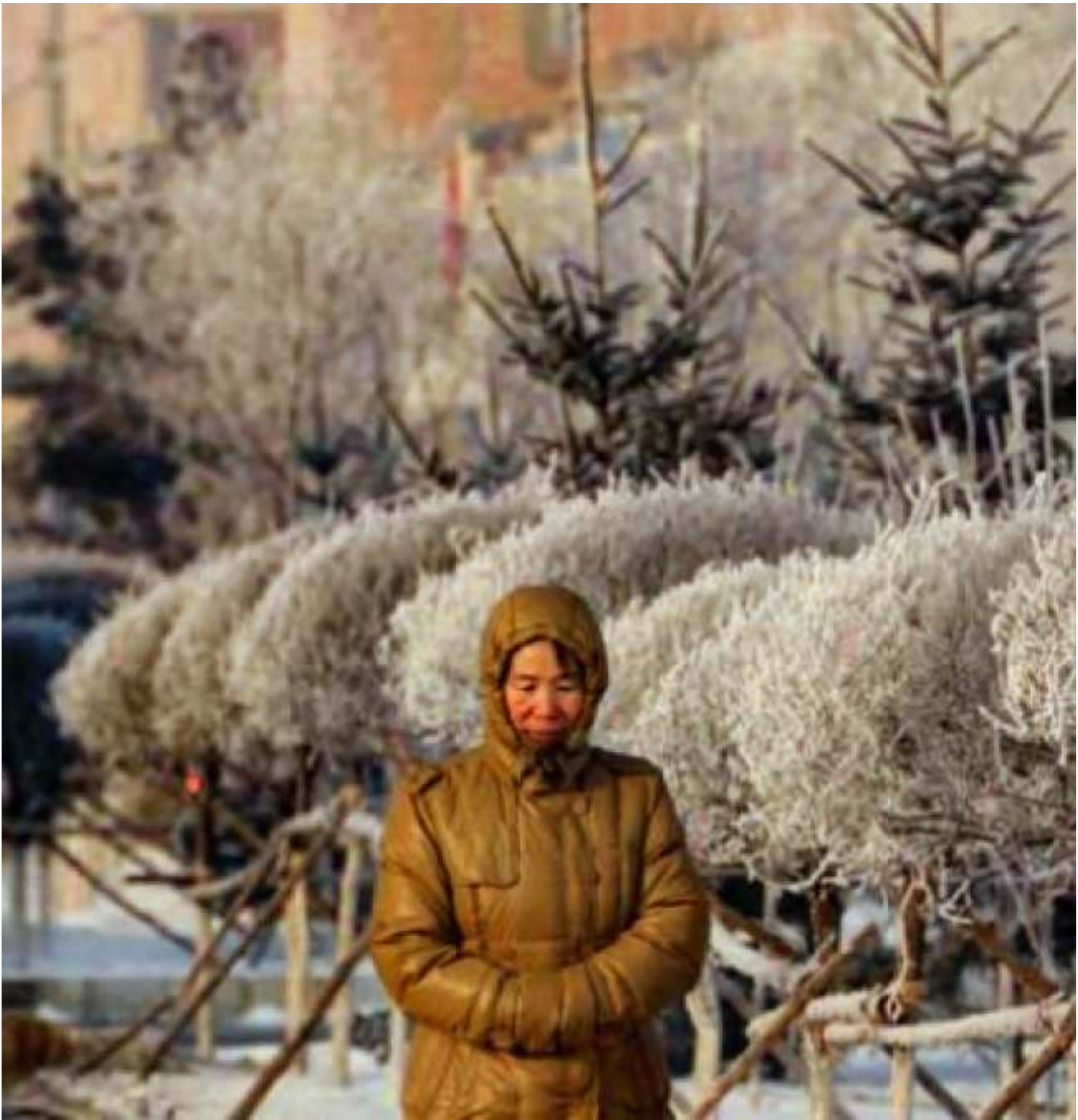}\vspace{2pt}
			\includegraphics[width=1\linewidth]{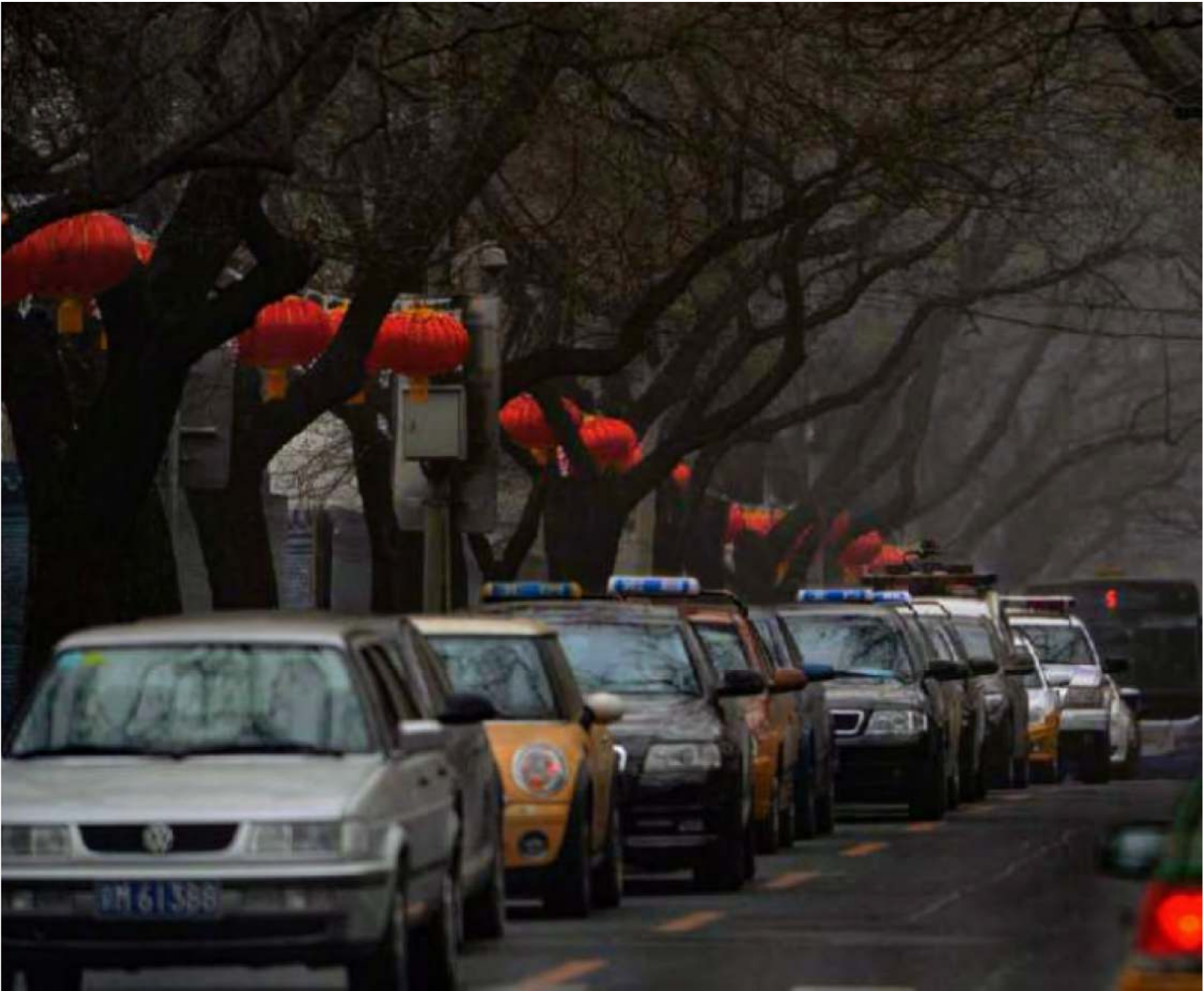}\vspace{2pt}
			\includegraphics[width=1\linewidth]{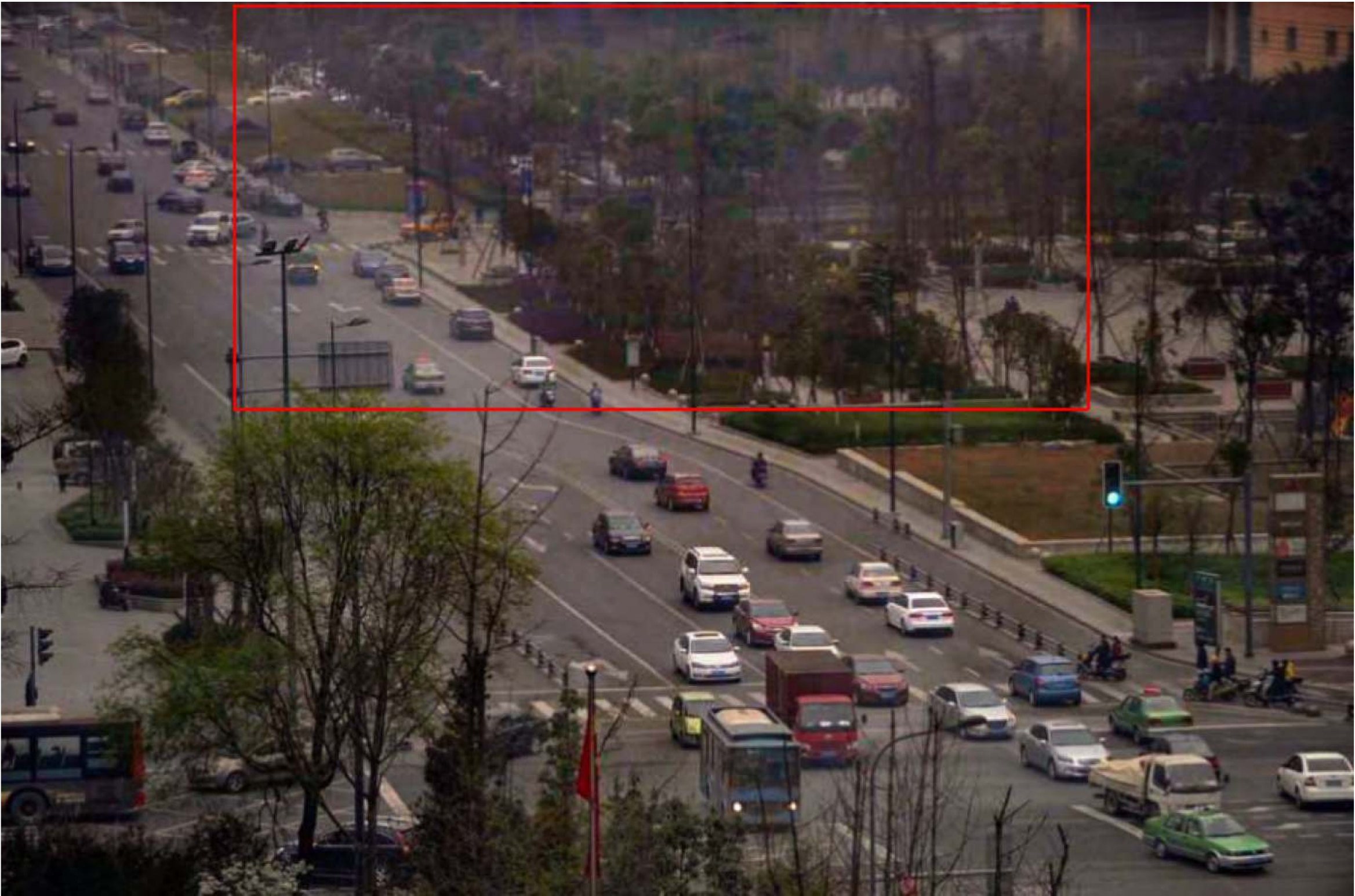}\vspace{2pt}
			\includegraphics[width=1\linewidth]{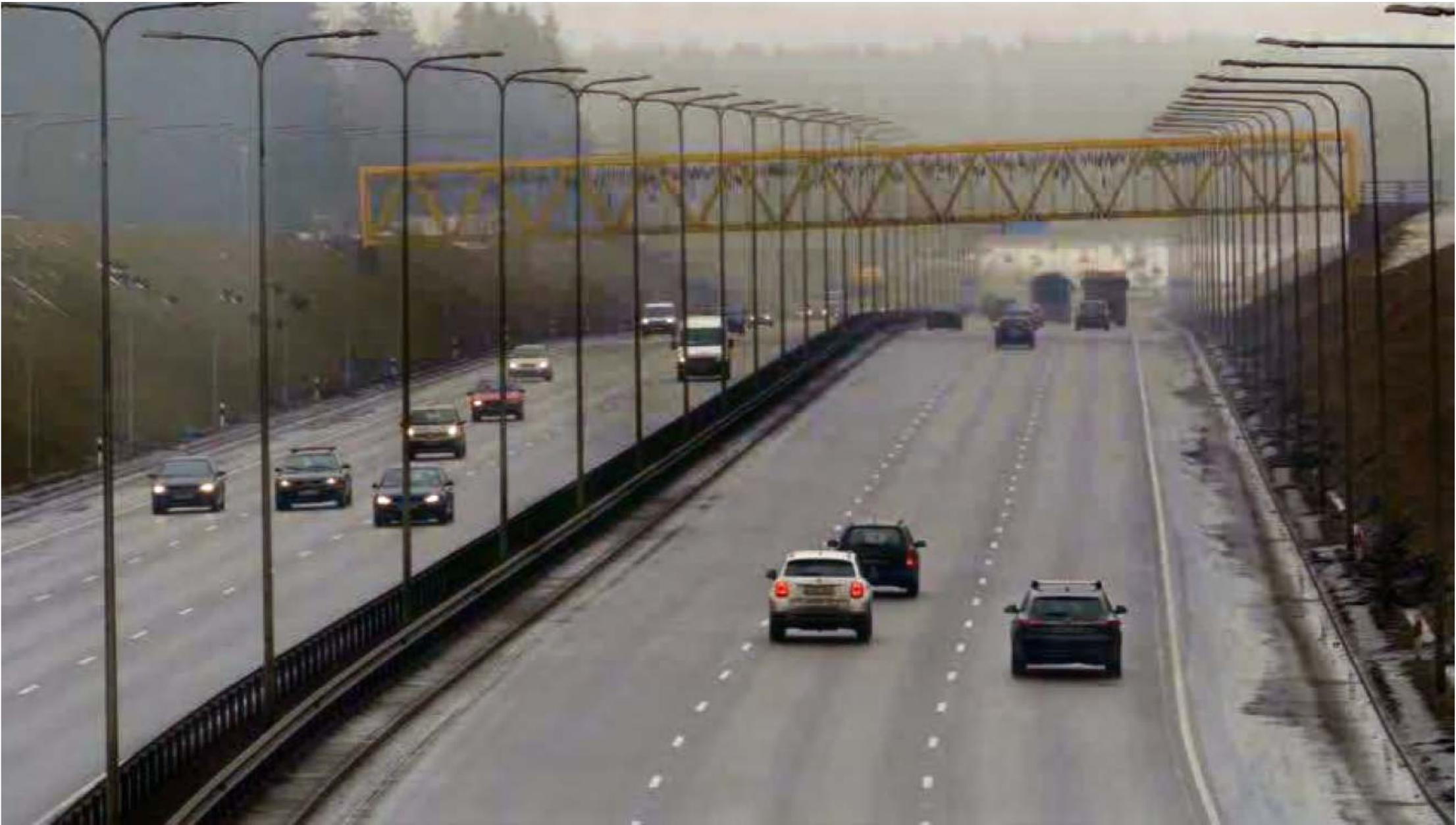}\vspace{2pt}
			\includegraphics[width=1\linewidth]{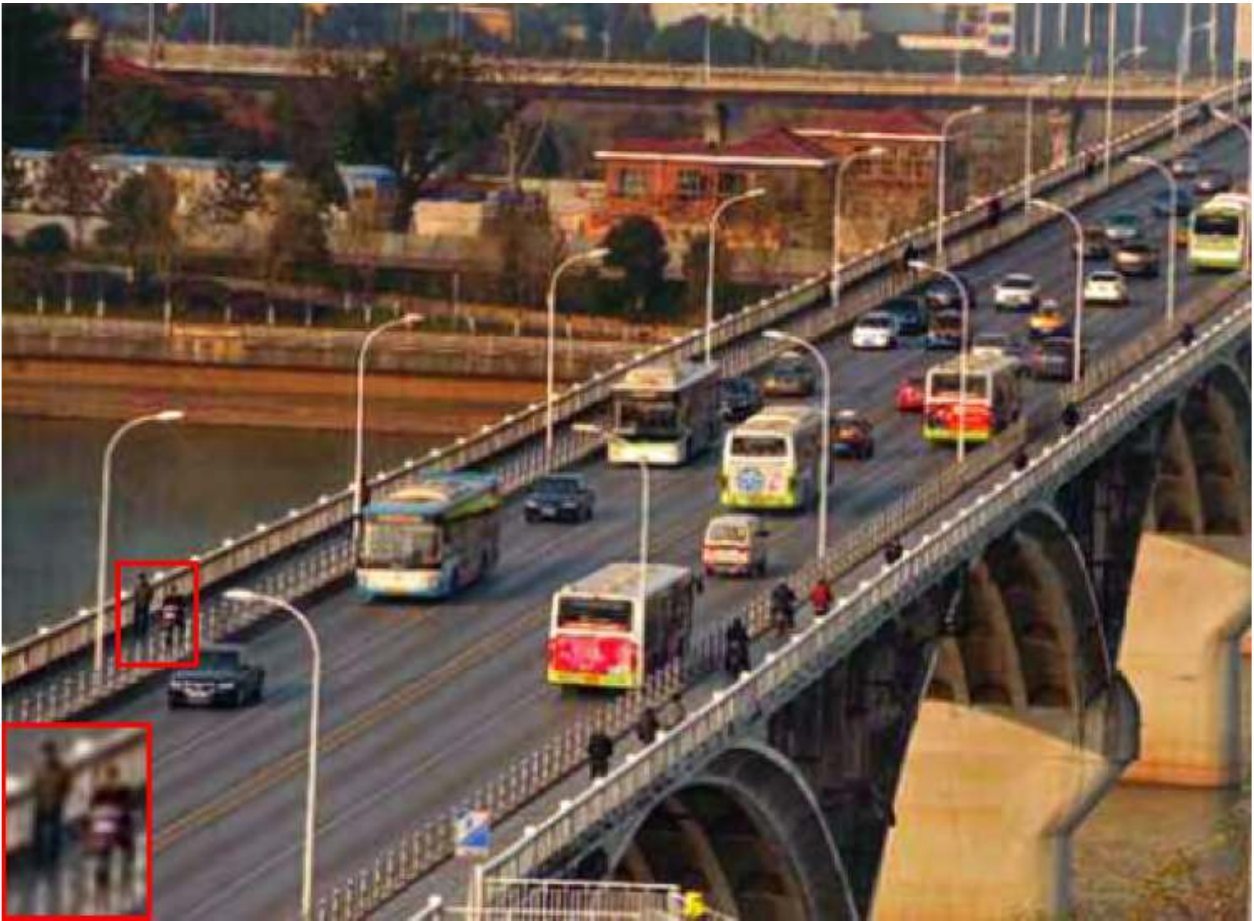}\vspace{2pt}
			\includegraphics[width=1\linewidth]{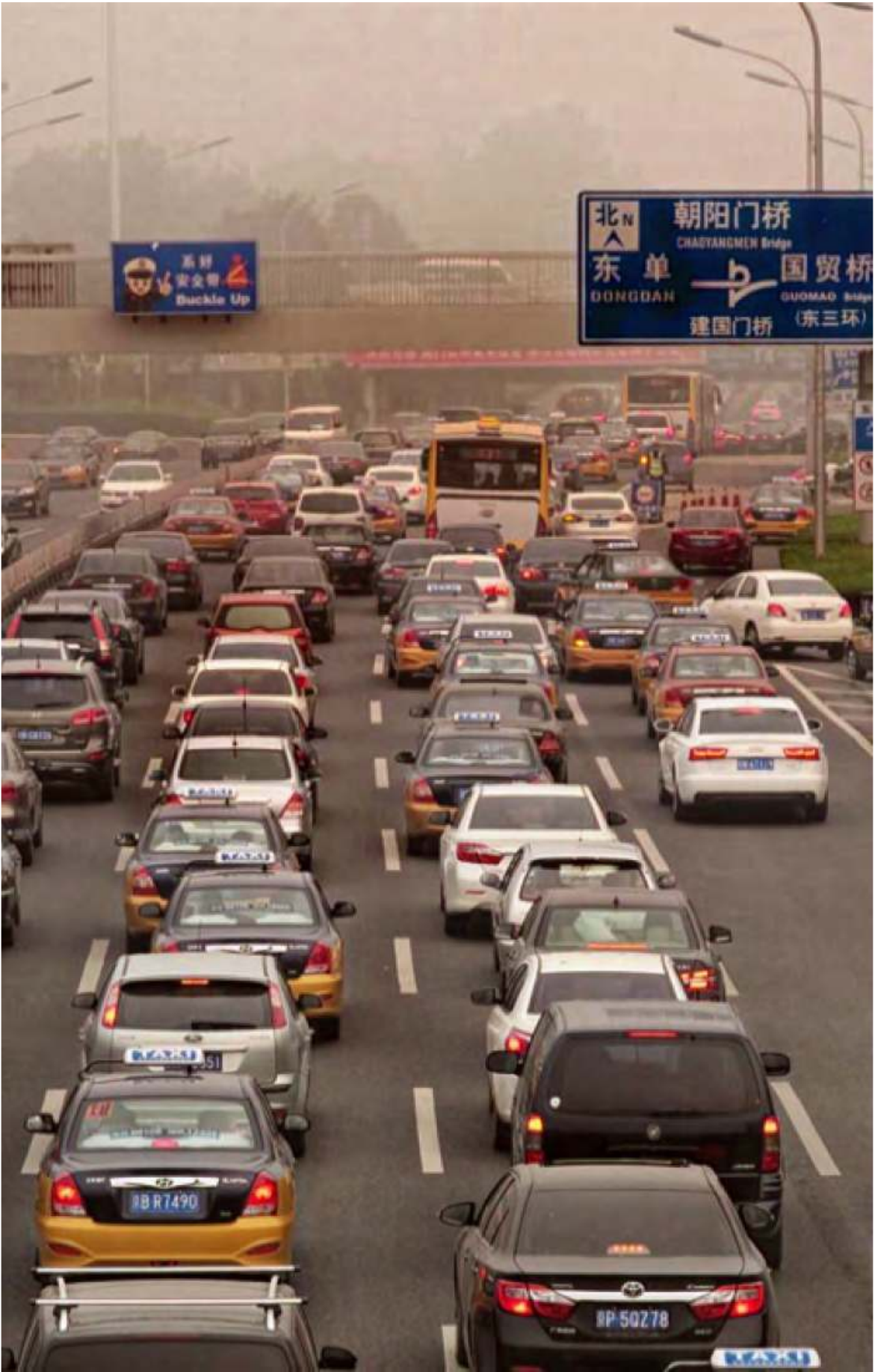}\vspace{2pt}
			\includegraphics[width=1\linewidth]{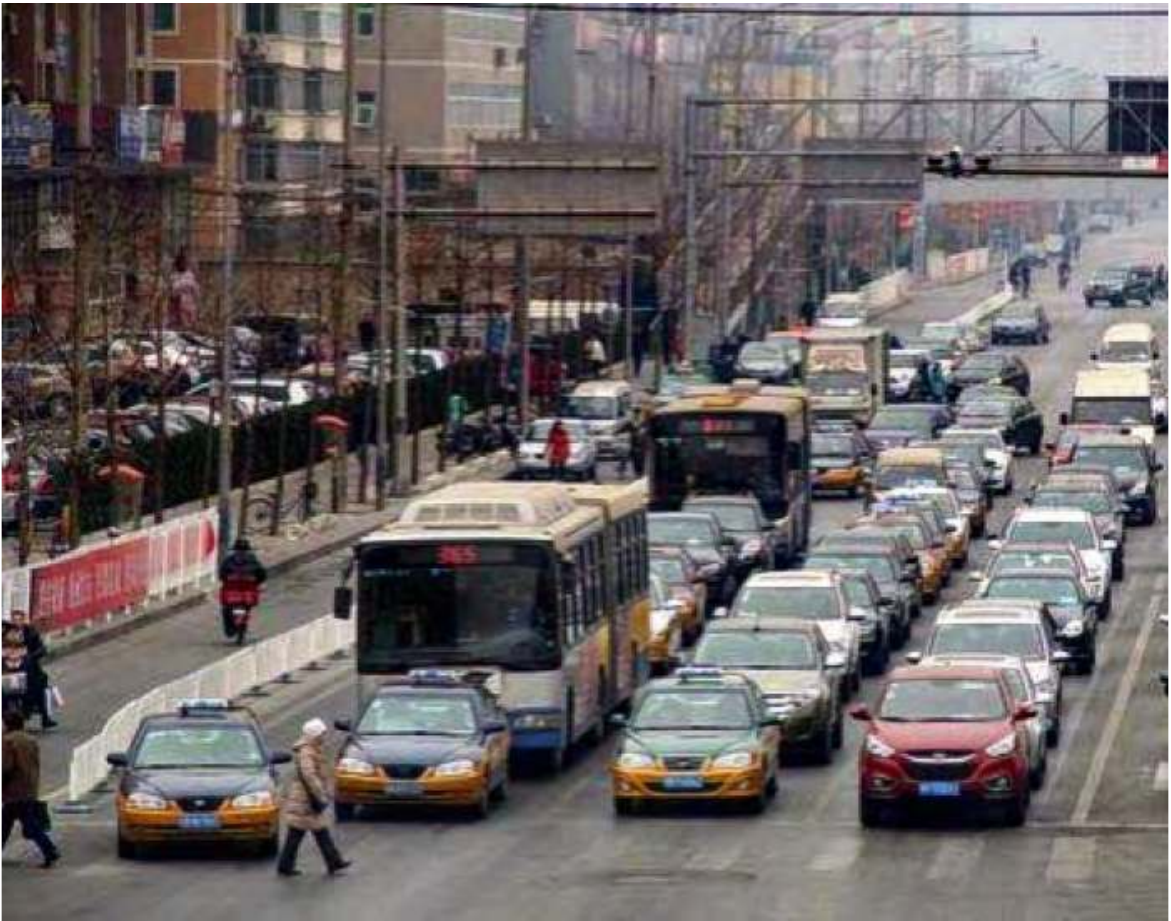}\vspace{2pt}
	\end{minipage}}\hspace{-0.45em}
	\subfigure[\scriptsize{DAdehazing~\cite{shao2020domain}}]{
		\begin{minipage}[b]{0.12\linewidth}
			\includegraphics[width=1\linewidth]{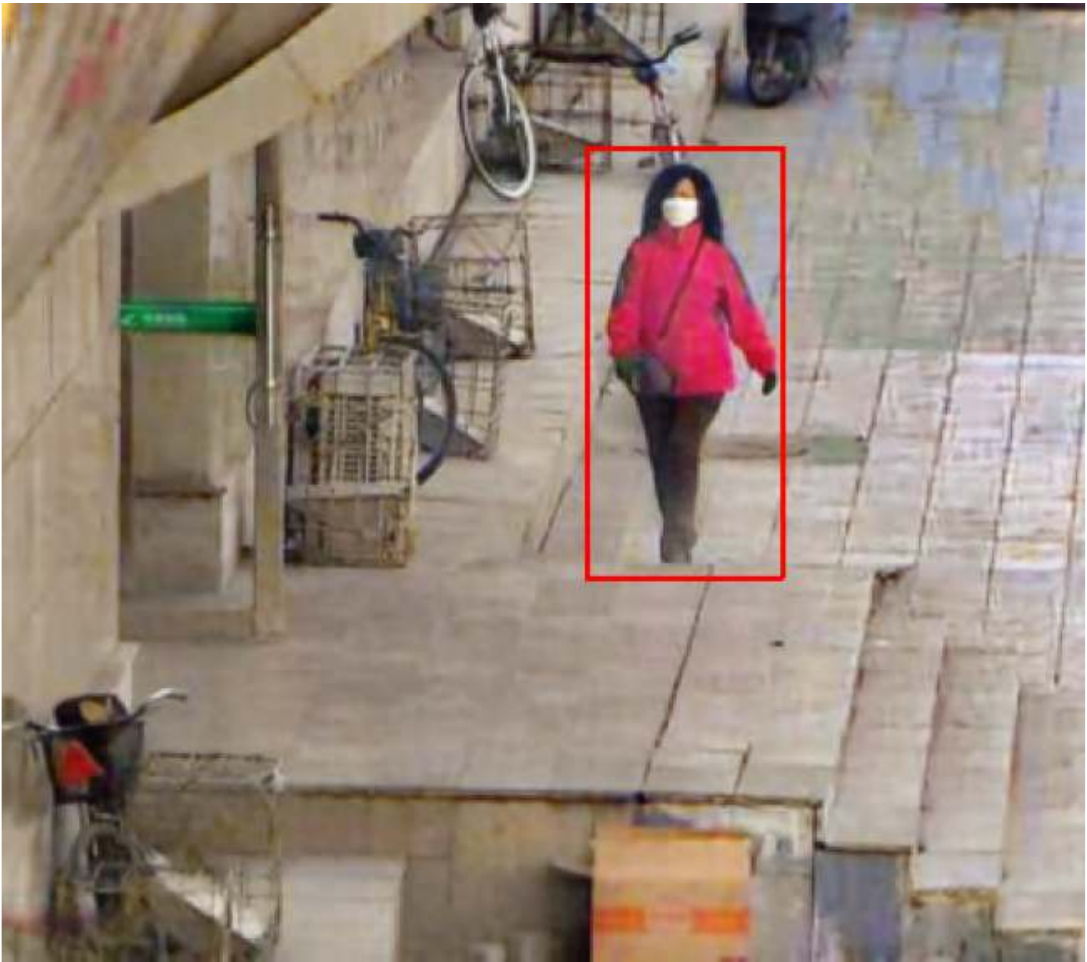}\vspace{2pt}
			\includegraphics[width=1\linewidth]{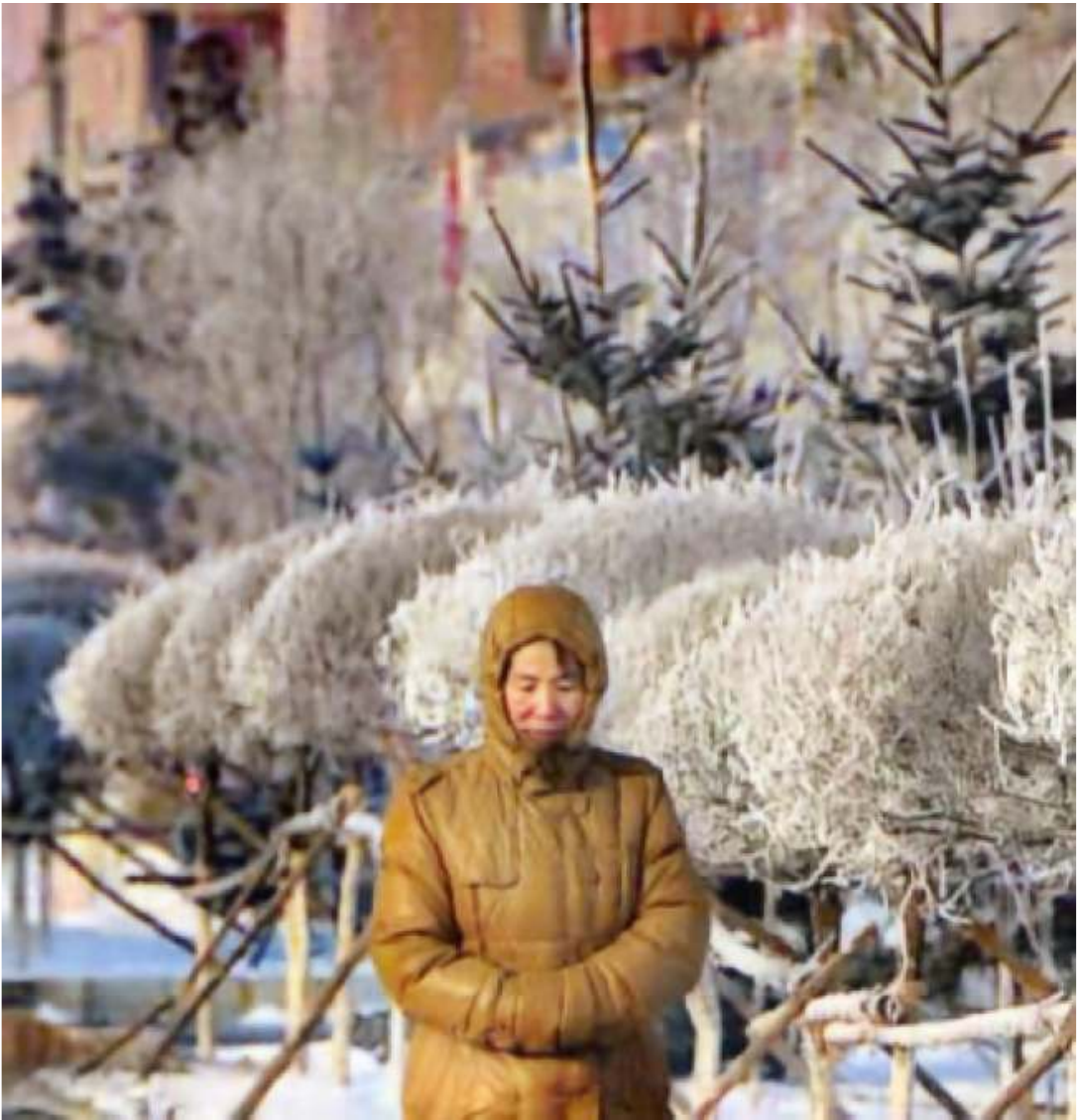}\vspace{2pt}
			\includegraphics[width=1\linewidth]{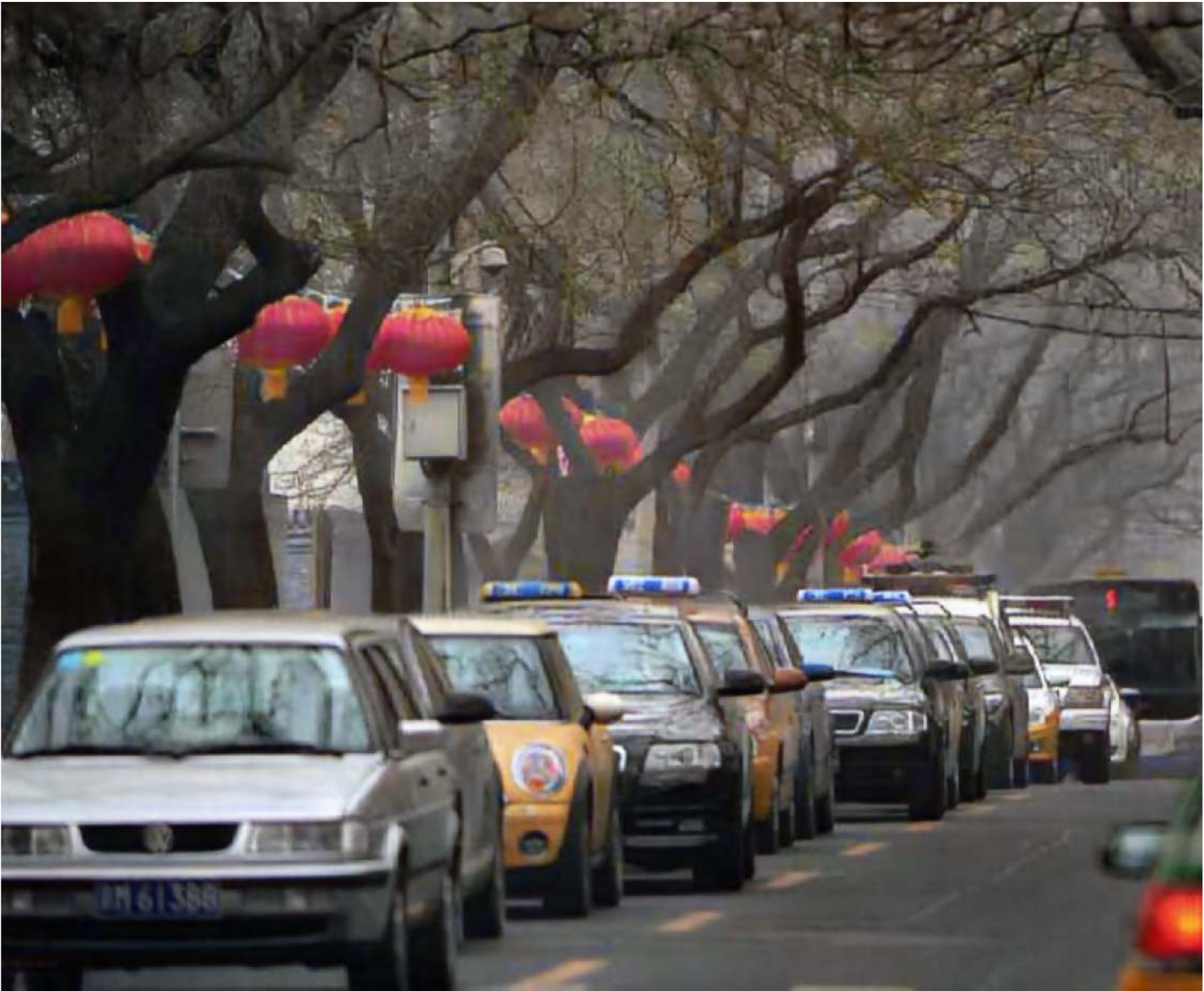}\vspace{2pt}
			\includegraphics[width=1\linewidth]{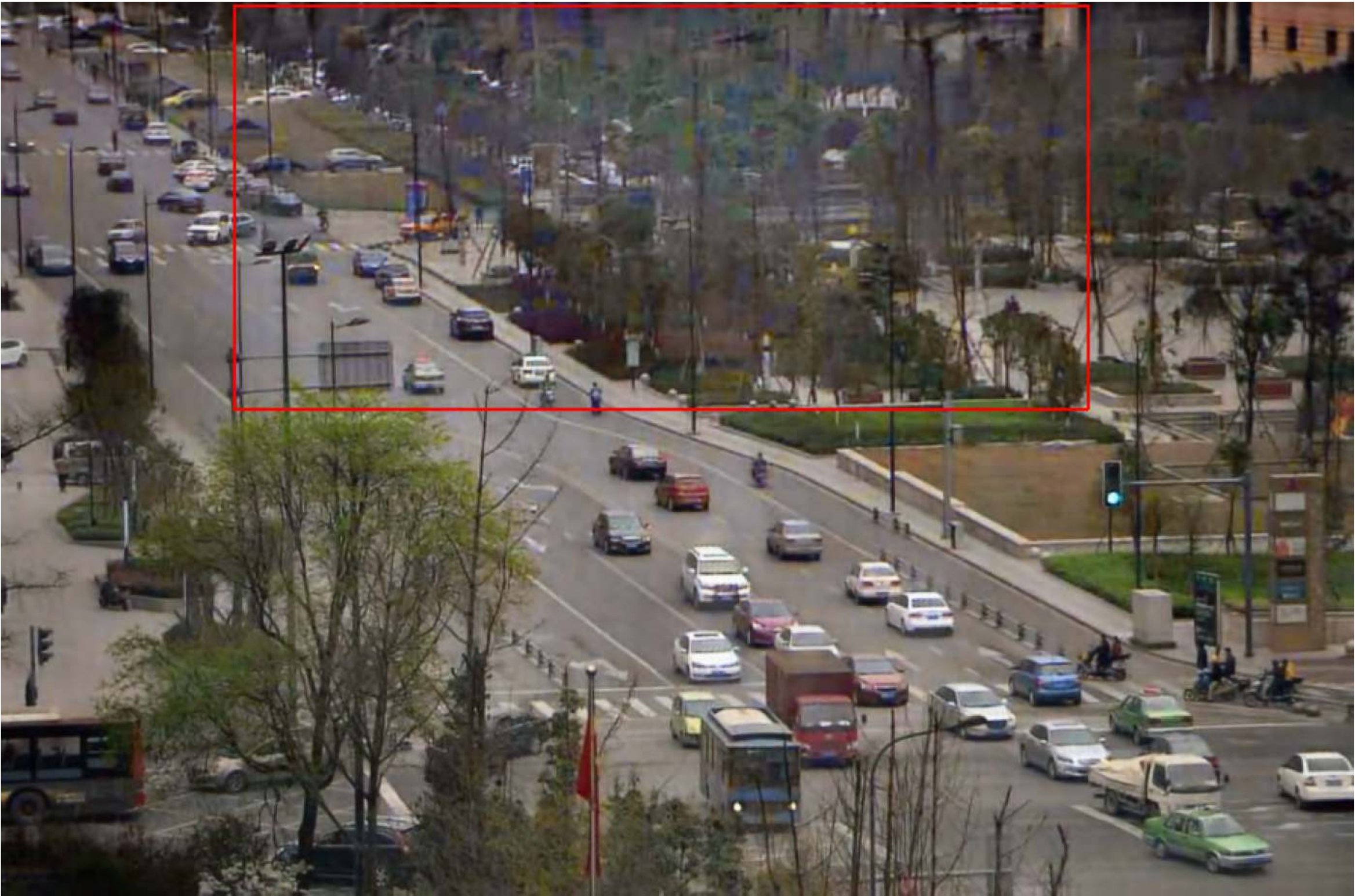}\vspace{2pt}
			\includegraphics[width=1\linewidth]{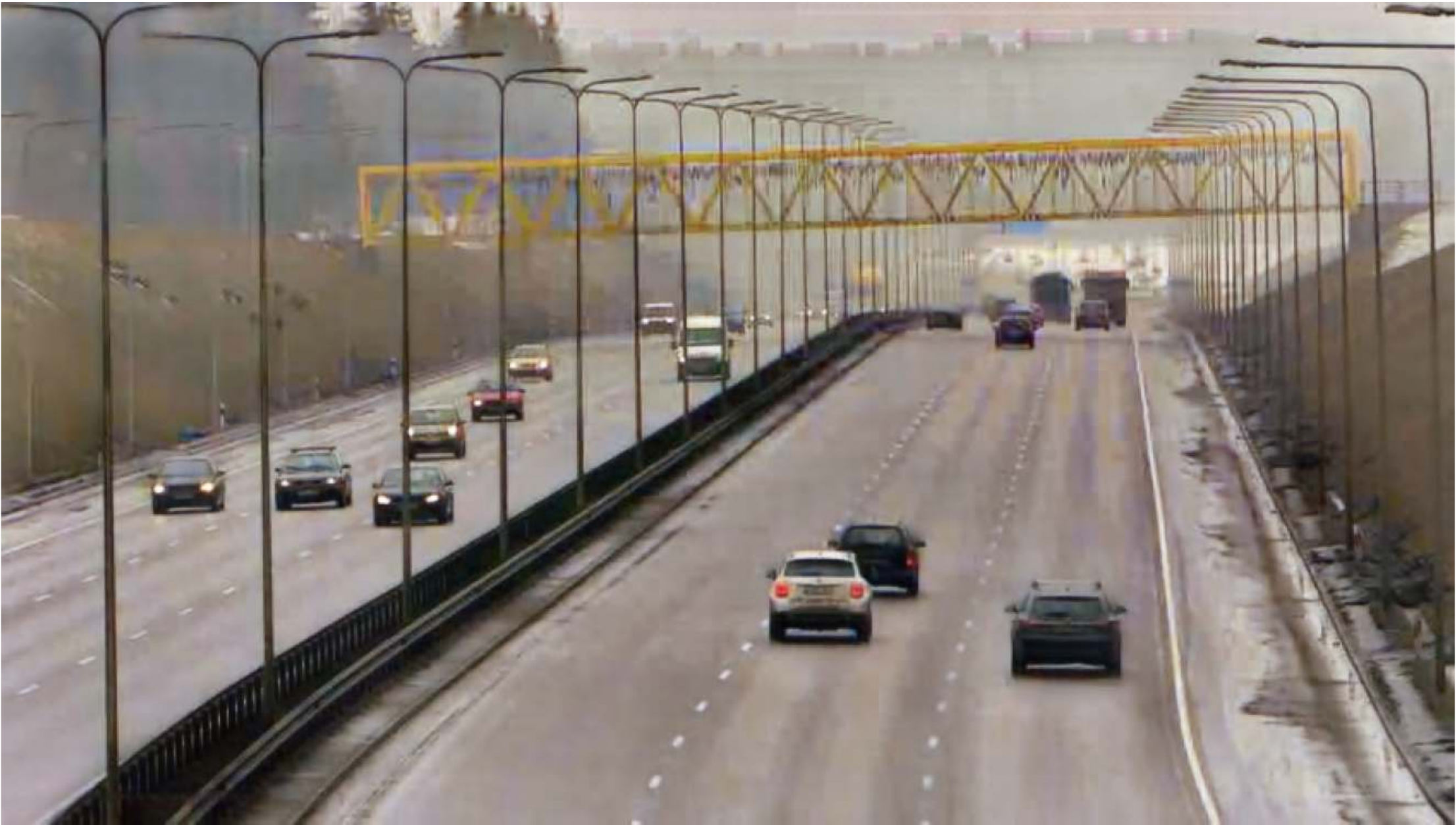}\vspace{2pt}
			\includegraphics[width=1\linewidth]{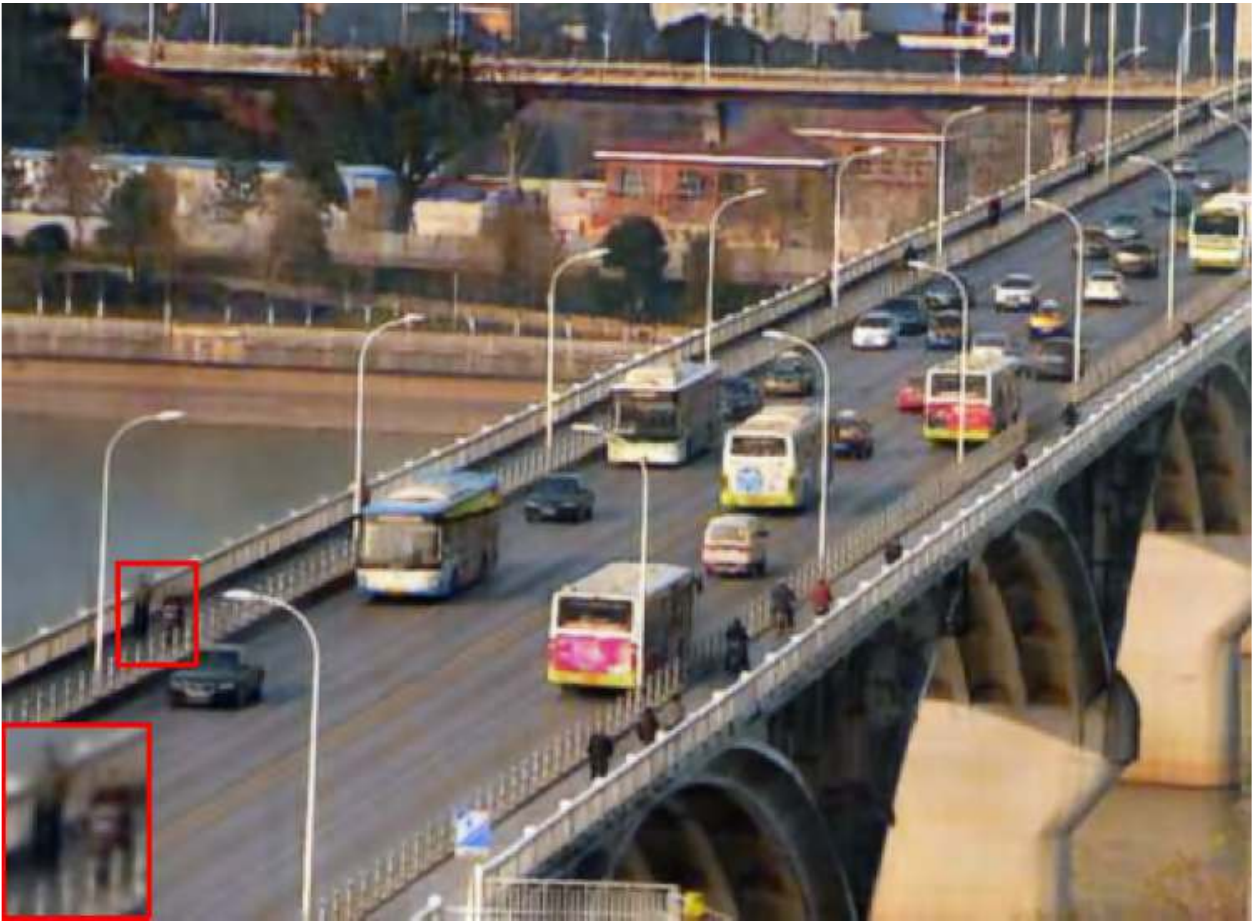}\vspace{2pt}
			\includegraphics[width=1\linewidth]{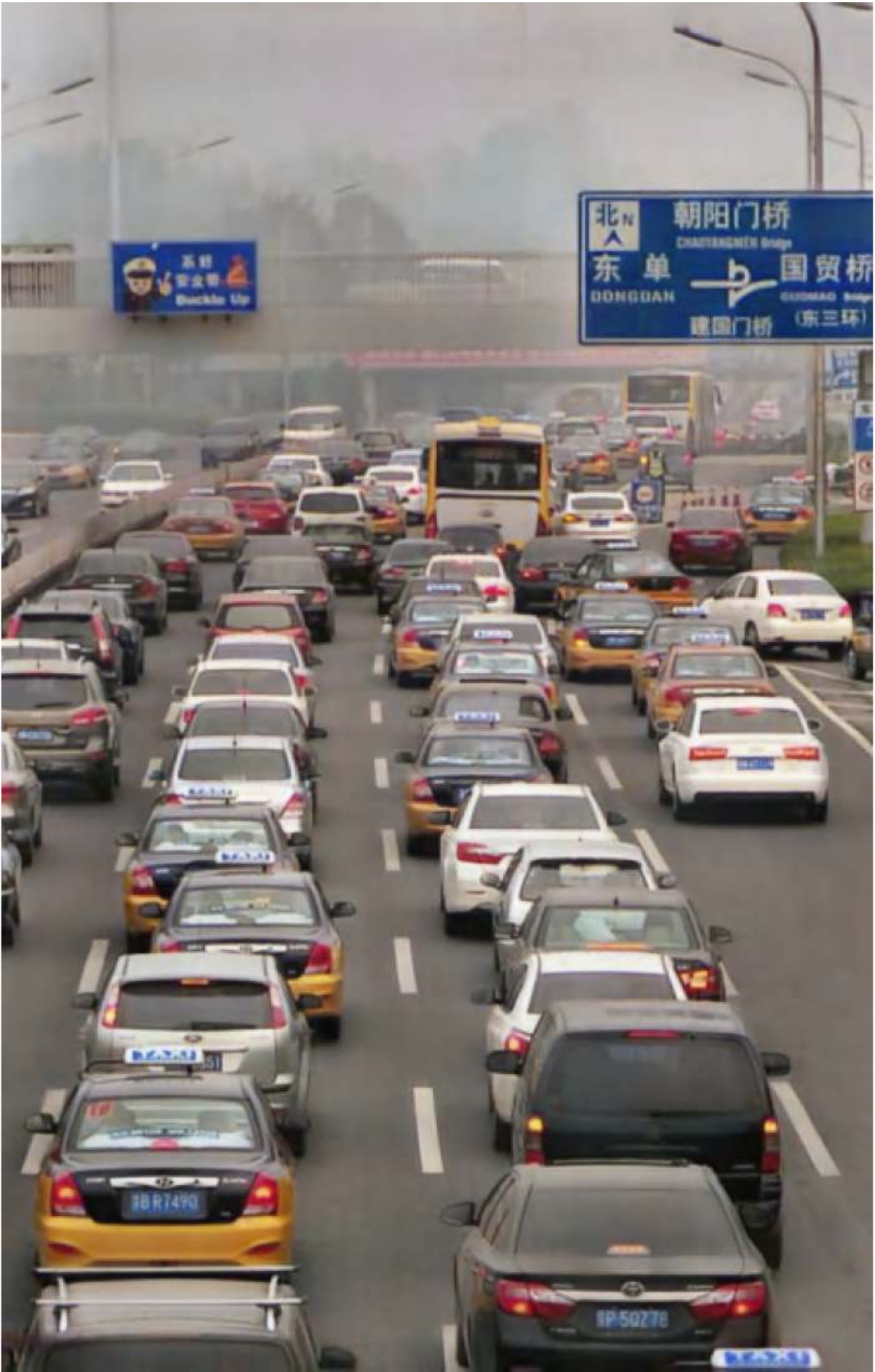}\vspace{2pt}
			\includegraphics[width=1\linewidth]{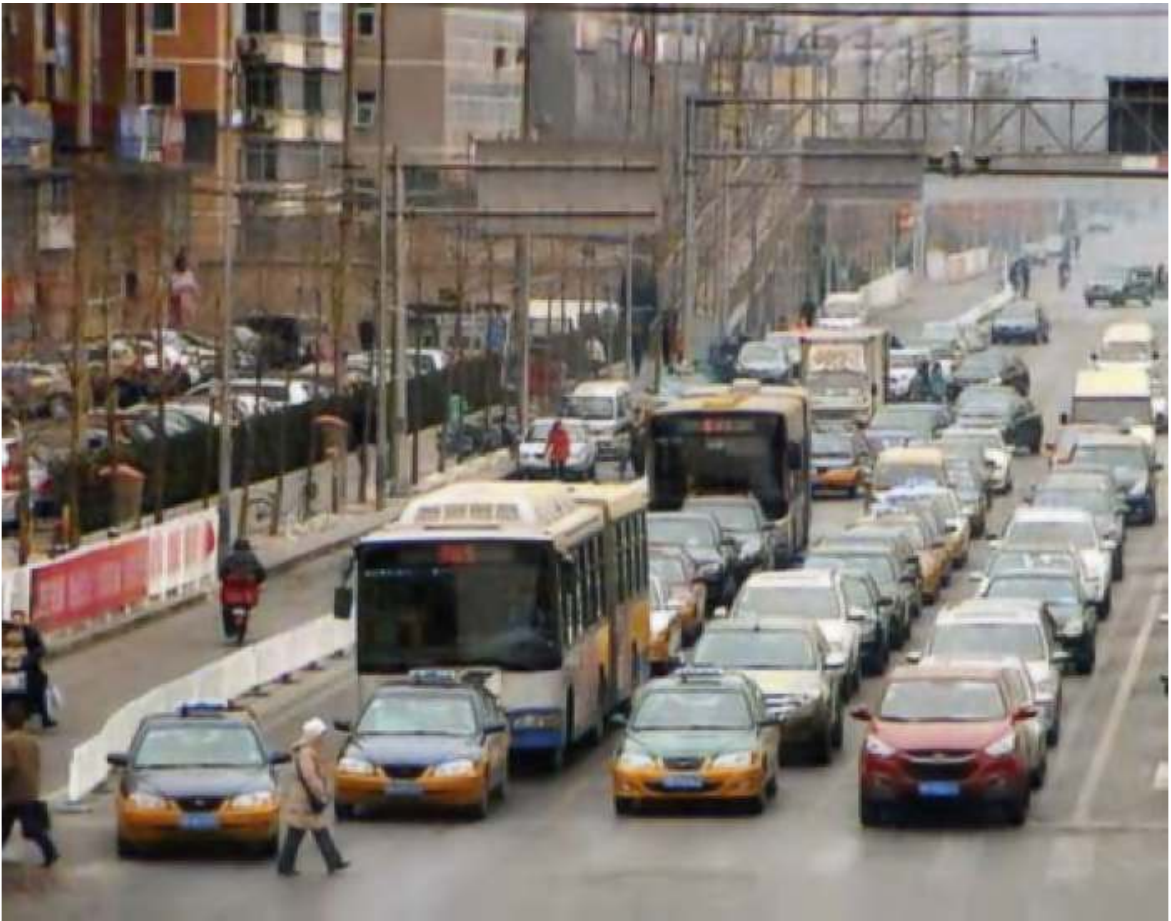}\vspace{2pt}
	\end{minipage}}\hspace{-0.45em}
	\subfigure[\scriptsize{FFANet~\cite{qin2020ffa}}]{
		\begin{minipage}[b]{0.12\linewidth}
			\includegraphics[width=1\linewidth]{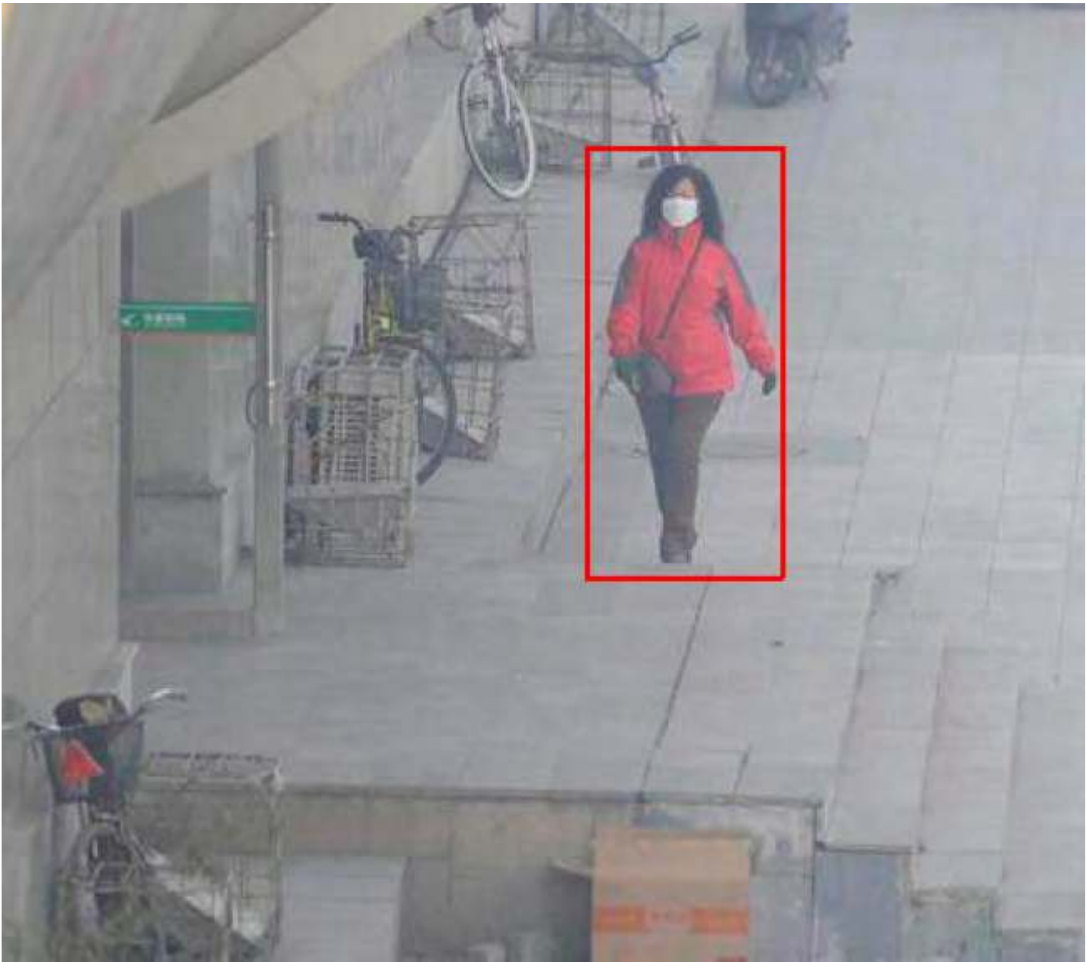}\vspace{2pt}
			\includegraphics[width=1\linewidth]{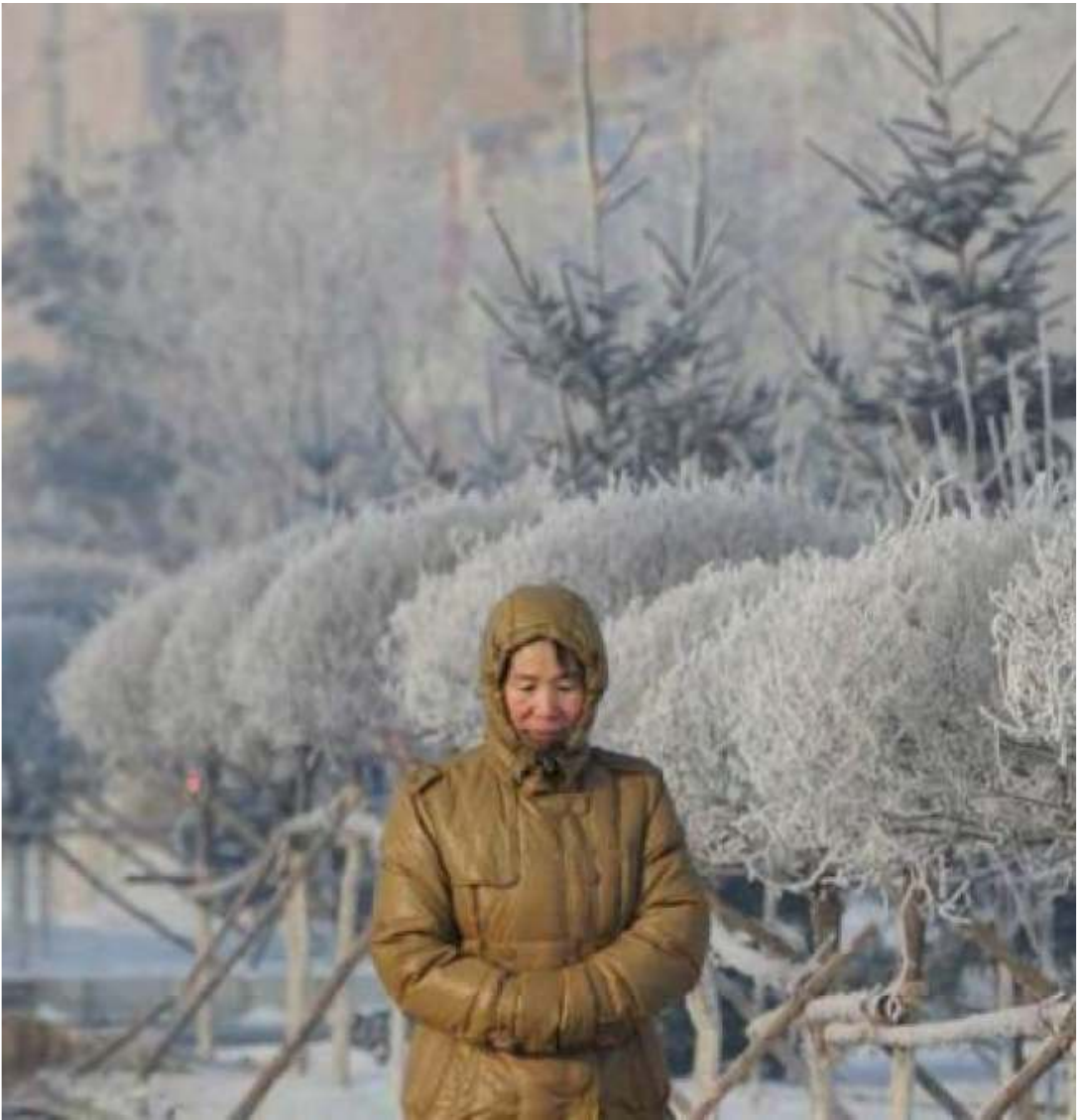}\vspace{2pt}
			\includegraphics[width=1\linewidth]{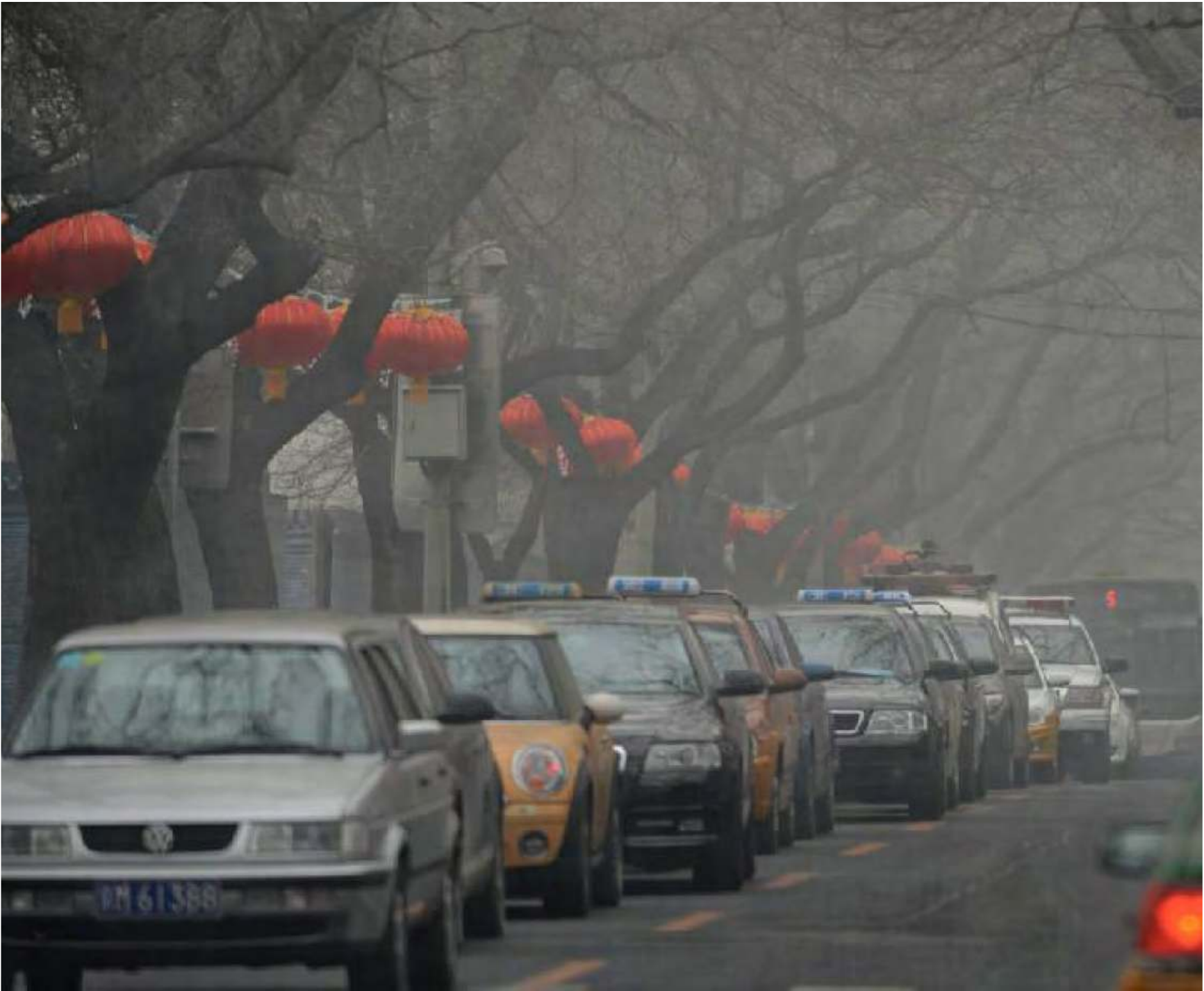}\vspace{2pt}
			\includegraphics[width=1\linewidth]{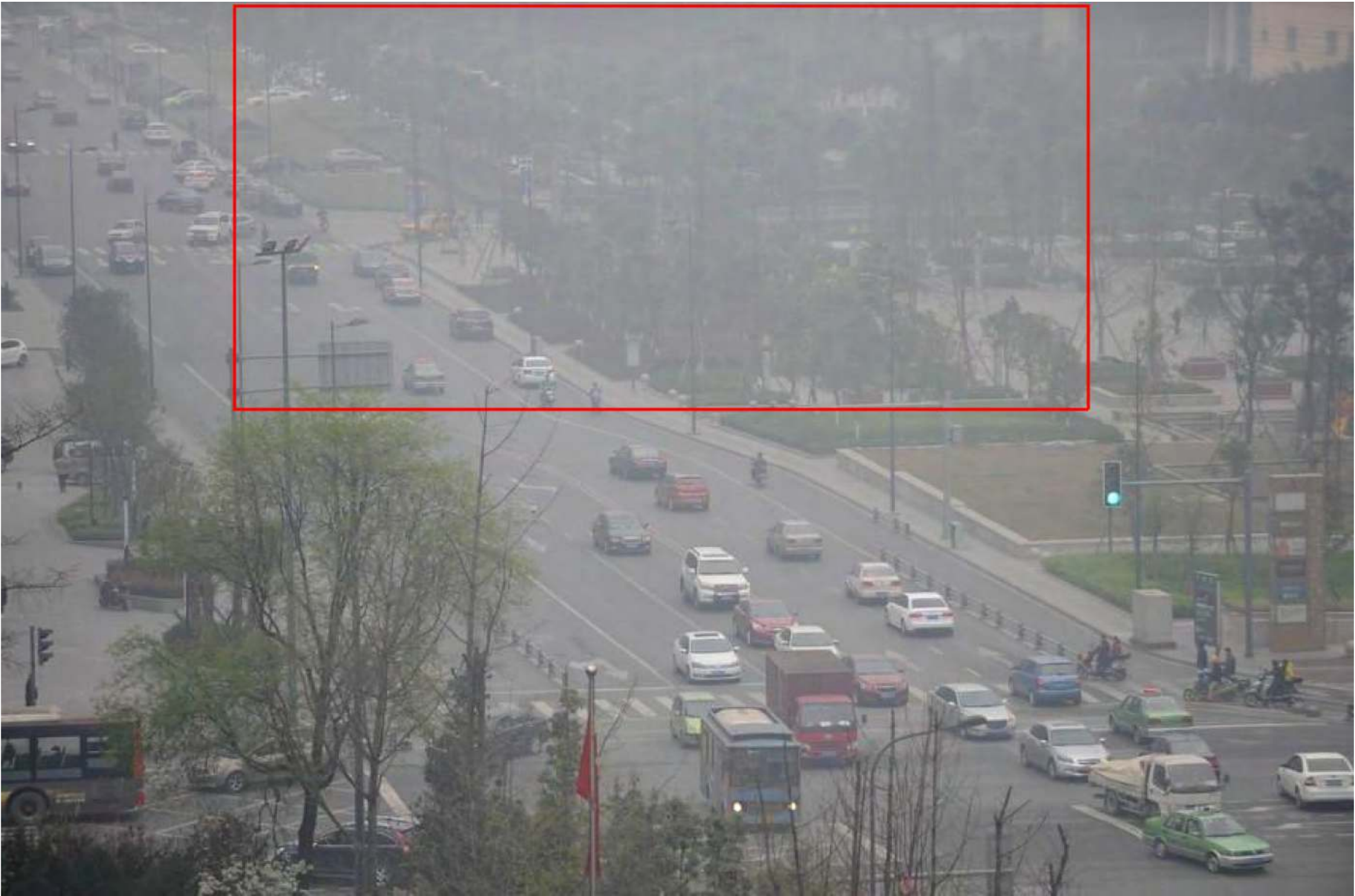}\vspace{2pt}
			\includegraphics[width=1\linewidth]{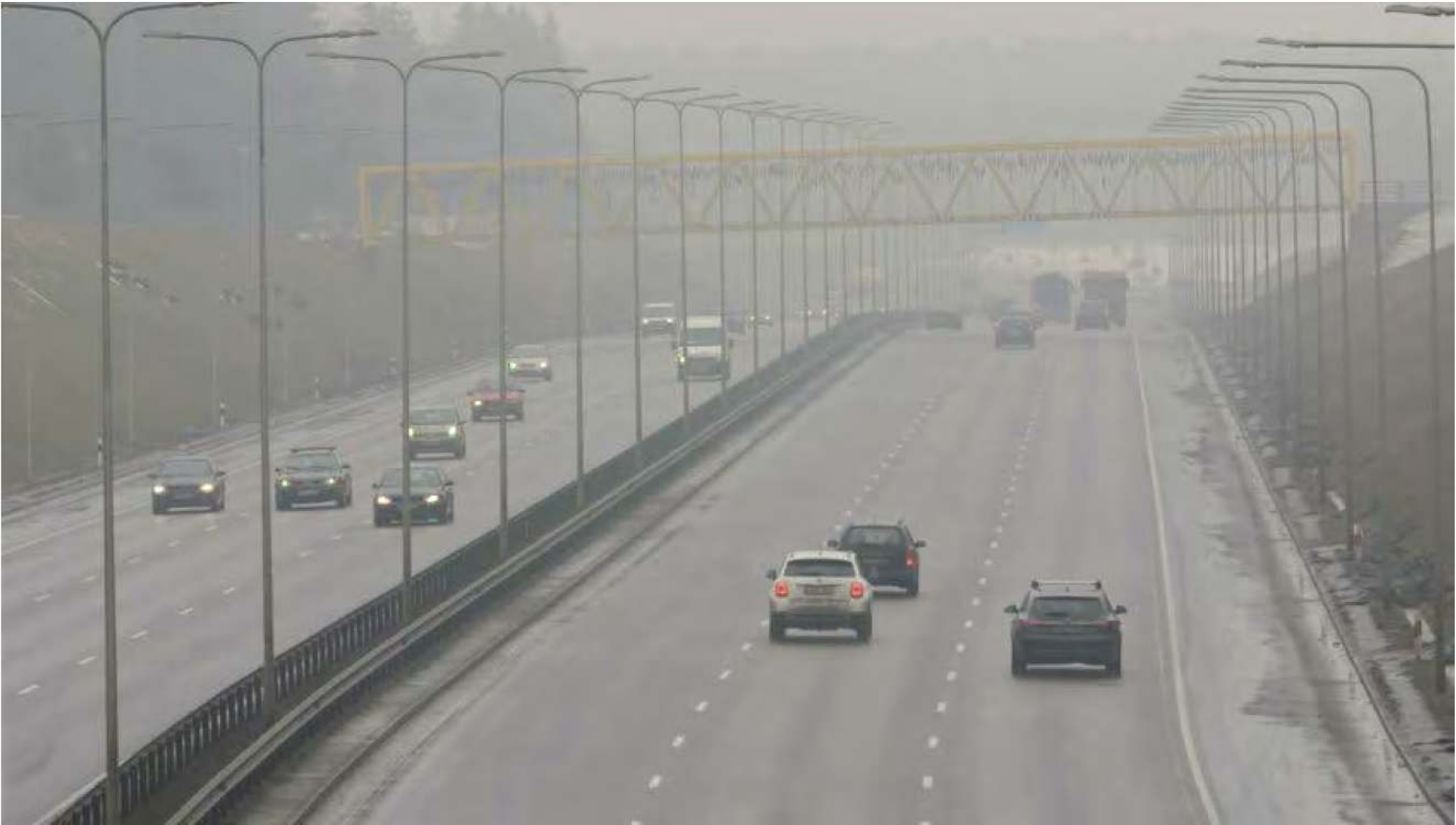}\vspace{2pt}
			\includegraphics[width=1\linewidth]{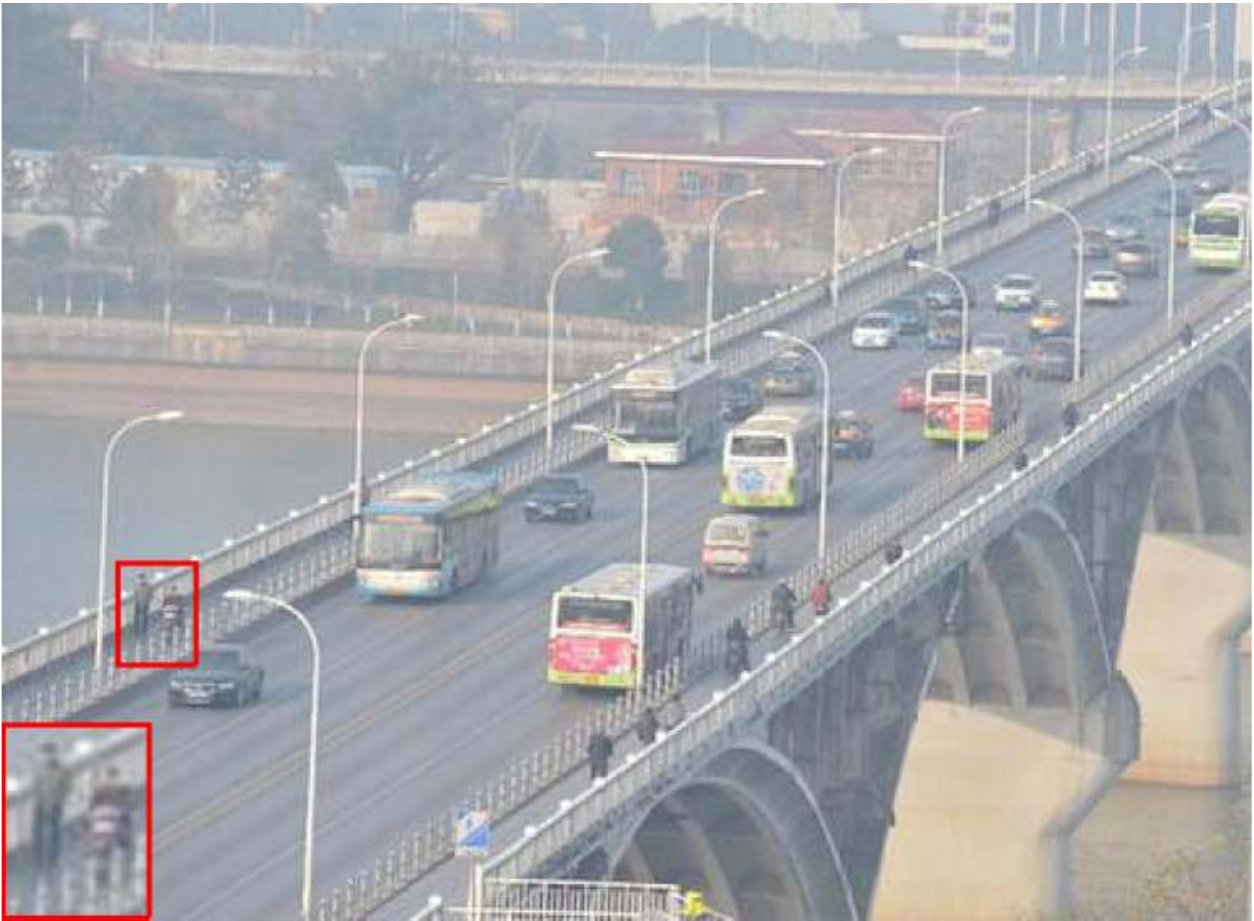}\vspace{2pt}
			\includegraphics[width=1\linewidth]{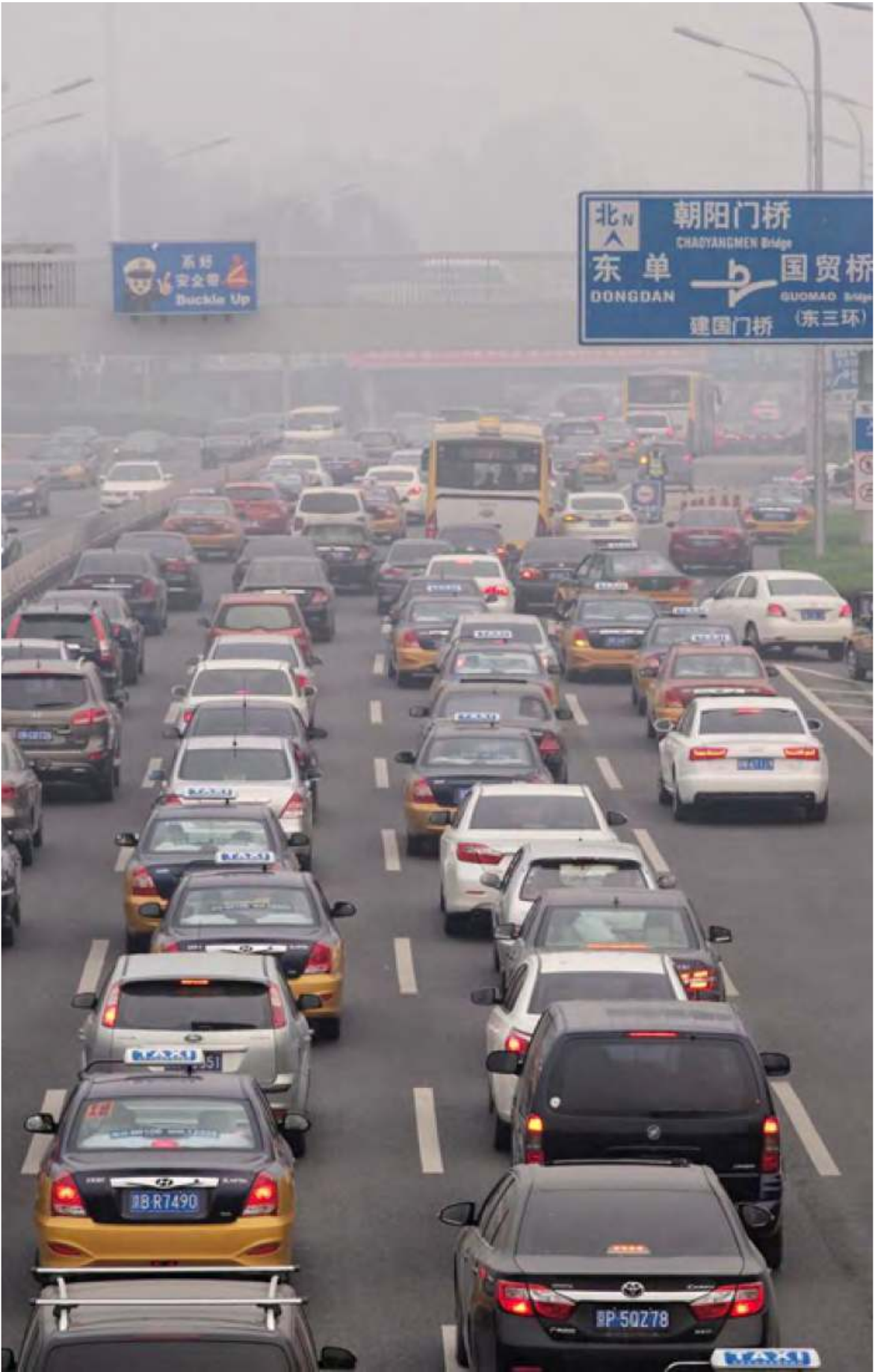}\vspace{2pt}
			\includegraphics[width=1\linewidth]{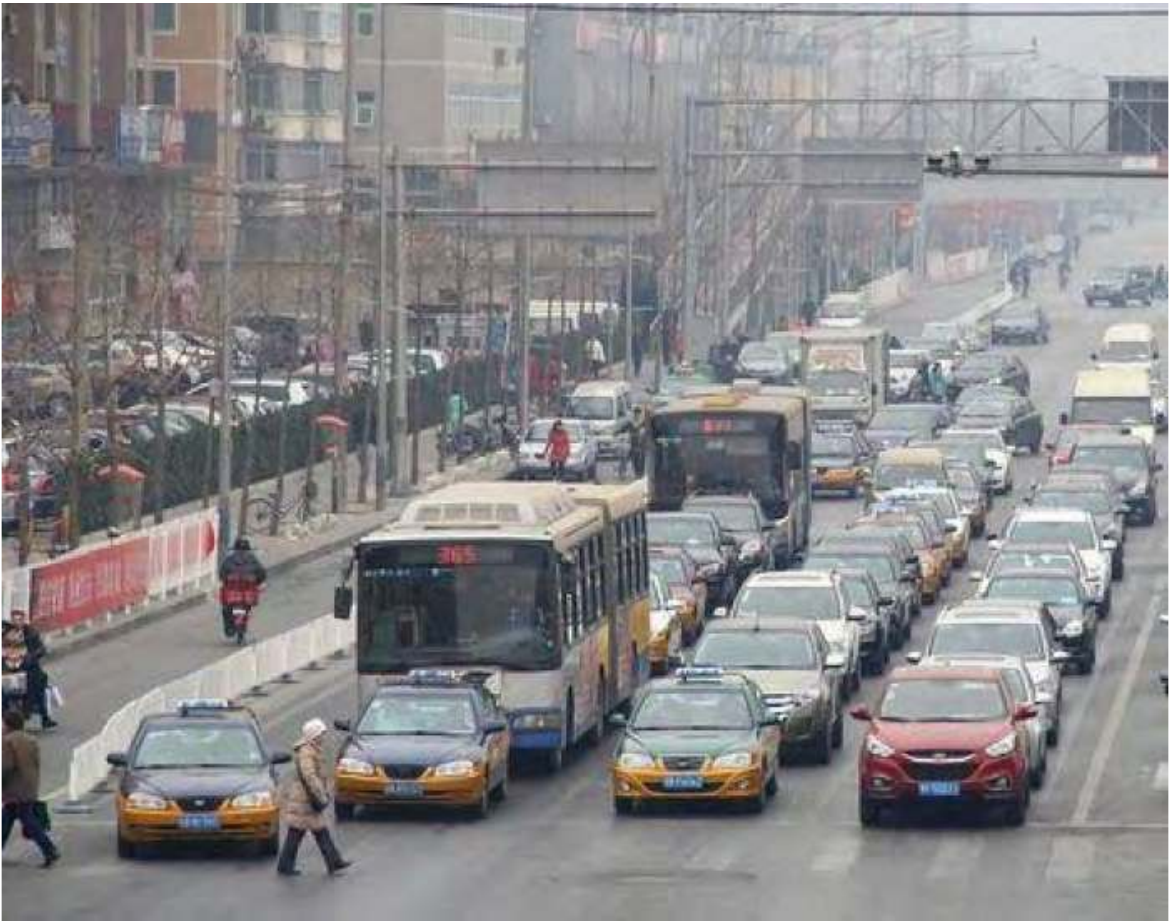}\vspace{2pt}
	\end{minipage}}\hspace{-0.45em}
	\subfigure[\scriptsize{MSBDN-DFF~\cite{dong2020multi}}]{
		\begin{minipage}[b]{0.12\linewidth}
			\includegraphics[width=1\linewidth]{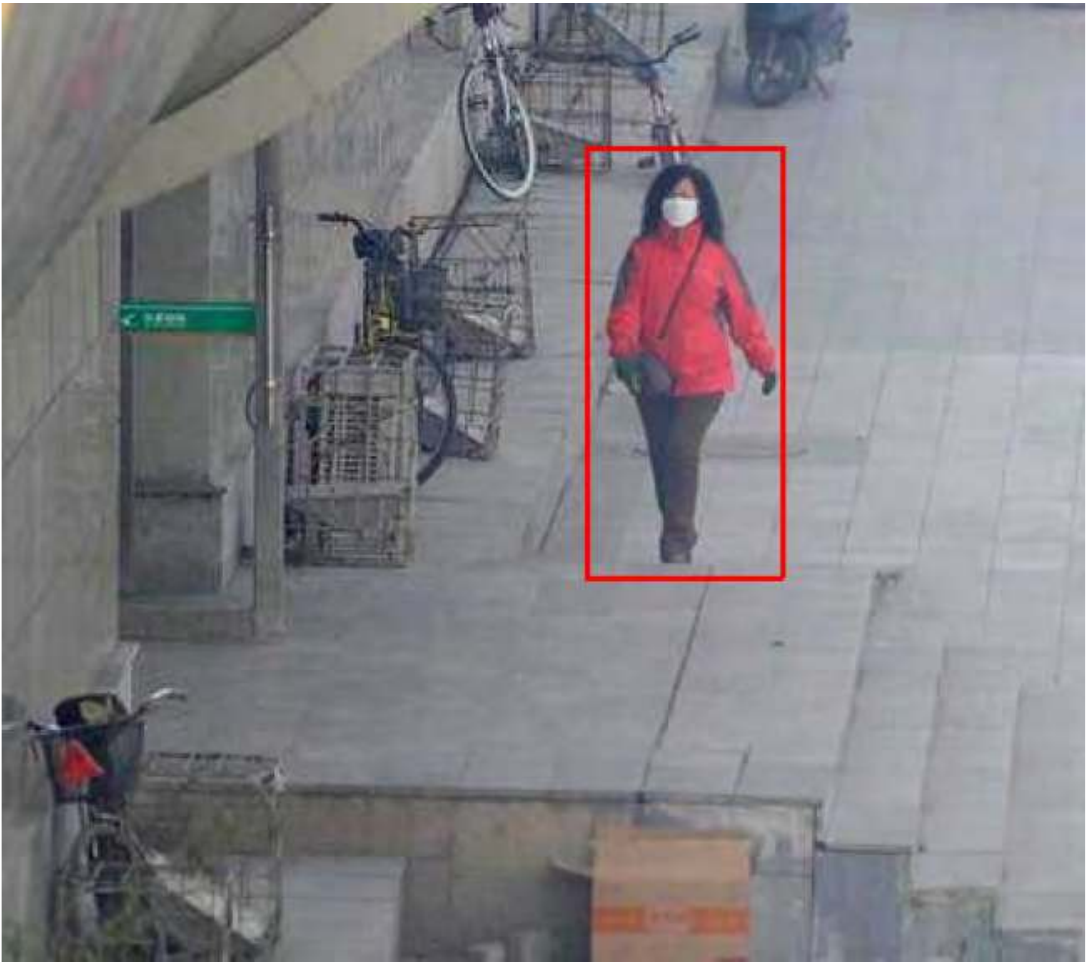}\vspace{2pt}
			\includegraphics[width=1\linewidth]{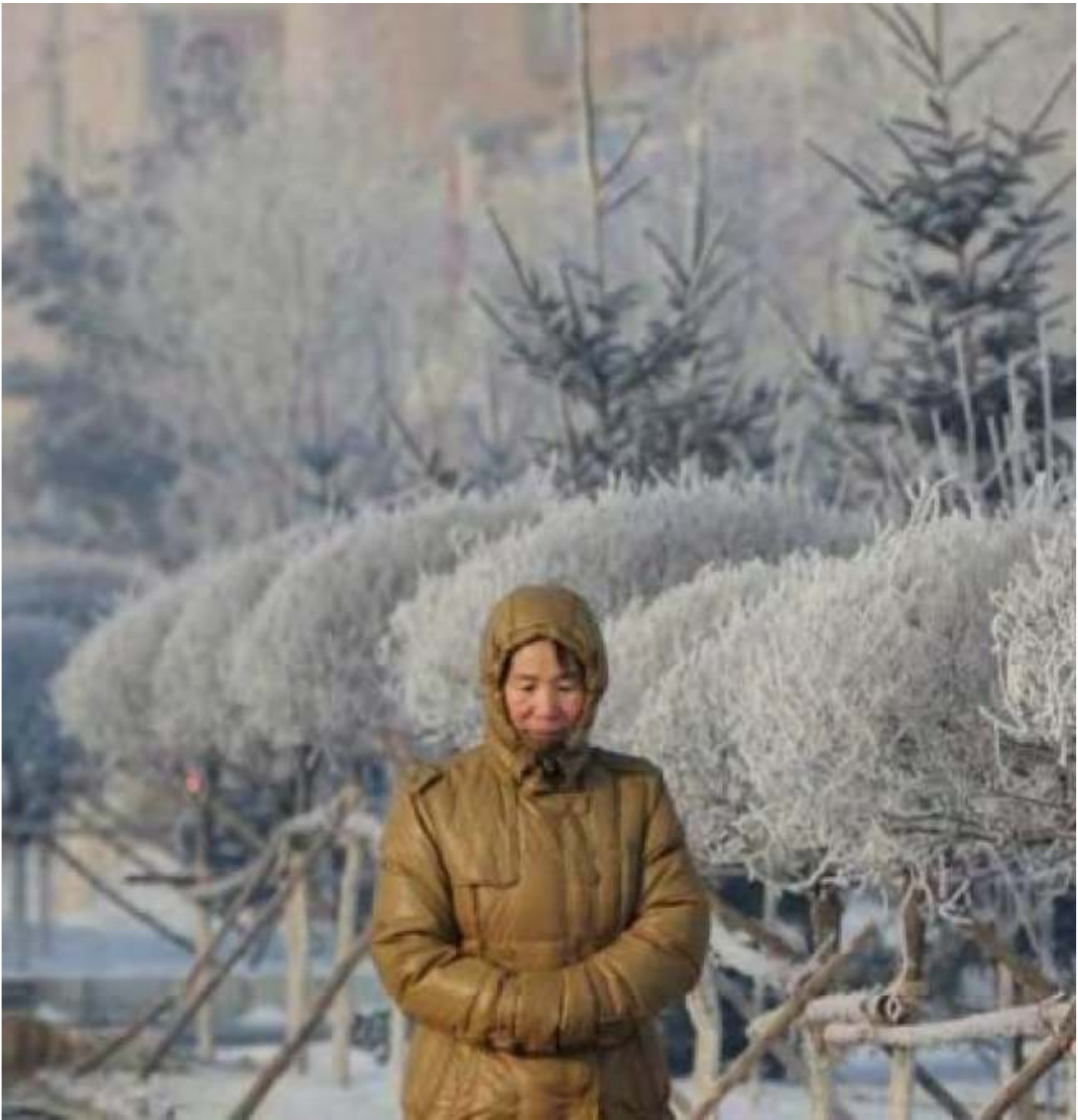}\vspace{2pt}
			\includegraphics[width=1\linewidth]{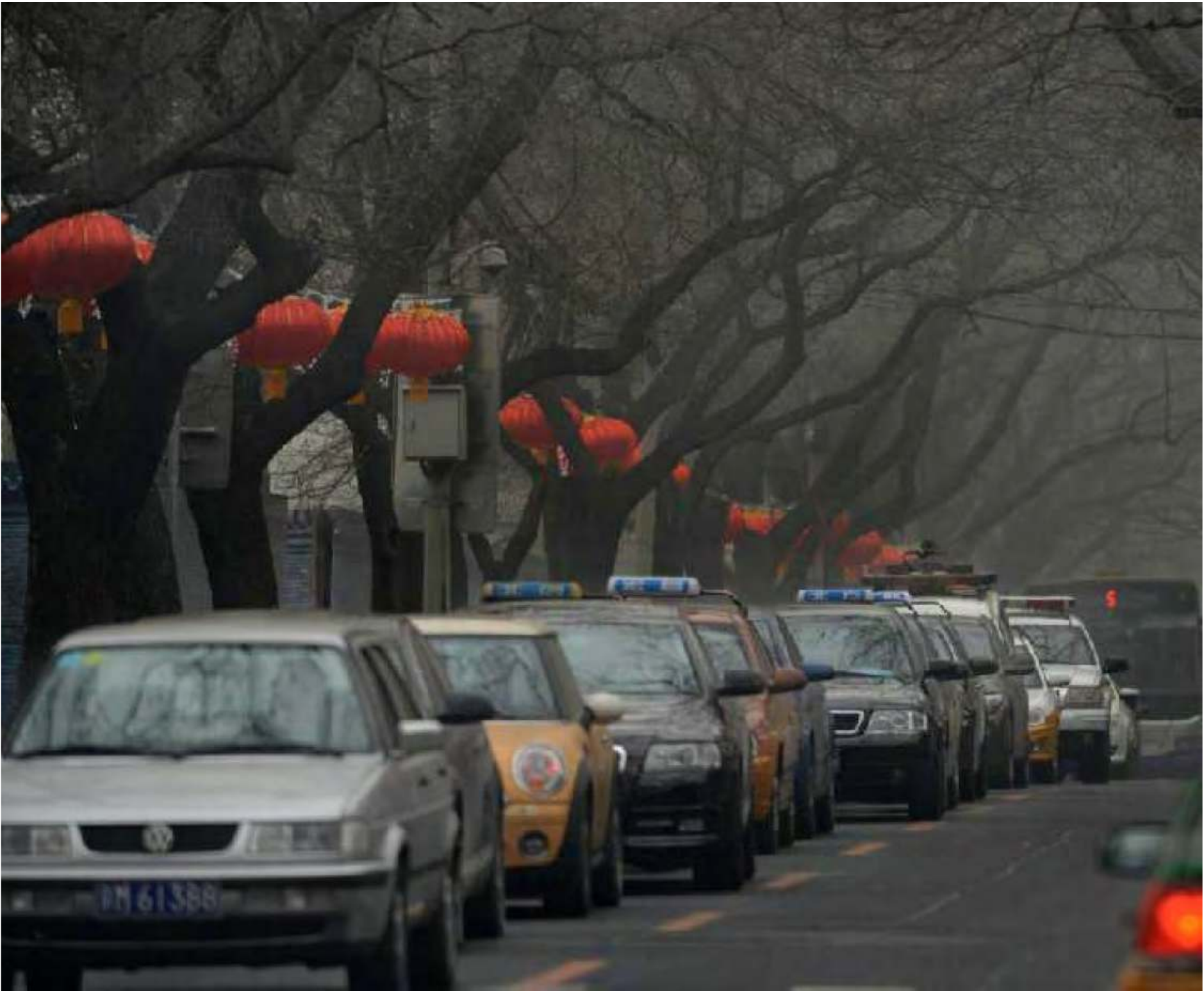}\vspace{2pt}
			\includegraphics[width=1\linewidth]{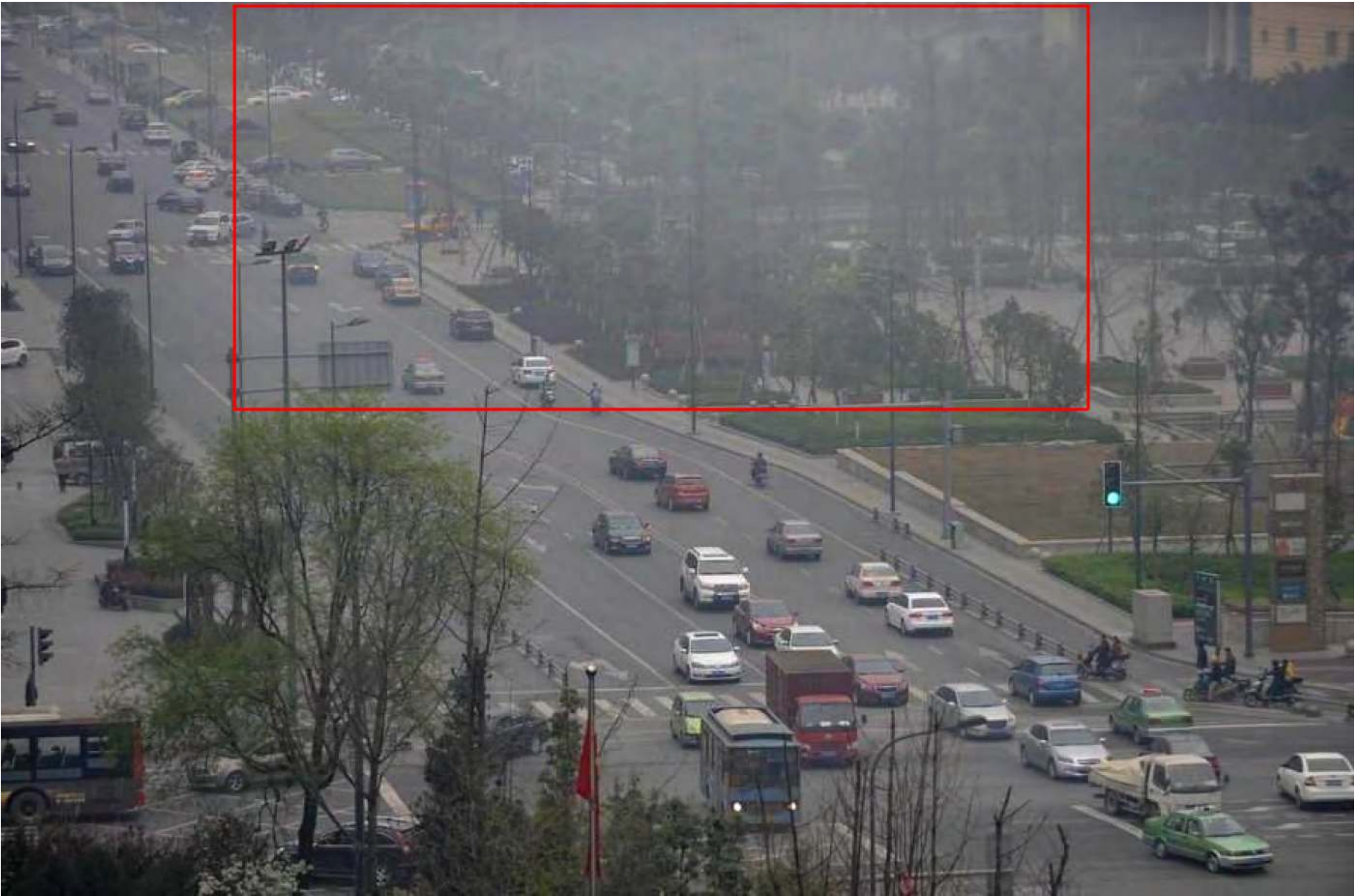}\vspace{2pt}
			\includegraphics[width=1\linewidth]{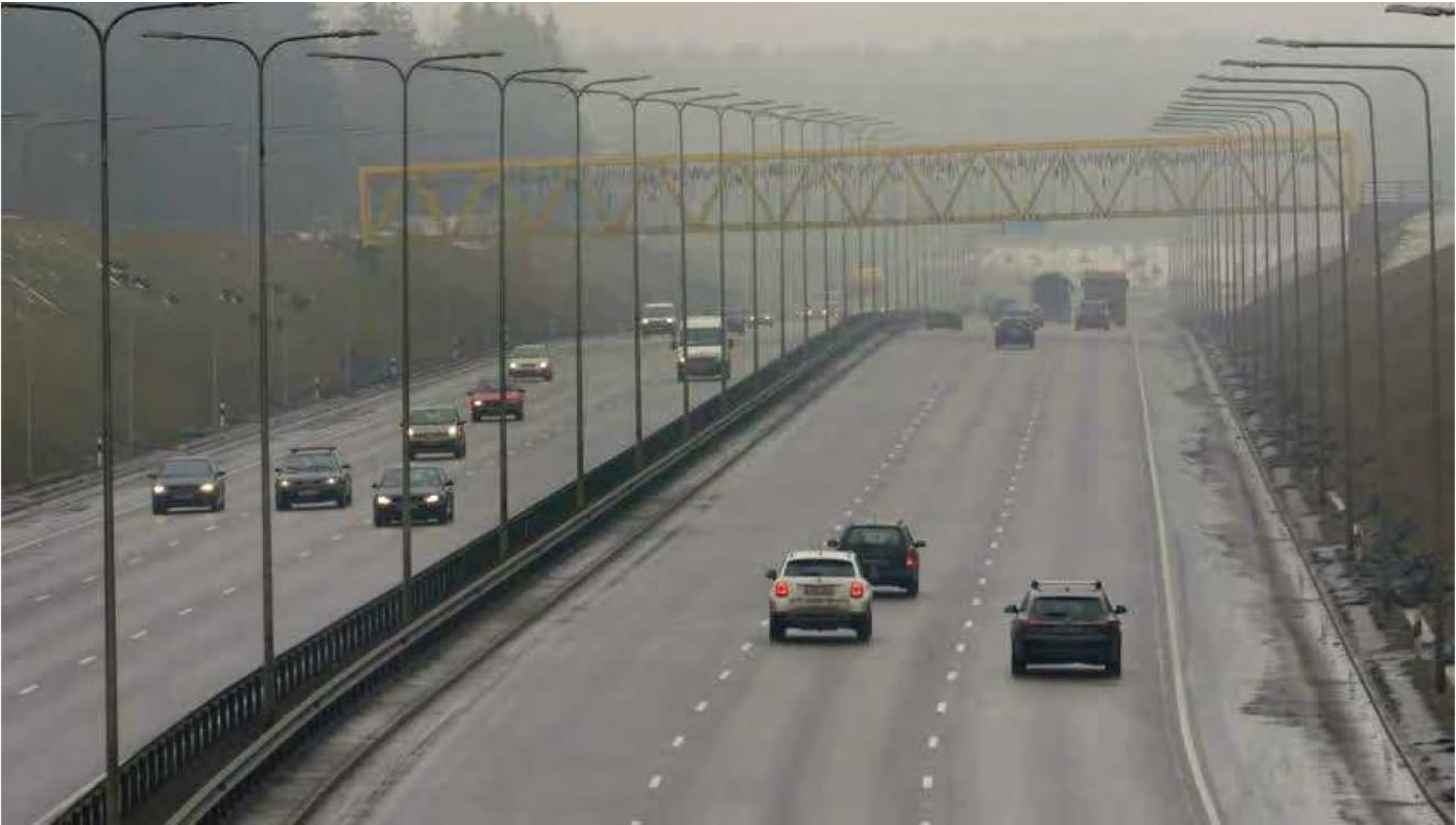}\vspace{2pt}
			\includegraphics[width=1\linewidth]{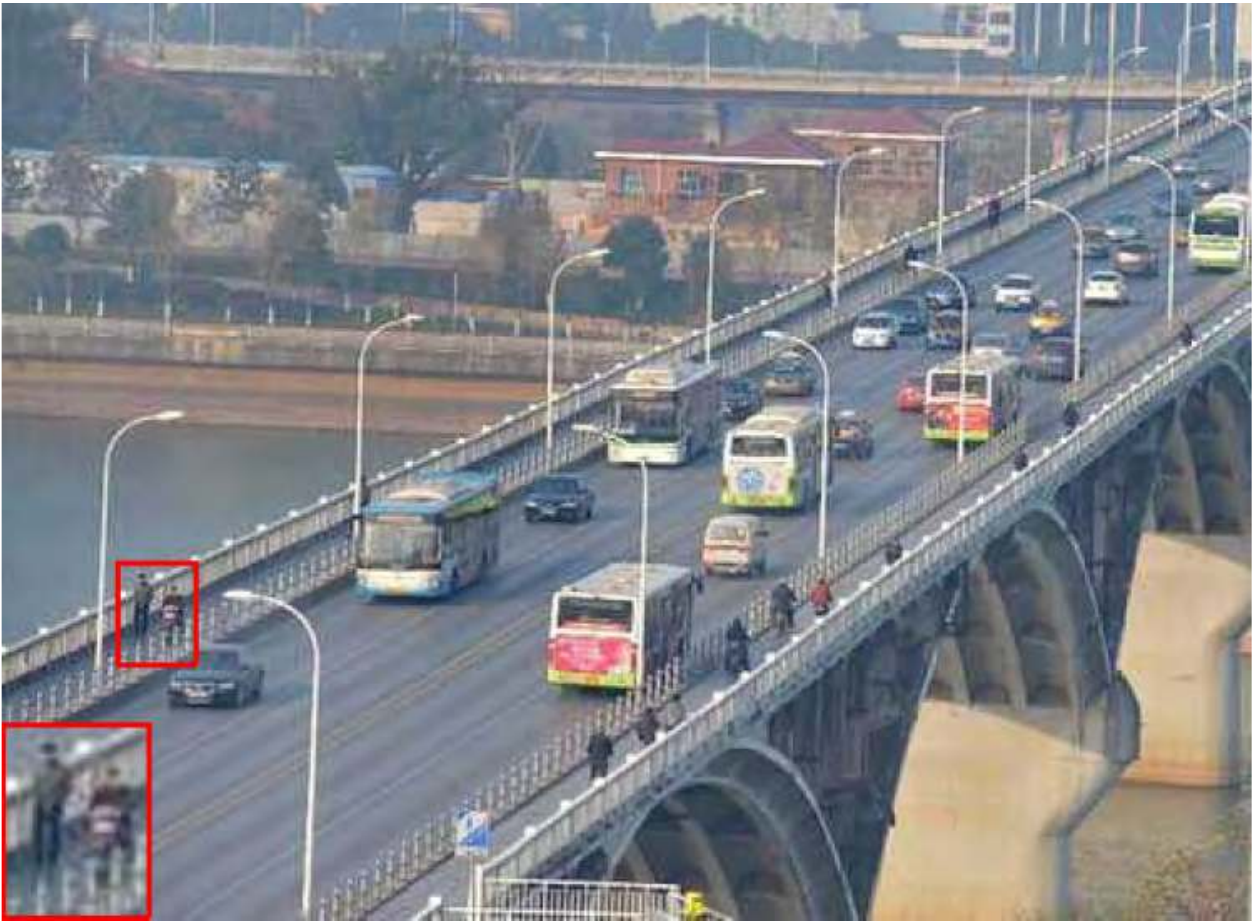}\vspace{2pt}
			\includegraphics[width=1\linewidth]{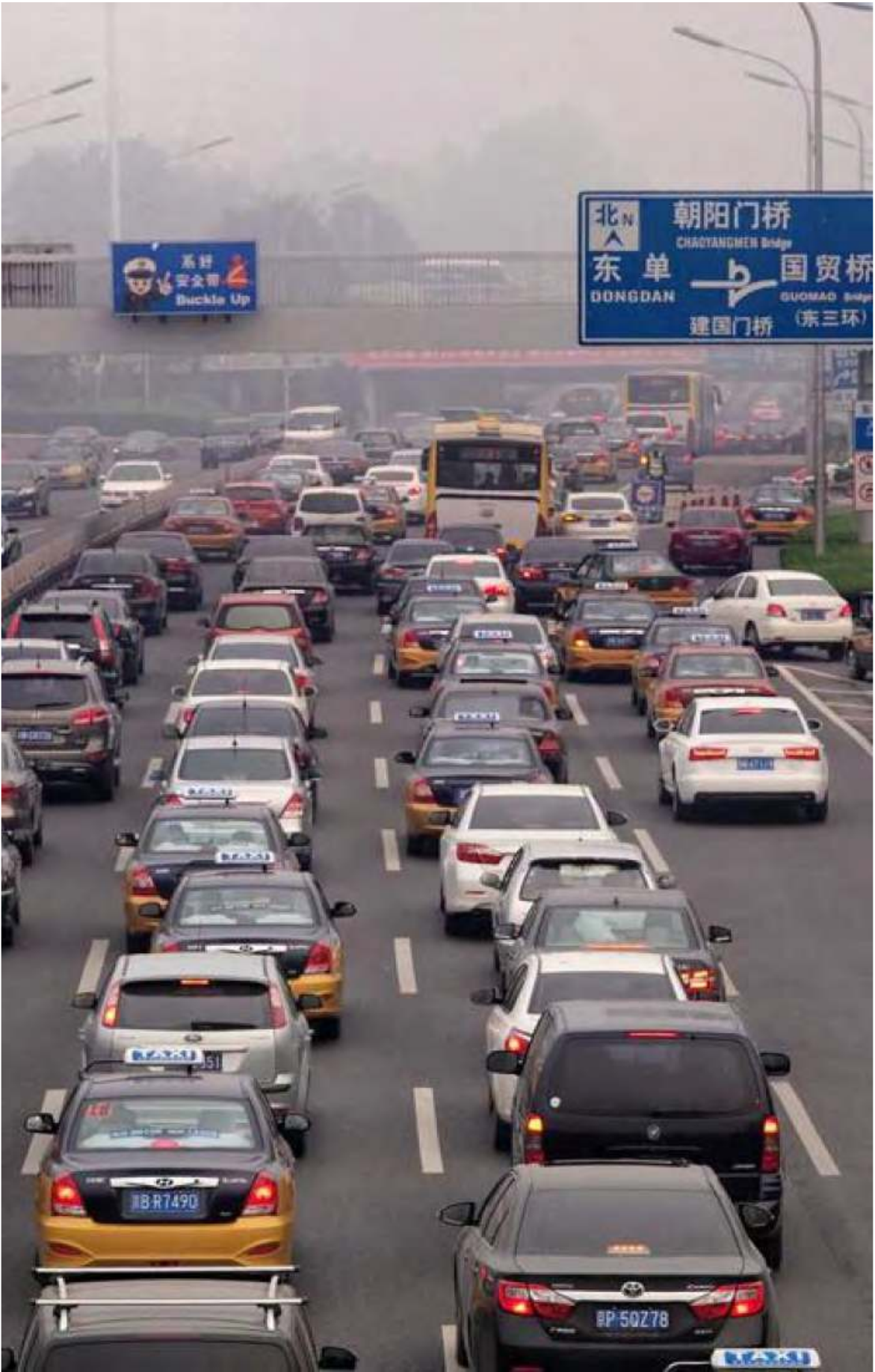}\vspace{2pt}
			\includegraphics[width=1\linewidth]{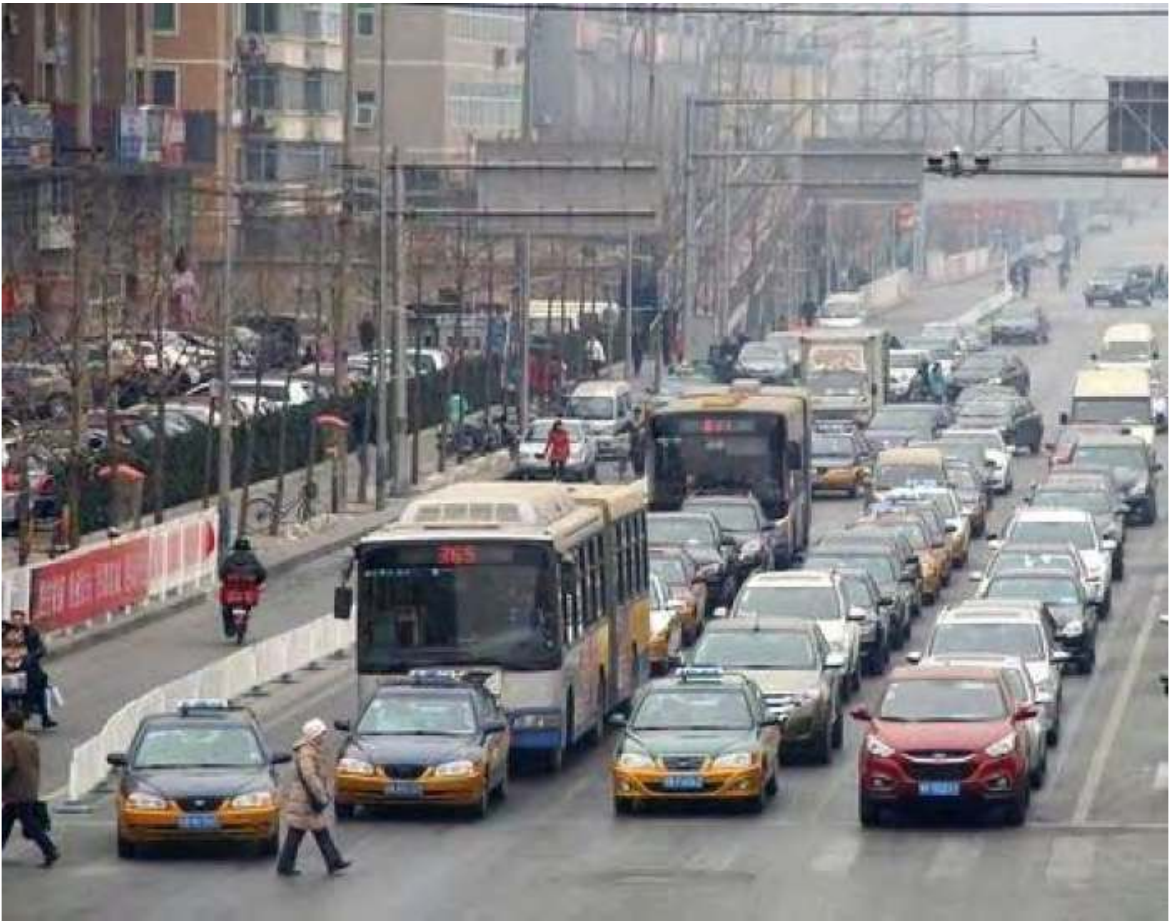}\vspace{2pt}
	\end{minipage}}\hspace{-0.45em}
	\subfigure[\scriptsize{Ours}]{
		\begin{minipage}[b]{0.12\linewidth}
			\includegraphics[width=1\linewidth]{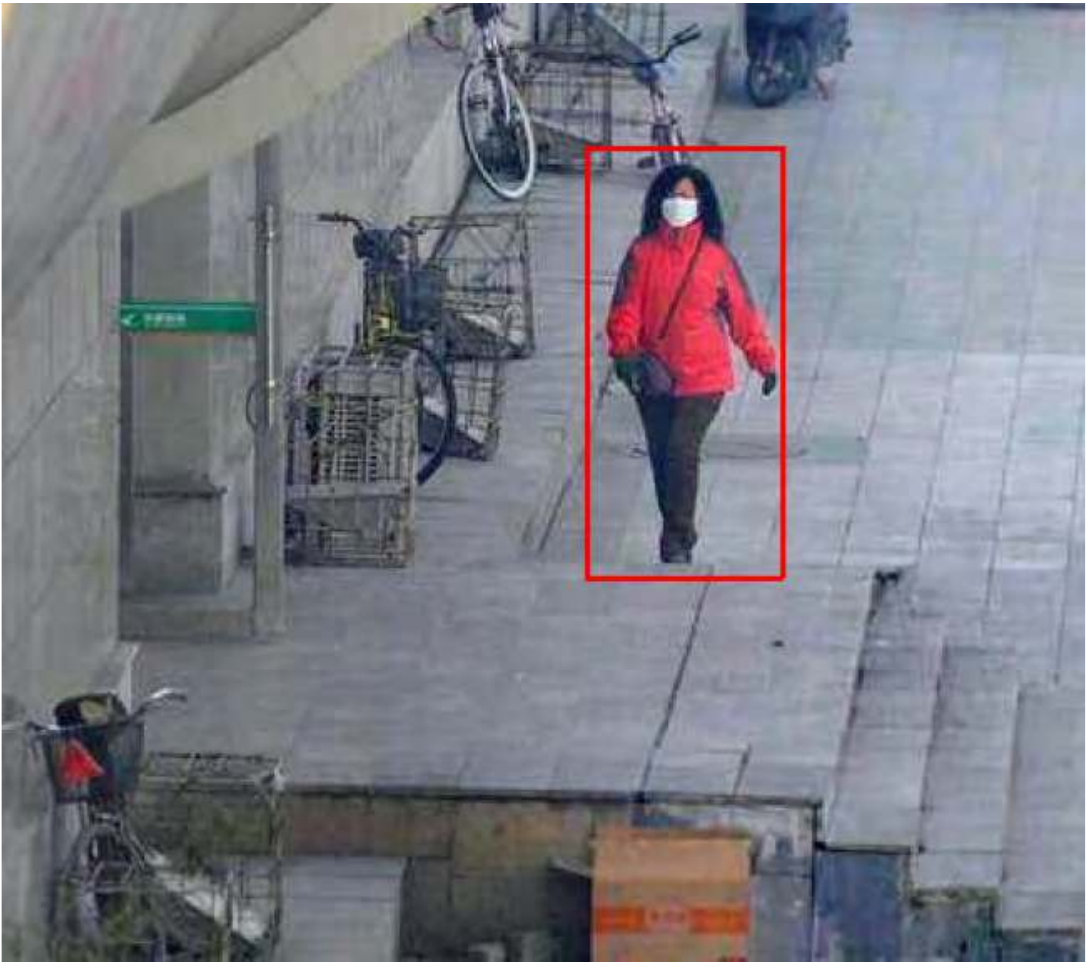}\vspace{2pt}
			\includegraphics[width=1\linewidth]{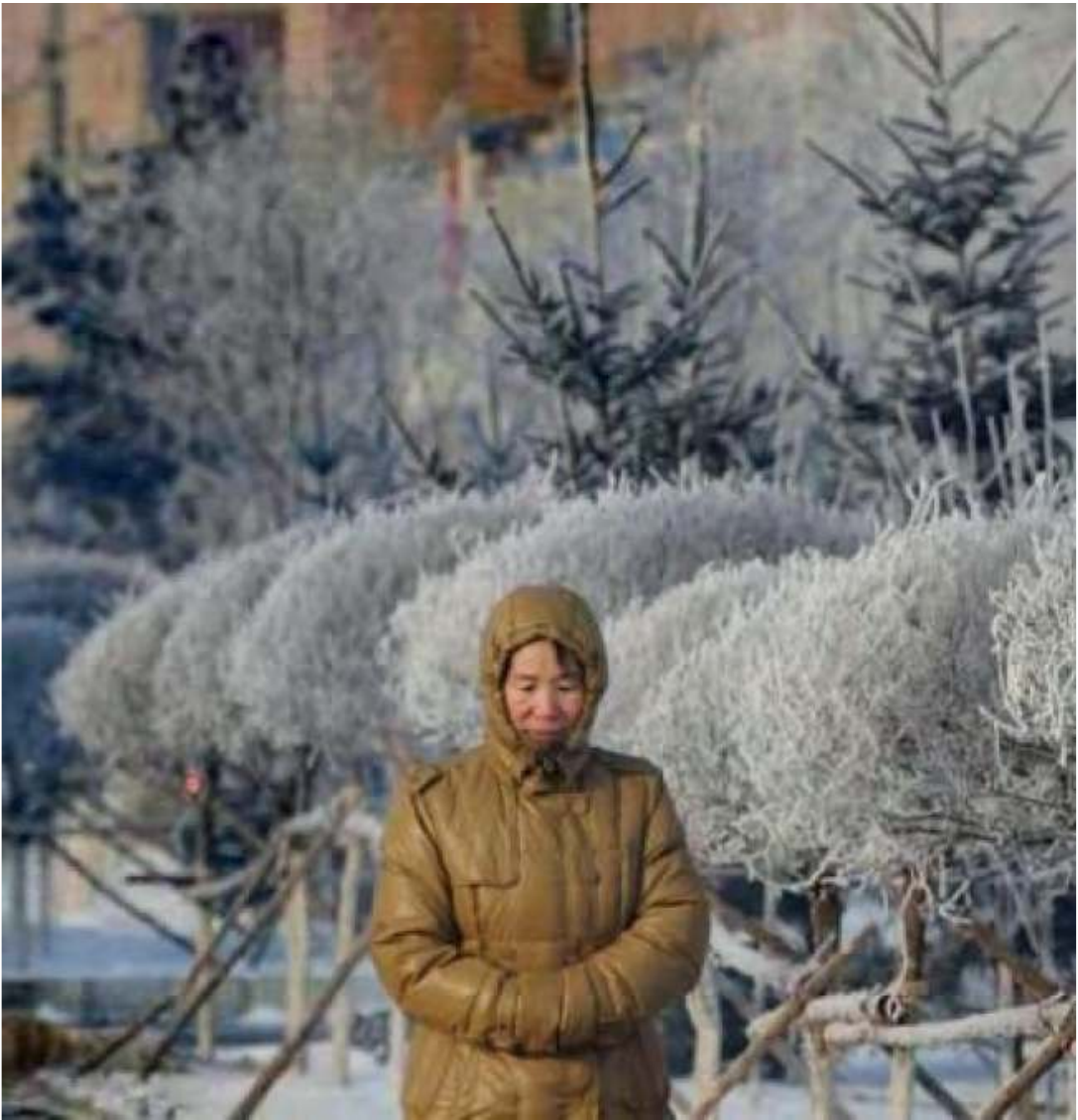}\vspace{2pt}
			\includegraphics[width=1\linewidth]{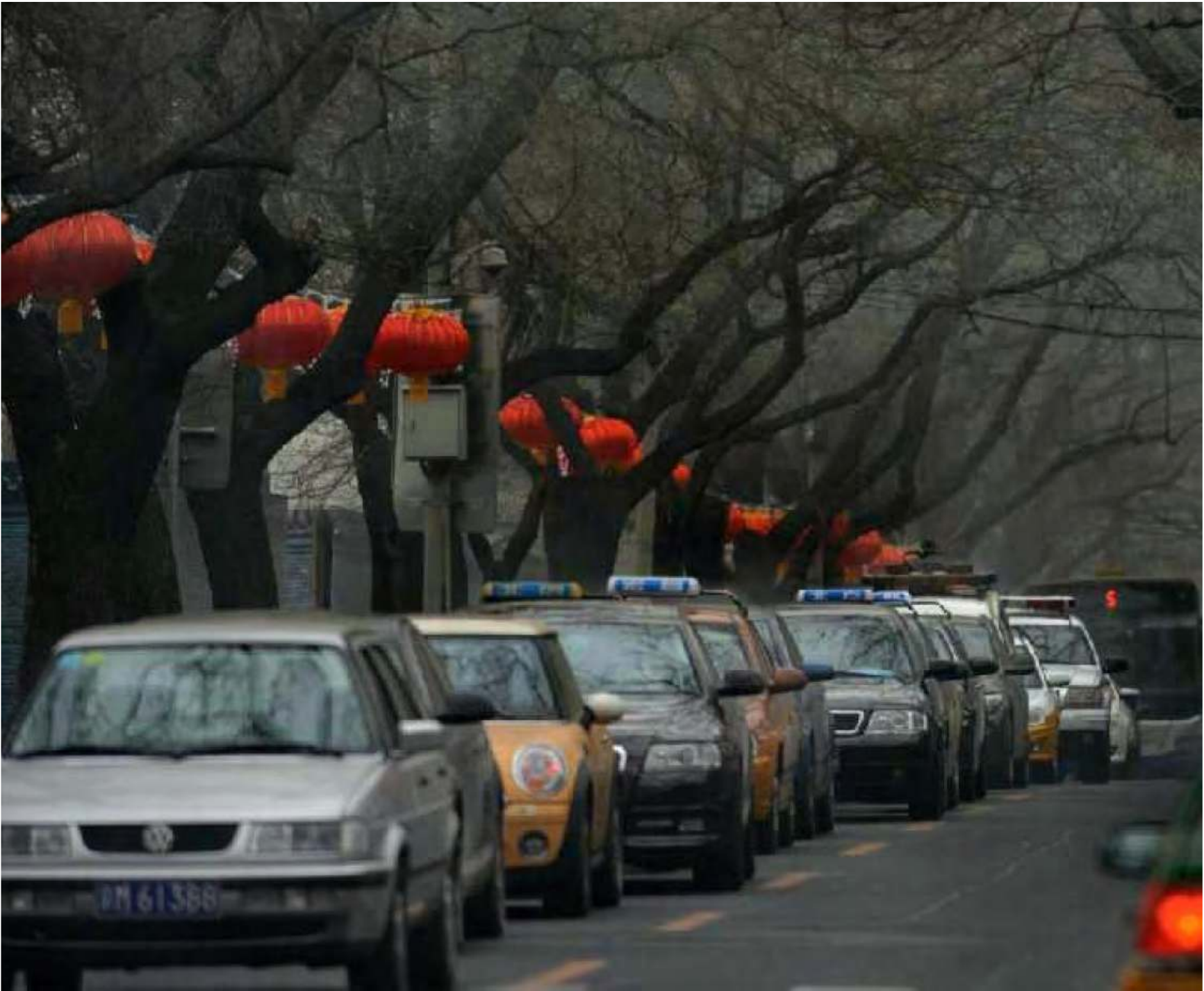}\vspace{2pt}
			\includegraphics[width=1\linewidth]{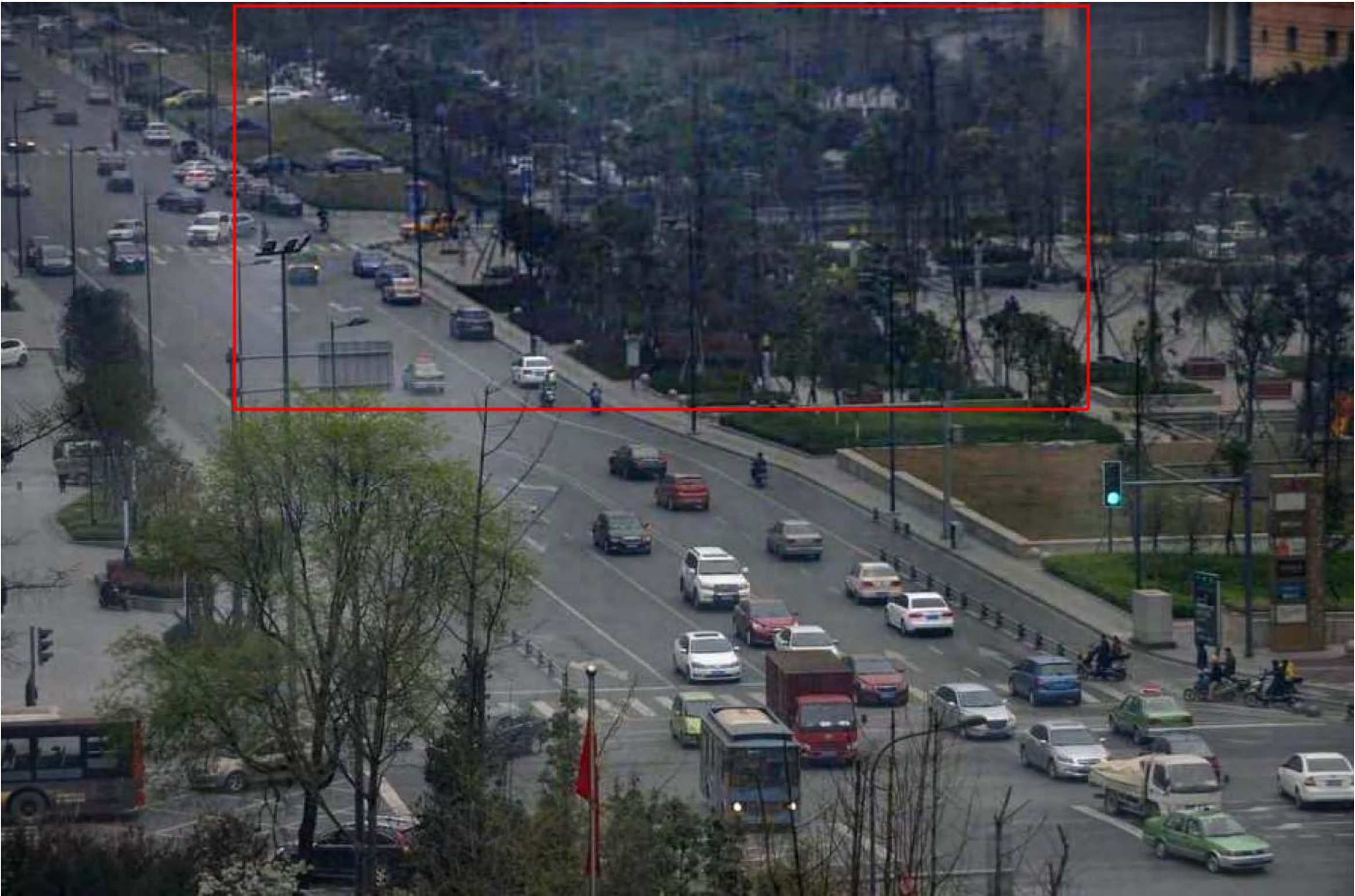}\vspace{2pt}
			\includegraphics[width=1\linewidth]{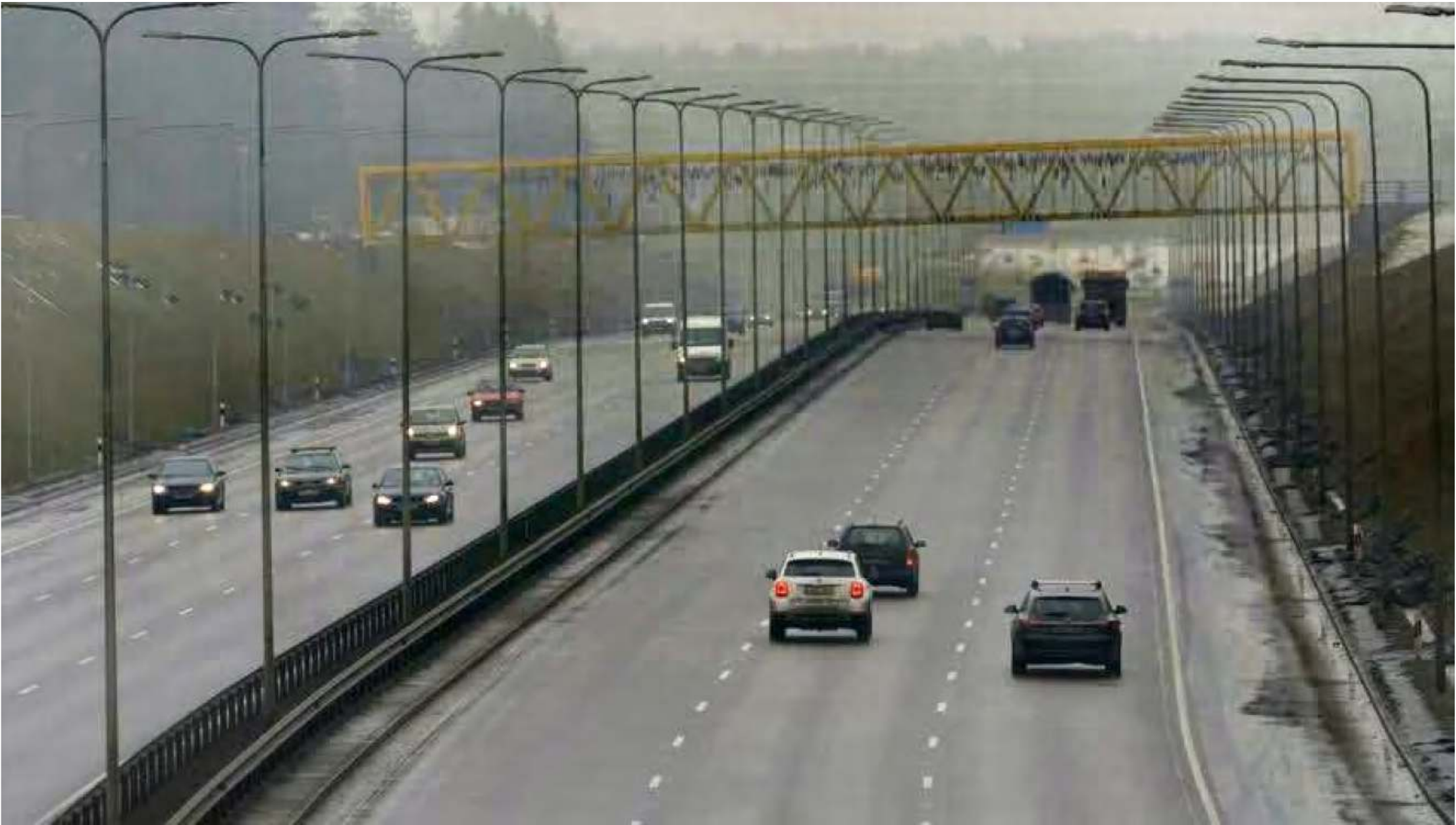}\vspace{2pt}
			\includegraphics[width=1\linewidth]{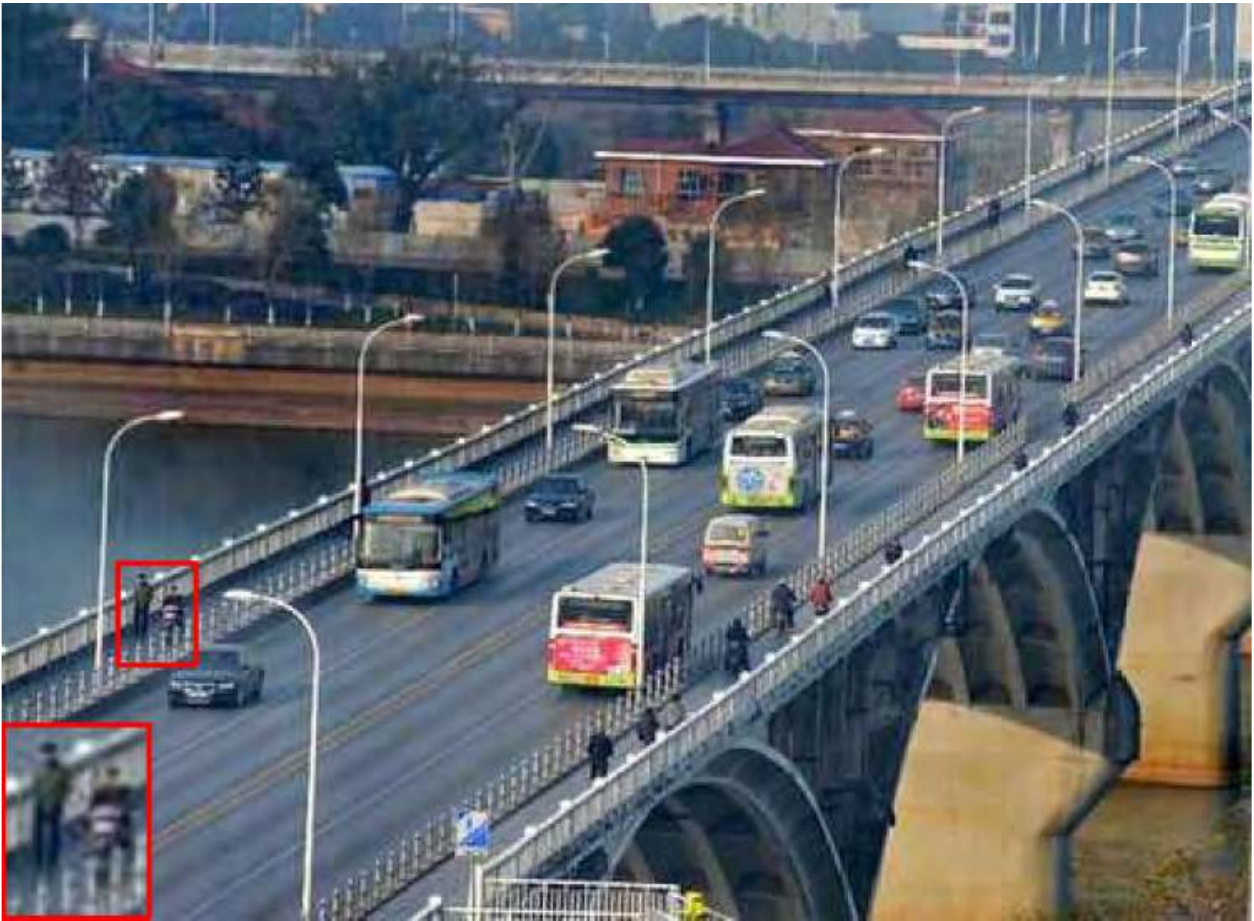}\vspace{2pt}
			\includegraphics[width=1\linewidth]{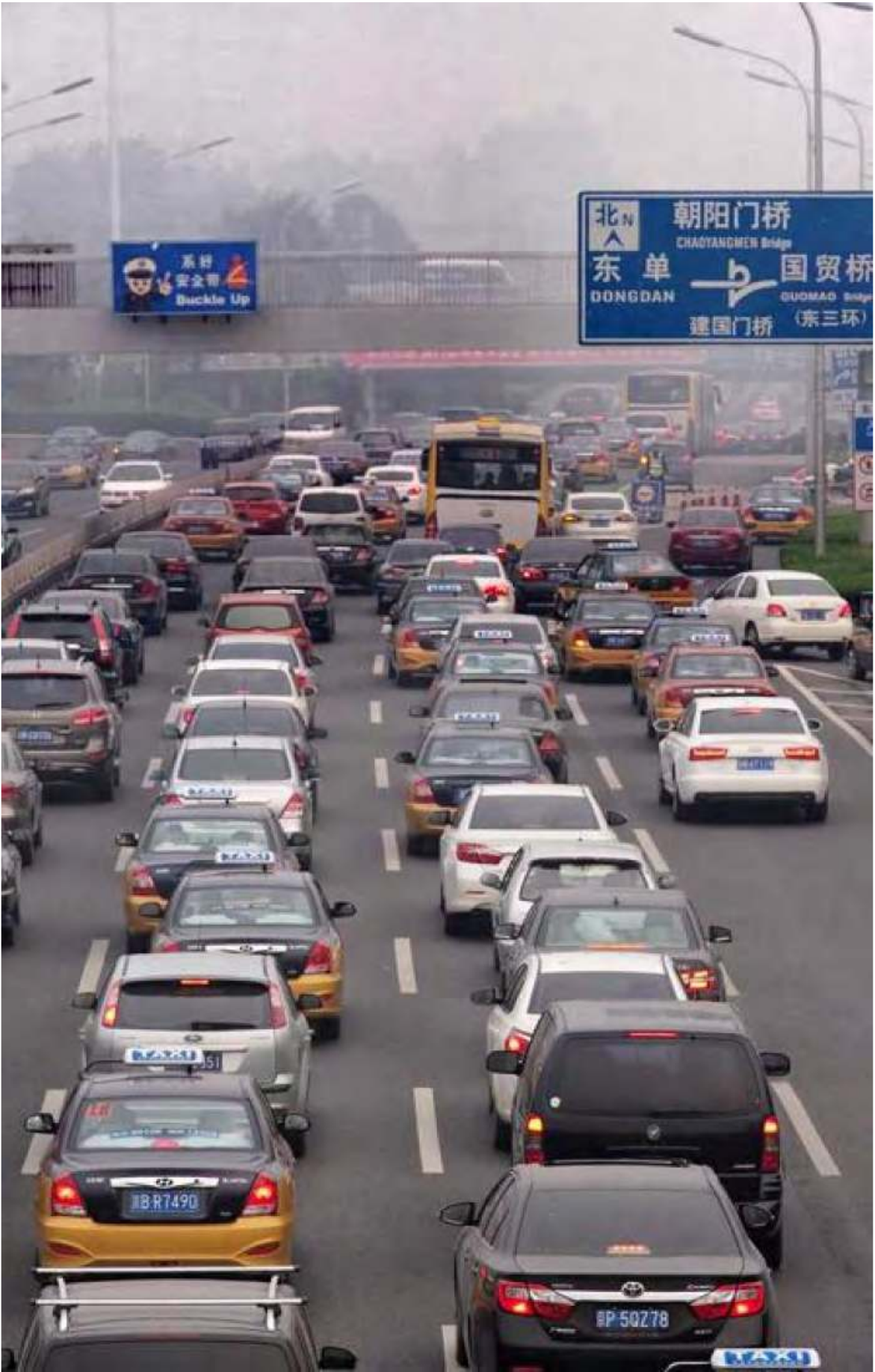}\vspace{2pt}
			\includegraphics[width=1\linewidth]{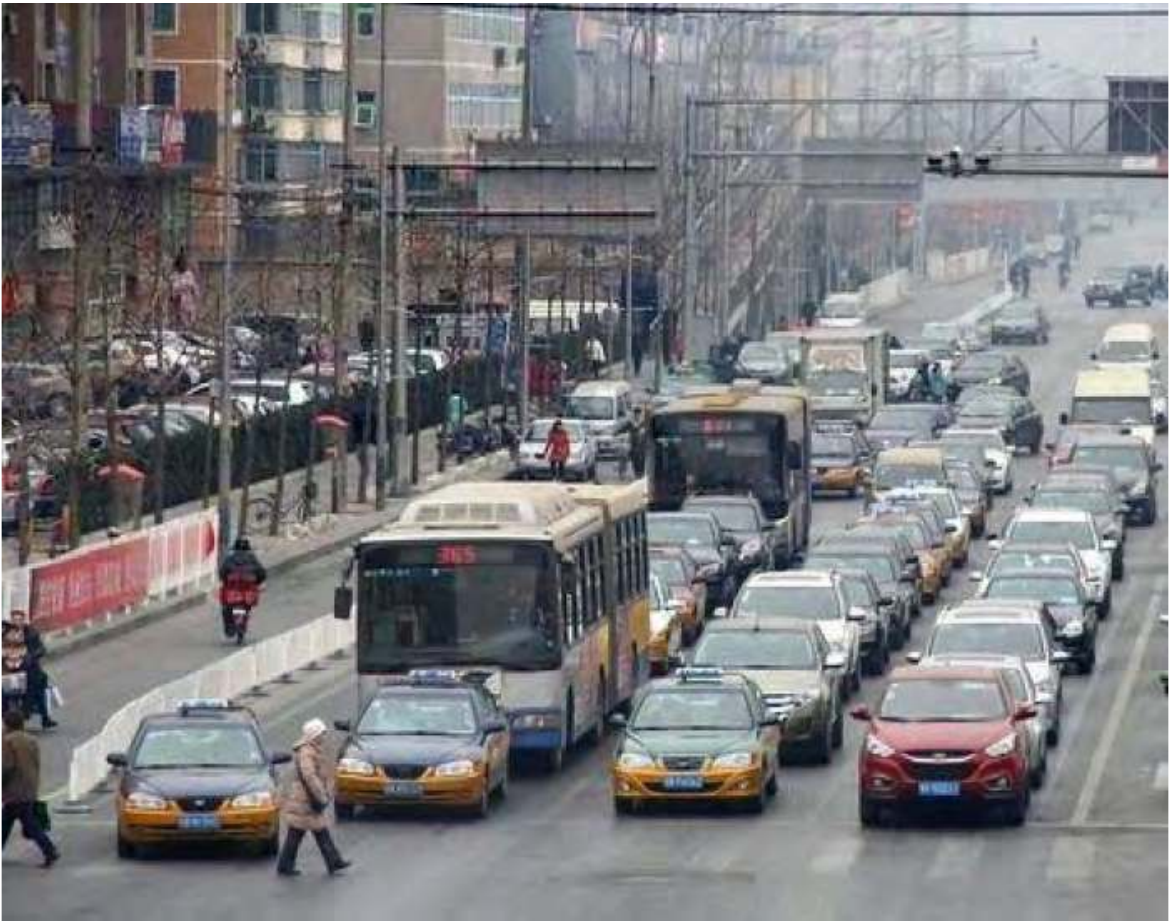}\vspace{2pt}
	\end{minipage}}
	\caption{Visual results of real world hazy images on RTTS~\cite{li2019benchmarking} dataset}
	\label{fig:contrast_inter_domain}
\end{figure*}

\subsection{Results on Synthetic Datasets}

To evaluate the effectiveness of our proposed intra-domain adaptation, we train our model on SOTS~\cite{li2019benchmarking} dataset and HazeRD~\cite{zhang2017hazerd} dataset and compare the results with other previous methods.

To verify the robustness in the case of distribution shifts, we test the dehazing performance of several models under different haze distributions, as shown in Figure \ref{figure:syn_its_contrast}. From those results, we can observe that previous algorithms all encounter the phenomenon of performance drop when facing different haze distributions, e.g., magnified area in the images. In other words, when the haze distribution is a hard sample, the dehazed image has higher chance that it remains haze in global or detail. Compared with previous methods, our approach generates clearer images under different haze situations which verifies the effectiveness of the intra-domain adaptation (more detailed proof can be found in \ref{section_ablation}).

The quantitative evaluation are shown in Table \ref{quantitative results}. Our method achieves the best performance on both PSNR and SSIM. Compared with the state-of-the-art method MSBDN-DFF~\cite{dong2020multi}, our method achieves performance gain on both SOTS~\cite{li2019benchmarking} and HazeRD~\cite{zhang2017hazerd}.

\begin{figure*}[htb]
	\subfigure[Range]{
		\includegraphics[width=0.3\linewidth]{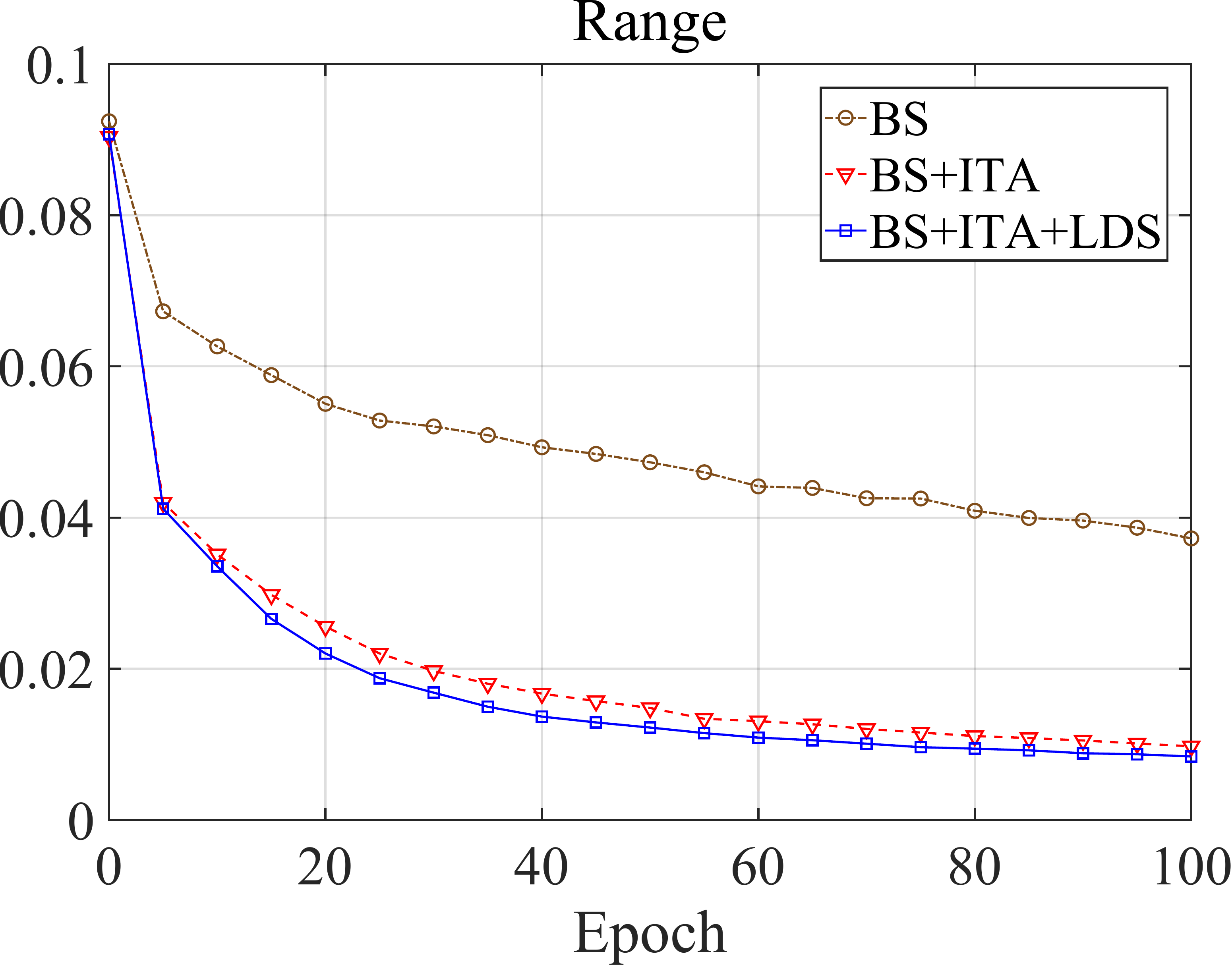}
	}
	\subfigure[Standard Deviation]{
		\includegraphics[width=0.3\linewidth]{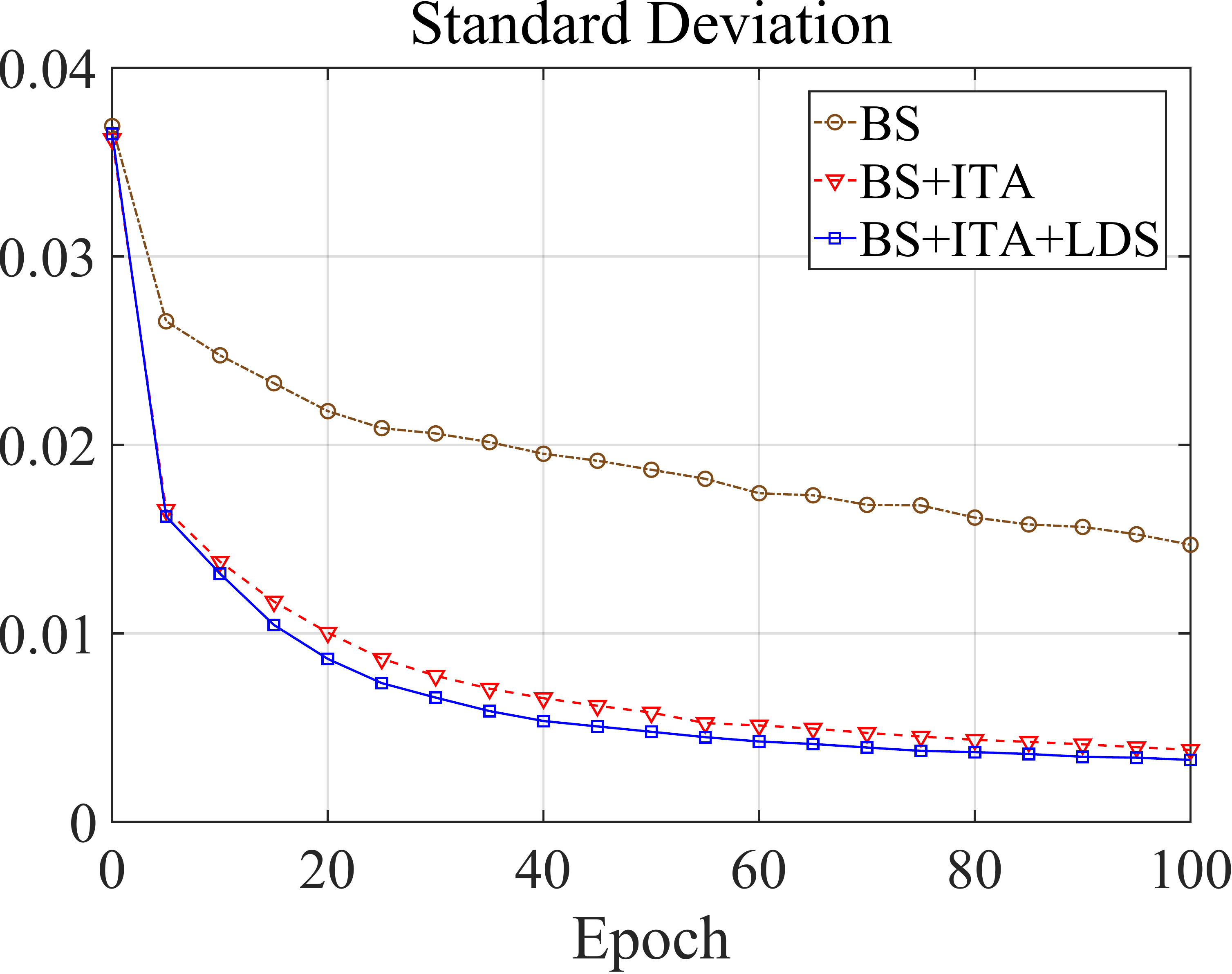}
	}
	\subfigure[Coefficient of Variation]{
		\includegraphics[width=0.3\linewidth]{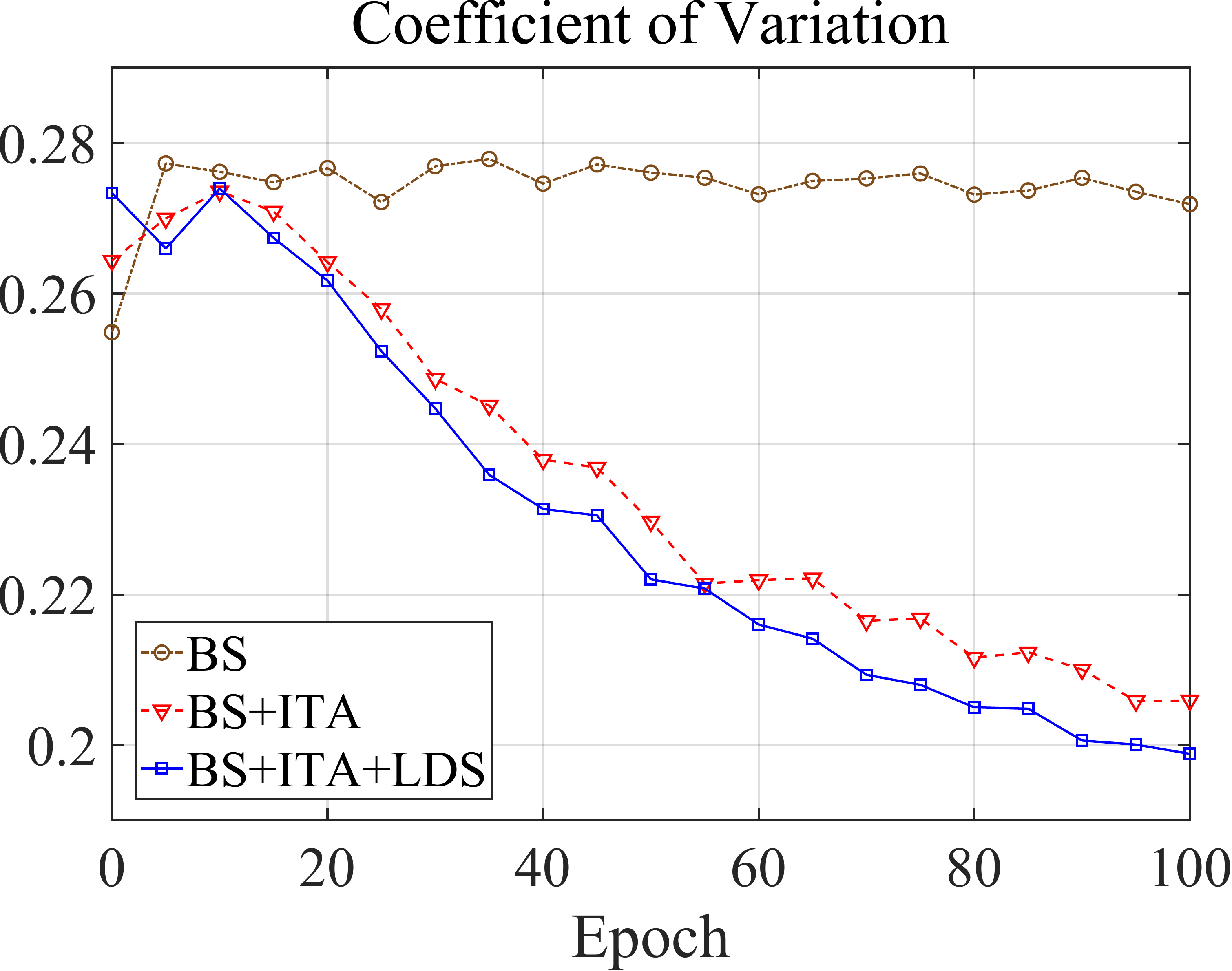}
	}
	\caption{The dispersion evaluation of the dehazing losses. Network with the intra-domain adaptation gets more compact dehazing losses which demonstrates that the features are aligned in feature space.}
	\label{fig:loss_var}
\end{figure*}

\begin{figure*}[htb]
	\subfigure[Input]{
		\centering
		\includegraphics[width=0.19\linewidth]{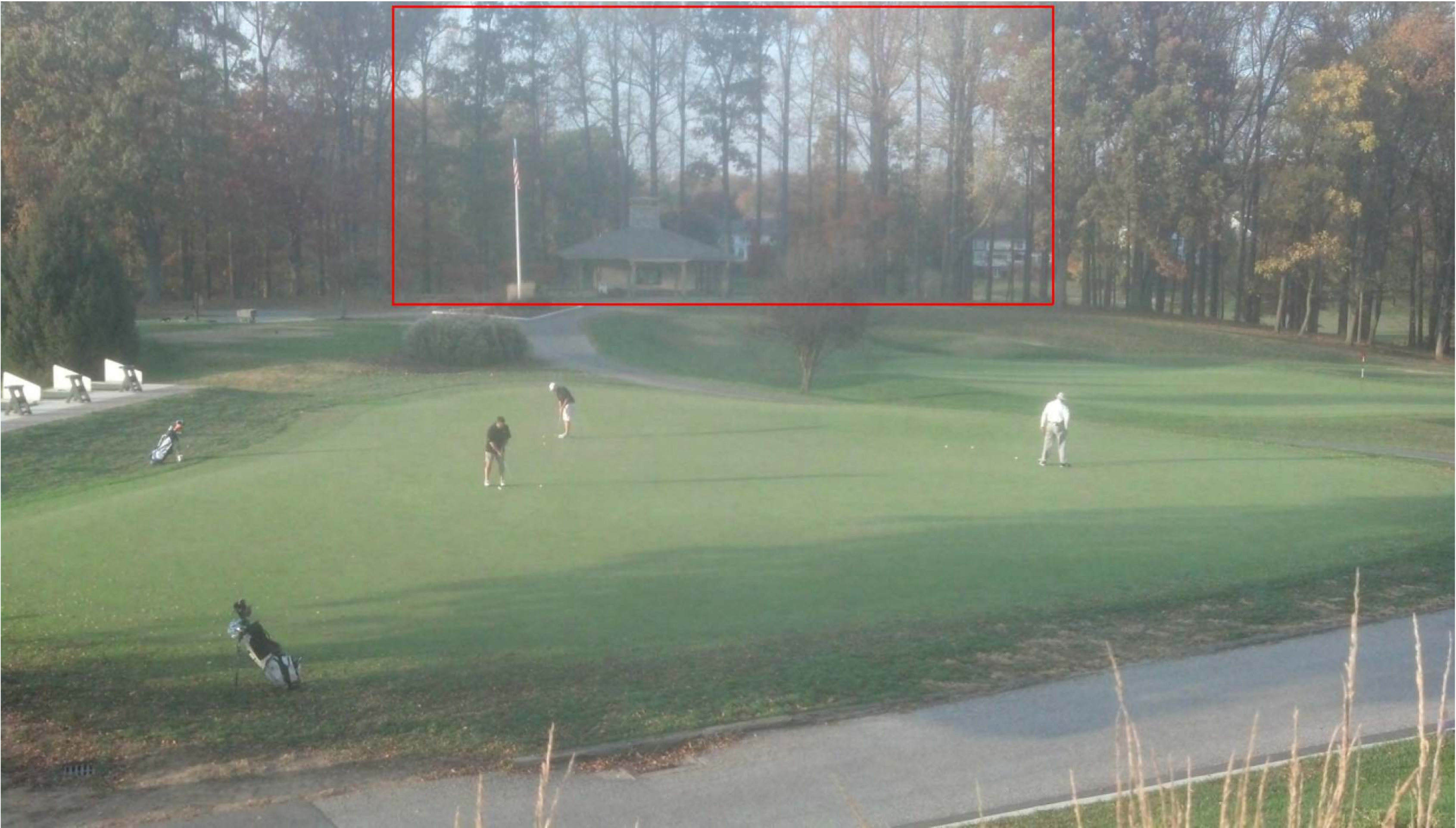}
	}\hspace{-0.55em}
	\subfigure[BS]{
		\centering
		\includegraphics[width=0.19\linewidth]{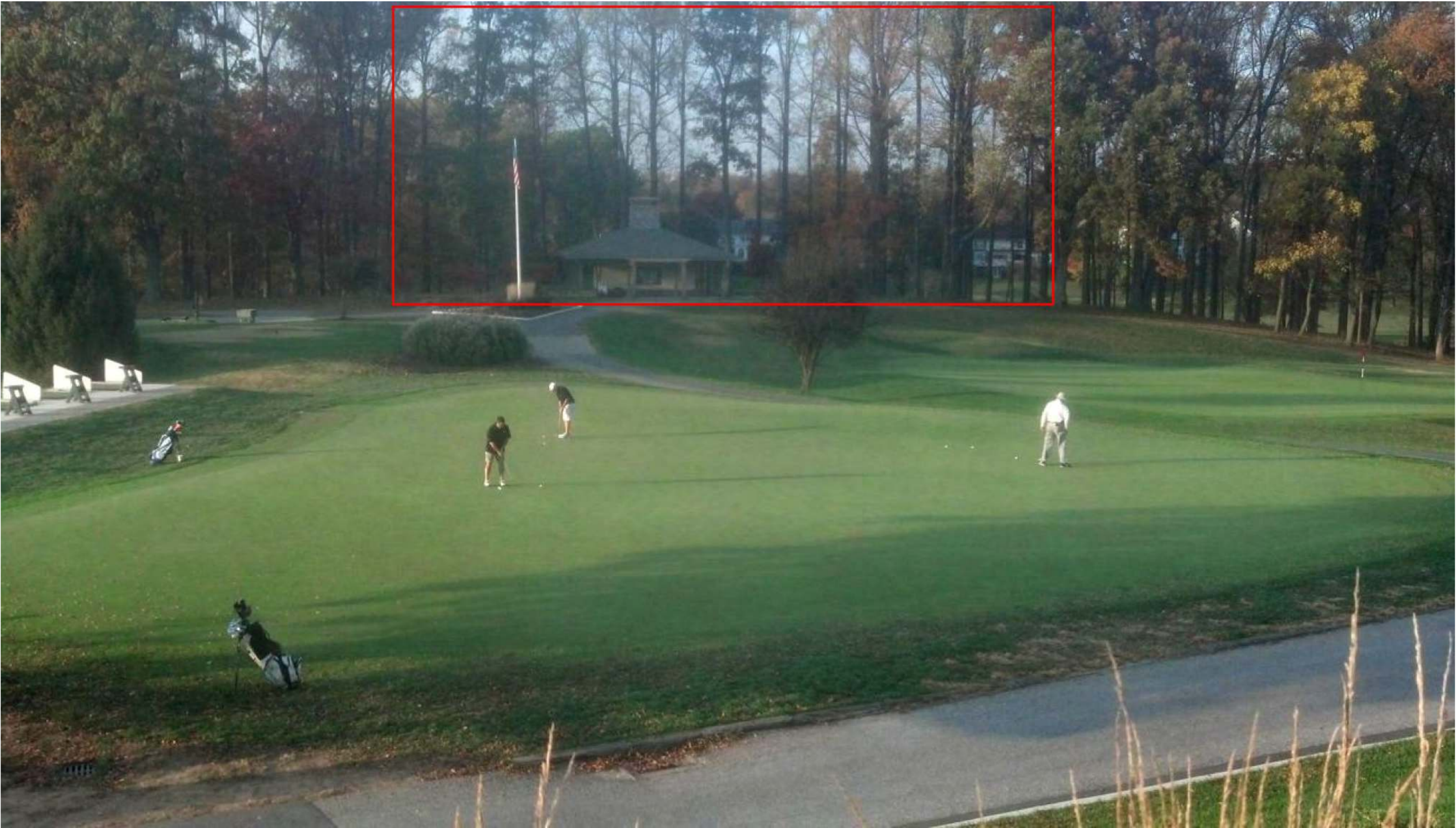}
	}\hspace{-0.55em}
	\subfigure[BS+ITE]{
		\centering
		\includegraphics[width=0.19\linewidth]{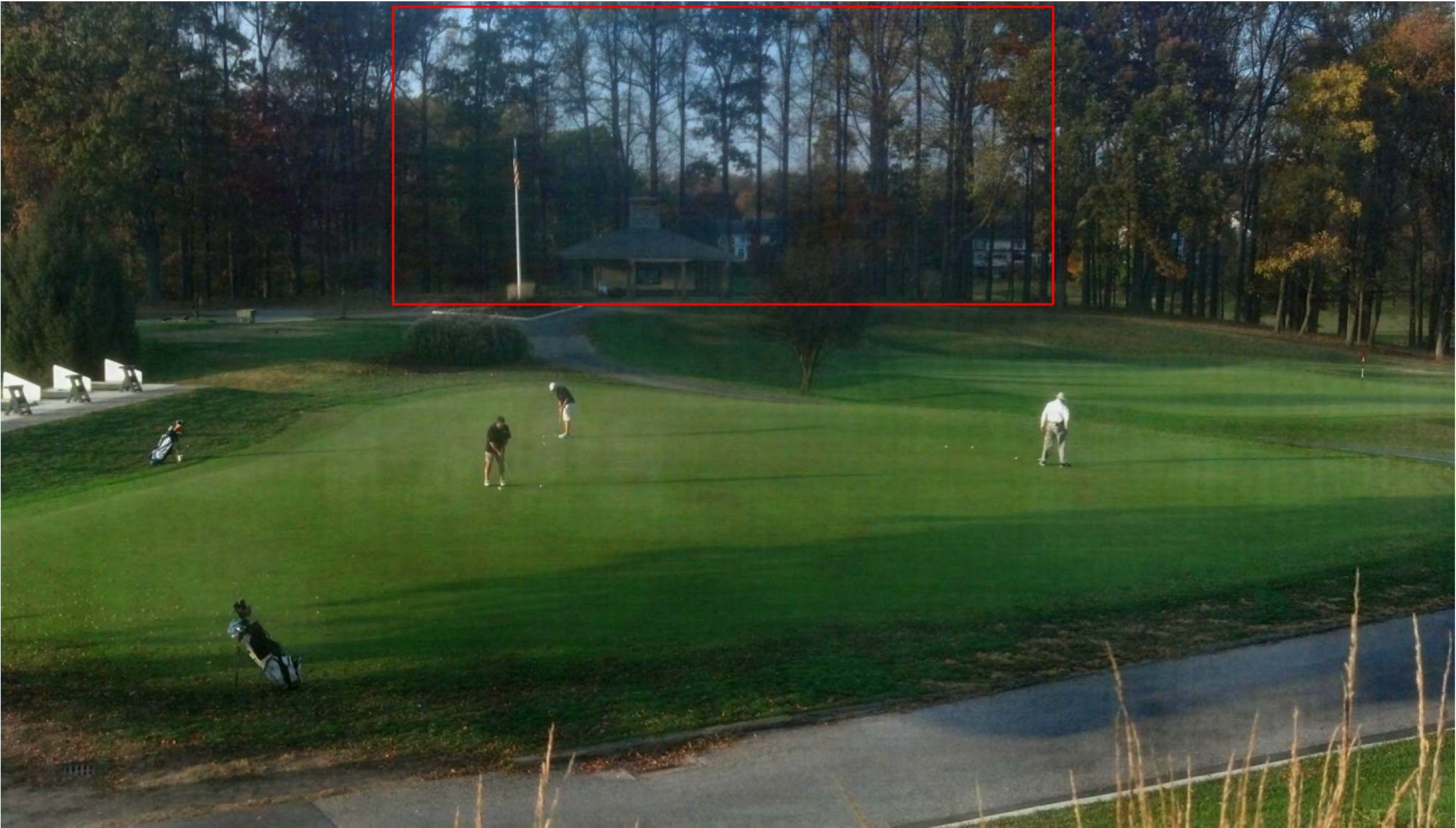}
	}\hspace{-0.55em}
	\subfigure[BS+ITA]{
		\centering
		\includegraphics[width=0.19\linewidth]{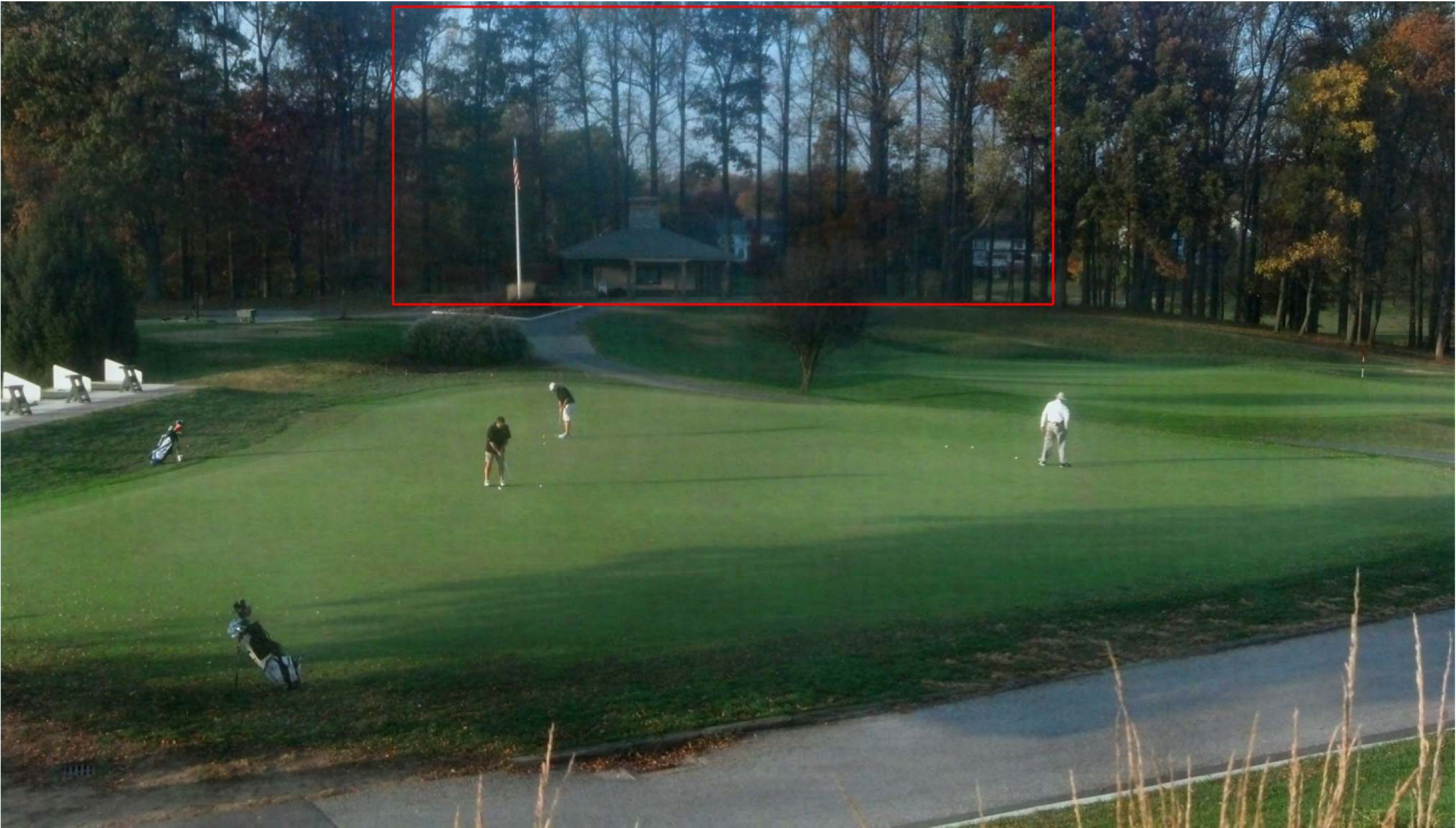}
	}\hspace{-0.55em}
	\subfigure[BS+ITA+ITE]{
		\centering
		\includegraphics[width=0.19\linewidth]{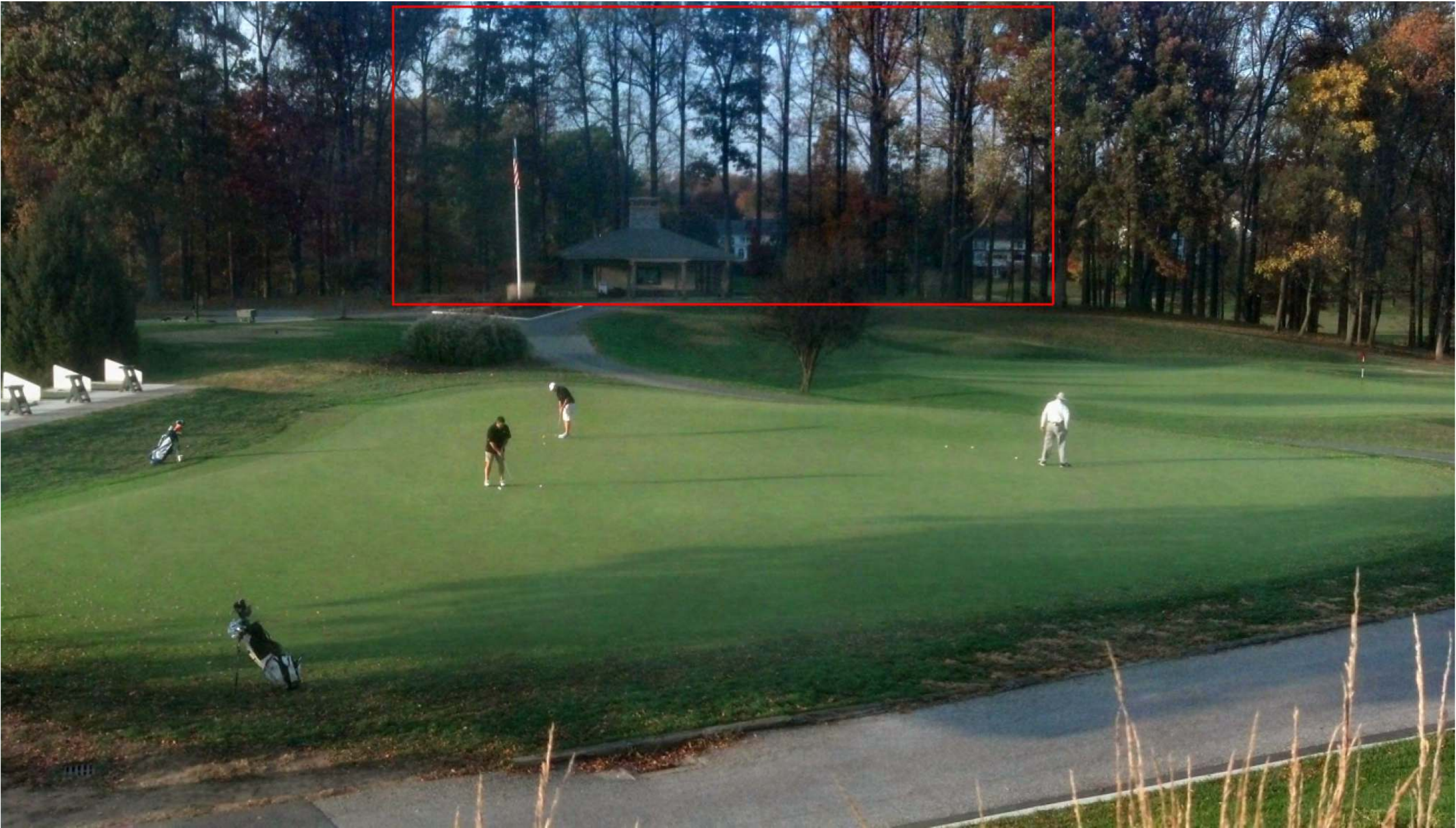}
	}
	\caption{Ablation study of real world hazy images on RTTS~\cite{li2019benchmarking} datset. }
	\label{fig:ablation_real}
\end{figure*}

\subsection{Results on Real Images}

To evaluate our proposed constrained inter-domain adaptation between the synthetic domain and the real domain, we conduct experiments on the real dataset RTTS~\cite{li2019benchmarking} and compare visual results with other previous methods.

The visual results are shown in Figure \ref{fig:contrast_inter_domain}. From the results, we can observe that previous dehazing methods have different limitations on real images. Specifically, DCP~\cite{he2010single} suffers from serious color distortion and overexposure, e.g., the first, third and fourth rows of Figure \ref{fig:contrast_inter_domain} (b). Besides, the dehazed results of Dehazenet~\cite{cai2016dehazenet}, FFA~\cite{qin2020ffa} and MSBDN-DFF~\cite{dong2020multi} all have residual haze, e.g., the first, fourth and sixth rows of Figure \ref{fig:contrast_inter_domain} (c), (f) and (g). In addition, the dehazed results of EPDN~\cite{qu2019enhanced} suffer from brightness issues (much darker), e.g., the third row and the traffic signs in the seventh row of Figure \ref{fig:contrast_inter_domain} (d) and color distortion (some results are more yellow than other methods), e.g., the sixth and seventh rows of Figure \ref{fig:contrast_inter_domain} (d). Furthermore, DAdehazing~\cite{shao2020domain} reaches better visual results than the other previous methods since they integrate the inter-domain adaptation in their approach. The overall brightness and the color are well-maintained during dehazing process. However, there is still some residual haze, e.g., the trees in the fourth row of Figure \ref{fig:contrast_inter_domain} (e). In addition, some results of DAdehazing become less realistic or blur, e.g., the people in the first row, the trees in the fourth row and the people in the sixth row of Figure \ref{fig:contrast_inter_domain} (e). Those problems are caused by only considering the inter-domain gap between the source and the target without considering the intra-domain gap of the target. when the distribution of the real domain are aligned to the hard samples of the synthetic domain, the difficulty of image dehazing is increased. Overall, the method we proposed achieves the best performance in removing haze, maintaining the color and brightness of the images, and restoring details.

\subsection{Ablation Study}
\label{section_ablation}

In order to verify the effectiveness of each module in our proposed method, we conduct ablation studies on the intra-domain adaptation and the inter-domain adaptation.

In the intra-domain adaptation part, we conduct ablation study using the following settings: 1) \textbf{BS}: base network; 2) \textbf{BS+ITA}: base network with the intra-domain adaptation; 3) \textbf{BS+ITA+LDS}: base network with the intra-domain adaptation and the loss-based deep supervision.

\begin{table}[tb]
	\caption{Ablation study on SOTS~\cite{li2019benchmarking} dataset. ``ITA" denotes the intra-domain adaptation, ``LDS" denotes the loss-based deeply supervision and ``Base" is the encoder-decoder architecture with residual blocks in Figure~\ref{fig:framework}.}
	\label{table:quentify_ablation}
	\resizebox{0.48\textwidth}{!}{%
		\begin{tabular}{cccccc}
			\toprule
			& Network   			& ITA                  & LDS        & PSNI                 & SSIM                 \\
			\midrule
			\multirow{6}{*}{SOTS}  &	Base &                      &           &         23.80            &   0.881                 \\
			&	Base	& \checkmark            &           &      27.32                 &    0.929                 \\
			&	Base	& \checkmark            & \checkmark &     28.13                 &       0.941              \\ 
			
			&	MSBDN-DFF	 &                      &           &         33.79            &    0.983                 \\
			&	MSBDN-DFF	& \checkmark            &           &          34.45            &    0.984                 \\
			&	MSBDN-DFF	& \checkmark            & \checkmark &     \textcolor{red}{35.26}                 &       \textcolor{red}{0.985}            \\
			\bottomrule
			
	\end{tabular}}
\end{table}

The quantitative results of the intra-domain adaptation are shown in Table \ref{table:quentify_ablation}, which demonstrates that base network with the intra-domain adaptation and the loss-based deep supervision achieves the best performance. To further prove that the improvement on performance is promoted by the intra-domain adaptation, we compare the intra-domain gap under all three methods. Instead of directly measuring the distribution similarity of the features which is not intuitive, we utilize dehazing losses to measure the intra-domain gap. In other words, if the dehazing losses of the same scene are less discrete, the features of the same scene are more closely aligned. Specifically, we calculate the range, the standard deviation and the coefficient of variation of the dehazing losses in each scene and take the average of all scenes. The results are shown in Figure \ref{fig:loss_var}. From the results, we can observe that dehazing losses decrease faster after we apply the intra-domain adaptation to base network. Moreover, dehazing losses are more compact in methods with the intra-domain adaptation which demonstrates that the features of the same scene images are aligned in feature space.

In the inter-domain part, we conduct ablation study with the following settings: 1) \textbf{BS}: base network; 2) \textbf{BS+ITE}: base network with the inter-domain adaptation; 3) \textbf{BS+ITA}: base network with the intra-domain adaptation; 4) \textbf{BS+ITA+ITE}: base network with the intra-domain and the inter-domain adaptation. 

The visual results are shown in Figure \ref{fig:ablation_real}. The result of base network has residual haze due to the domain gap. This phenomenon is alleviated by the intra-domain adaptation or the inter-domain adaptation. However, color distortion appears if the inter-domain adaptation is directly applied because the network is sensitive to the haze distribution of the input image. Base network with intra-domain adaptation and inter-domain adaptation achieve the best performance.

\section{Conclusion}

In this paper, we propose a two-step dehazing network (TSDN) which consists of an intra-domain adaptation step and a constrained inter-domain adaptation step. First, we subdivide the distributions within the synthetic domain into subsets and mine the optimal subset (easy samples) by loss-based supervision. Then, we propose the intra-domain adaptation within the synthetic domain to alleviate the distribution shift. Specifically, we align features with different haze distributions to base feature (easy sample) by adversarial learning. Finally, we conduct the constrained inter-domain adaptation from the real domain to the optimal subset of the synthetic domain, alleviating the domain shift between domains as well as the distribution shift within the real domain. Moreover, when the distribution is aligned to the easy sample, the difficulty of image dehazing is reduced, which enhances the performance. Extensive experimental results demonstrate that our method performs favorably against the state-of-the-art algorithms both on the synthetic datasets and the real datasets.


%

\ifCLASSOPTIONcaptionsoff
  \newpage
\fi



%
%
%

\bibliographystyle{./IEEEtran}
\bibliography{./IEEEabrv,./IEEEexample}

%








\end{document}